\documentclass[journal]{IEEEtran}
\ifCLASSINFOpdf
\else
\fi

\usepackage{amsmath}
\usepackage{amssymb}
\usepackage{booktabs}

\usepackage{graphicx}
\usepackage{float}
\usepackage{subfig}
\usepackage{multirow}
\usepackage{xcolor}

\usepackage[capitalize]{cleveref}
\crefname{section}{Sec.}{Secs.}
\Crefname{section}{Section}{Sections}
\Crefname{table}{Table}{Tables}
\crefname{table}{Tab.}{Tabs.}

\usepackage{cite}

\begin{document}
%
\title{Don't worry about mistakes! Glass Segmentation Network via Mistake Correction}
%
%
%

\author{Chengyu~Zheng,  Peng~Li, Xiao-Ping~Zhang, \textit{Fellow}, \textit{IEEE}, Xuequan~Lu, \textit{Senior Member}, \textit{IEEE},  Mingqiang~Wei, \textit{Senior Member}, \textit{IEEE}
\thanks{C. Zheng, P. Li and M. Wei are with the School of Computer Science and Technology, Nanjing University of Aeronautics and Astronautics, Nanjing, China (zhengcyluda@gmail.com; xiaolizi12625@163.com; mingqiang.wei@gmail.com).}
\thanks{X.-P. Zhang is with the Department of Electrical, Computer and Biomedical Engineering, Ryerson University, Toronto, Canada (xzhang@ee.ryerson.ca).}
\thanks{X. Lu is with the Schoo of Information Technology, Deakin University, Victoria, Australia (xuequan.lu@deakin.edu.au).}
}

%
%

\markboth{Journal of \LaTeX\ Class Files,~Vol.~14, No.~8, August~2015}%
{Shell \MakeLowercase{\textit{et al.}}: Bare Demo of IEEEtran.cls for IEEE Journals}
%



\maketitle

\begin{abstract}
Recall one time when we were in an unfamiliar mall. We might mistakenly think that there exists or does not exist a piece of glass in front of us. Such mistakes will remind us to walk more safely and freely  at the same or a similar place next time. To absorb the human mistake correction wisdom,  
  we propose a novel glass segmentation network to detect transparent glass, dubbed GlassSegNet. 
  Motivated by this human behavior, GlassSegNet utilizes two key stages: the identification stage (IS) and the correction stage (CS).
  The IS is designed to simulate the detection procedure of human recognition for identifying transparent glass by global context 
  and edge information. 
  The CS then progressively refines the coarse prediction by correcting mistake regions based on gained experience. 
  Extensive experiments show clear improvements of our GlassSegNet over thirty-four state-of-the-art methods on three benchmark datasets.
\end{abstract}

\begin{IEEEkeywords}
GlassSegNet, mistake correction, glass segmentation, identification and correction
\end{IEEEkeywords}

%
\IEEEpeerreviewmaketitle

\section{Introduction}
%
%
%
%
\label{sec:intro}
Transparent glass allows light to flow through, leading to its appearance inherited from the covered scene. 
To facilitate downstream applications (e.g., depth estimation, robotic navigation),
the glass in captured images should be well segmented in advance. 
Glass segmentation seeks to identify transparent glass regions in images \cite{mei2020don, xie2020segmenting}.

Recall one time when we entered an unfamiliar environment (e.g., a shopping mall). 
We might feel nervous that we mistakenly think there is a piece of glass in front of us, or maybe we are careless and do not realize there is a piece of glass. \textbf{\textit{Why does it happen?}} 
In fact, there exist two simple yet objective facts for such human illusions: (1) the high variability of unknown backgrounds covered by the glass, and (2) the mutual characteristic between the glass and non-glass regions. 
The two types of mistakes will remind us to walk more safely and freely at the same or a similar place next time. 
This human behavior can be referred to as the \textit{mistake correction} principle. 
We raise an intriguing question – can imitating the mistake correction behavior of humans benefit the performance boost of glass segmentation models? 
We attempt to answer this question by proposing a novel glass segmentation network motivated by the mistake correction principle. 

\begin{figure*}[t]
      \centering
     \subfloat[Input]{
      \begin{minipage}[t]{0.23\textwidth}
            \centering
            \includegraphics[width=1\linewidth]{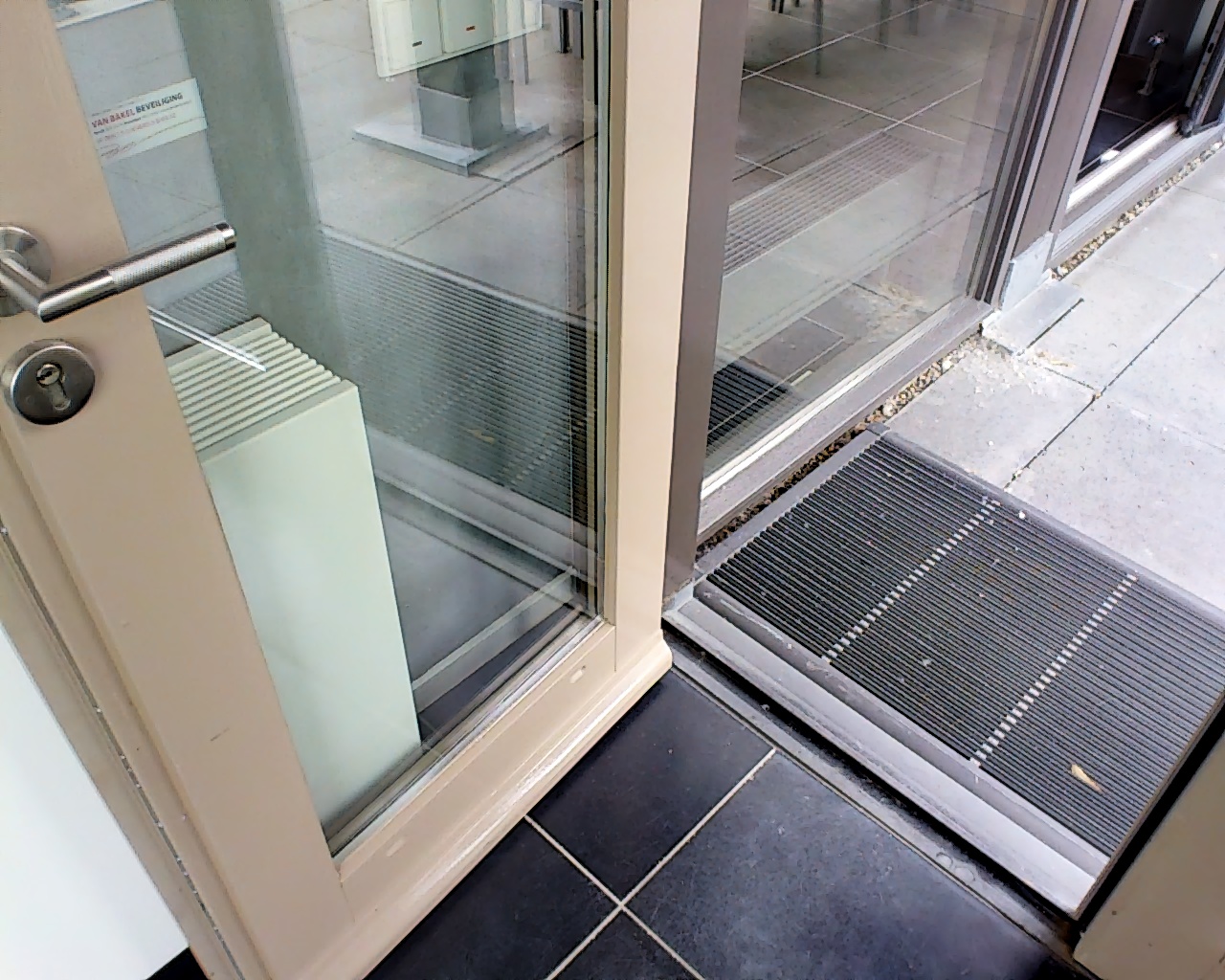}
      \end{minipage}
      }
     \subfloat[GT]{
      \begin{minipage}[t]{0.23\textwidth}
            \centering
            \includegraphics[width=1\linewidth]{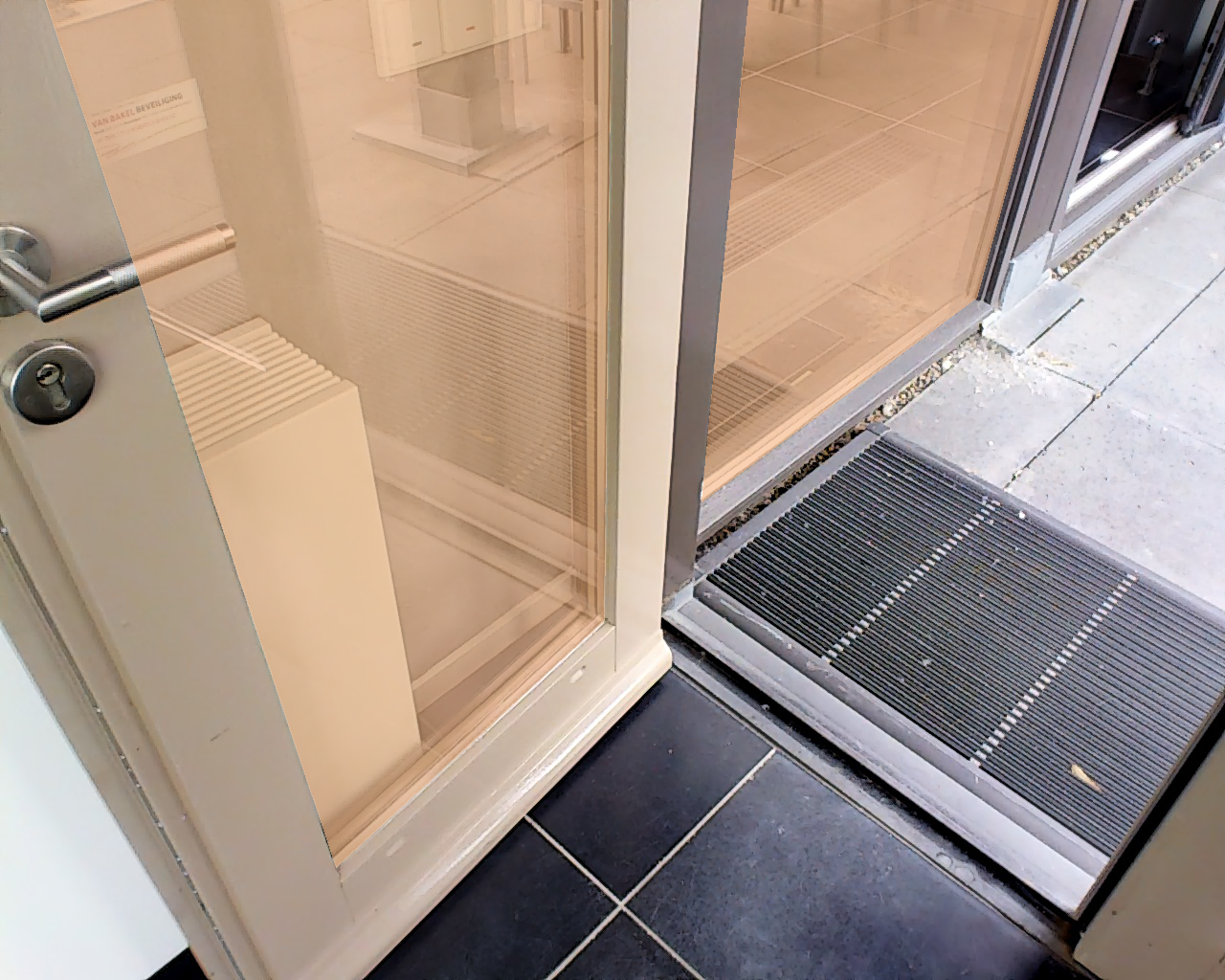}
      \end{minipage}
      }   
      \subfloat[GDNet \cite{mei2020don}]{
      \begin{minipage}[t]{0.23\textwidth}
            \centering
            \includegraphics[width=1\linewidth]{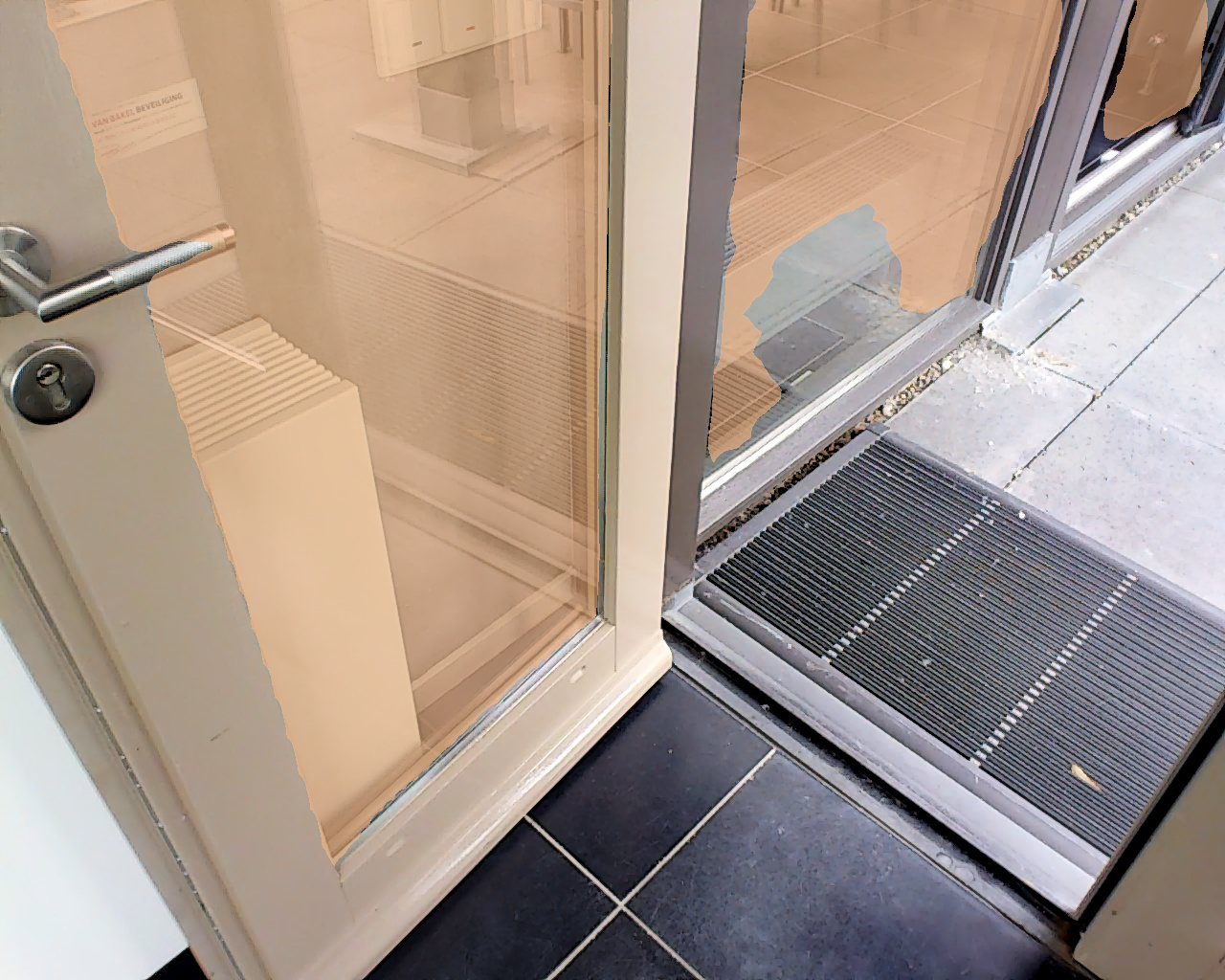}
      \end{minipage}
      } 
      \subfloat[GlassSegNet (ours)]{
      \begin{minipage}[t]{0.23\textwidth}
            \centering
            \includegraphics[width=1\linewidth]{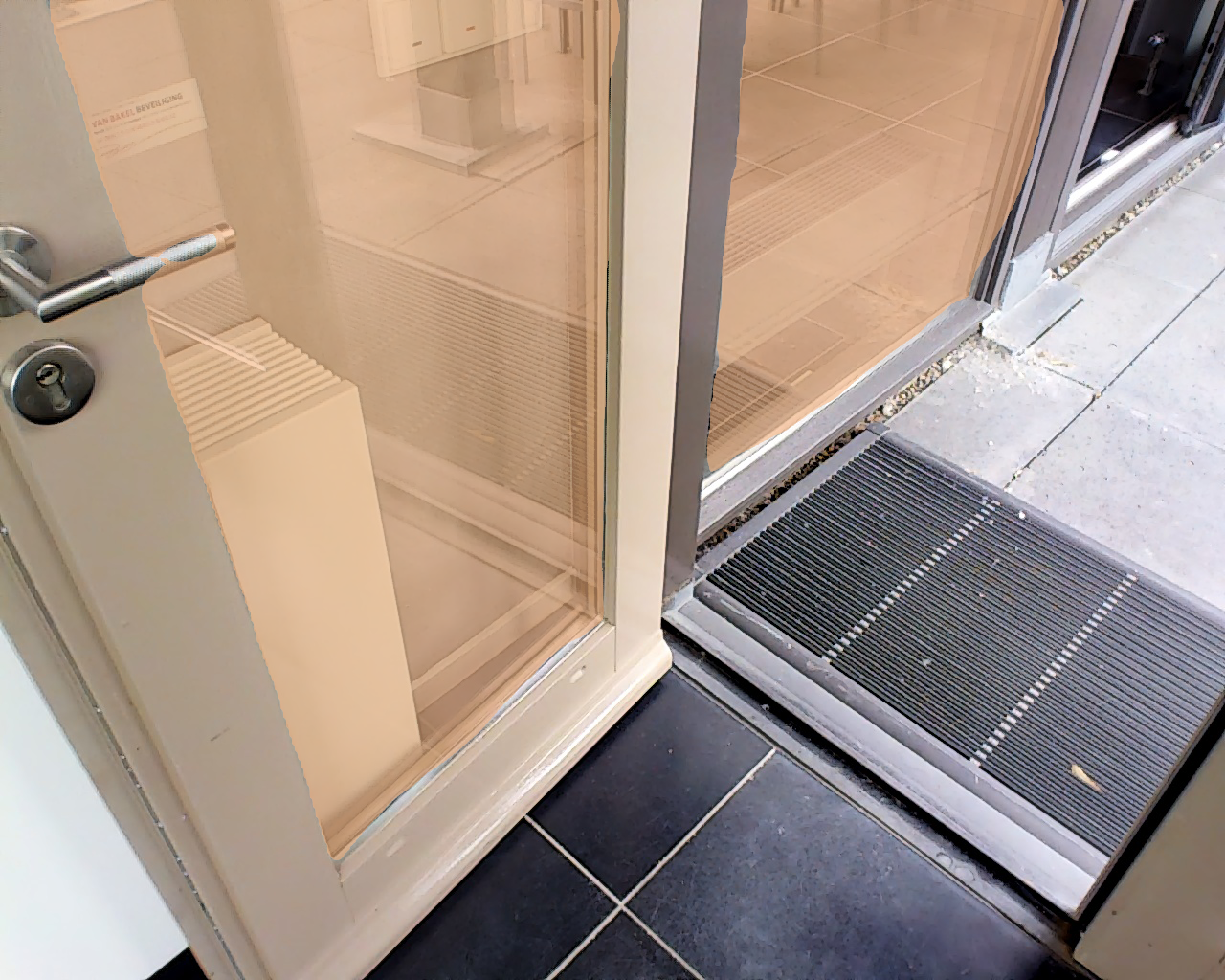}
      \end{minipage}
      } 
   
      \subfloat[TP]{\label{TP}
      \begin{minipage}[t]{0.23\textwidth}
            \centering
            \includegraphics[width=1\linewidth]{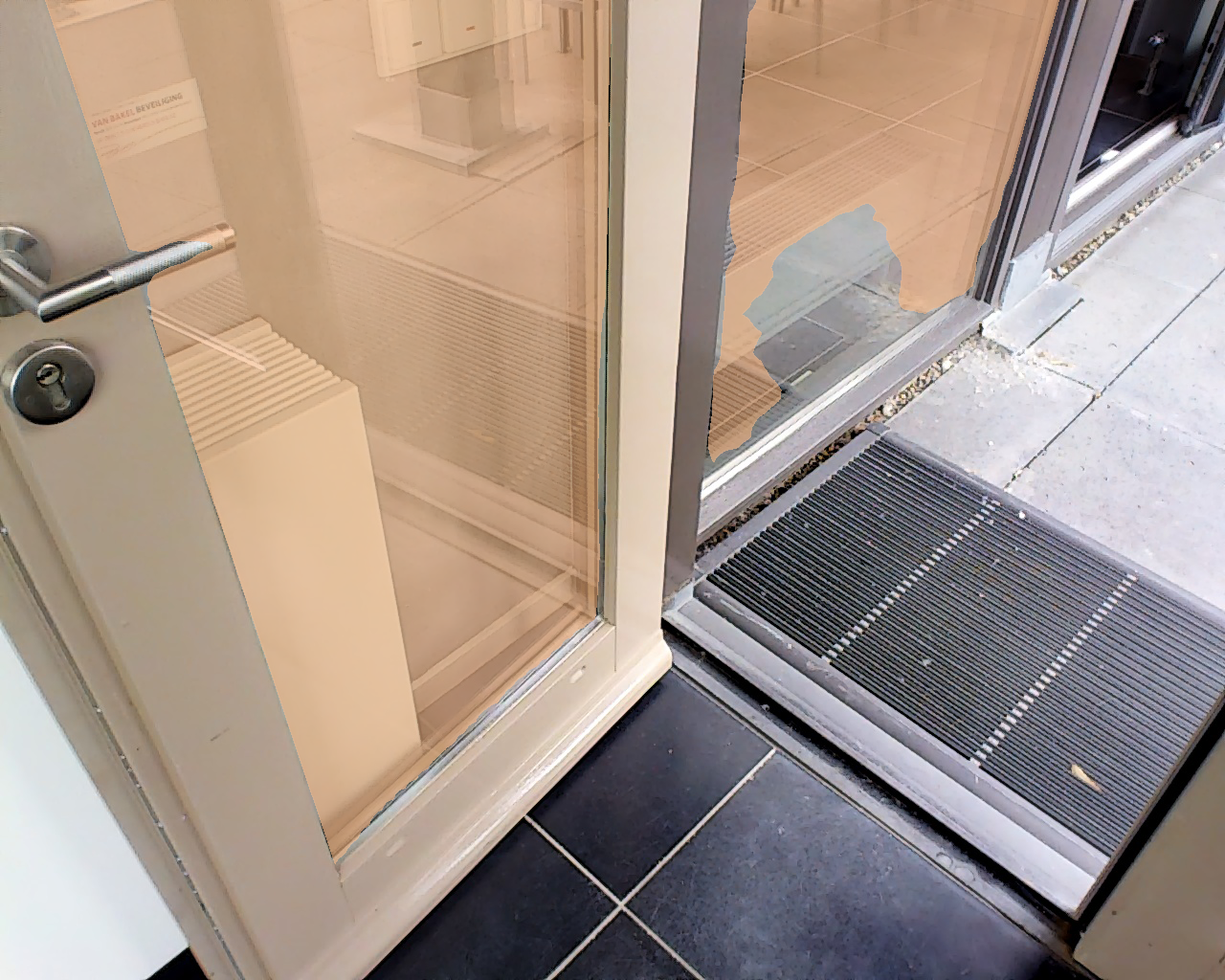}
      \end{minipage}
      }
      \subfloat[TN]{\label{TN}
      \begin{minipage}[t]{0.23\textwidth}
            \centering
            \includegraphics[width=1\linewidth]{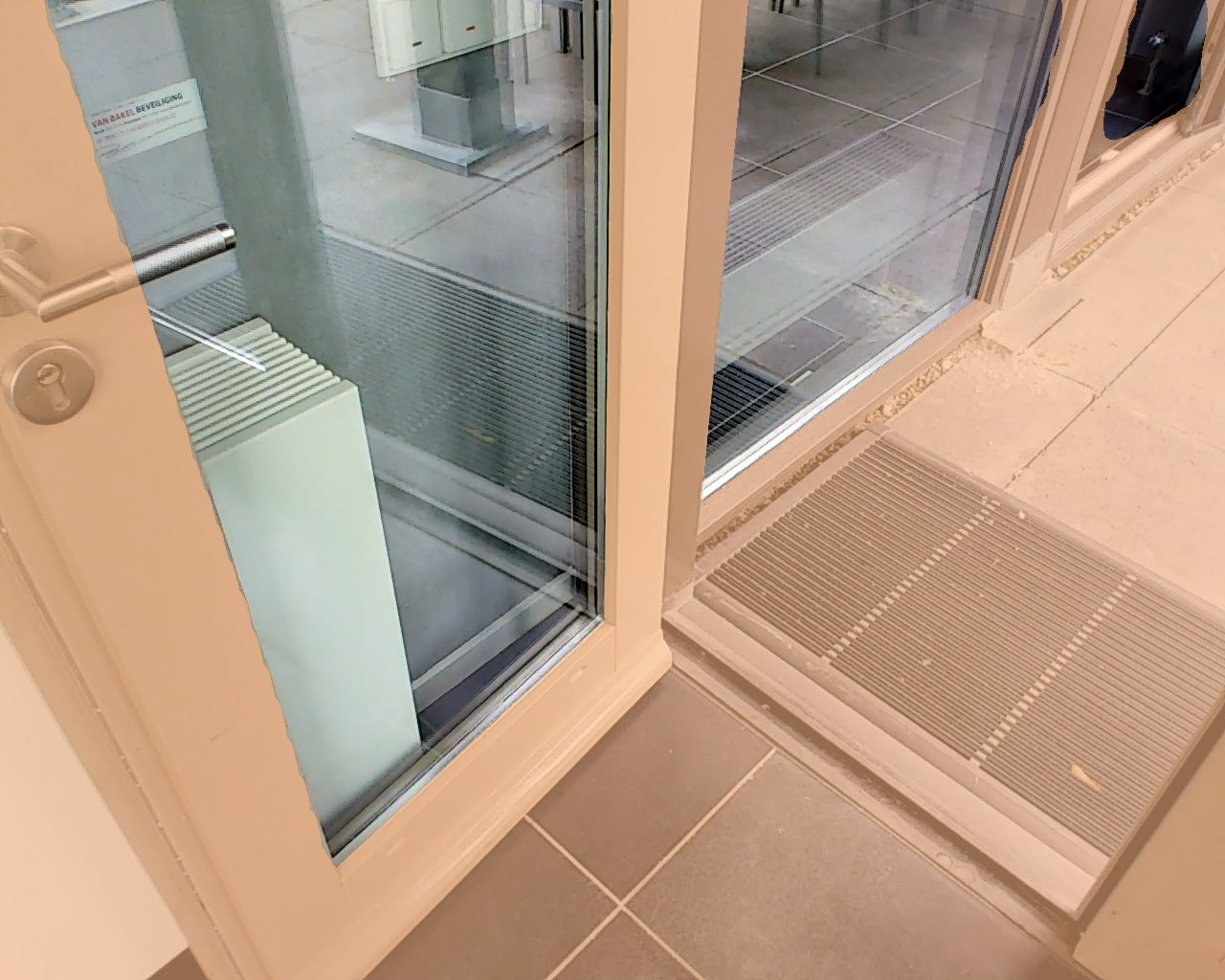}
      \end{minipage}
      }          
      \subfloat[FP]{\label{FP}
      \begin{minipage}[t]{0.23\textwidth}
            \centering
            \includegraphics[width=1\linewidth]{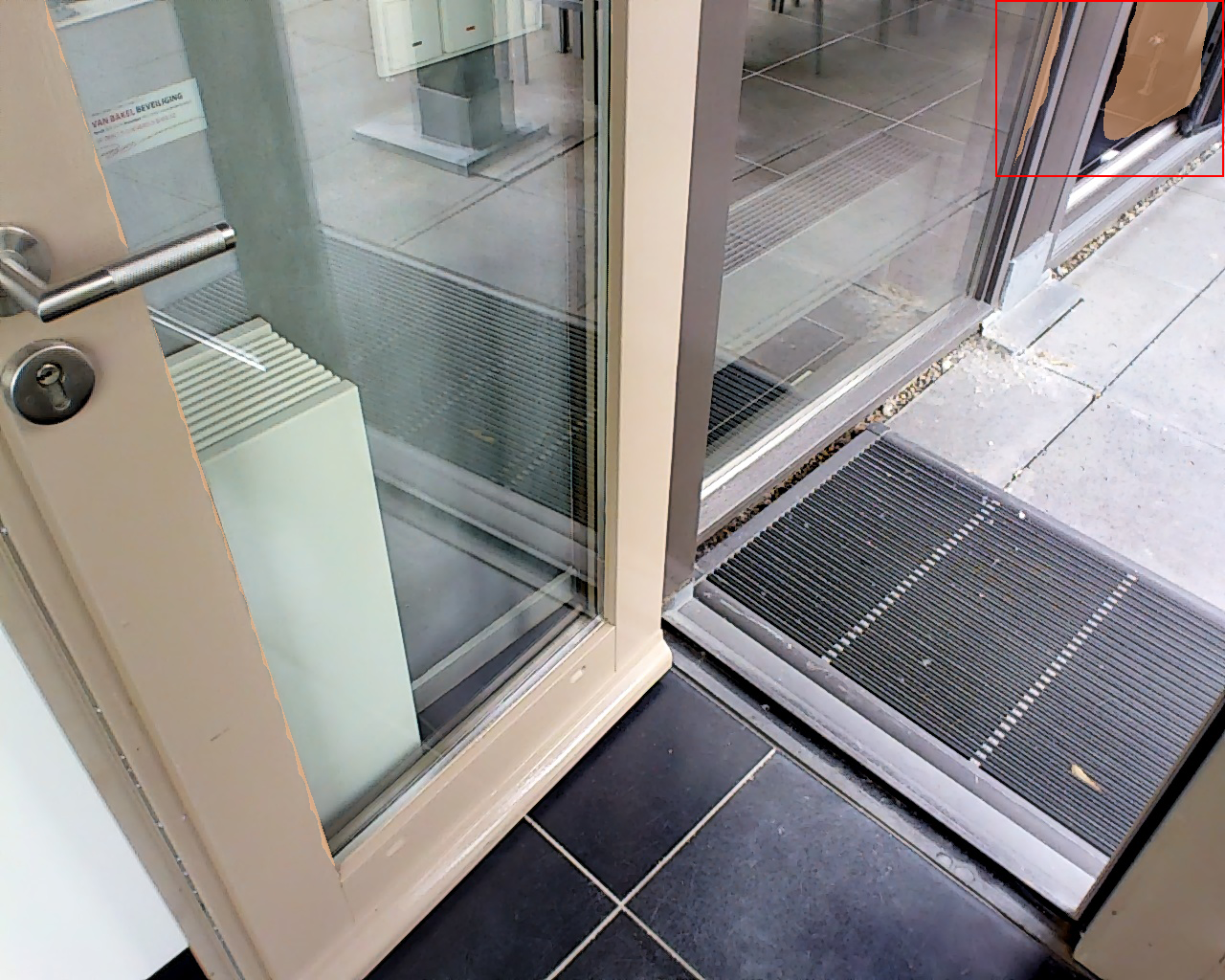}
      \end{minipage}
      }
      \subfloat[FN]{\label{FN}
      \begin{minipage}[t]{0.23\textwidth}
            \centering
            \includegraphics[width=1\linewidth]{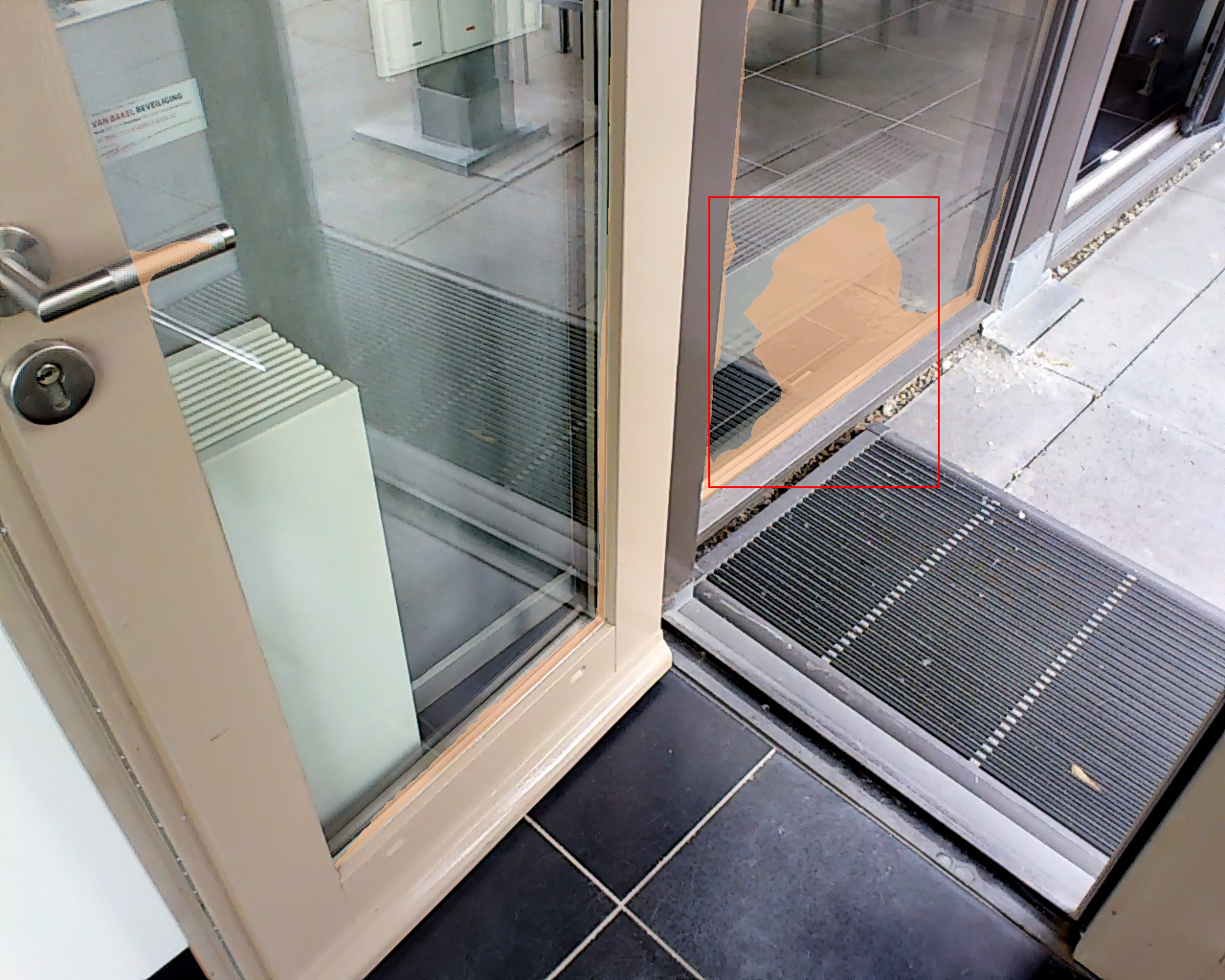}
      \end{minipage}
      }   
      \caption{Glass segmentation is a challenging vision task. 
      Existing methods, e.g., GDNet \cite{mei2020don}, can achieve a glass segmentation result which is decomposed into some of the four segmentation types: 
      \textbf{TP}: true positive glass regions, i.e., correctly detecting the glass regions as glass; \textbf{TN}: true negative non-glass regions, i.e., correctly detecting non-glass regions as non-glass; \textbf{FP}: false positive glass regions, i.e., incorrectly detecting non-glass regions as glass; \textbf{FN}:  false negative non-glass regions, i.e., incorrectly detecting glass regions as non-glass. 
      In the four prediction types, FP and FN are two mistake cases.
      By absorbing the human mistake correction wisdom, our GlassSegNet can largely avoid the FP and FN cases. }
      \label{fig:distra}
\end{figure*}

We empirically observe four types of glass segmentation results by existing methods: 1) true positive glass regions (TP), 2) true negative non-glass regions (TN), 3) false positive glass regions (FP), and 4) false negative non-glass regions (FN).  
The last two types, i.e., the false cases, are the main factors leading to poor glass segmentation for existing methods (see \cref{fig:distra}). 
By mimicking the mistake correction behavior of humans, we surprisingly find that these two types of false cases can reinforce the glass segmentation ability of our GlassSegNet.

GlassSegNet is a novel glass segmentation network to detect transparent glass in images. 
First, GlassSegNet identifies the transparent glass regions using the global context and edge information in the identification stage. Specifically, a criss-cross strip attention (CCSA) module is designed to capture long-range semantic dependencies in terms of spatial position in a non-local way. CCSA can infer the initial location of the transparent glass from a global perspective. Meanwhile, we utilize an edge block to explicitly model glass edge information to locate the glass regions and preserve the glass boundaries. Second, a mistake correction module is proposed in the correction stage. It discovers the false-positive (false-negative) mistakes and then removes these mistakes to yield better representations of the glass: 
it connects features at different levels to extract FN and FP mistake features induced by visual ambiguities, and generates FN maps and FP maps by gained experience for supervision; 
the FN map is added to the input features to obtain FN-augmented features, enabling the model to better discriminate the FN mistakes; then the FP map is subtracted from the FN-augmented features to make the model less vulnerable to the FP mistakes. 
Such a mistake correction strategy is applied to different levels of features to refine the segmentation results progressively. 

Experimental results show our GlassSegNet outperforms state-of-the-art methods on three public benchmark datasets. 
To summarize, we have the main contributions as follows:
\begin{itemize}
\item We introduce the concept of human mistake correction behavior to glass segmentation, and accordingly propose an effective glass segmentation network to yield quality segmentation results. To our knowledge, this is the first time to do so. 
\item We propose GlassSegNet, which first identifies the transparent glass regions using the global context and edge information and then focuses on mistake discovery and removal to progressively refine the segmentation results. 
\item Extensive experiments verify that the imitation of mistake correction from humans is very promising to improve the performance  of glass segmentation. 
As a result, our GlassSegNet soundly exhibits clearer improvements than 34 state-of-the-art methods on three benchmark datasets. 
\end{itemize}

\section{Related Work}
\label{sec:relat}
Glass segmentation is 
similar to binary segmentation, such as salient object detection (SOD), camouflaged object detection (COD). It only assigns one of the two categories to each pixel: one is the foreground (the region that needs to detect, i.e., glass), and the other is the background (the other region, i.e., non-glass).
However, it is challenging to apply existing binary segmentation techniques to detecting glass, since glass regions are highly transparent. 


\subsection{Semantic Segmentation} 
Semantic segmentation is a key problem in computer vision, aiming to assign a semantic class label to each pixel in the given image. 
With the development of deep neural networks, an end-to-end training architecture method called fully convolutional networks (FCNs) \cite{long2015fully} has been proposed to solve this problem, which uses multi-scale context fusion to achieve high segmentation performance. 
Specifically,  multi-level encoder features are combined with their corresponding decoder features by  concatenation \cite{yang2021densely}. 
Chen et al. \cite{chen2017deeplab} introduced an atrous spatial pyramid pooling module (ASPP) with multi-scale dilation convolutions for contextual information aggregation. 
Zhao et al. \cite{zhao2017pyramid} further proposed PSPNet to capture a wider range of contextual information by using a pooling operation and the pyramid structure. 
Wang et al. \cite{wang2018non} introduced non-local networks utilizing a self-attention mechanism \cite{vaswani2017attention, cheng2016long}, which calculates the relationship between each pixel and all other pixels in an image, thus capturing global contextual information. 
Chen \cite{chen2021net} developed a dual-branch joint network that utilizes the global context guidance generated from the first branch to refine the second branch.

However, applying existing segmentation methods for glass segmentation (i.e., treating glasses as one of the object categories) cannot solve the fundamental problem of glass segmentation, in that the area of glass may also be segmented. 

\subsection{Salient Object Detection (SOD)}
SOD aims at identifying the most visually distinctive objects or regions in an image. 
Most of the early SOD methods \cite{li2015visual, wang2015deep, zhao2015saliency, liu2016dhsnet} in deep learning directly used the features from the backbone network to produce the saliency map. 
Furthermore, there are also some methods achieving remarkable performance by combining multi-level CNN features in different ways \cite{zhang2017amulet, zhang2018bi, wang2019iterative, pang2020multi, xu2021exploring, lin2021residual, tian2022learning}. 
Recently, boundary information has been leveraged to enhance salient object detection performance \cite{li2018contour, feng2019attentive, qin2019basnet, zhao2019egnet, su2019selectivity, wu2019stacked, zhou2020interactive, wei2020label, chen2020contour}. 
Nguyen et al. \cite{nguyen2019deepusps} proposed a two-stage mechanism for robust unsupervised salient object detection.
Zhang et al. \cite{zhang2020adaptive} presented a novel adaptive graph convolutional network with attention graph clustering for co-saliency detection. 

Applying these SOD methods may not be appropriate for glass segmentation, as the background region in glass scenes may or may not be salient. 
Even if it is salient, it is likely that only part of the glass is salient. 

\subsection{Camouflaged Object Detection (COD)}
COD \cite{lin2022frequency} aims to identify objects that ``perfectly'' assimilate into their surroundings, which has a wide range of valuable applications. 
Fan et al. \cite{fan2020camouflaged} proposed a large-scale dataset and a robust network called SINet for camouflaged object detection, which mimics the process of predation in nature to segment camouflaged objects. 
Mei et al. \cite{mei2021camouflaged} developed the Positioning and Focus Network (PFNet), which mines distraction information to refine the coarse prediction. 
Li et al. \cite{li2021uncertainty} proposed a paradigm of leveraging contradictory information to enhance the ability of both salient object detection and camouflaged object detection. 
Although camouflaged object detection and glass segmentation are very similar, the purpose of glass segmentation is to identify the glass as a whole in a complex and changeable environment. Therefore, the techniques for COD might not be optimal in glass segmentation.

\subsection{Specific Region Segmentation}
\textbf{Shadow Detection.}
Shadow detection is also a binary segmentation task.
Khan et al. \cite{panagopoulos2009robust} applied CNN to detect shadow regions.  
Zhu et al. \cite{zhu2018bidirectional} presented a network that explores and combines global/local context in deep/shallow layers of the deep convolutional neural network (CNN) to detect shadow regions.  
Hu et al. \cite{hu2018direction} analyzed image context in a direction-aware manner for shadow detection. 
Zheng et al.  \cite{zheng2019distraction} proposed a distraction-aware shadow detection network, which explicitly learns and integrates the semantics of visual distraction regions in an end-to-end framework.  
However, compared to shadow regions, the information contained in glass regions is more complex. Therefore, directly using the shadow detection methods may not be effective.

\textbf{Mirror Detection.} 
Similar to other image detection tasks, mirror detection aims at segmenting mirror regions in single images. 
Yang et al. \cite{yang2019my} made the first attempt to automatically detect mirrors and proposed MirrorNet by utilizing inconsistencies between the inside and outside of the mirror region, called contextual contrasted features, to segment mirrors from the real scene.
Lin et al. \cite{lin2020progressive} introduced a model to progressively learn the content similarity between the inside and outside of the mirror while explicitly detecting the mirror boundaries. 
Tan et al. \cite{tan2022mirror} re-thought the image-level visual chirality property and reformulated it as a learnable pixel level cue for mirror detection.
Mei et al. \cite{mei2022mirror} proposed a network, called MirrorNet+, for mirror segmentation, by modeling both contextual contrasts and semantic associations. 
Glass detection is very similar to mirror detection which also has the problem of similar foreground and background.

\textbf{Transparent Object Detection (TOD).} 
Transparent object detection (TOD) aims to assign a transparent object  class label to each pixel in the given image. 
It is more similar to the semantic segmentation task, as the transparent object class is not single but includes multiple  classes, e.g., stuff and thing.
Xie et al. \cite{xie2020segmenting} proposed a large-scale dataset named Trans10K and a novel boundary-aware segmentation method called TransLab for TOD. 
Furthermore, they propose a novel transformer-based segmentation pipeline for TOD termed as Trans2Seg, and a new fine-grained transparent object segmentation dataset, termed as Trans10K-v2 \cite{xie2021segmenting}. 
Since the transparent object detection task is similar to semantic segmentation, we experimentally show that the transparent object detection method is not optimal for the binary glass segmentation task.

\textbf{Glass Segmentation.} 
There are no fixed patterns for the glass regions in the images since they depend on what appears behind the glass. This means the content of glass regions is the same as the content of the background region. 
By exploring abundant contextual features from a large receptive field, Mei et al. \cite{mei2020don} pioneered  a novel glass detection network (GDNet).
Lin et al. \cite{lin2021rich} proposed a rich context aggregation module (RCAM) to extract multi-scale boundary features and a reflection-based refinement module (RRM) to detect reflection. Yu et al. \cite{yu2022progressive} developed a novel feature fusion strategy for excavating useful information via both focus and exploration between features. In addition, recent work \cite{mei2022glass, huo2022glass} has improved the accuracy of glass segmentation tasks by using multimodal information.

\section{Proposed Methodology}
\label{sec:metho}
\textbf{Motivation}. Humans may have an illusion when they enter an unfamiliar environment that contains glass. That is, they mistakenly think there is a piece of glass in front of them, or they do not realize there is a piece of glass in front of them. But next time, they are very likely to avoid such mistakes in the same or similar places since they have learned such knowledge. We consider that a glass segmentation network can also behave like humans who have ever entered an environment more than one time. 
Thus, inspired by the behavior of human mistake correction, we design the GlassSegNet. 
GlassSegNet will produce more accurate glass segmentation maps by finding and removing mistake regions.

\subsection{Overview}
\begin{figure*}[htbp]
  \centering
  \includegraphics[width=0.9\linewidth]{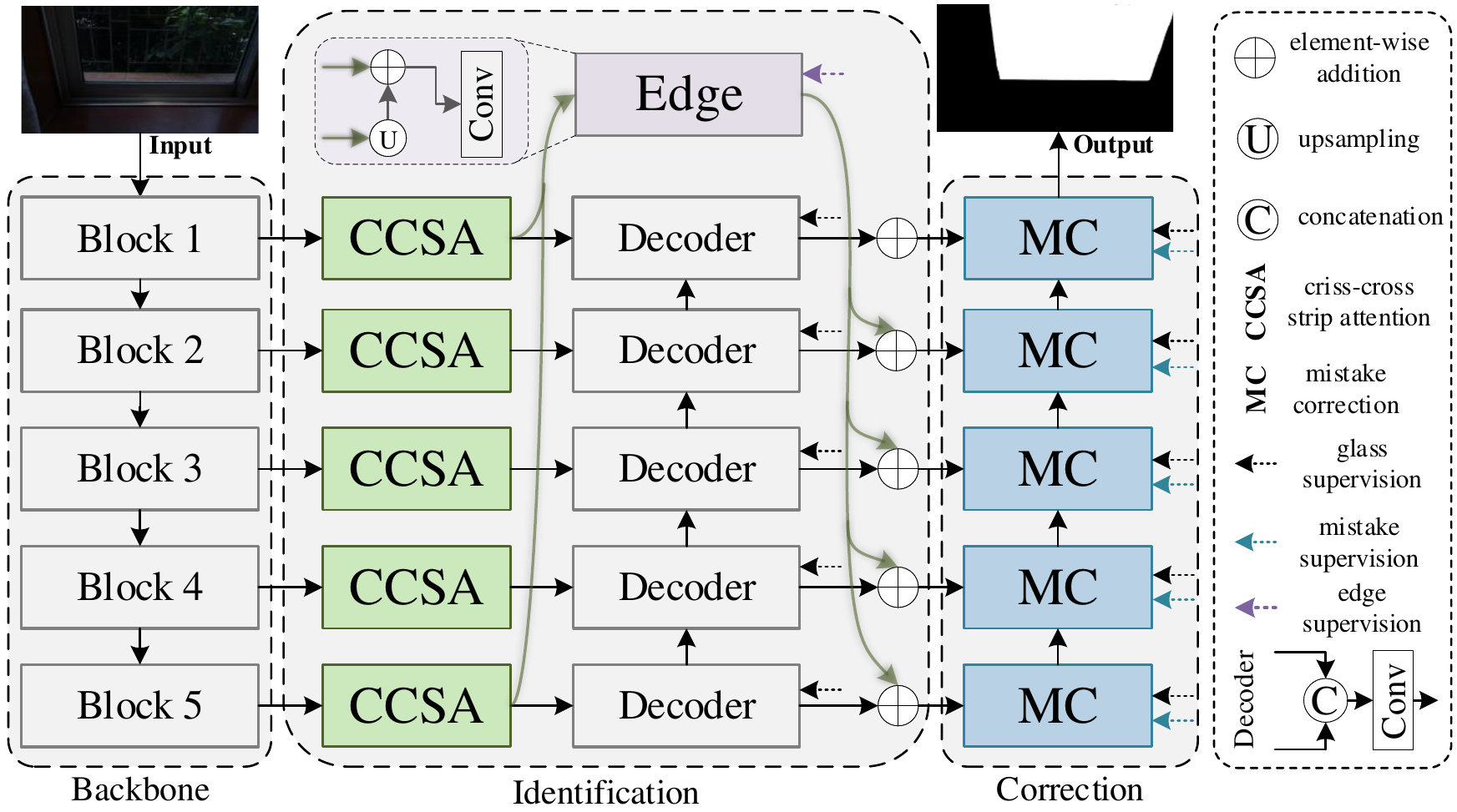}
  \caption{Network architecture of GlassSegNet. It mainly consists of three stages. In the first stage, ResNet-50 \cite{he2016deep} is applied to obtain multi-scale features, and the channel numbers of the obtained features are reduced by $3\times3$ convolution layers. In the identification stage, the transparent glass region is identified by the criss-cross strip attention (CCSA) module and the edge block (Edge). In the correction stage, a mistake correction (MC) module is used to perform the mistake correction for progressively refining the coarse prediction via correcting mistake regions using existing experience.    
}
\label{fig:overview}
\end{figure*}

\Cref{fig:overview} illustrates the pipeline of our proposed network, which uses ResNet-50 \cite{he2016deep} as the backbone. 
First, we feed a single RGB image $I$ into the backbone network to extract multi-scale backbone features $F_{ba}=\{f_i|i=0,1,2,3,4\}$. 
In order to reduce the computation, we decrease the number of channels of $F_{ba}$ using four $3 \times 3$ convolutional layers. 
Then, in the identification stage, the criss-cross strip attention (CCSA) module is proposed to locate the potential transparent glass regions based on the multi-level features, while the edge block (Edge) is used to locate the glass regions and preserve the glass boundary.
Finally, we leverage a mistake correction (MC) module in the correction stage to discover the false-positive (false-negative) mistakes and then remove these mistakes to get better representations of the transparent glass. 

\subsection{Identification Stage}
\label{sec:IS}
The identification stage contains two key modules, i.e., the CCSA module and the edge block (Edge), to locate the transparent glass regions. Specifically, CCSA is proposed to capture long-range semantic dependencies in terms of spatial position in a non-local way to infer the initial position of the transparent glass from a global perspective, and Edge is utilized to explicitly model glass edge information to help locate glass areas and preserve glass boundaries.

\begin{figure*}[htbp]
  \centering
  \includegraphics[width=0.9\linewidth]{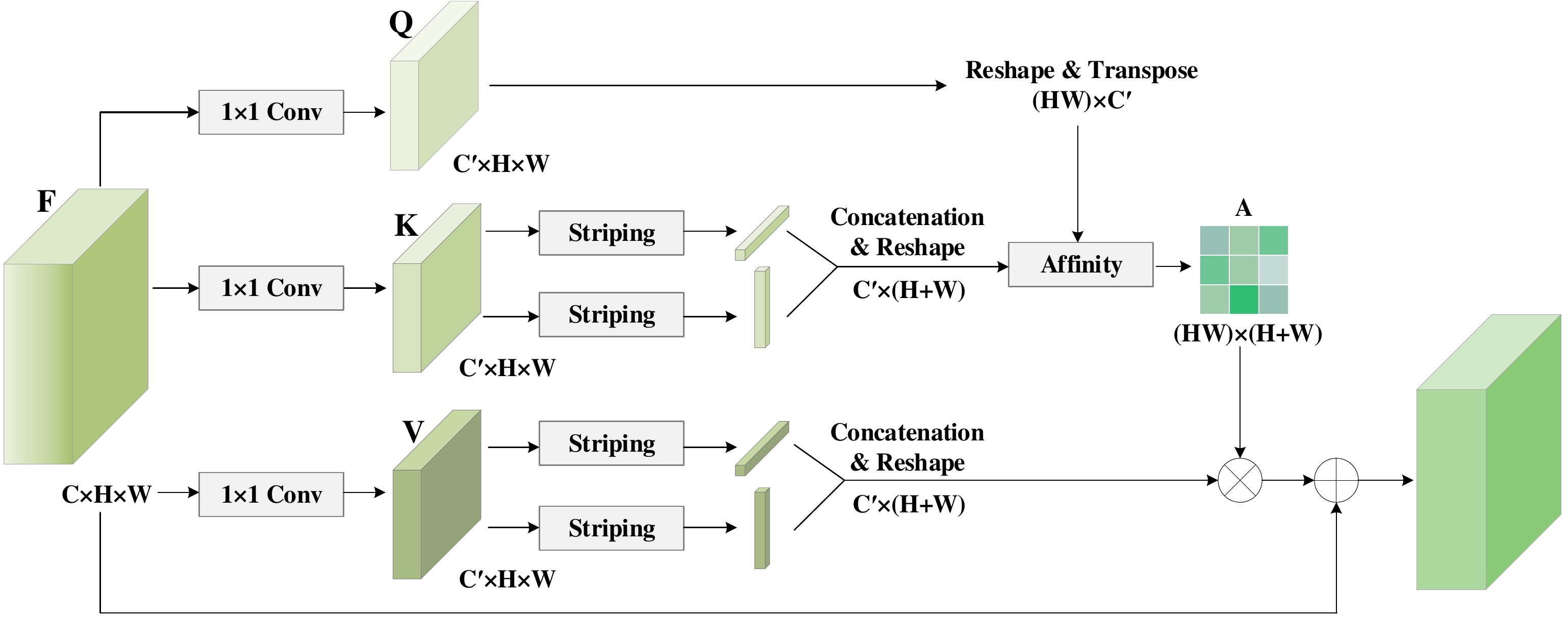}
   
  \caption{
      The details of Criss-Cross Strip Attention Module.
  }
  \label{fig:CCSA}
\end{figure*}

\textbf{Criss-Cross Strip Attention (CCSA) Module.} \Cref{fig:CCSA} illustrates the detailed structure of our designed CCSA module. Given the input features, the CCSA module aims to learn semantic-enhanced high-level features and further generate the initial segmentation map. It is implemented in a non-local way, to capture long-range dependencies in terms of channel and spatial position, for enhancing the semantic representation of the highest-level features from a global perspective. To reduce the computational complexity in time and space, we introduce strip attention and criss-cross operations. 
We apply a striping operation to collect contextual information in horizontal and vertical directions and further use a criss-cross attention operation to gather global affinity information between each pixel to enhance the pixel-wise representative capability.

Specifically, given the input features $F_{ba} \in \mathbb{R}^{C\times H\times W}$, where $C$ is the number of channels, $H$ and $W$ are the spatial dimensions of the input tensor. We first feed $F_{ba}$ into two convolution layers with $1\times 1$ filters to generate two new features $Q$ and $K$ respectively, where$\left\{Q, K\right\} \in \mathbb{R}^{C'\times H \times W}$. $C'$ is the channel number of features, which is less than $C$ for dimension reduction. We then apply a Striping operation on the horizontal and vertical directions of the feature $K$, representing average pooling with a pooling window of size $W \times 1$ and $H \times 1$, to generate striped features by encoding the context representation in different directions. Finally, the generated strip features are concatenated to obtain the local features $K' \in \mathbb{R}^{C'\times (H + W)}$.

We reshape and transpose the local features $Q$ into $\mathbb{R}^{(HW) \times C'}$ to obtain the feature $Q'$. After that, we perform an Affinity operation between $Q'$ and $K'$ to further calculate the attention map $A \in \mathbb{R}^{(HW) \times (H+W)}$ along the horizontal direction. The Affinity operation is defined as follows:
\begin{equation}
      A_{j,i} = \frac{exp(Q'_i \cdot K'_j)}{\sum_{i=1}^{N}exp(Q'_i \cdot K'_j)}, 
\end{equation}
where $A_{j,i} \in A$ denotes the degree of correlation between $Q'_i$ and $K'_j$.

Meanwhile, we feed $F_{ba}$ into another convolutional layer with a kernel size of $1 \times 1$ to generate the feature map $V\in \mathbb{R}^{C'\times H \times W}$. Similar to the above operation, for $V$ we can obtain the representation maps in the vertical dimension and in the horizontal dimension, with the spatial dimensions of $C \times 1 \times W$ and $C \times H \times 1$, respectively. Then $V' \in \mathbb{R}^{(H+W) \times C'}$ is obtained by using concatenation and reshape operations on them. We perform a matrix multiplication between $A$ and $V'$, and reshape the result to $\mathbb{R}^{C'\times H \times W}$. Finally, we perform an element-wise sum operation with $F_{ba}$ to obtain the final output $F_{ccsa} \in \mathbb{R}^{C\times H \times W}$ as follows:
\begin{equation}
      F_{ccsa}^j = \sum_{i=1}^NA_{j,i}\cdot V_i+F_{ba}^j
\end{equation}
where $F_{ccsa}^j$ is a feature vector in the output feature map $F_{ccsa}$ at the position $j$. The final feature $F_{ccsa}$ models the long-range semantic dependencies between all positions and thus is more discriminative than the input feature $F_{ba}$.

\textbf{Edge Block.} 
The edge information is useful and is widely used  in existing salient object detection methods.  Meanwhile, the edges are also an important identification feature of the glass region.  We propose an Edge Block to explicitly model the transparent glass information and glass edge information that are complementary, 
in order to preserve the transparent glass boundaries. At the same time, the glass edge features are also helpful for localization.

In this block, we aim to model the glass edge information and extract the glass edge features. Since the feature $F_{ccsa1}$ preserves better edge information, we extract local edge information from it.  However, in order to get glass edge features, only local information is not sufficient. High-level semantic information or location information is also needed.  The receptive field of the feature $F_{ccsa5}$ is the largest among other level features $F_{ccsai}, i=1,2,3,4$, and the location is the most accurate, thus we use a top-down location strategy to propagate the top-level location information to the  feature $F_{ccsa1}$ to restrain the non-glass edges. The fused features $F_{edge}$ can be denoted as:
\begin{equation}
      F_{edge} = Conv(F_{ccsa1} + \phi(F_{ccsa5})),
\end{equation}
where $\phi$ is bilinear interpolation operation that aims to up-sample $F_{ccsa5}$ to the same size as $F_{ccsa5}$. $Conv$ is a convolution block consisting of convolutional layer with a kernel size of $3 \times 3$,  $BatchNorm$ layer, and $ReLU$ activation function. To model the glass edge feature explicitly, we add an extra glass edge supervision to supervise the glass edge features.

After obtaining the complementary glass edge features and glass features, we aim to leverage the glass edge features to guide the glass features to perform better on both segmentation and localization. A simple way is to fuse $F_{edge}$ with the decoder features $F_{de}$ at each level, which will  sufficiently leverage the multi-resolution glass features. Specifically, we add sub-side paths for $F_{de}=\{f_{dei}|i=1,2,3,4, 5\}$. In each sub-side path, by fusing the glass edge features into enhanced glass features, we make the location of high-level predictions more accurate, and more importantly, the segmentation details become better. The process of fusion is denoted as:
\begin{equation}
      F_{en} = F_{de} + \phi(F_{ccsa}).
\end{equation}

\subsection{Correction Stage}
\label{sec:CS}
According to the human mistake correction behavior, we first need to extract the mistake features (i.e., FP and FN), and then remove them to obtain  accurate glass prediction.
Thus, we design a Mistake Correct (MC) module to generate the mistake feature through mistake supervision and remove them according to their characteristics. 

\begin{figure*}[htbp]
  \centering
  \includegraphics[width=0.8\linewidth]{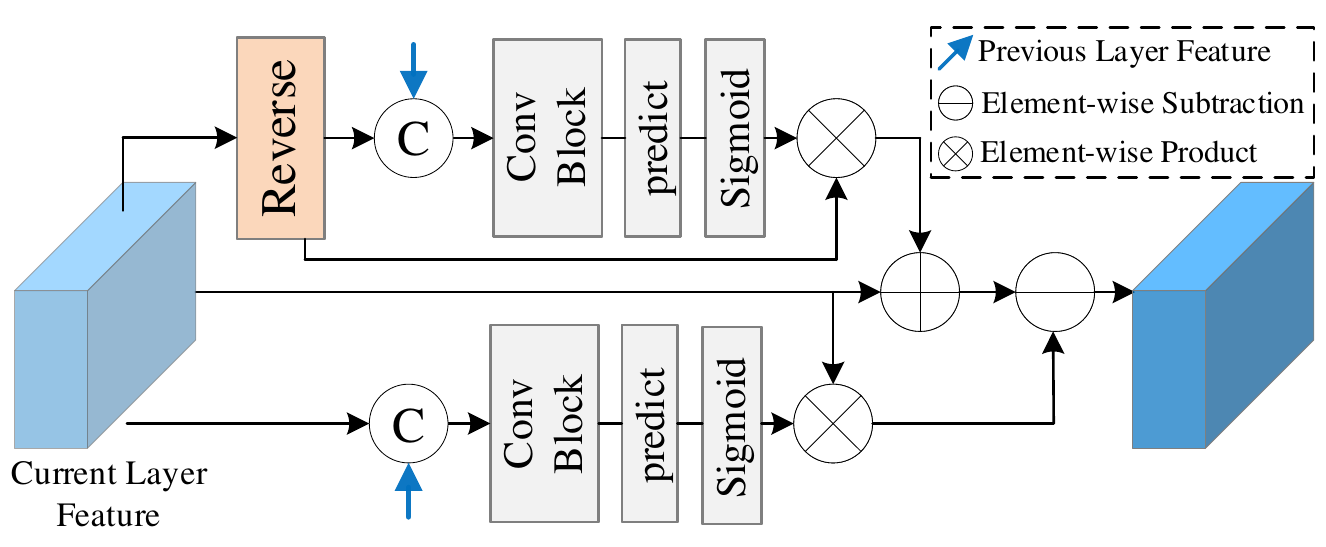}
   
  \caption{
      The details of Mistake Correct Module.
  }
  \label{fig:MCM}
\end{figure*}

The detailed structure of the MC is shown in \Cref{fig:MCM}, which contains two branches: the False Positive (FP) branch and the False Negative (FN) branch. 
The MC module is designed to explicitly learn semantic features of the potential mistake regions and fuse the  mistake features with the input image features to produce mistake-aware features that will be used for glass segmentation. The FN branch and FP branch are discussed in detail below.

\textbf{FN Branch.}  \quad It is designed to learn FN mistake maps, which are used to augment the input enhanced features $F_{en}$. We first apply a reverse operation to the enhanced features and concatenate them to higher-level FN features.  It then employs a ConvBlock, consisting of convolutional layer with a kernel size of $3 \times 3$,  $BatchNorm$ layer, and $ReLU$ activation function, to extract the FN mistake features. To force the FN mistake features to capture the semantics necessary to recognize potential FN mistake regions, we use the FN mistake features for FN mistake prediction, by estimating a soft binary map indicating the possible FN mistake locations on the input image. Then, the masked FN mistake features are obtained by multiplying reverse enhanced features with  soft binary map in an element-wise manner. To enhance the feature activation on the FN mistake region, the masked FN mistake features are added to input features to produce FN-augmented features. The attention mechanism is designed to enable the network to quickly focus on and augment the features around possible FN mistake regions. 

\textbf{FP Branch.}  \quad Similar to the FN branch, the FP branch is used to learn the FP mistake maps, which are used to further enhance the FN-argumented features. We directly concatenate the input features to higher-level FP features. It also adopts a ConvBlock with the same architecture as that of the FN branch to extract the FP mistake features. To force the FP mistake features to capture useful semantics of potential FP mistake regions, we predict a soft binary map of the FP mistake in the same way as the FN branch. We then multiply the input features with the soft binary map (element-wise) to obtain the masked FP mistake features. 
Finally, we subtract the masked FP mistake features from FN-augmented features to eliminate the negative effect of FP features on detection. This would make the network less susceptible to possible FP distraction.

\subsection{Mistake Loss}
Since we supervise the MG block to generate the FP and FN mistake features, their corresponding ground-truth FP and FN maps are often necessary. 
Unfortunately, annotating ground-truth FP and FN maps will be laborious and subjective. 
Currently, there are two main techniques for FP and FN supervision. 
One is implicit supervision, which has been widely used in previous works \cite{chen2018reverse, huang2017semantic, xiao2018deep, zhu2018bidirectional, mei2021camouflaged}.
They design a network that exploits some mistake integration strategies, such as element-wise addition and element-wise subtraction, and just supervise the final prediction of the network. 
The other is explicit supervision \cite{zheng2019distraction}.
It utilizes existing models to predict glass maps and compare them with the corresponding GT glass maps to generate the FP and FN GT maps. 
The network can capture more accurate mistake features by explicit supervision. 

To address the aforementioned issue, 
we thus utilize the strategy proposed by DSDNet \cite{zheng2019distraction} to acquire the approximate FN and FP ground truth based on the differences between the predictions from existing glass segmentation models and their ground truth.
Specifically, we choose two baseline glass segmentation models \cite{mei2020don, lin2021rich} to generate the predictions. 
For each image, we first compute a difference map between the prediction from each of the models and the corresponding ground truth glass map.
Then, we separate the false positive region and false negative region from the difference maps to form the final ground-truth FP and FN maps. 
The computation is defined as: 
\begin{equation}
\begin{aligned}
& Mistake_i = predict_i - GT, \\
& G^{FP}_i = (Mistake_i > 0), \\
& G^{FN}_i = (Mistake_i < 0),
\end{aligned}
\end{equation}
where $\text{predict}_i$ is the prediction map generated by two baseline glass segmentation models, and $GT$ is the corresponding ground truth glass map. 

The mistake loss is formulated as $L_{mistake} = L_{FP} + L_{FN}$. $L_{FP}$ and $L_{FN}$ are the FP and FN mistake losses in the corresponding branch respectively defined as:
\begin{equation}
\begin{aligned}
L_{FP} = -\sum\limits_{j}[G^{FP}_jlog(P_j^{FP}) + (1-G^{FP}_j)log(1-P_j^{FP})],
\end{aligned}
\end{equation}
and
\begin{equation}
\begin{aligned}
L_{FN} = -\sum\limits_{j}[G^{FN}_jlog(P_j^{FN}) +(1-G^{FN}_j)log(1-P_j^{FN})],
\end{aligned}
\end{equation}
where $P_j^{FP}$ and $P_j^{FN}$ denote the value at the $j^{th}$ pixel in the predicted FP and FN maps, respectively.

\subsection{Loss Function}
We train our network to jointly optimize the predicted glass map, edge map, FN map, and FP map by minimizing:
\begin{equation}
L=  \sum_{i} \lambda_i L_{glass}^i + \sum_{i} \lambda_i L_{FN}^i + \sum_{i} \lambda_i L_{FP}^i + L_{edge},
\end{equation}
where $L_{glass}^i$, $L_{FN}^i$ and $L_{FP}^i$ are the loss of the glass map, FN map, and FP map at the $i^{th}$ scale, respectively. 
$L_{edge}$ is the loss of the edge map, and $\lambda_i$ is the weighting factor that is set to 4, 4, 2, and 1,  respectively.

$L_{glass}$ consists of two parts: the weighted BCE loss $l_{wbce}$ and the weighted IoU loss $l_{wiou}$ \cite{wei2020f3net}, i.e., $L_{glass}=l_{wbce}+l_{wiou}$. 
The definition of $l_{wbce}$ is:
\begin{equation}
\begin{aligned}
L_{wbce} = & - \frac{\sum\limits_{j}W_j[G_jlog(P_j) + (1-G_j)log(1-P_j)]
}{\sum\limits_{j}(W_j-1)}. 
\end{aligned}
\end{equation}

$l_{wiou}$ makes the network to focus on the global structure: 
\begin{equation}
\begin{aligned}
l_{wiou} = & 1 - \frac{\sum\limits_{j}(G_jP_j)W_j}{\sum\limits_{j}(G_j + P_j - G_jP_j)W_j}, 
\end{aligned}
\end{equation}
where $G_j$ and $P_j$ are the values at the $j^{th}$ pixel in the GT and  predicted glass maps, respectively. $W_j$ is a weight function:
\begin{equation}
\begin{aligned}
W_j = (1 + \gamma \alpha_j),
\end{aligned}
\end{equation}
where $\gamma$ is a hyper-parameter and set to 5. $\alpha$ is 
calculated by: 
\begin{equation}
\alpha_j = \left| \frac{\sum\limits_{j \in A_j}G_j}{\sum\limits_{j \in A_j} 1} - G_j\right|,
\end{equation}
where $A_j$ is the area surrounding of the $j^{th}$ pixel. $\alpha_j \in [0,1]$ represents the dissimilarity between the $j^{th}$ pixel and its surrounding pixels. 
In addition, $L_{FP}$, $L_{FN}$ and $L_{edge}$ also use the weighted BCE loss $l_{wbce}$.

\section{Experiments}
\label{sec:exper}
\subsection{Experimental Settings}
\textbf{Implementation Details.} 
We use PyTorch \cite{paszke2019pytorch} to implement the proposed network. 
The data augmentation methods are random horizontal flip, color jitter, and random crop. The image is resized to $416\times416$. 
We use the ResNeXt-101 model \cite{xie2017aggregated}  pre-trained on ImageNet \cite{deng2009imagenet} to initialize the parameters of the backbone. The other parameters of the proposed model are initialized randomly. 
We train the whole network by using the stochastic gradient descent (SGD) scheme with a momentum of 0.9 and a weight decay of $ 5\times 10^{-4}$. 
The initial learning rate is set to 0.001. We use poly decay strategies \cite{yu2018bisenet} with a power of 0.9 to adjust the learning rate. The batch size is set to 12, and the network is trained on an NVIDIA GTX 1080 Ti graphics card. 

During testing, the input images are resized to $416 \times 416$ for inference without any pre-processing such as the fully connected conditional random field (CRF) \cite{krahenbuhl2011efficient}.

\textbf{Datasets.} 
We train our method on three glass segmentation datasets: GDD \cite{mei2020don}, Trans10K-Stuff \cite{xie2020segmenting} and HSO \cite{yu2022progressive}.
GDD consists of 2,980 training images and 936 testing images. 
Trans10K is a large-scale transparent object segmentation dataset, containing two categories of objects, i.e., stuff and things. As the glass segmentation task we strive to address is exclusively about grouping rather than identifying object categories, we only use the `stuff' images and the corresponding annotations in Trans10K. In our experiment, 2,455 glass image/mask pairs in Trans10K-Stuff are used for training and 1,771 images in both validation and testing sets are used for testing.
HSO is a glass segmentation dataset \cite{yu2022progressive} to diversify the patterns of glass, especially the glass in home scenes, for advancing household robot applications and in-depth research on this topic, which consists of 3,070 training images and 1,782 test images.

\textbf{Evaluation Metrics.} 
We adopt three metrics widely used in computer vision tasks to evaluate  our model and state-of-the-art methods. 
In addition to the popular metric of the intersection of union (IoU), we also apply 
mean absolute error (MAE) metric which is used in \cite{chen2018reverse, li2018contour, liu2018picanet, hou2017deeply}. 
The intersection over union (IoU) is widely used in the segmentation field, which is defined as:
\begin{equation}
      IoU = \frac{TP}{TP+FP+FN},
\end{equation}
where $TP$, $FP$, and $FN$ are the numbers of true positives, false positives, and false negatives pixels, respectively.


MAE is the mean absolute error, i.e., the mean value of the absolute error between the prediction and the ground truth, which is defined as:
\begin{equation}
MAE = \frac{1}{H\times W} \sum\limits_{(x,y)}^{(H,W)}{|p(x,y)-g(x,y) \vert},
\end{equation}
where $g(x,y)\in [0,1]$ is the ground-truth label of the pixel $(x, y)$ and $p(x, y) \in [0, 1]$ is the predicted probability of being glass at the pixel $(x, y)$. 

In addition, we select the balanced error rate (BER) \cite{vicente2015leave} from the shadow detection task as another metric:
\begin{equation}
BER = 100\times(1-\frac{1}{2}(\frac{TP}{N_p}+\frac{TN}{N_n})),
\end{equation}
where $TP$, $TN$, $N_n$, and $N_p$ are the numbers of true positives, true negatives, glass, and non-glass pixels, respectively.

Note that for $IoU$ and $F_\beta$, it is ``higher is better'', while for $MAE$ and $BER$, it is ``lower the better''.

\textbf{Compared Methods.}
Glass segmentation is a relatively new research topic. Currently, there are only a few deep learning-based methods for glass segmentation. 
We verify the effectiveness of our method by comparing it with thirty methods selected from other related fields and four glass segmentation methods. 
Specifically, we choose semantic segmentation methods ICNet \cite{zhao2018icnet}, PSPNet \cite{zhao2017pyramid}, DeepLabv3+ \cite{chen2017deeplab}, DenseASPP \cite{yang2018denseaspp}, BiSeNet \cite{yu2018bisenet}, EGNet \cite{zhao2019egnet}, DANet \cite{fu2019dual}, CCNet \cite{huang2019ccnet}, GFFNet \cite{li2020gated}, SFNet \cite{li2020semantic} and FaPN \cite{huang2021fapn}; salient object detection methods DSS \cite{hou2017deeply}, PiCANet \cite{liu2018picanet}, RAS \cite{chen2018reverse}, CPD \cite{wu2019cascaded}, EGNet\cite{zhao2019egnet}, F$^3$Net \cite{wei2020f3net}, MINet-R \cite{pang2020multi}, ITSD \cite{zhou2020interactive} and LDF \cite{wei2020label}; shadow detection methods DSC \cite{hu2018direction}, DSD \cite{zheng2019distraction} and BDRAR \cite{zhu2018bidirectional}; medical image segmentation method PraNet \cite{fan2020pranet}; camouflaged object detection methods SINet \cite{fan2020camouflaged}, PFNet \cite{mei2021camouflaged} and RankNet \cite{lv2021simultaneously}; mirror segmentation methods MirrorNet \cite{yang2019my} and PMD \cite{lin2020progressive}; transparent object segmentation methods TransLab \cite{xie2020segmenting} and Trans2Seg \cite{xie2021segmenting}; and glass segmentation methods GDNet \cite{mei2020don}, GSD \cite{lin2021rich}, EBLNet \cite{he2021enhanced} and PGSNet \cite{yu2022progressive}.
For fair comparisons, we retrain their released codes on the benchmark datasets.

\subsection{Comparison Results}

\Cref{tab:example} reports the quantitative results of our method against the 34 state-of-the-art methods on three benchmark datasets. We can see that our method outperforms all the other methods on all three challenging datasets under all three standard metrics. Notably, compared with the state-of-the-art glass segmentation method PGSNet \cite{yu2022progressive}, our method improves IoU by 5.27\% and reduces BER by 0.60\% on the HSO dataset. 

\begin{table*}[htbp]
\centering
\begin{tabular}{c|c|ccc|ccc|ccc}
\hline
\multirow{3}{*}{Methods} & \multirow{3}{*}{Pub.'Year} & \multicolumn{3}{c|}{GDD}                                                       & \multicolumn{3}{c|}{Trans 10K-stuff}                                           & \multicolumn{3}{c}{HSO}                                                       \\ \cline{3-11} 
                         &                            & \multicolumn{3}{c|}{Trainset:2980 Testset:936}                                 & \multicolumn{3}{c|}{Trainset:2455 Testset:1771}                                & \multicolumn{3}{c}{Trainset:3070 Testset:1782}                                \\ \cline{3-11} 
                         &                            & \multicolumn{1}{c|}{IoU $\uparrow$}        & \multicolumn{1}{c|}{MAE$\downarrow$}   & BER$\downarrow$   & \multicolumn{1}{c|}{IoU$\uparrow$}        & \multicolumn{1}{c|}{MAE$\downarrow$}   & BER $\downarrow$  & \multicolumn{1}{c|}{IoU$\uparrow$}        & \multicolumn{1}{c|}{MAE$\downarrow$}   & BER $\downarrow$  \\ \hline
      ICNet$\bullet$                    & ECCV'18                    & \multicolumn{1}{c|}{69.59}  & \multicolumn{1}{c|}{0.164} & 16.10 & \multicolumn{1}{c|}{74.94}  & \multicolumn{1}{c|}{0.110} & 10.92 & \multicolumn{1}{c|}{62.15}  & \multicolumn{1}{c|}{0.165} & 17.07 \\
      PSPNet$\bullet$                   & CVPR'17                    & \multicolumn{1}{c|}{84.06} & \multicolumn{1}{c|}{0.084} & 8.79  & \multicolumn{1}{c|}{87.89}  & \multicolumn{1}{c|}{0.045} & 5.46  & \multicolumn{1}{c|}{77.60}  & \multicolumn{1}{c|}{0.095} & 10.57 \\
      DeepLabv3+$\bullet$               & ECCV'18                    & \multicolumn{1}{c|}{69.95} & \multicolumn{1}{c|}{0.147} & 15.49 & \multicolumn{1}{c|}{51.52}  & \multicolumn{1}{c|}{0.229} & 23.80 & \multicolumn{1}{c|}{64.47} & \multicolumn{1}{c|}{0.149} & 16.03 \\
      DenseASPP$\bullet$                & CVPR'18                    & \multicolumn{1}{c|}{83.68}  & \multicolumn{1}{c|}{0.081} & 8.66  & \multicolumn{1}{c|}{86.34}  & \multicolumn{1}{c|}{0.051} & 6.12  & \multicolumn{1}{c|}{75.94}  & \multicolumn{1}{c|}{0.096} & 11.34 \\
      BiSeNet$\bullet$                  & ECCV'18                    & \multicolumn{1}{c|}{80.00}  & \multicolumn{1}{c|}{0.106} & 11.04 & \multicolumn{1}{c|}{85.82}  & \multicolumn{1}{c|}{0.056} & 6.11  & \multicolumn{1}{c|}{75.85}  & \multicolumn{1}{c|}{0.101} & 11.04 \\
      DANet$\bullet$                    & CVPR'19                    & \multicolumn{1}{c|}{84.15}  & \multicolumn{1}{c|}{0.089} & 8.96  & \multicolumn{1}{c|}{88.18}  & \multicolumn{1}{c|}{0.045} & 5.28  & \multicolumn{1}{c|}{77.69}  & \multicolumn{1}{c|}{0.091} & 10.60 \\
      CCNet$\bullet$                    & ICCV'19                    & \multicolumn{1}{c|}{84.29}  & \multicolumn{1}{c|}{0.085} & 8.63  & \multicolumn{1}{c|}{88.20}  & \multicolumn{1}{c|}{0.044} & 5.15  & \multicolumn{1}{c|}{78.17}  & \multicolumn{1}{c|}{0.092} & 10.34 \\
      GFFNet$\bullet$                   & AAAI'20                    & \multicolumn{1}{c|}{82.41} & \multicolumn{1}{c|}{0.855}  & 9.11  & \multicolumn{1}{c|}{69.29}  & \multicolumn{1}{c|}{0.143} & 14.19 & \multicolumn{1}{c|}{77.34}  & \multicolumn{1}{c|}{0.094} & 9.69  \\
      SFNet$\bullet$                    & ECCV'20                    & \multicolumn{1}{c|}{80.96} & \multicolumn{1}{c|}{0.102} & 10.23 & \multicolumn{1}{c|}{71.27}  & \multicolumn{1}{c|}{0.133} & 13.14 & \multicolumn{1}{c|}{77.48}  & \multicolumn{1}{c|}{0.091} & 10.79 \\
      FaPN$\bullet$                     & ICCV'21                    & \multicolumn{1}{c|}{86.65}  & \multicolumn{1}{c|}{0.062} & 5.69  & \multicolumn{1}{c|}{89.09}  & \multicolumn{1}{c|}{0.042} & 4.80  & \multicolumn{1}{c|}{78.05}  & \multicolumn{1}{c|}{0.089} & 9.51  \\ \hline
      DSS$\circ$                      & TPAMI'19                   & \multicolumn{1}{c|}{80.24}  & \multicolumn{1}{c|}{0.123} & 9.73  & \multicolumn{1}{c|}{84.77}  & \multicolumn{1}{c|}{0.075} & 6.42  & \multicolumn{1}{c|}{73.08}  & \multicolumn{1}{c|}{0.135} & 12.04 \\
      PiCANet$\circ$                  & CVPR'18                    & \multicolumn{1}{c|}{83.74}  & \multicolumn{1}{c|}{0.093} & 8.24  & \multicolumn{1}{c|}{83.99}  & \multicolumn{1}{c|}{0.077} & 7.03  & \multicolumn{1}{c|}{71.66} & \multicolumn{1}{c|}{0.148} & 13.31 \\
      RAS$\circ$                      & ECCV'18                    & \multicolumn{1}{c|}{80.96}  & \multicolumn{1}{c|}{0.106} & 9.48  & \multicolumn{1}{c|}{85.40}  & \multicolumn{1}{c|}{0.062} & 6.20  & \multicolumn{1}{c|}{74.63}  & \multicolumn{1}{c|}{0.116} & 11.24 \\
      CPD$\circ$                      & CVPR'19                    & \multicolumn{1}{c|}{82.52}  & \multicolumn{1}{c|}{0.095} & 8.87  & \multicolumn{1}{c|}{86.08}  & \multicolumn{1}{c|}{0.064} & 5.89  & \multicolumn{1}{c|}{76.16}  & \multicolumn{1}{c|}{0.111} & 10.58 \\
      EGNet$\circ$                    & ICCV'19                    & \multicolumn{1}{c|}{85.05} & \multicolumn{1}{c|}{0.083} & 7.43  & \multicolumn{1}{c|}{84.57}  & \multicolumn{1}{c|}{0.068} & 6.59  & \multicolumn{1}{c|}{74.29}  & \multicolumn{1}{c|}{0.119} & 11.58 \\
      F$^3$Net$\circ$                    & AAAI'20                    & \multicolumn{1}{c|}{84.79}  & \multicolumn{1}{c|}{0.082} & 7.38  & \multicolumn{1}{c|}{86.23}  & \multicolumn{1}{c|}{0.061} & 5.81  & \multicolumn{1}{c|}{76.84}  & \multicolumn{1}{c|}{0.105} & 10.58 \\
      MINet-R$\circ$                  & CVPR'20                    & \multicolumn{1}{c|}{82.03} & \multicolumn{1}{c|}{0.092} & 8.55  & \multicolumn{1}{c|}{85.88} & \multicolumn{1}{c|}{0.060} & 6.03  & \multicolumn{1}{c|}{76.61}  & \multicolumn{1}{c|}{0.104} & 10.33 \\
      ITSD$\circ$                     & CVPR'20                    & \multicolumn{1}{c|}{83.72}  & \multicolumn{1}{c|}{0.087} & 7.77  & \multicolumn{1}{c|}{85.44}  & \multicolumn{1}{c|}{0.063} & 6.26  & \multicolumn{1}{c|}{74.33} & \multicolumn{1}{c|}{0.123} & 11.39 \\
      LDF$\circ$                      & CVPR'20                   & \multicolumn{1}{c|}{83.40}    & \multicolumn{1}{c|}{0.084}      &     7.97  & \multicolumn{1}{c|}{84.54}            & \multicolumn{1}{c|}{0.065}      & 6.65      & \multicolumn{1}{c|}{76.98}     & \multicolumn{1}{c|}{0.102}      & 10.60      \\ \hline
      DSC$\triangle$                      & CVPR'19                    & \multicolumn{1}{c|}{83.56}  & \multicolumn{1}{c|}{0.090} & 7.97  & \multicolumn{1}{c|}{86.37}  & \multicolumn{1}{c|}{0.058} & 5.31  & \multicolumn{1}{c|}{71.93}  & \multicolumn{1}{c|}{0.128} & 13.11 \\
      DSD$\triangle$                      & CVPR'19                   & \multicolumn{1}{c|}{85.53}   & \multicolumn{1}{c|}{0.071}      &     7.17  & \multicolumn{1}{c|}{86.22}          & \multicolumn{1}{c|}{0.052}      &   5.57    & \multicolumn{1}{c|}{76.49}    & \multicolumn{1}{c|}{0.101}      &    11.07   \\
      BDRAR$\triangle$                    & ECCV'18                    & \multicolumn{1}{c|}{80.01}  & \multicolumn{1}{c|}{0.098} & 9.87  & \multicolumn{1}{c|}{85.00}  & \multicolumn{1}{c|}{0.061} & 6.04  & \multicolumn{1}{c|}{75.32}  & \multicolumn{1}{c|}{0.101} & 11.13 \\ \hline
      PraNet$\S$                   & MICCAI'20                  & \multicolumn{1}{c|}{82.06}  & \multicolumn{1}{c|}{0.098} & 9.33  & \multicolumn{1}{c|}{87.15}  & \multicolumn{1}{c|}{0.058} & 5.31  & \multicolumn{1}{c|}{71.93}  & \multicolumn{1}{c|}{0.128} & 13.11 \\ \hline
      SINet$\ast$                    & CVPR'20               & \multicolumn{1}{c|}{79.28}       & \multicolumn{1}{c|}{0.133} & 10.14 & \multicolumn{1}{c|}{85.01}           & \multicolumn{1}{c|}{0.064}      & 6.37  & \multicolumn{1}{c|}{76.07}          & \multicolumn{1}{c|}{0.113}      &    10.92   \\
      PFNet$\ast$                     & CVPR'21               & \multicolumn{1}{c|}{85.96}      & \multicolumn{1}{c|}{0.072}  & 6.50    & \multicolumn{1}{c|}{85.68}           & \multicolumn{1}{c|}{0.059}      &  5.97  & \multicolumn{1}{c|}{76.79}          & \multicolumn{1}{c|}{0.109}      &     10.38  \\
      RankNet$\ast$                   & CVPR'21               & \multicolumn{1}{c|}{84.54}          & \multicolumn{1}{c|}{0.085}   &  7.70     & \multicolumn{1}{c|}{86.55}            & \multicolumn{1}{c|}{0.055}      & 5.77   & \multicolumn{1}{c|}{76.88}            & \multicolumn{1}{c|}{0.105}      &   10.59    \\ \hline
      MirrorNet$\blacktriangle $                & ICCV'19                    & \multicolumn{1}{c|}{85.07}  & \multicolumn{1}{c|}{0.083} & 7.67  & \multicolumn{1}{c|}{88.30}  & \multicolumn{1}{c|}{0.047} & 4.95  & \multicolumn{1}{c|}{78.82}  & \multicolumn{1}{c|}{0.102} & 9.93  \\
      PMD$\blacktriangle $                      & CVPR'20                    & \multicolumn{1}{c|}{87.12}        & \multicolumn{1}{c|}{0.066}      &     5.97  & \multicolumn{1}{c|}{88.10}          & \multicolumn{1}{c|}{0.047}      &  5.02     & \multicolumn{1}{c|}{80.31}         & \multicolumn{1}{c|}{0.089}      &   8.81    \\ \hline
      TransLab$\diamond $                 & ECCV'20                    & \multicolumn{1}{c|}{81.64}  & \multicolumn{1}{c|}{0.097} & 9.70  & \multicolumn{1}{c|}{87.10}  & \multicolumn{1}{c|}{0.051} & 5.44  & \multicolumn{1}{c|}{74.32}  & \multicolumn{1}{c|}{0.123} & 12.00 \\
      Trans2Seg$\diamond $                & IJCAI'21                   & \multicolumn{1}{c|}{84.41} & \multicolumn{1}{c|}{0.078} & 7.36  & \multicolumn{1}{c|}{74.98}  & \multicolumn{1}{c|}{0.124} & 10.73 & \multicolumn{1}{c|}{77.98}  & \multicolumn{1}{c|}{0.095} & 9.65  \\ \hline
      GDNet$\star $                    & CVPR'20                    & \multicolumn{1}{c|}{87.63}  & \multicolumn{1}{c|}{\textcolor{blue}{0.063}} & \textcolor{blue}{5.62}  & \multicolumn{1}{c|}{88.68}  & \multicolumn{1}{c|}{0.046} & 4.72  & \multicolumn{1}{c|}{78.73}  & \multicolumn{1}{c|}{0.097} & 9.32  \\
      GSD$\star $                      & CVPR'21                    & \multicolumn{1}{c|}{87.53}  & \multicolumn{1}{c|}{0.066} & 5.90  & \multicolumn{1}{c|}{89.67}  & \multicolumn{1}{c|}{\textcolor{red}{0.042}} & 4.52  & \multicolumn{1}{c|}{78.86}  & \multicolumn{1}{c|}{0.103} & 9.79  \\
      EBLNet$\star $                   & ICCV'21                    & \multicolumn{1}{c|}{84.85}            & \multicolumn{1}{c|}{0.079}      &   7.60    & \multicolumn{1}{c|}{\textcolor{blue}{89.92}}            & \multicolumn{1}{c|}{0.047}      &    \textcolor{blue}{4.29}   & \multicolumn{1}{c|}{79.21}        & \multicolumn{1}{c|}{0.094}      & 9.48      \\
      PGSNet$\star $                   & IEEE TIP'22                & \multicolumn{1}{c|}{\textcolor{blue}{87.81}}  & \multicolumn{1}{c|}{\textcolor{red}{0.062}} & \textcolor{red}{5.56}  & \multicolumn{1}{c|}{89.79}  & \multicolumn{1}{c|}{\textcolor{red}{0.042}} & 4.39  & \multicolumn{1}{c|}{\textcolor{blue}{80.06}}  & \multicolumn{1}{c|}{\textcolor{blue}{0.089}} & \textcolor{blue}{9.08 } \\ \hline
      GlassSegNet              & Ours                        & \multicolumn{1}{c|}{\textcolor{red}{90.86}}          & \multicolumn{1}{c|}{\textcolor{red}{0.062}}      &  \textcolor{blue}{5.62}     & \multicolumn{1}{c|}{\textcolor{red}{92.67}}            & \multicolumn{1}{c|}{\textcolor{blue}{0.045}}      &  \textcolor{red}{4.21}     & \multicolumn{1}{c|}{\textcolor{red}{85.33}}            & \multicolumn{1}{c|}{\textcolor{red}{0.088}}      &    \textcolor{red}{8.40}   \\ \hline
      \end{tabular}
      \caption{Quantitative comparison to the state-of-the-art methods on the datasets of GDD \cite{mei2020don}, Trans10k-stuff \cite{xie2020segmenting} and HSO \cite{yu2022progressive}. All the methods are re-trained on the corresponding training set. 
      $\bullet$: semantic segmentation methods. 
      $\circ$: salient object detection methods.
      $\triangle$: shadow detection methods.
      $\S$: medical image segmentation method.
      $\ast$: camouflaged object detection
      $\blacktriangle$: mirror segmentation methods. 
      $\diamond$: transparent object segmentation methods. 
      $\star$: glass segmentation methods. 
      The first and second best results are marked in red and blue, respectively. Our method achieves the best performance on all three challenging datasets under three standard metrics.
      }
      \label{tab:example}
\end{table*}

\begin{figure*}[htbp]
      \centering
        \subfloat[Image]{
          \begin{minipage}[t]{0.08\textwidth}
                \centering

                \includegraphics[width=1\linewidth]{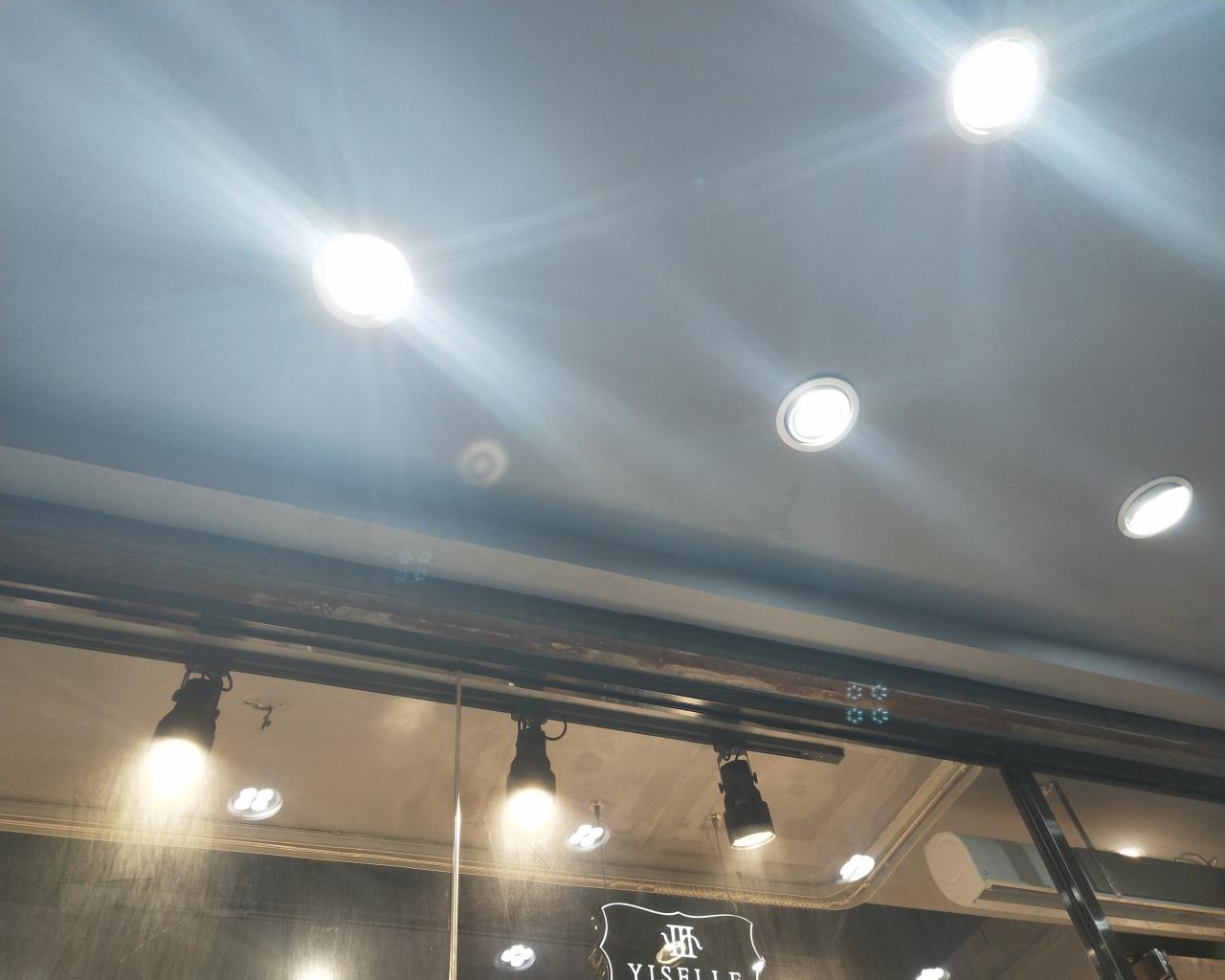}
                
                \includegraphics[width=1\linewidth]{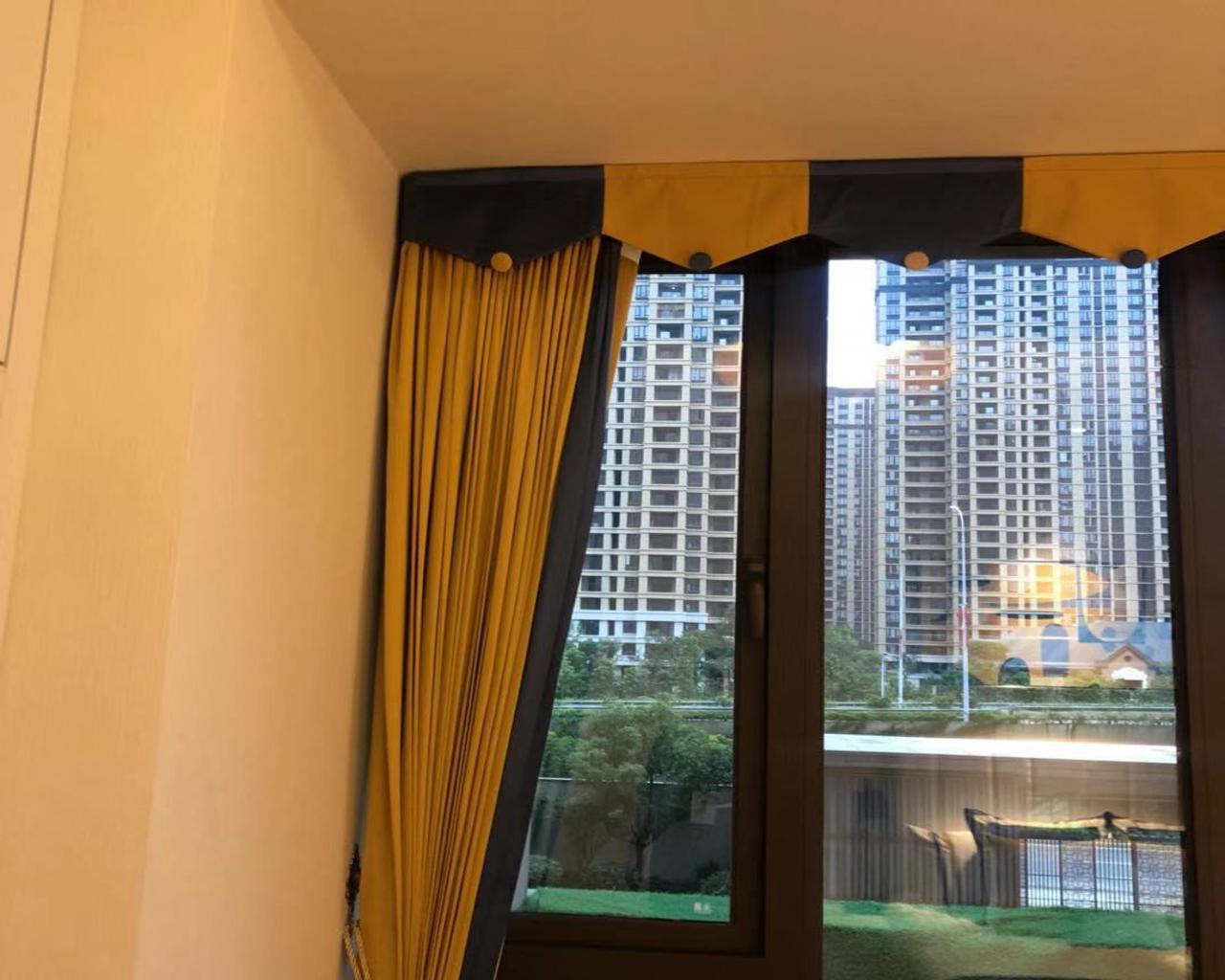}
    
                \includegraphics[width=1\linewidth]{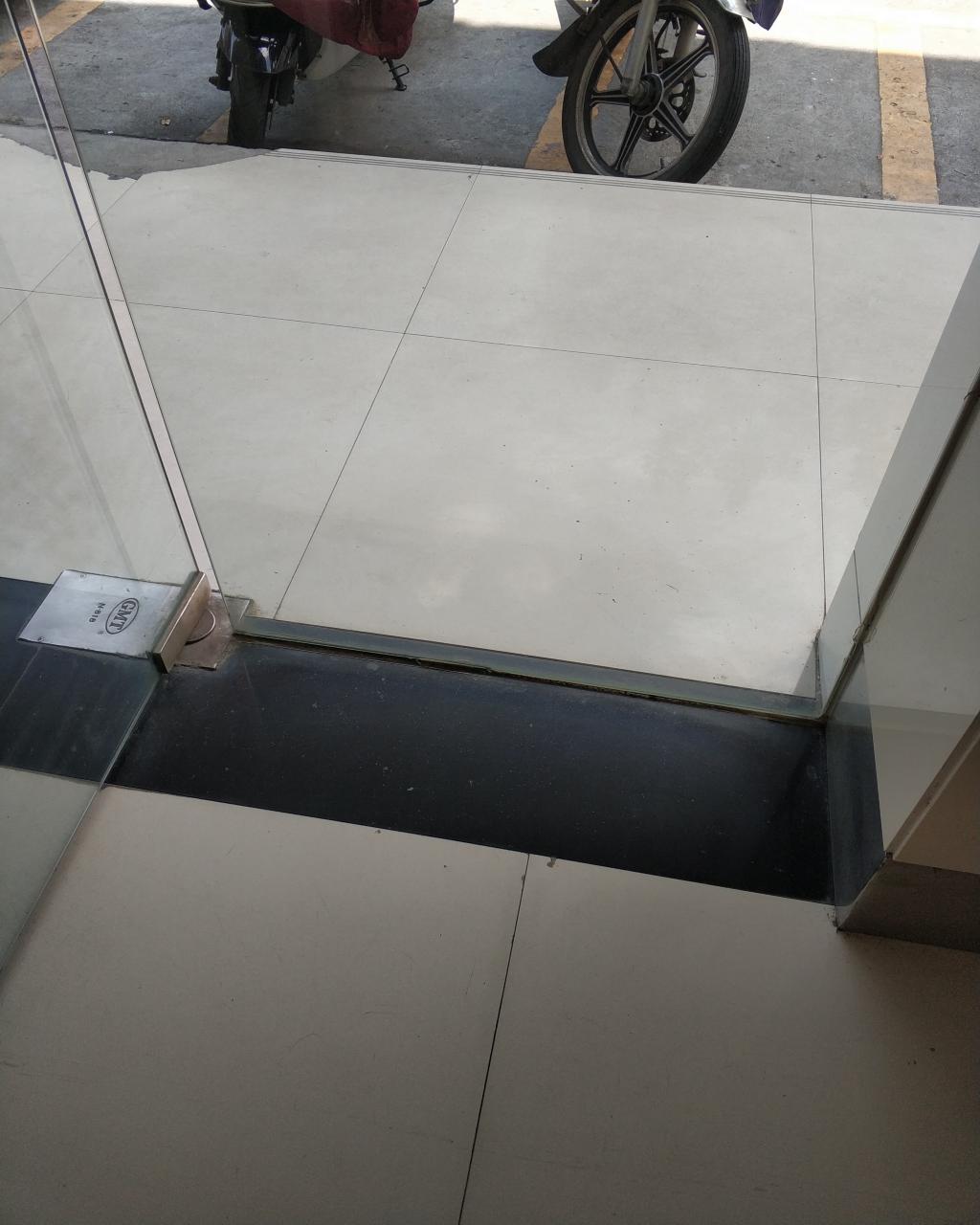}
                
                \includegraphics[width=1\linewidth]{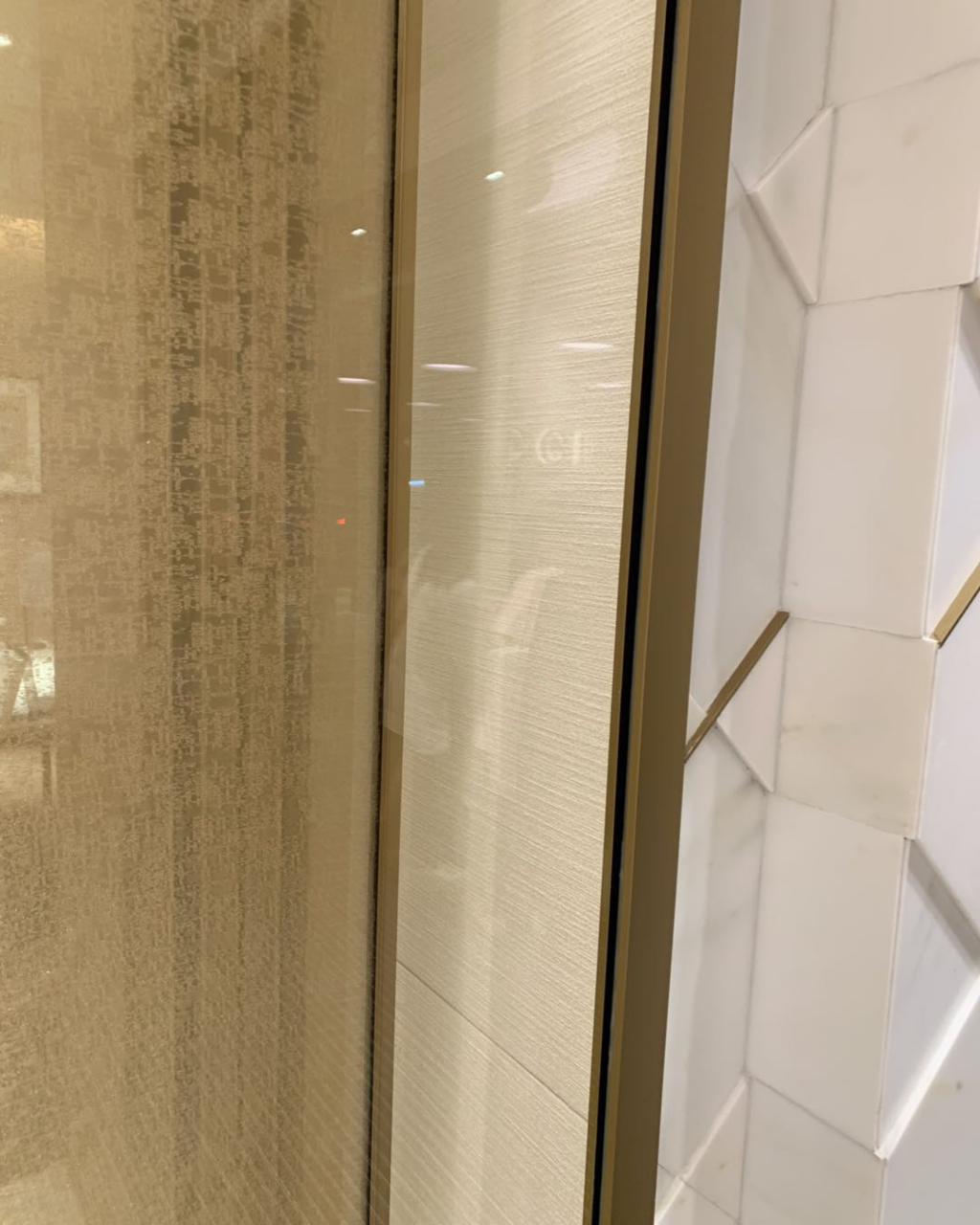}
    
                \includegraphics[width=1\linewidth]{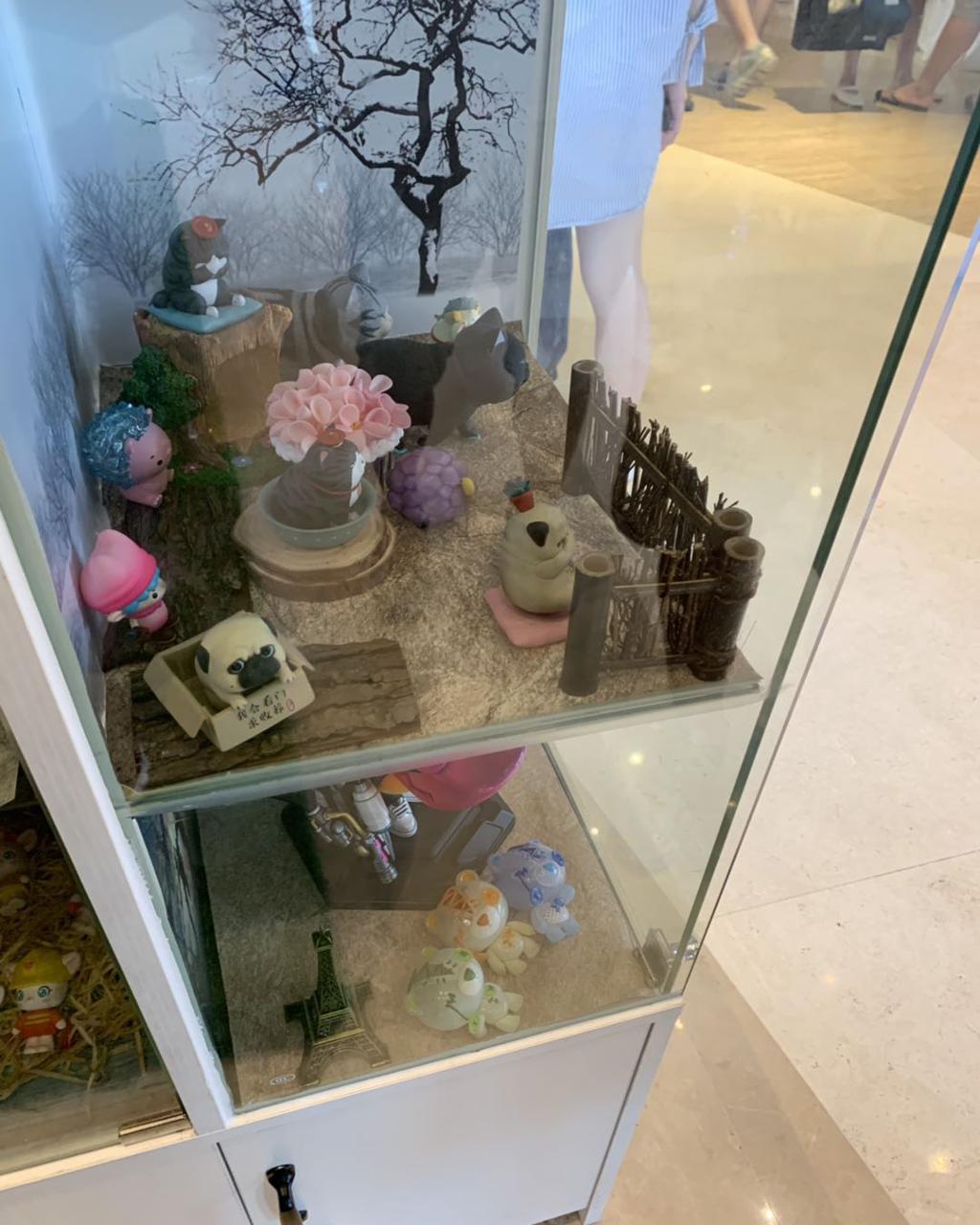}
    
                \includegraphics[width=1\linewidth]{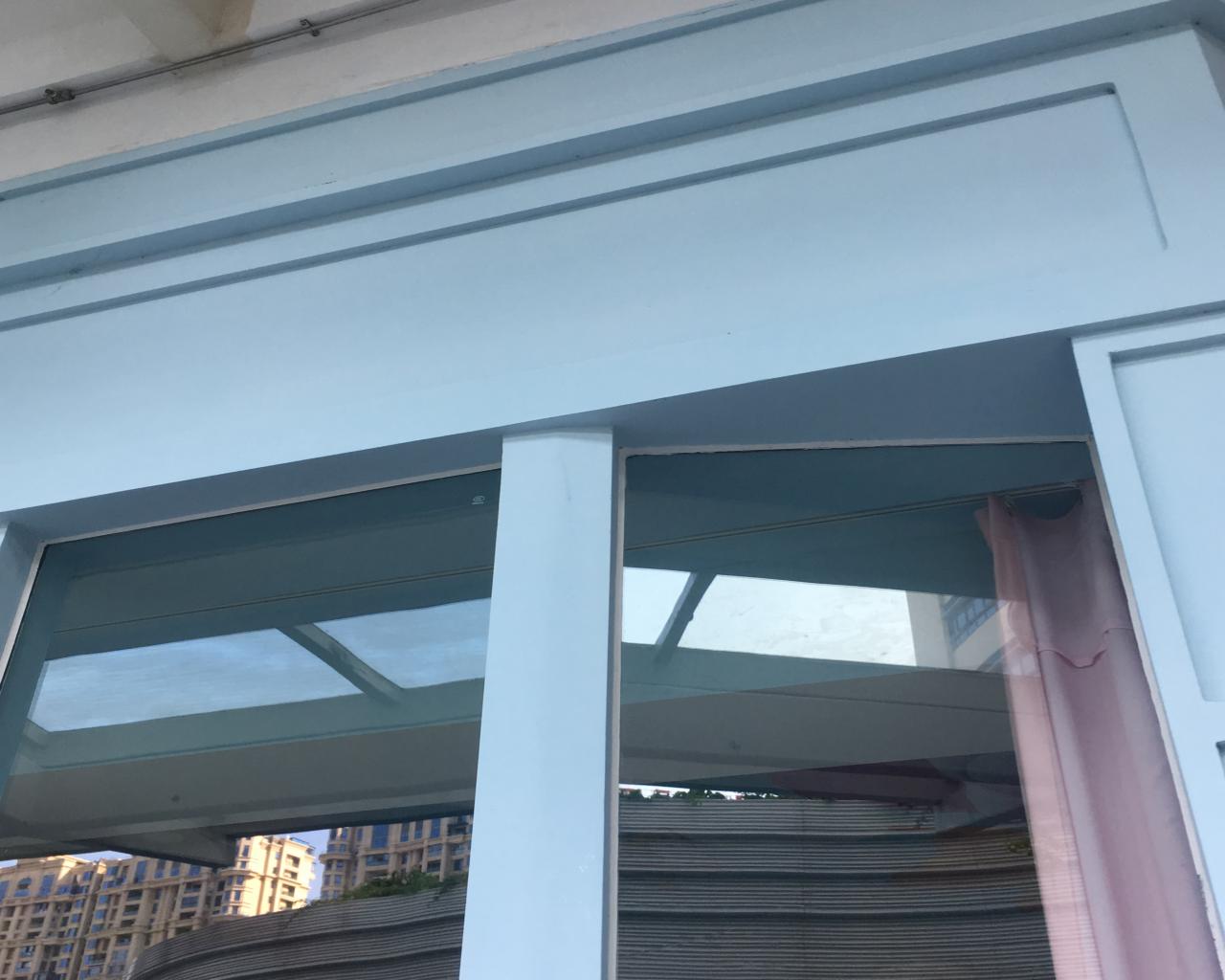}
                
                \includegraphics[width=1\linewidth]{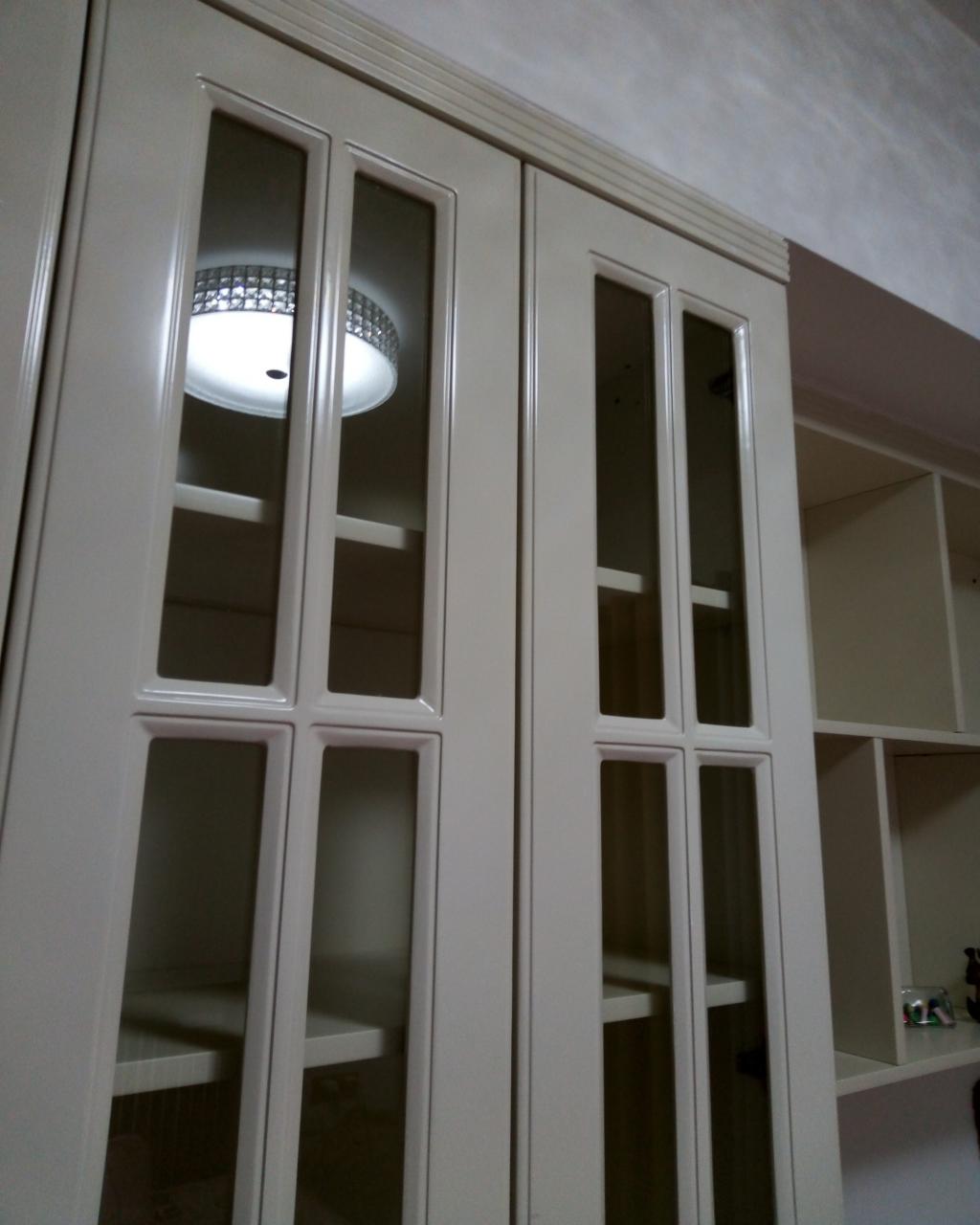}

                \includegraphics[width=1\linewidth]{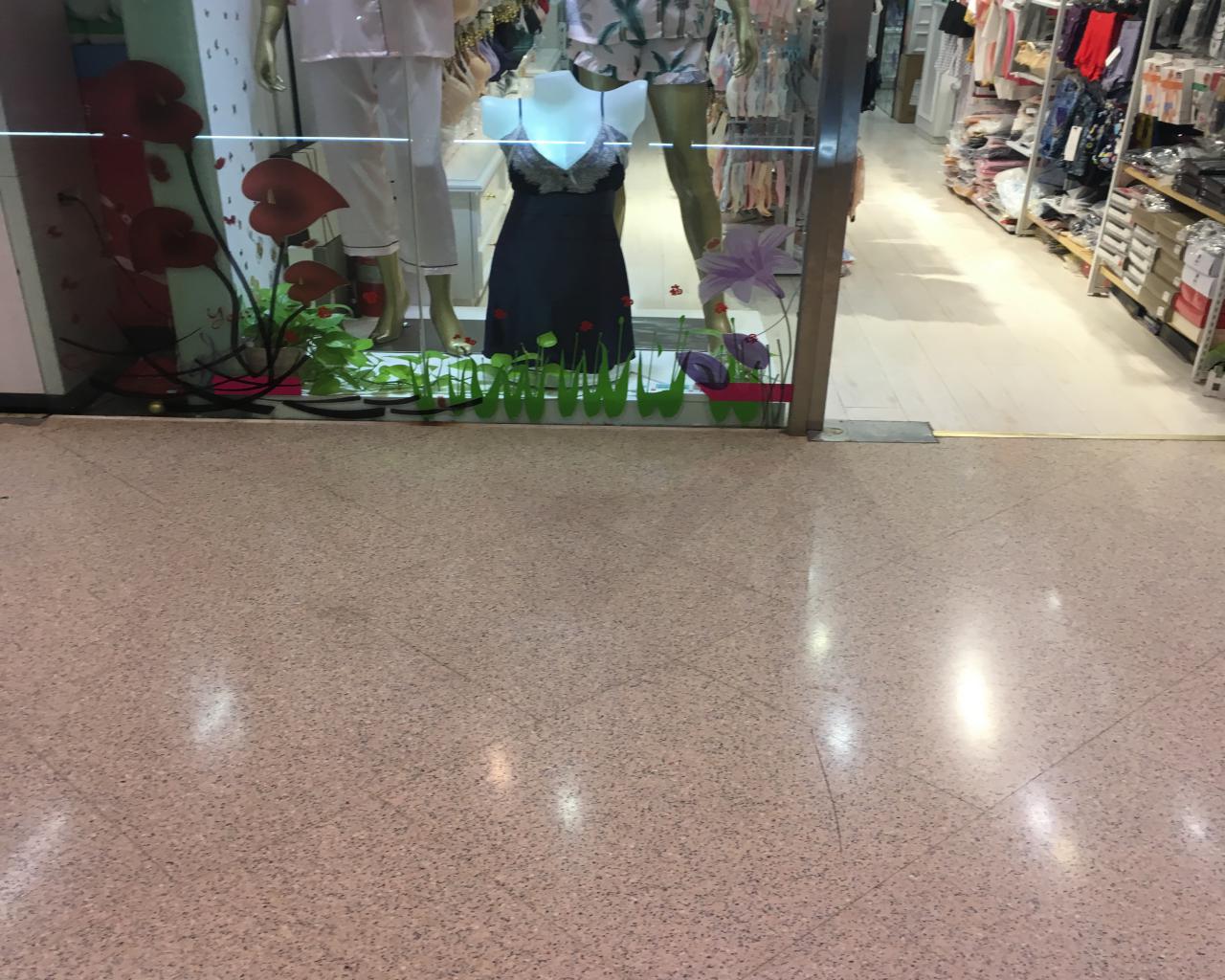}
    
                \includegraphics[width=1\linewidth]{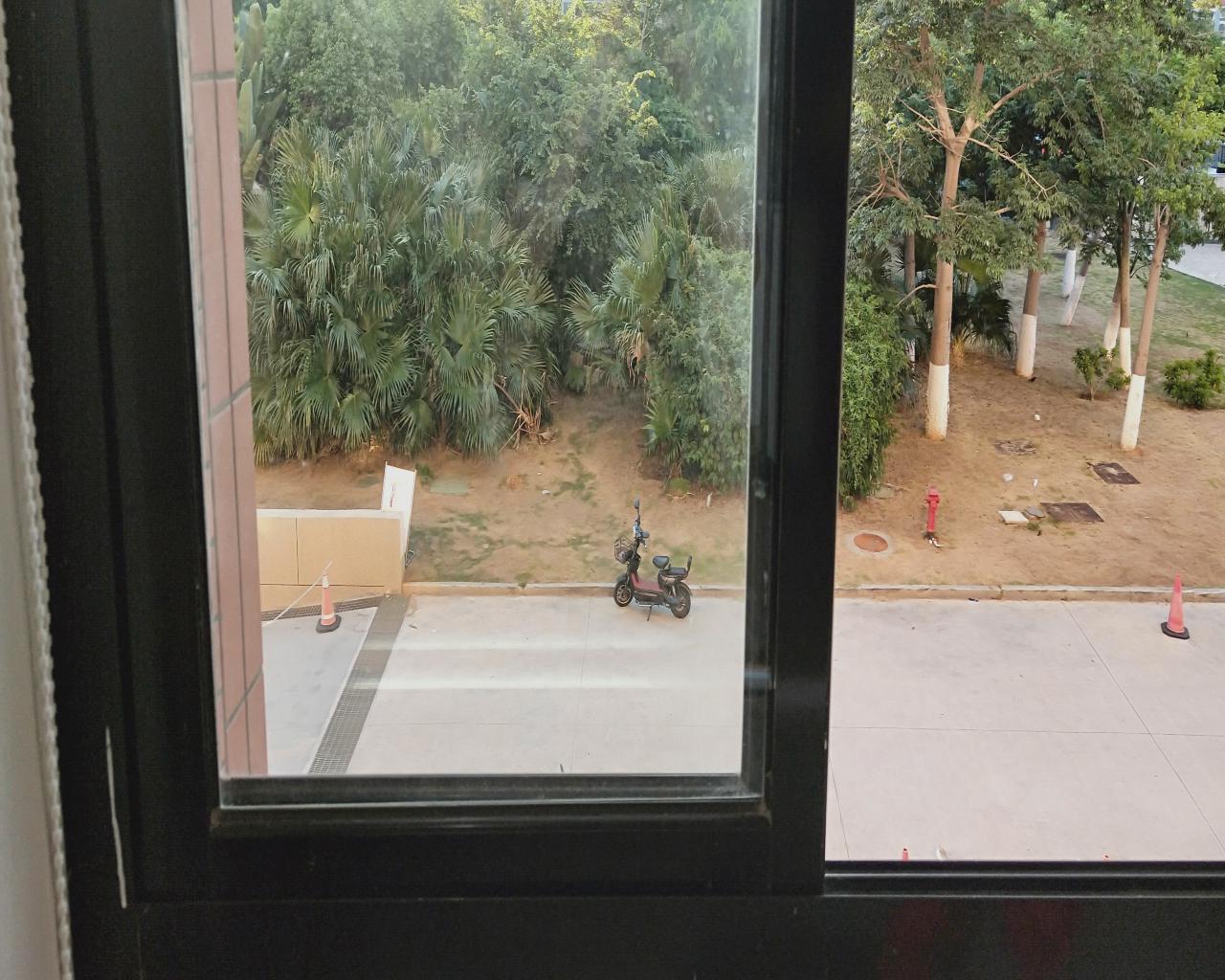}
                
                \includegraphics[width=1\linewidth]{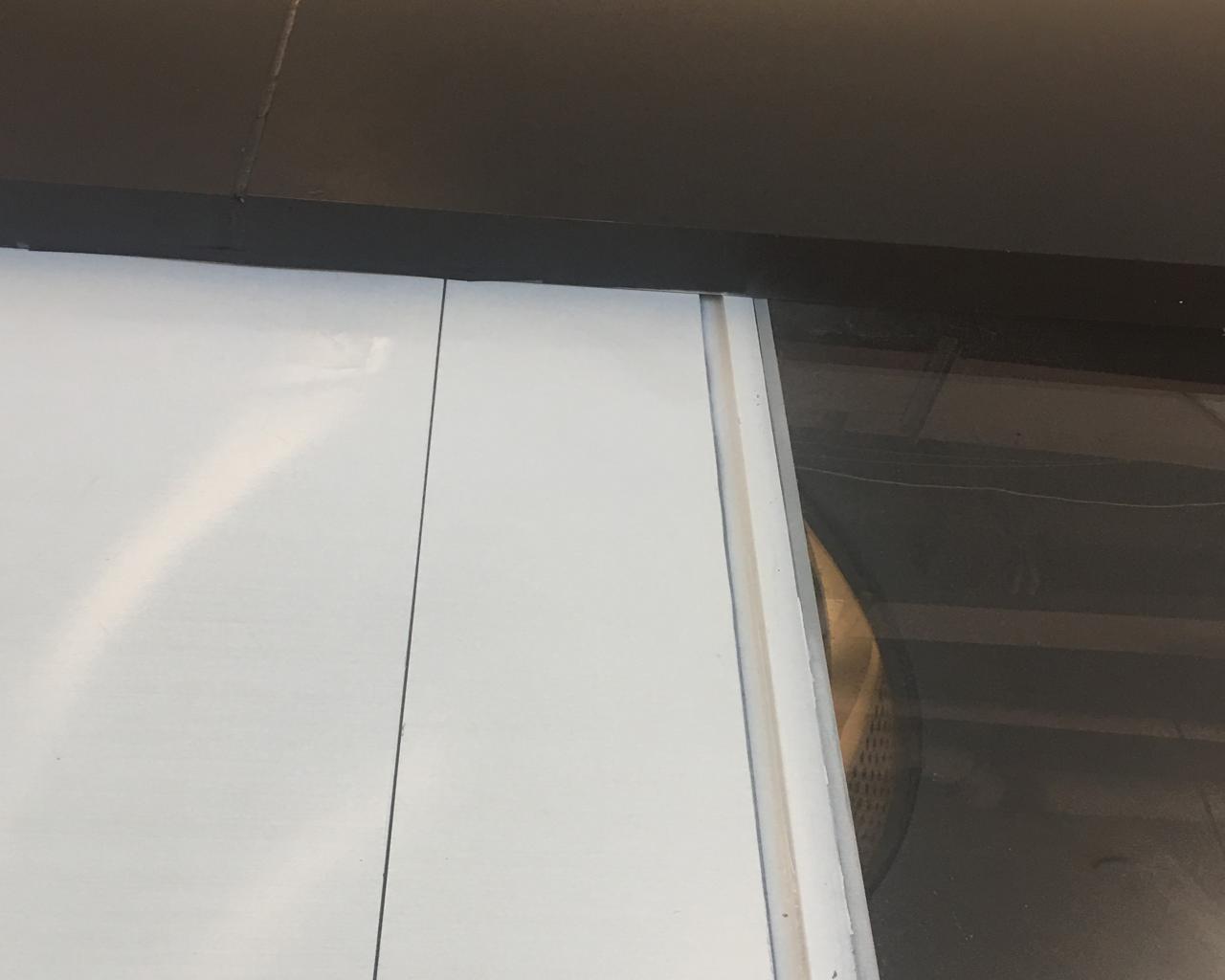}
    
                \includegraphics[width=1\linewidth]{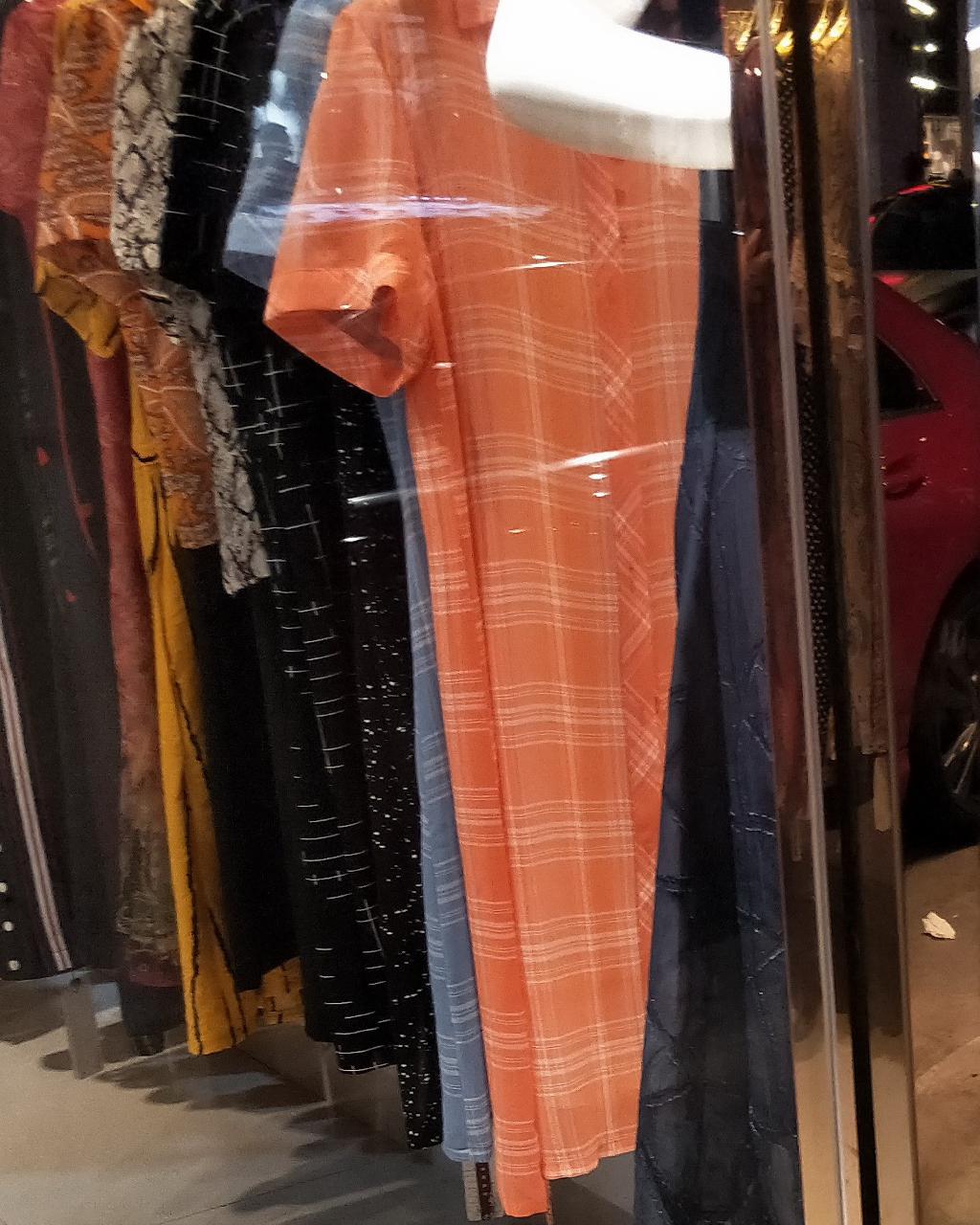}
    
                \includegraphics[width=1\linewidth]{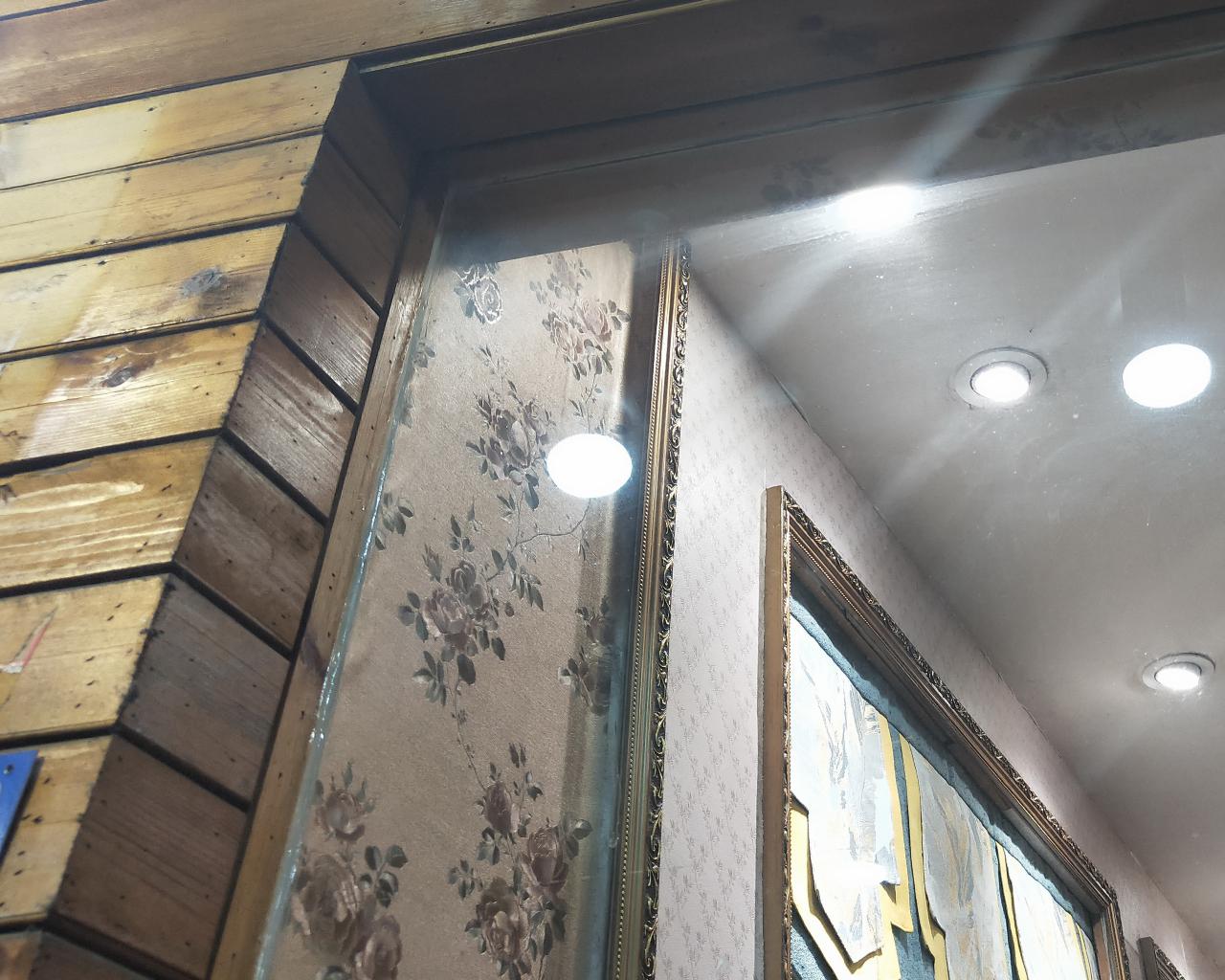}

          \end{minipage}
          }
        \subfloat[EGNet]{
          \begin{minipage}[t]{0.08\textwidth}
                \centering

                \includegraphics[width=1\linewidth]{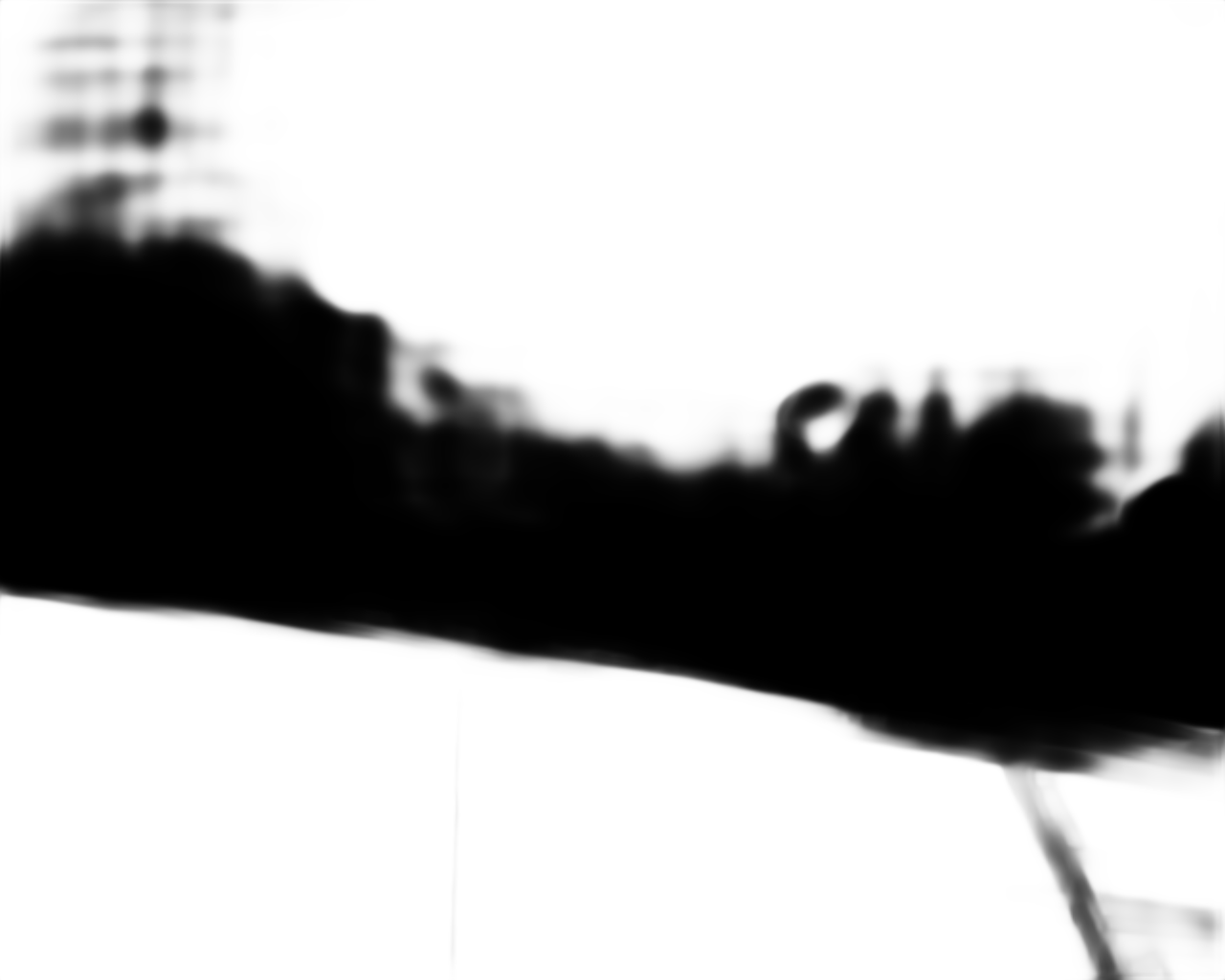}
                
                \includegraphics[width=1\linewidth]{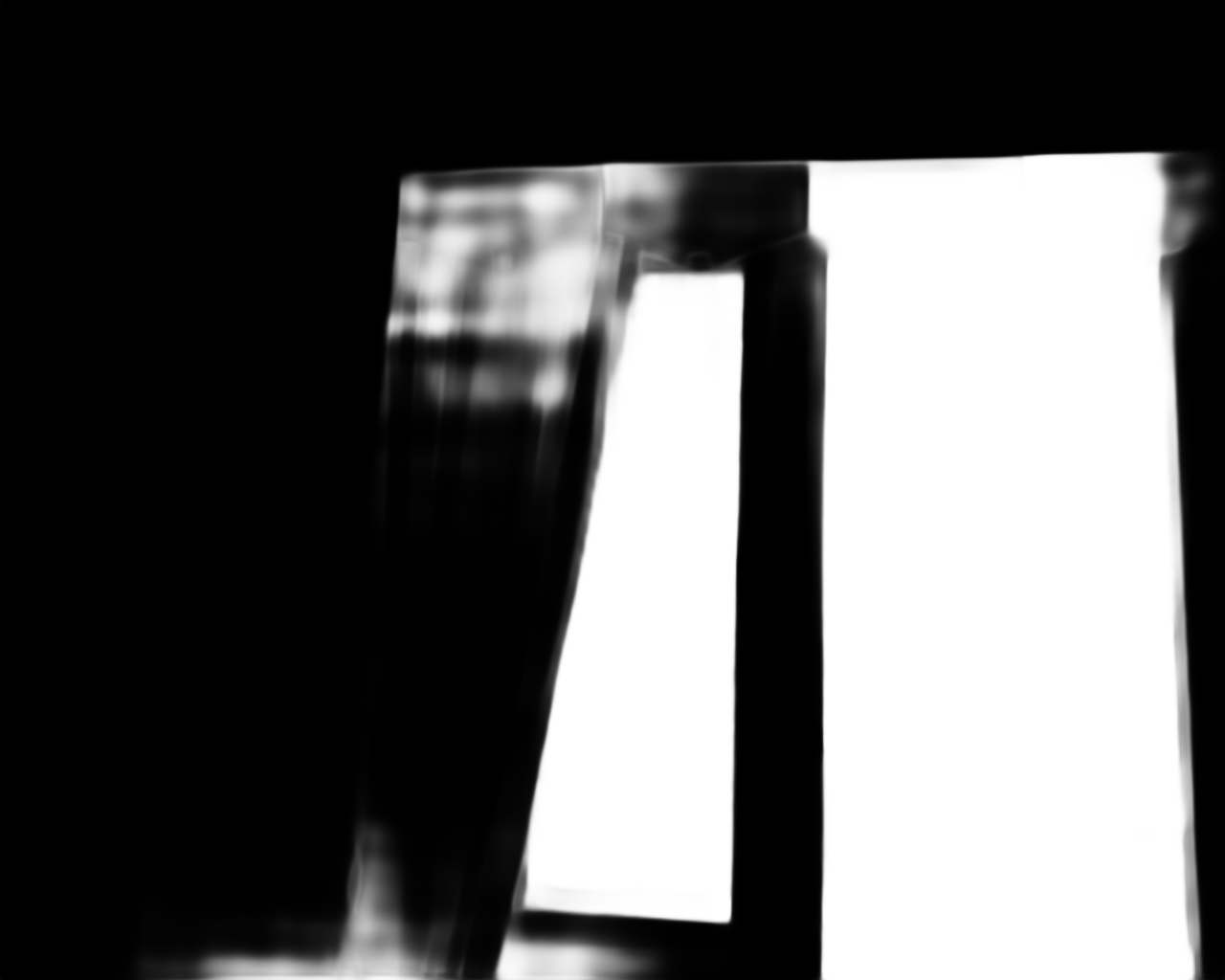}
    
                \includegraphics[width=1\linewidth]{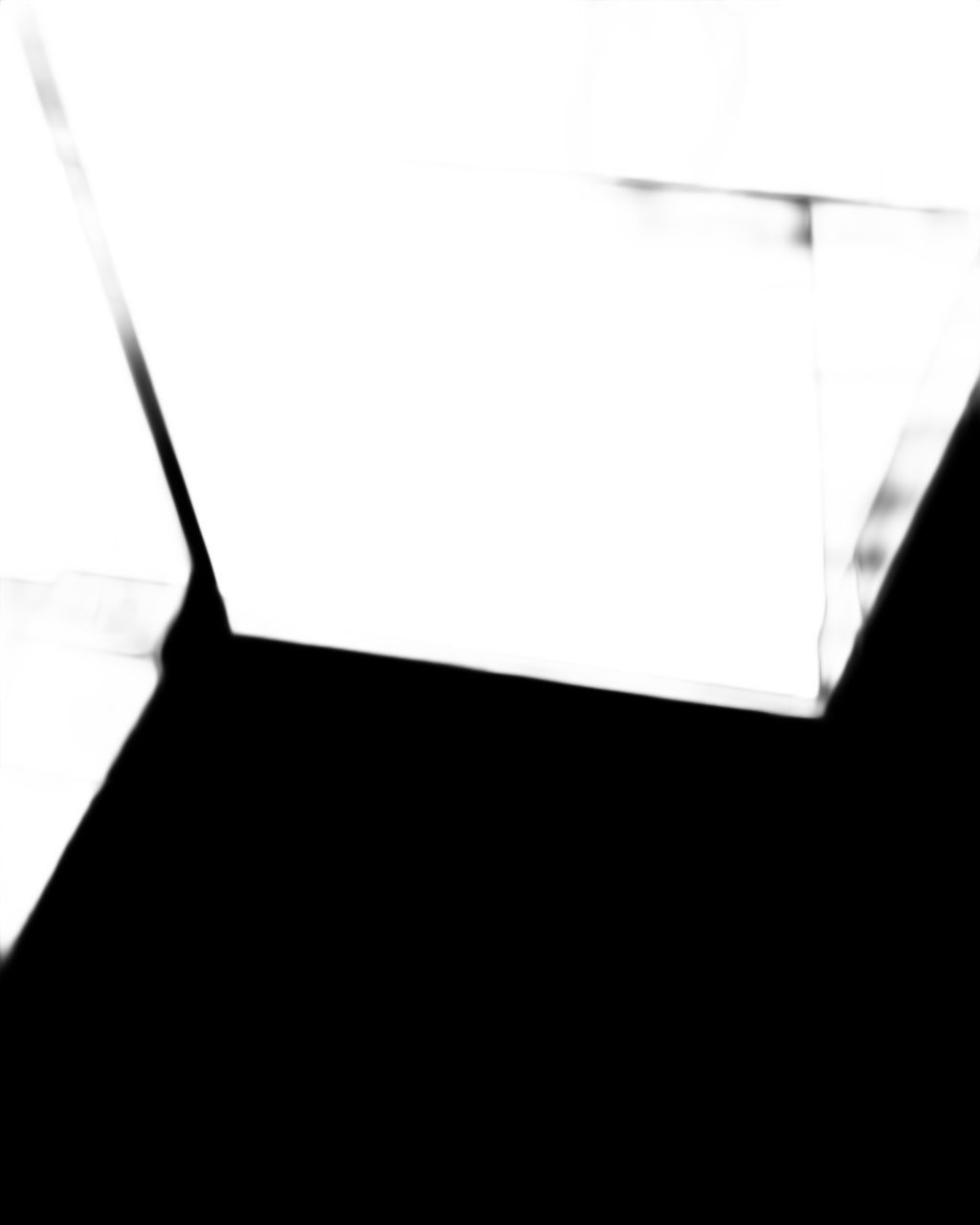}
                
                \includegraphics[width=1\linewidth]{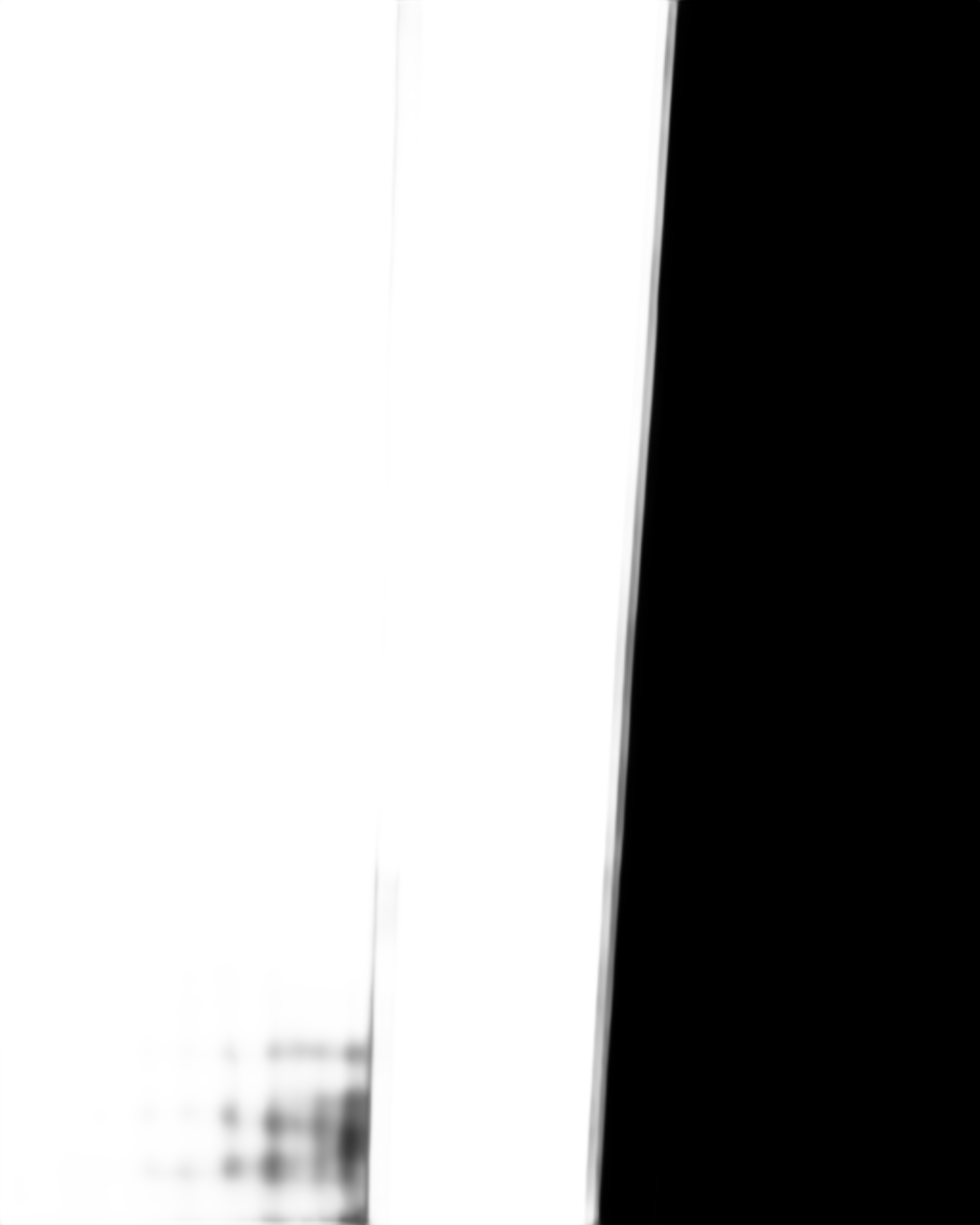}
    
                \includegraphics[width=1\linewidth]{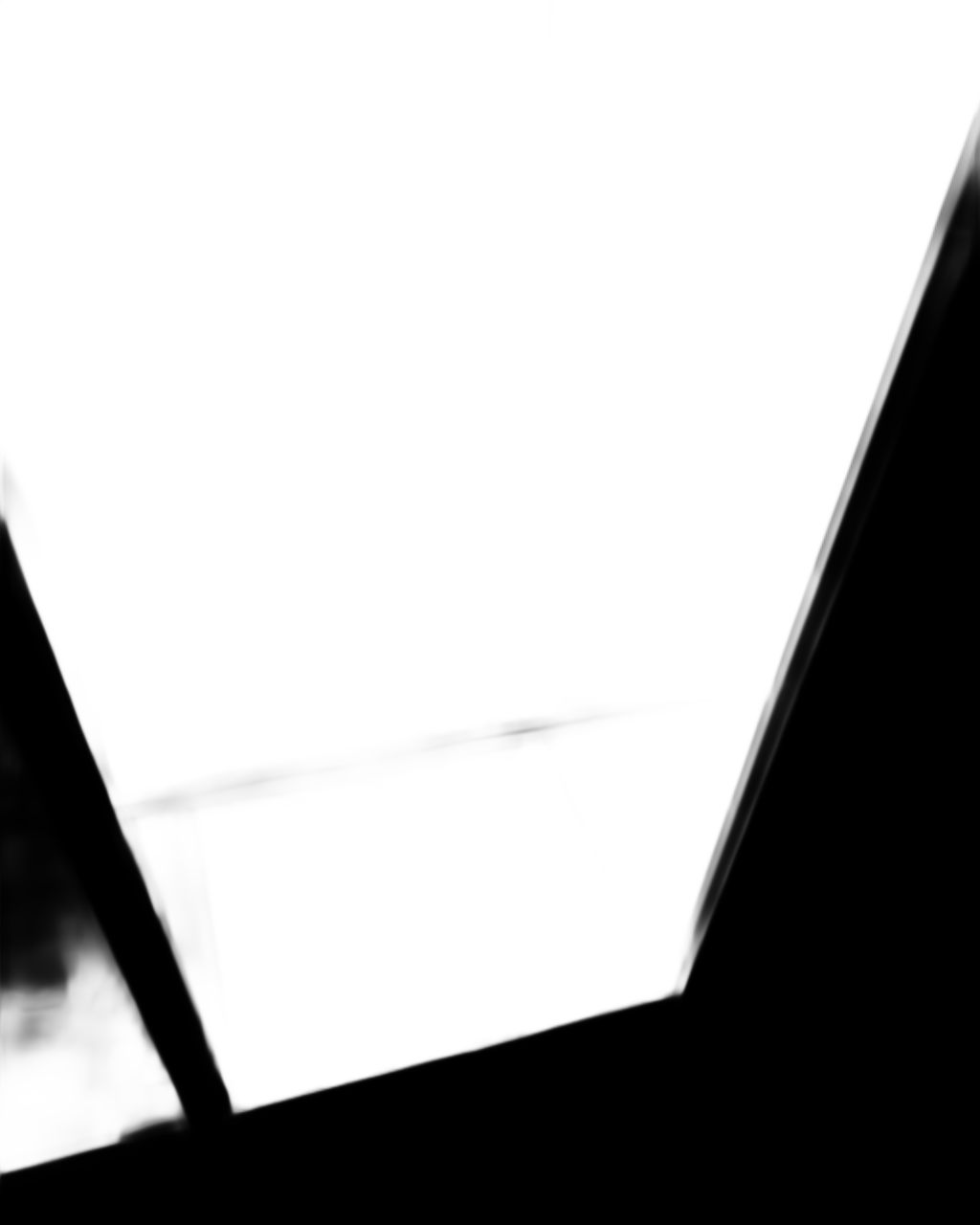}
    
                \includegraphics[width=1\linewidth]{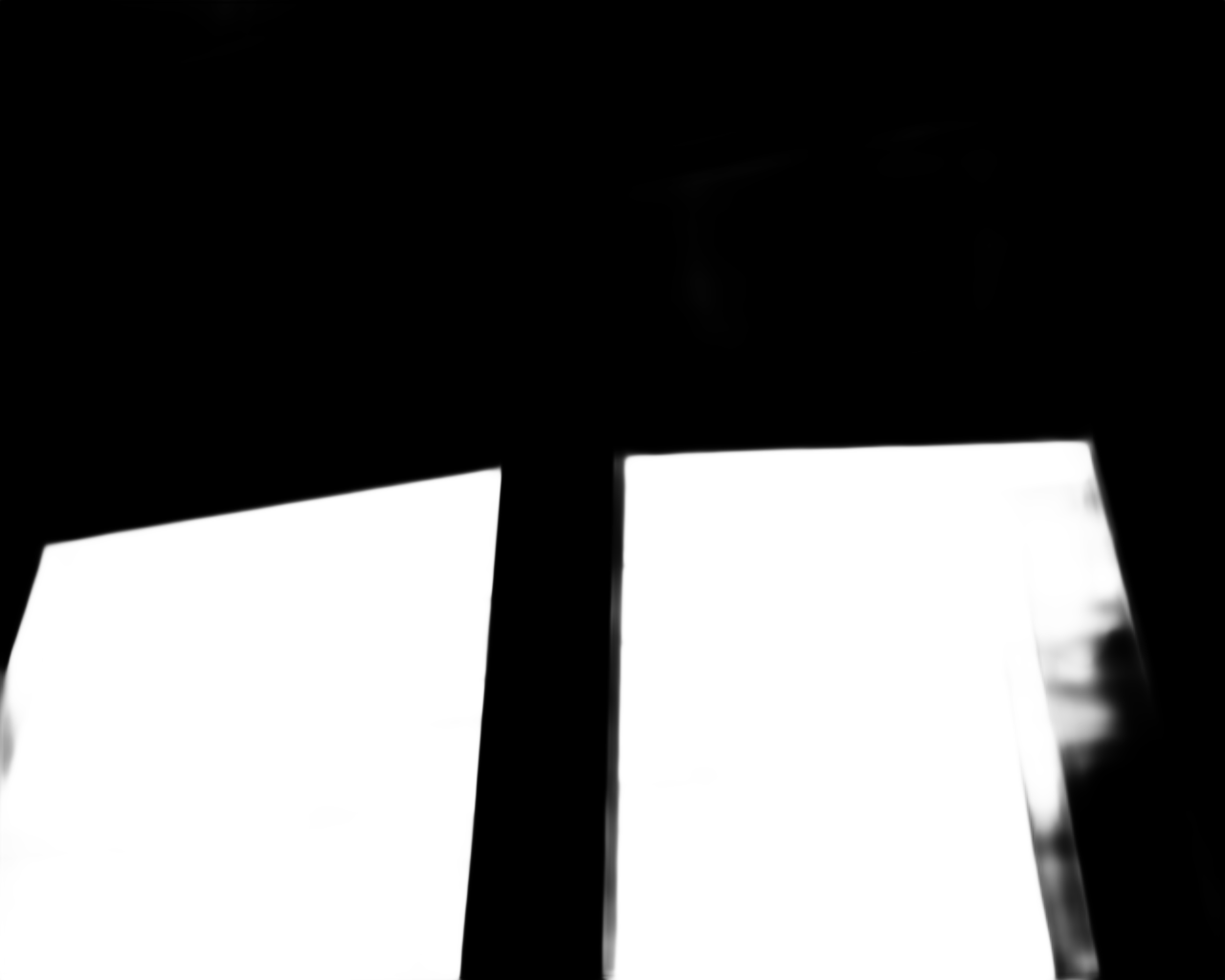}
                
                \includegraphics[width=1\linewidth]{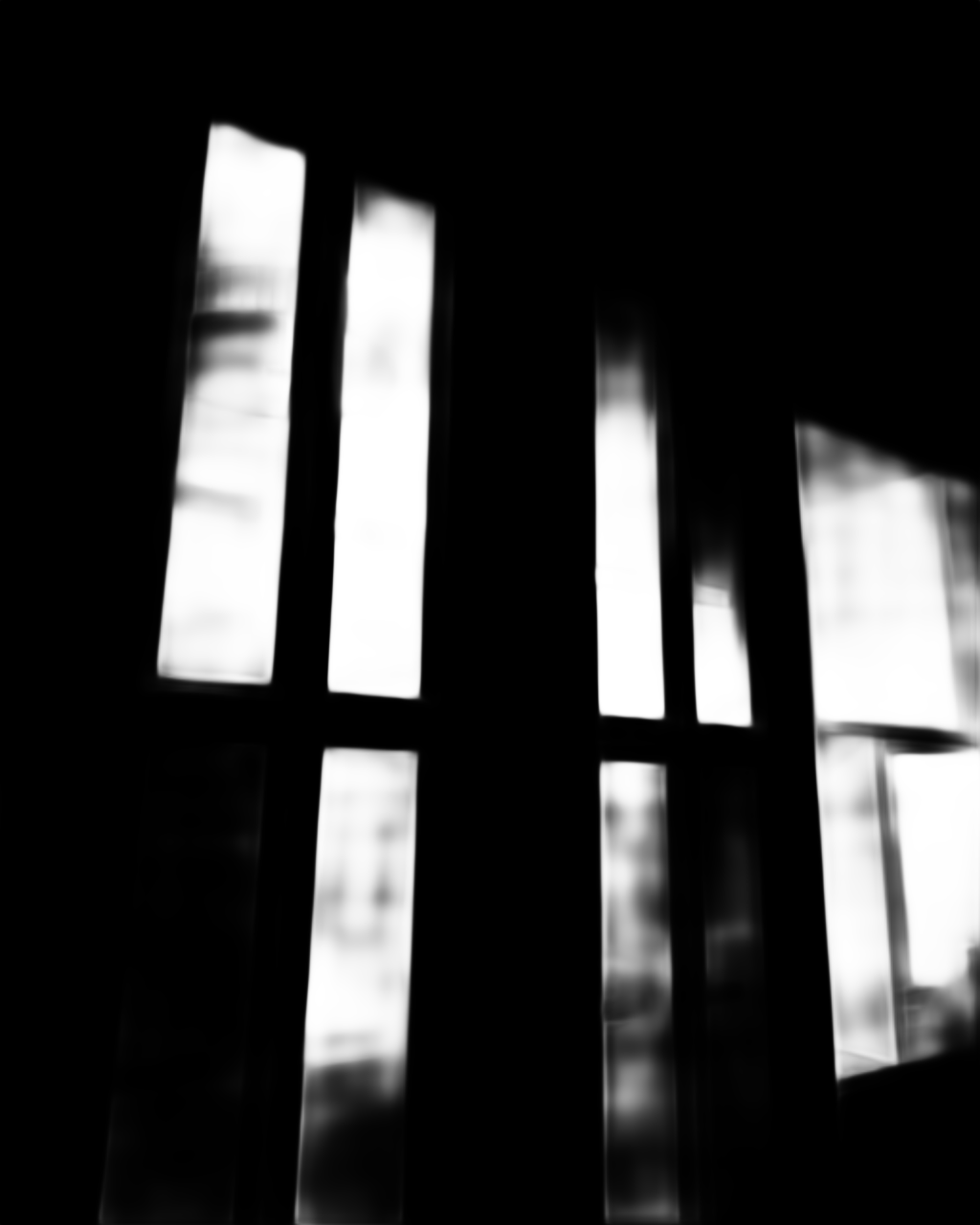}

                \includegraphics[width=1\linewidth]{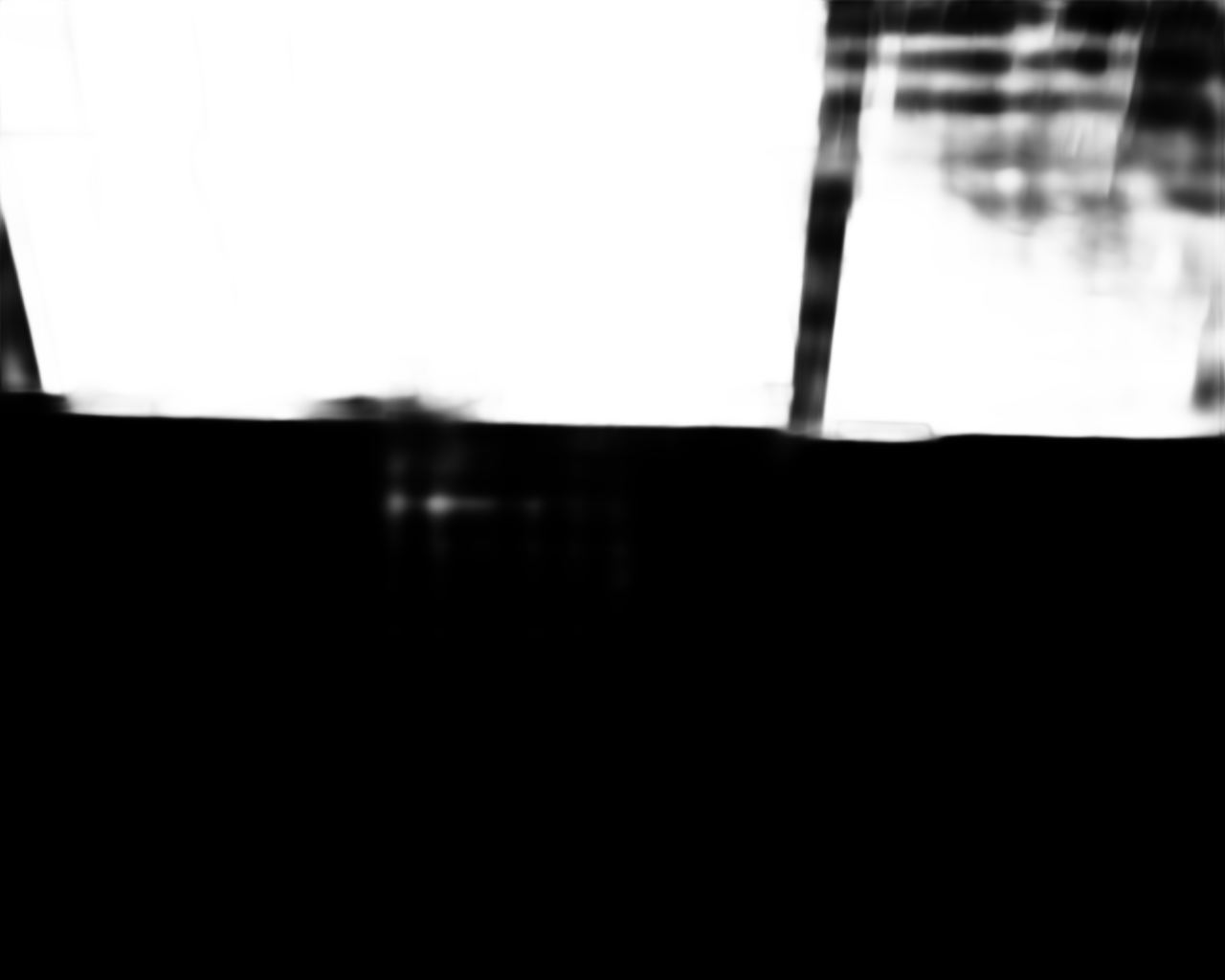}
    
                \includegraphics[width=1\linewidth]{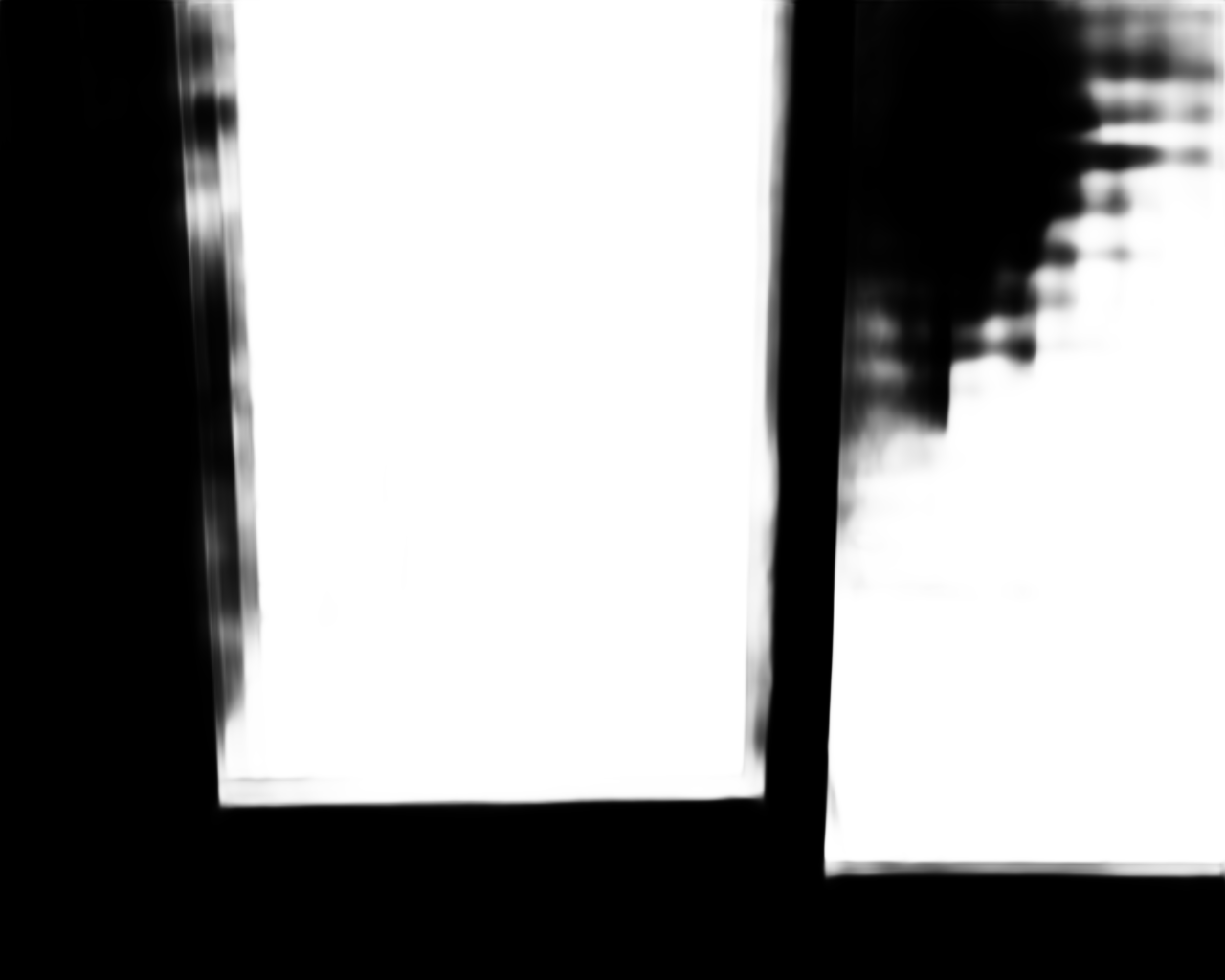}
                
                \includegraphics[width=1\linewidth]{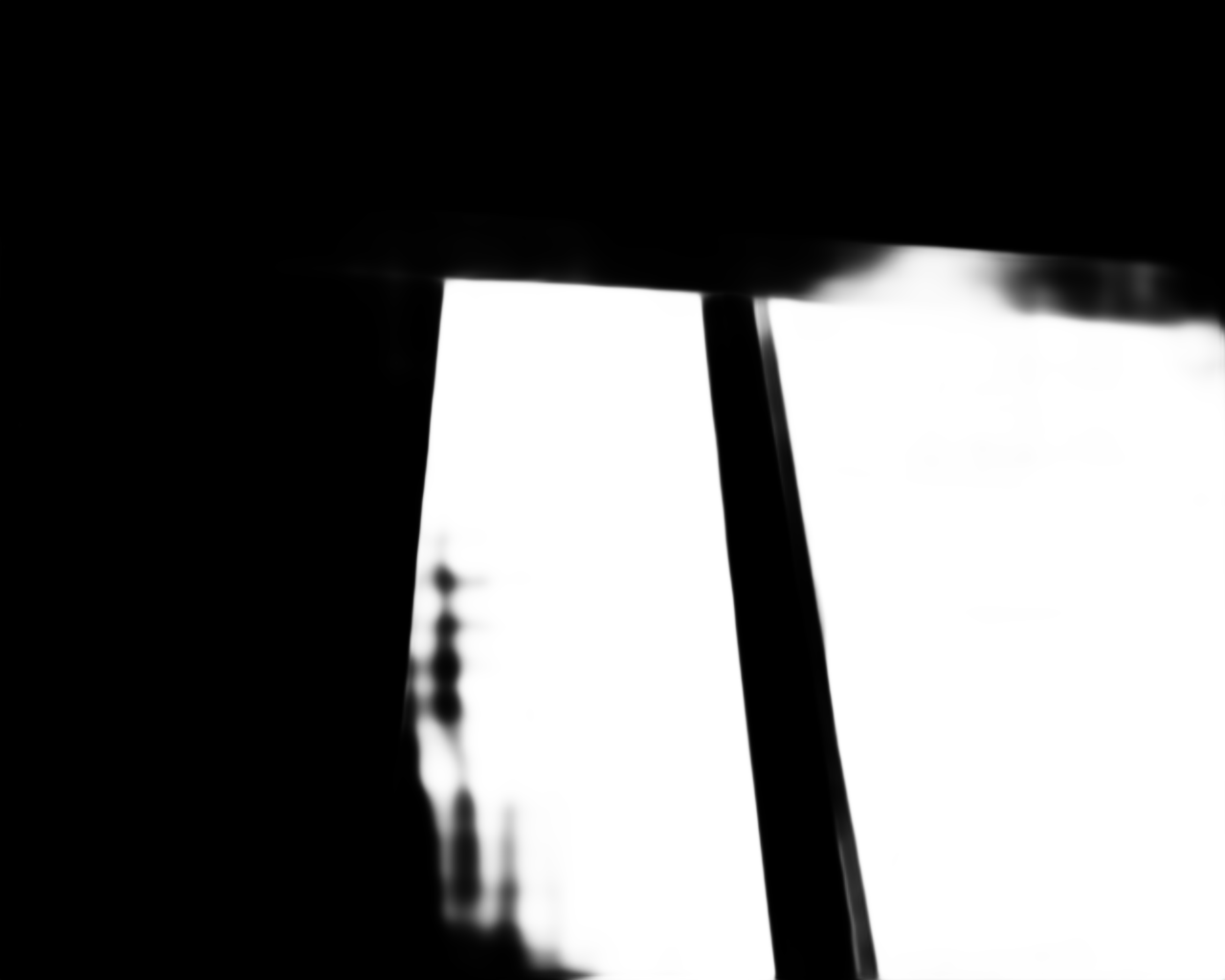}
    
                \includegraphics[width=1\linewidth]{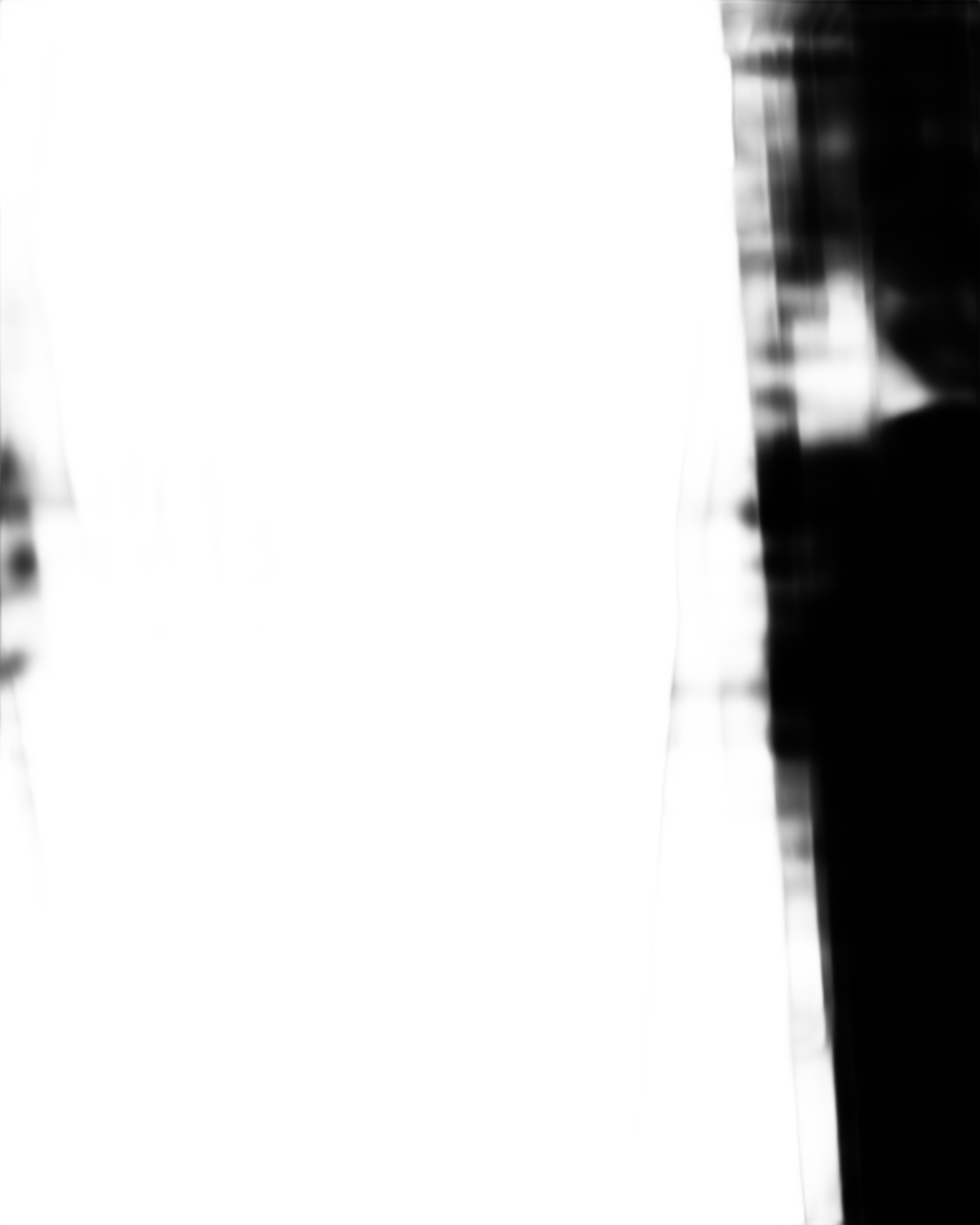}
    
                \includegraphics[width=1\linewidth]{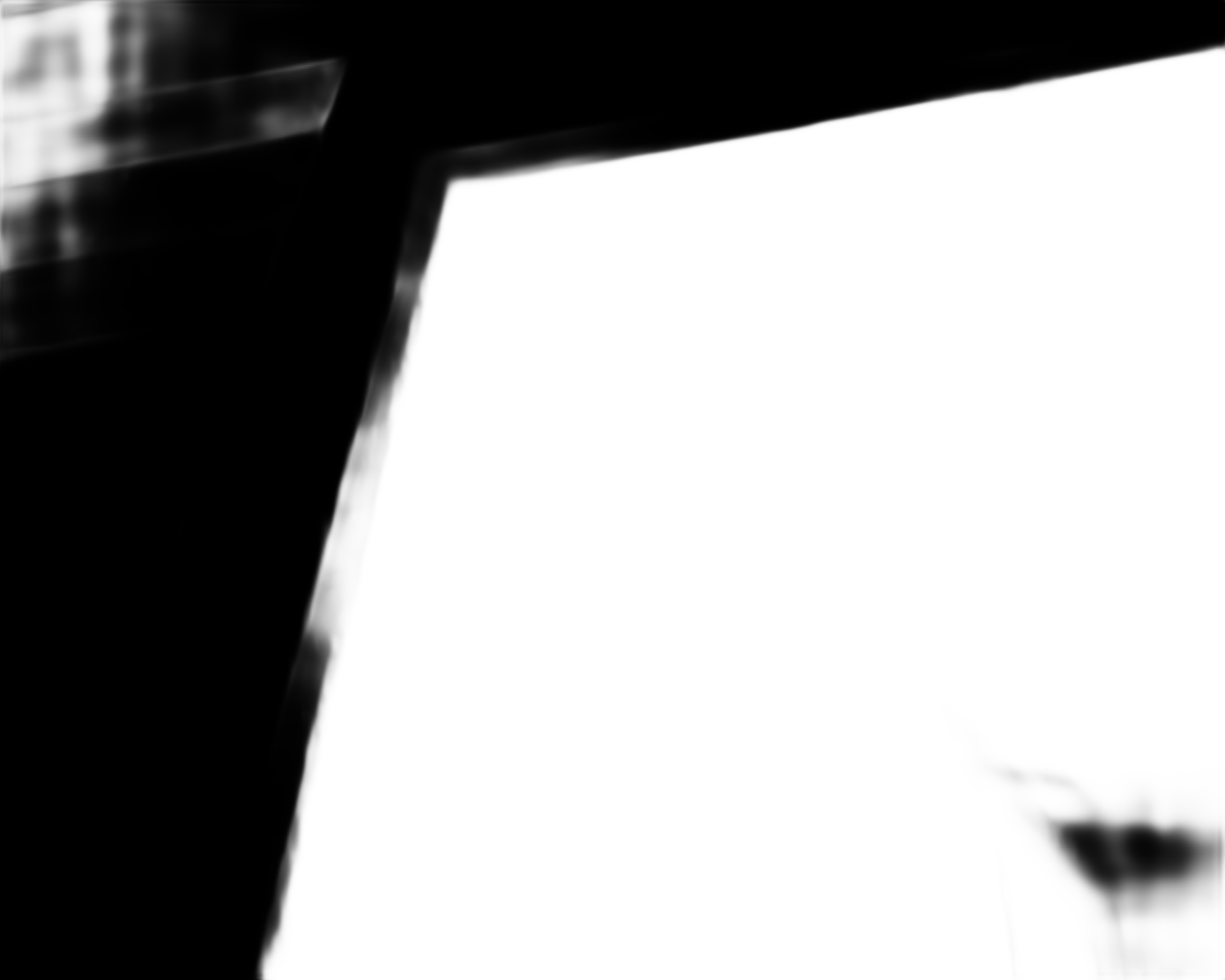}

          \end{minipage}
          }         
          \subfloat[LDF]{
          \begin{minipage}[t]{0.08\textwidth}
                \centering

                \includegraphics[width=1\linewidth]{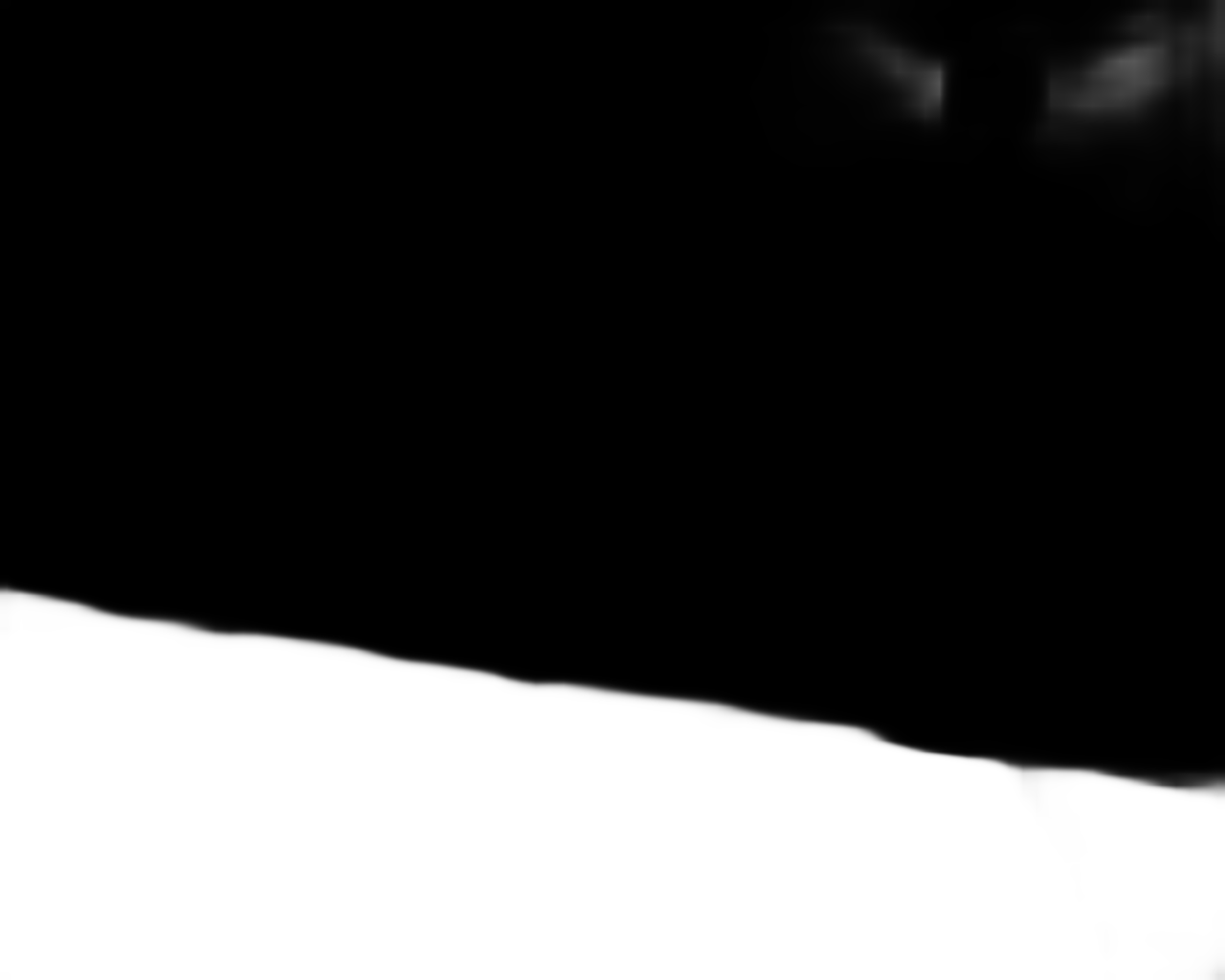}
                
                \includegraphics[width=1\linewidth]{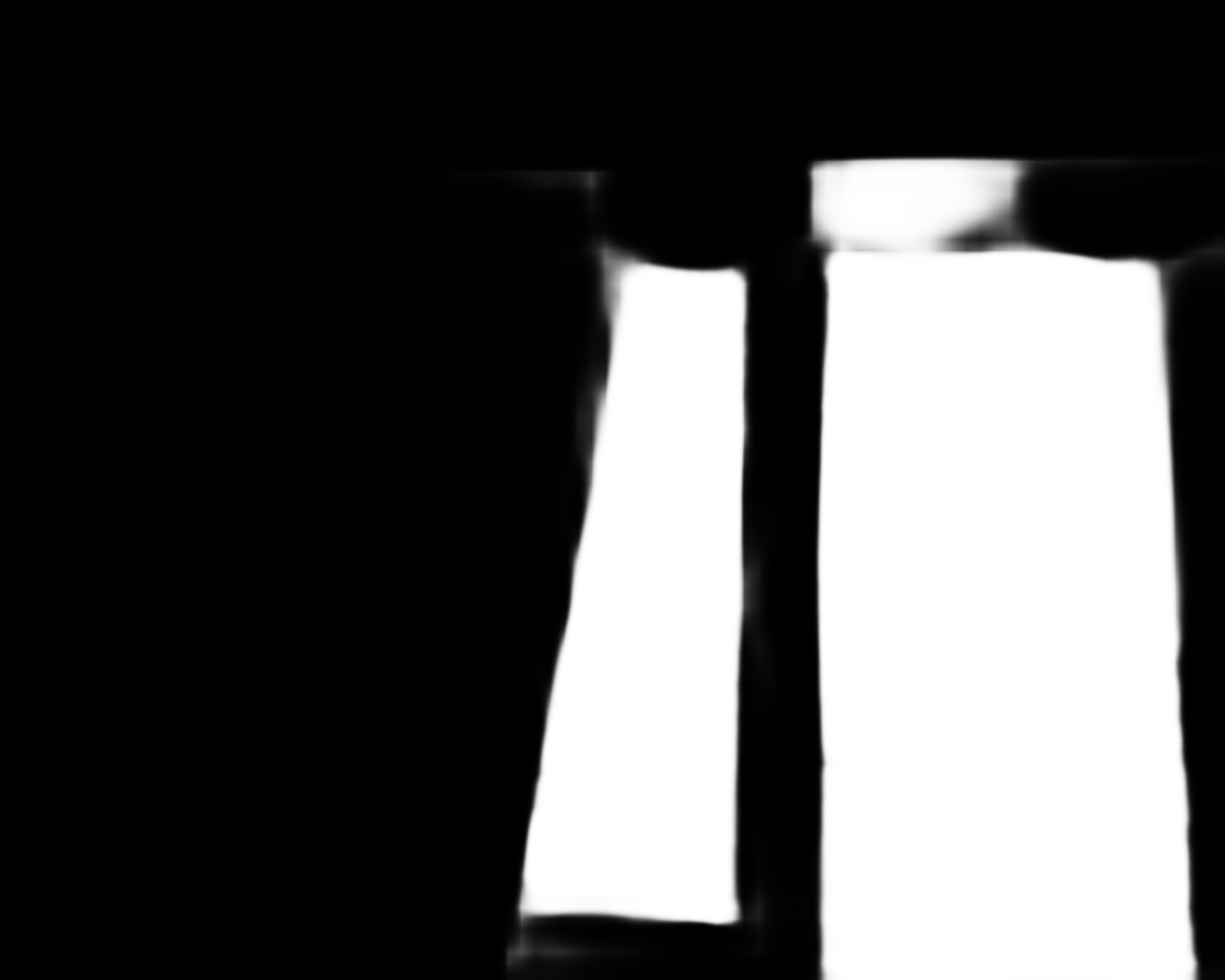}
    
                \includegraphics[width=1\linewidth]{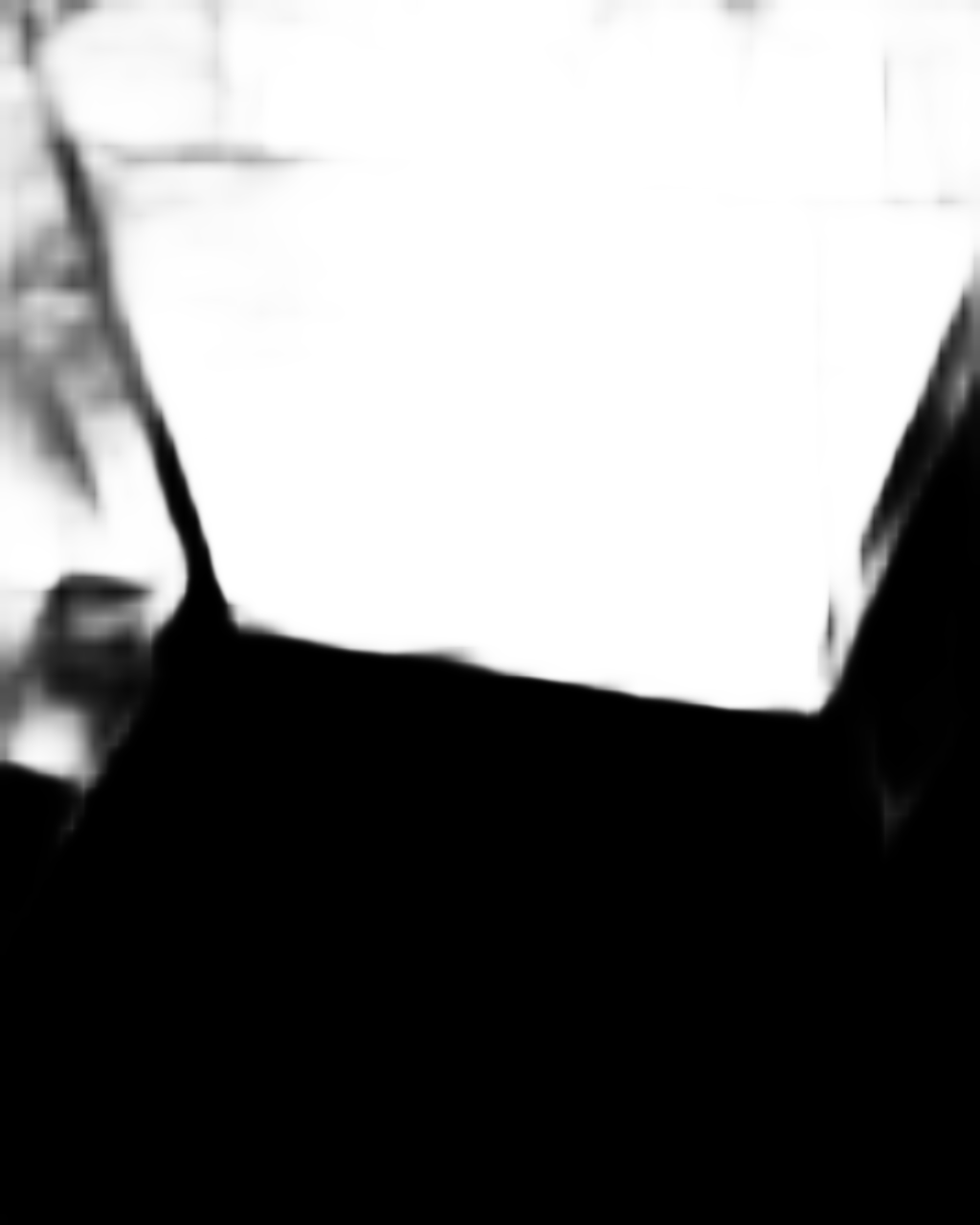}
                
                \includegraphics[width=1\linewidth]{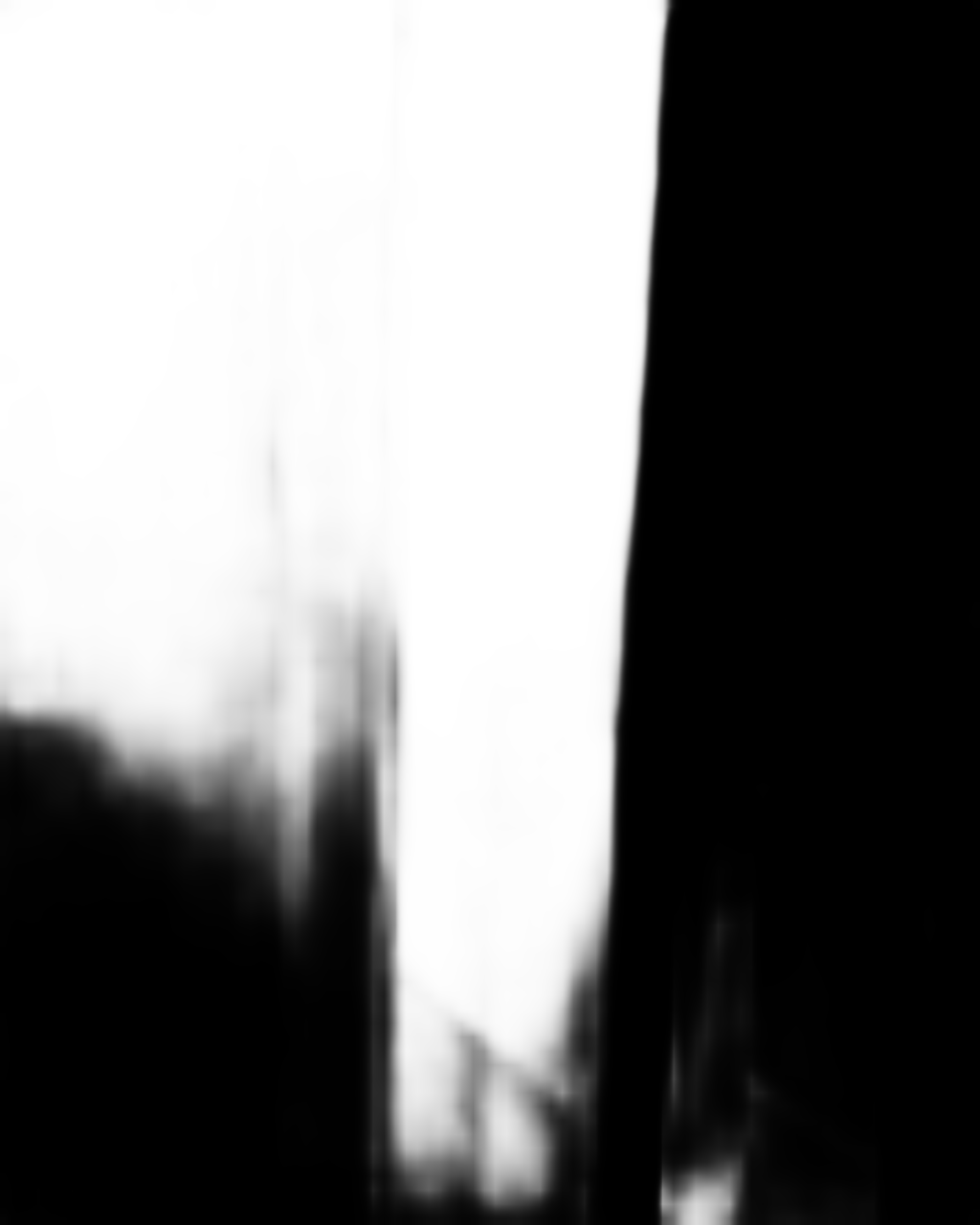}
    
                \includegraphics[width=1\linewidth]{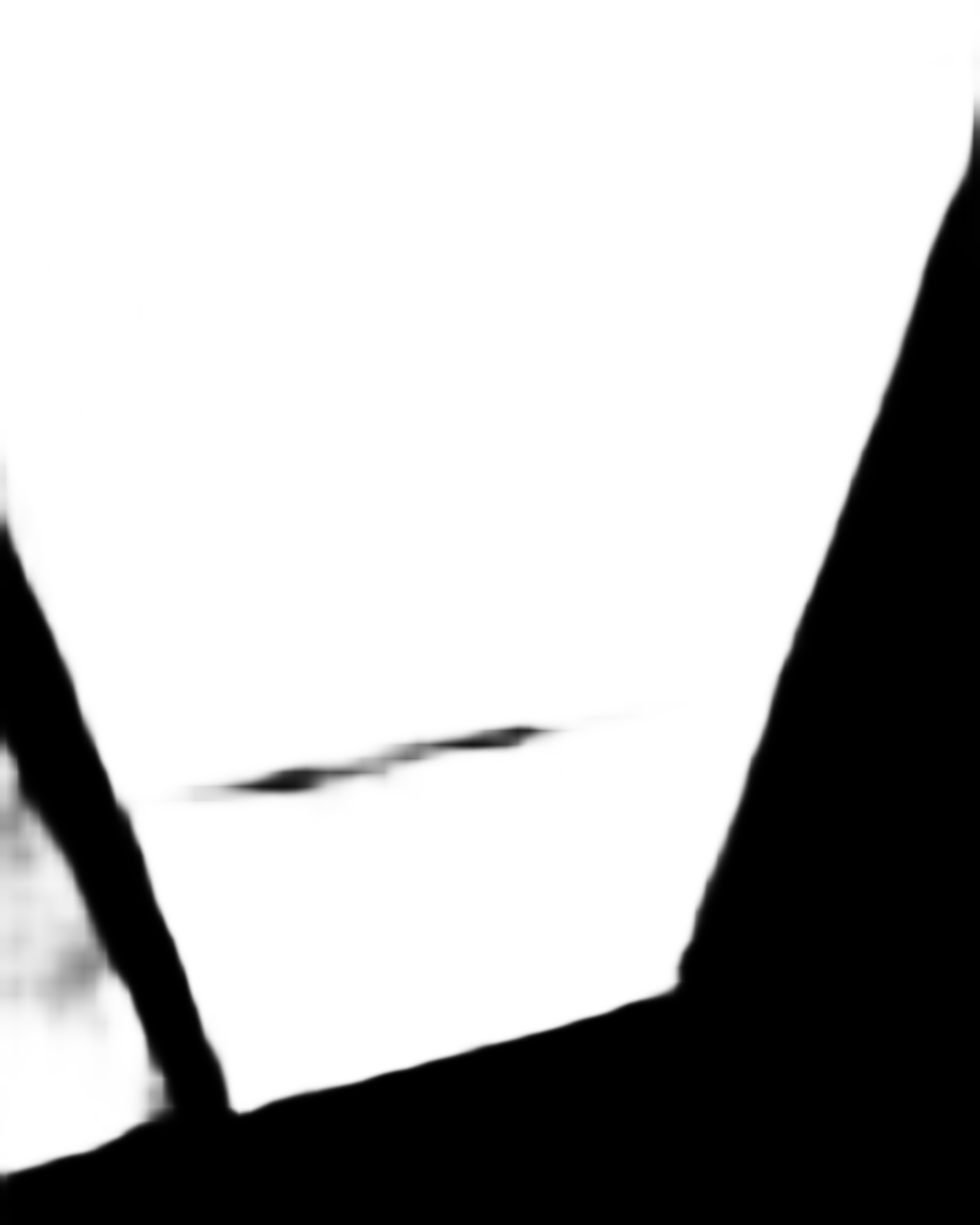}
    
                \includegraphics[width=1\linewidth]{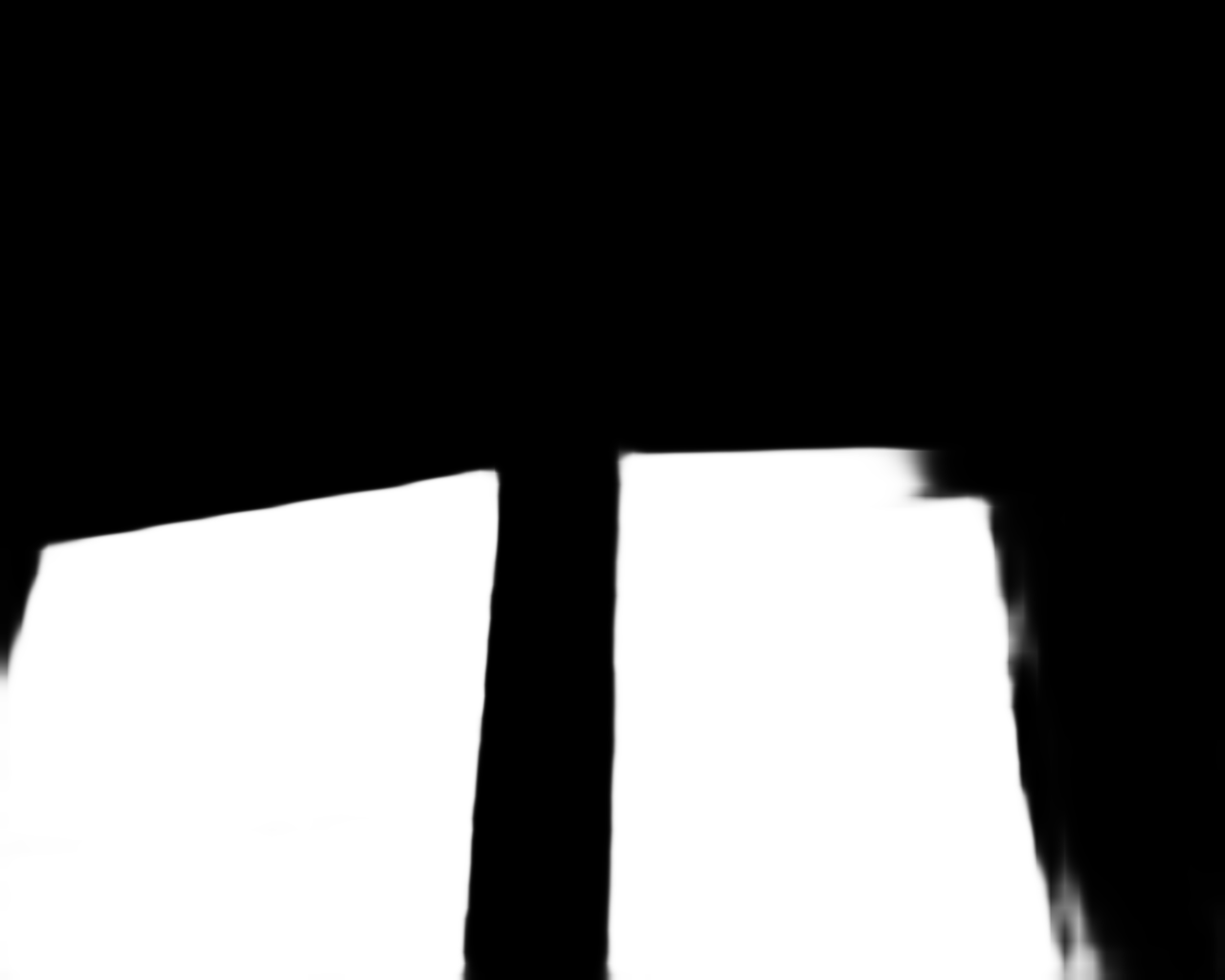}
                
                \includegraphics[width=1\linewidth]{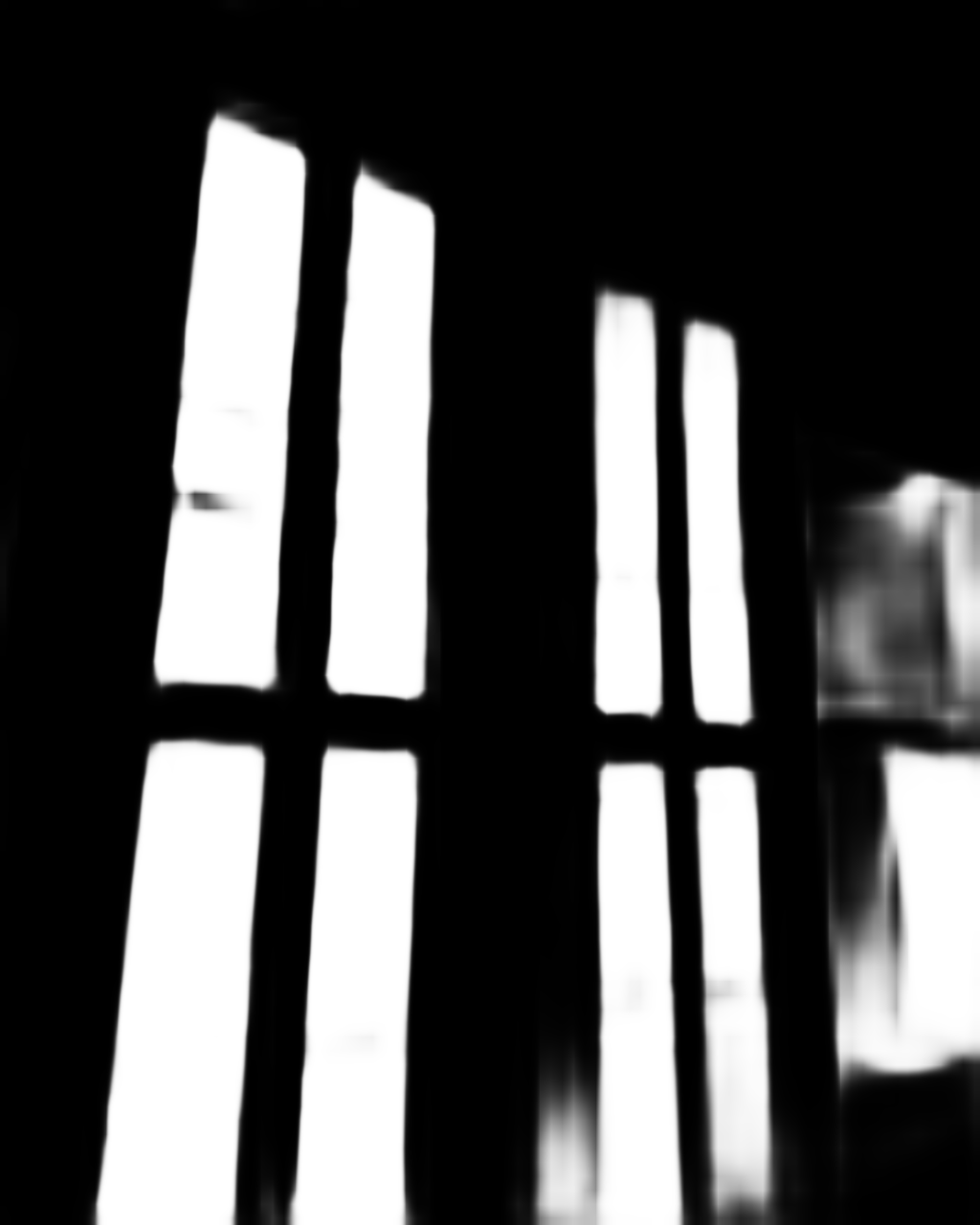}

                \includegraphics[width=1\linewidth]{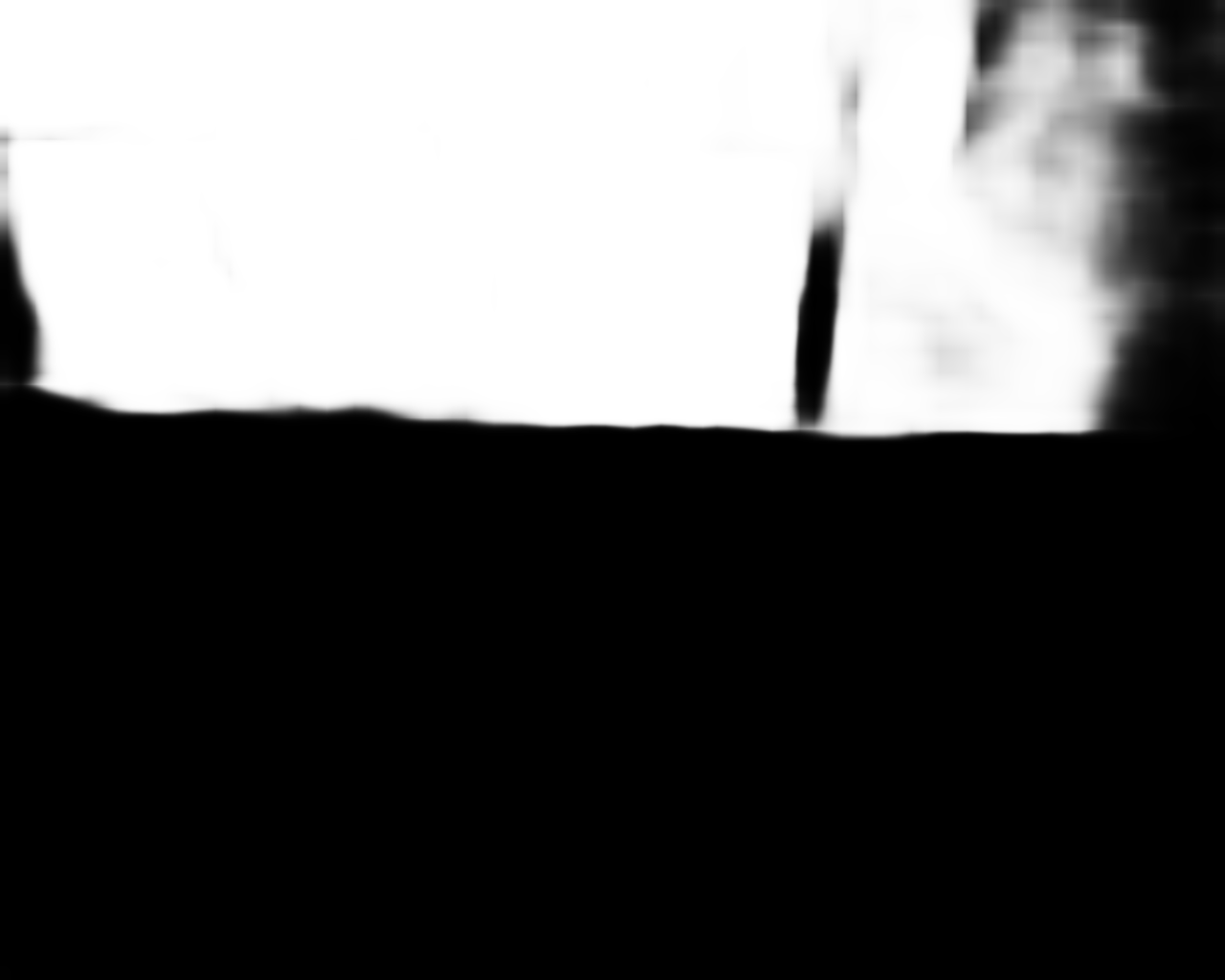}
    
                \includegraphics[width=1\linewidth]{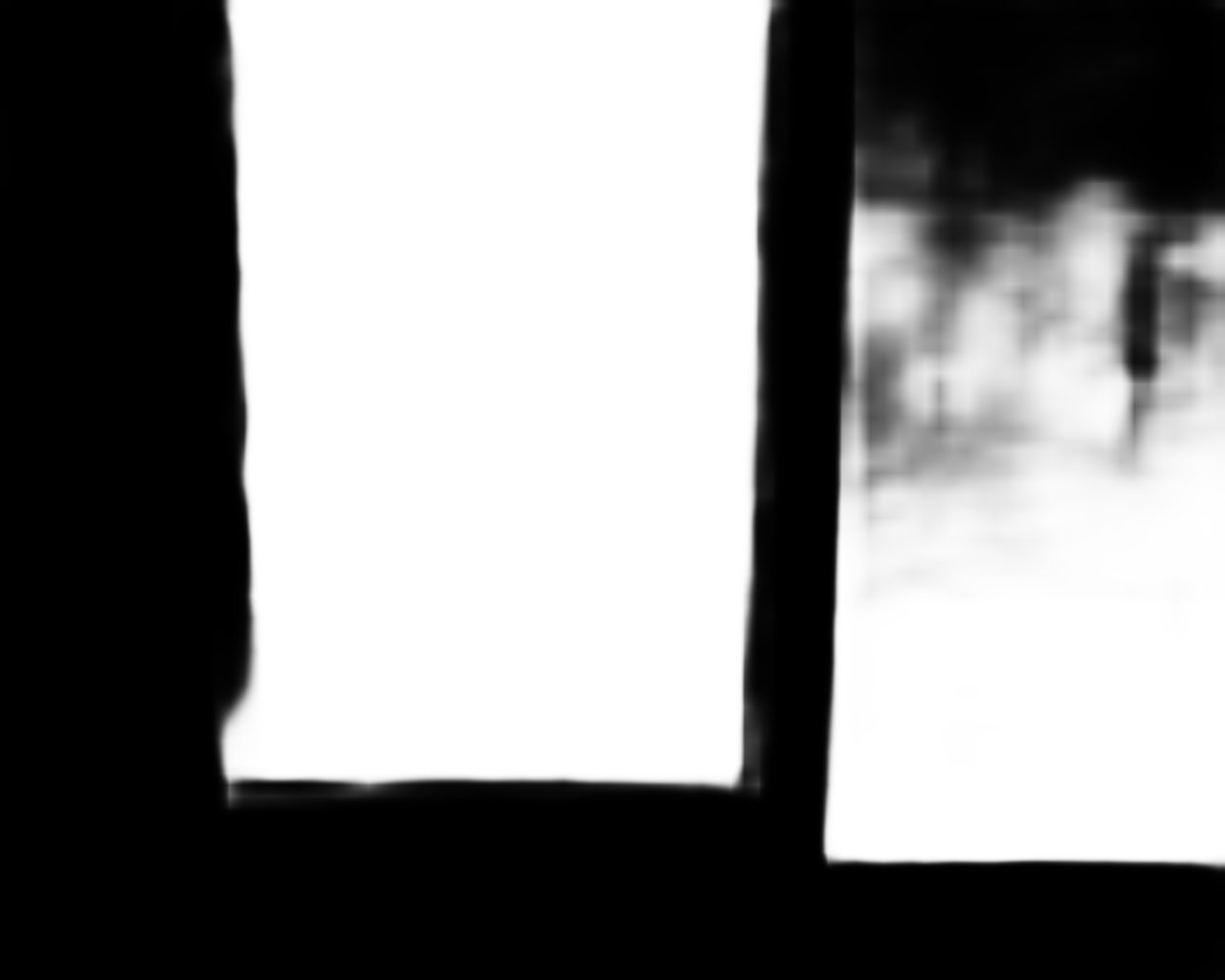}
                
                \includegraphics[width=1\linewidth]{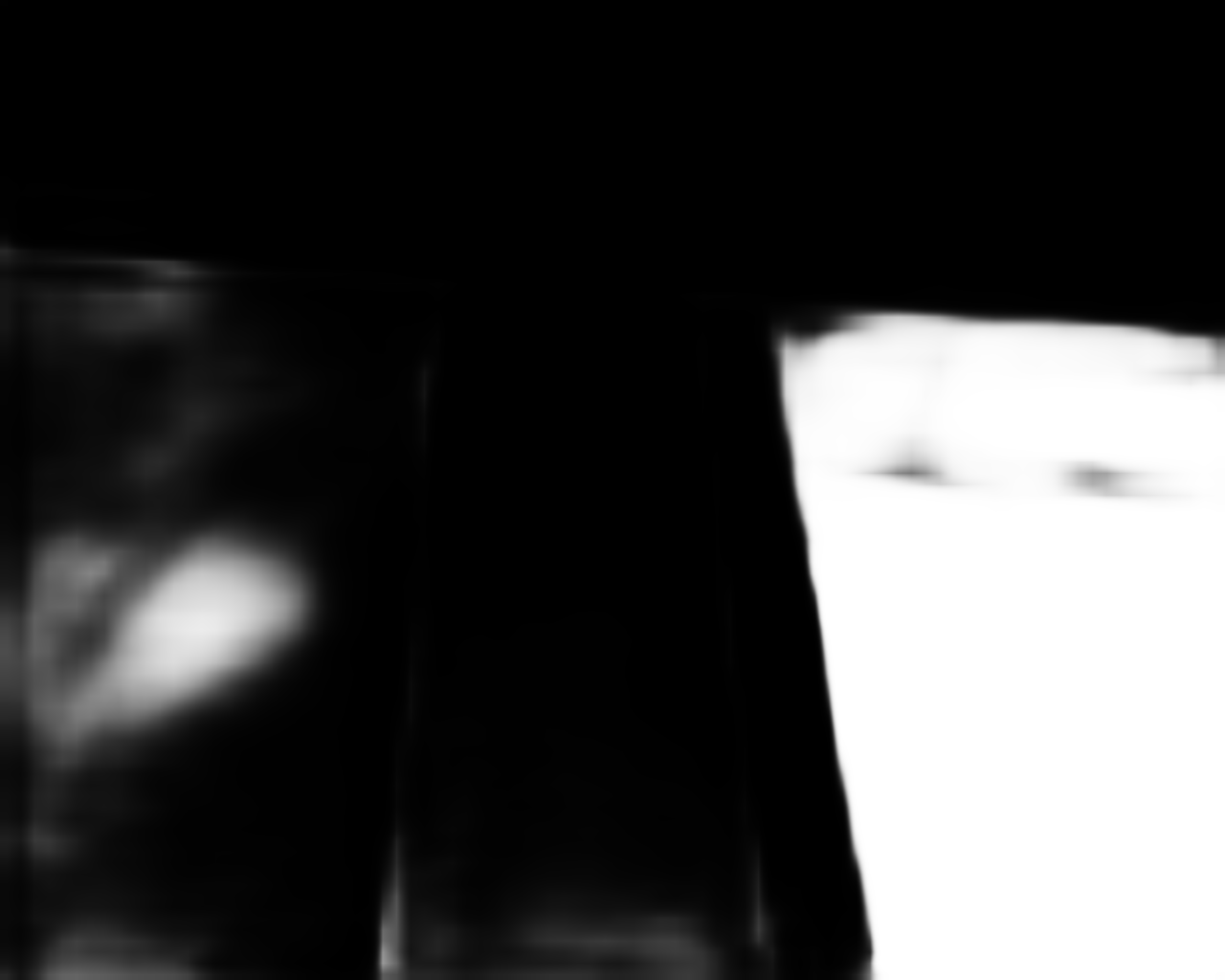}
    
                \includegraphics[width=1\linewidth]{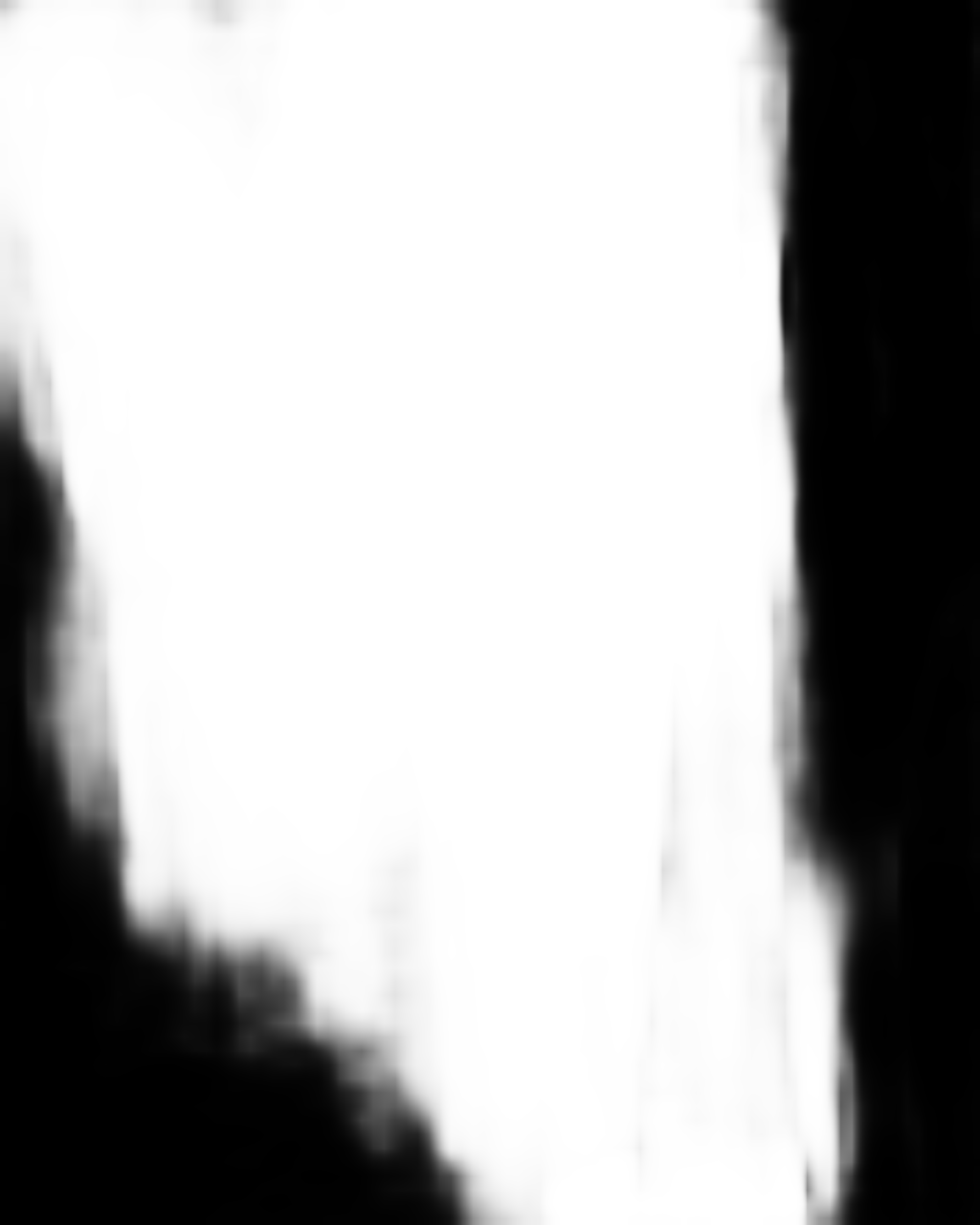}
    
                \includegraphics[width=1\linewidth]{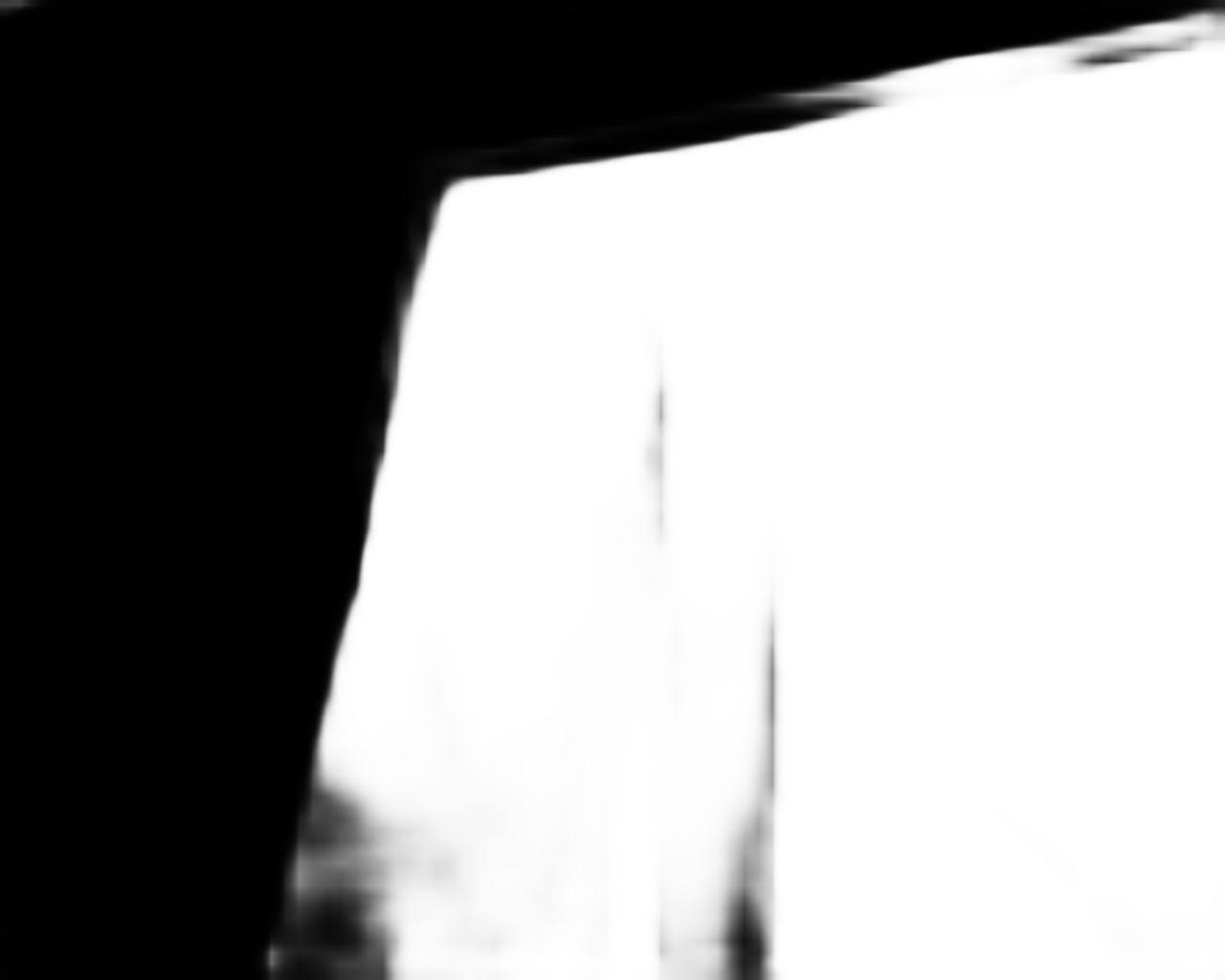}

          \end{minipage}
          }  
          \subfloat[DSC]{
          \begin{minipage}[t]{0.08\textwidth}
                \centering

                \includegraphics[width=1\linewidth]{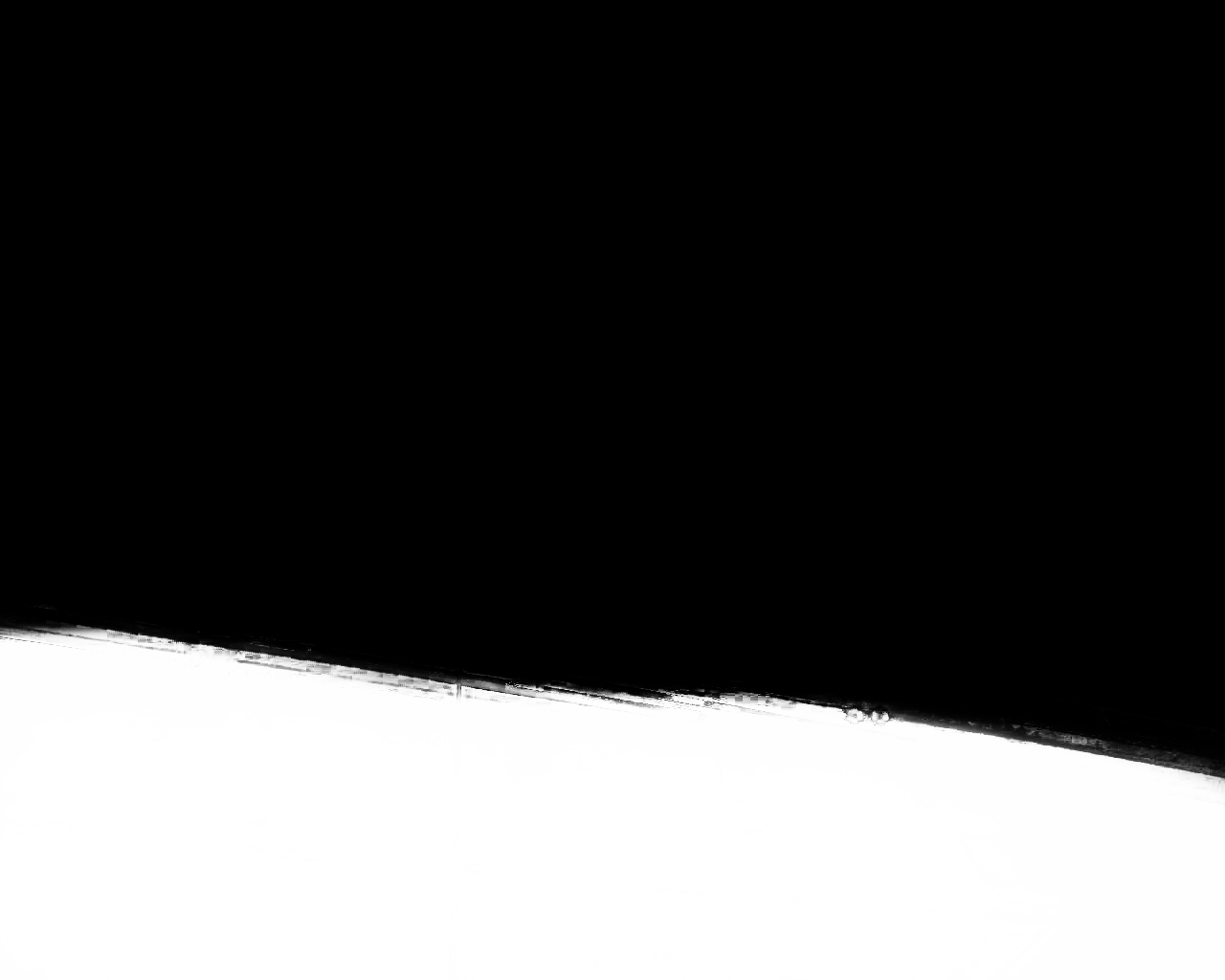}
                
                \includegraphics[width=1\linewidth]{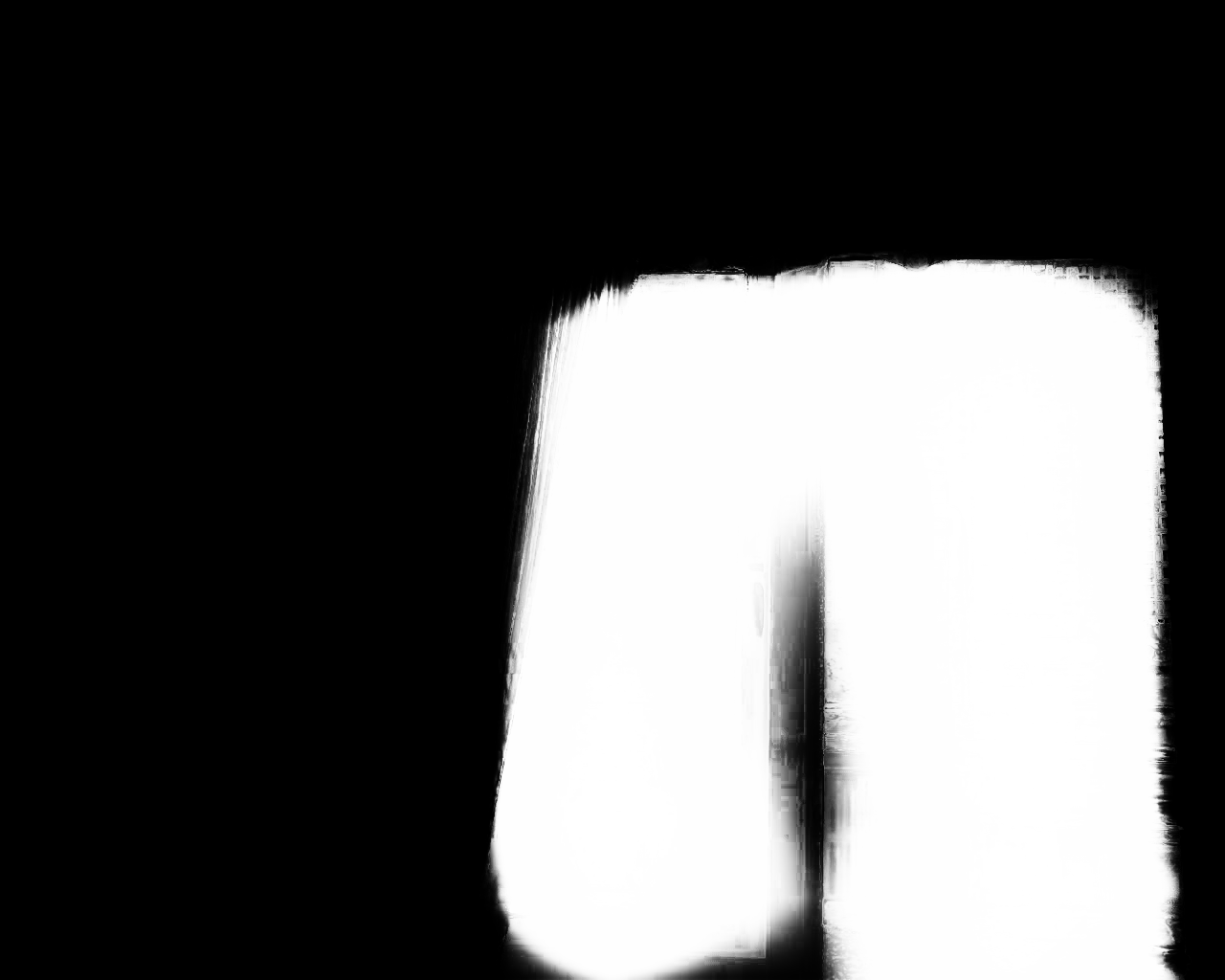}
    
                \includegraphics[width=1\linewidth]{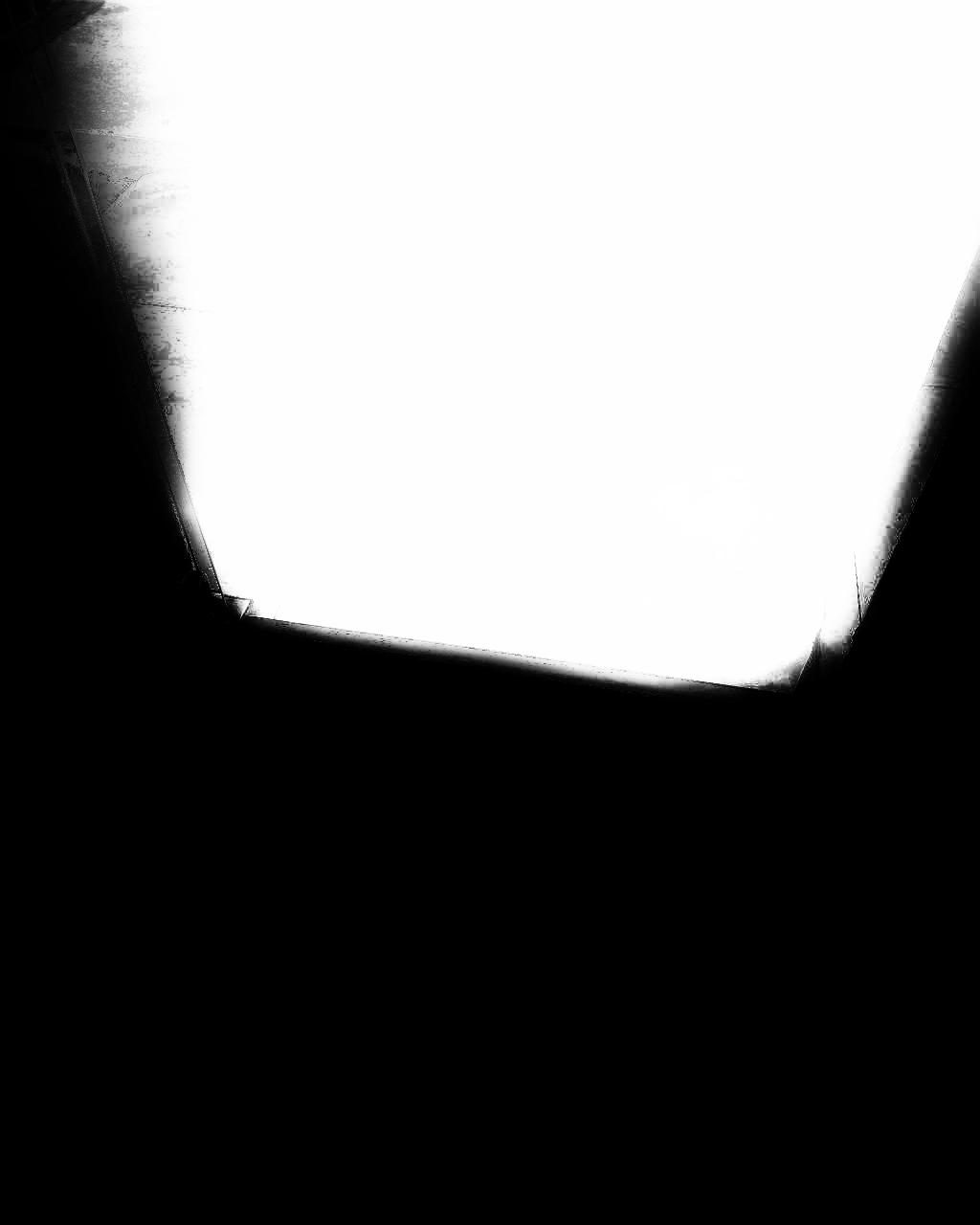}
                
                \includegraphics[width=1\linewidth]{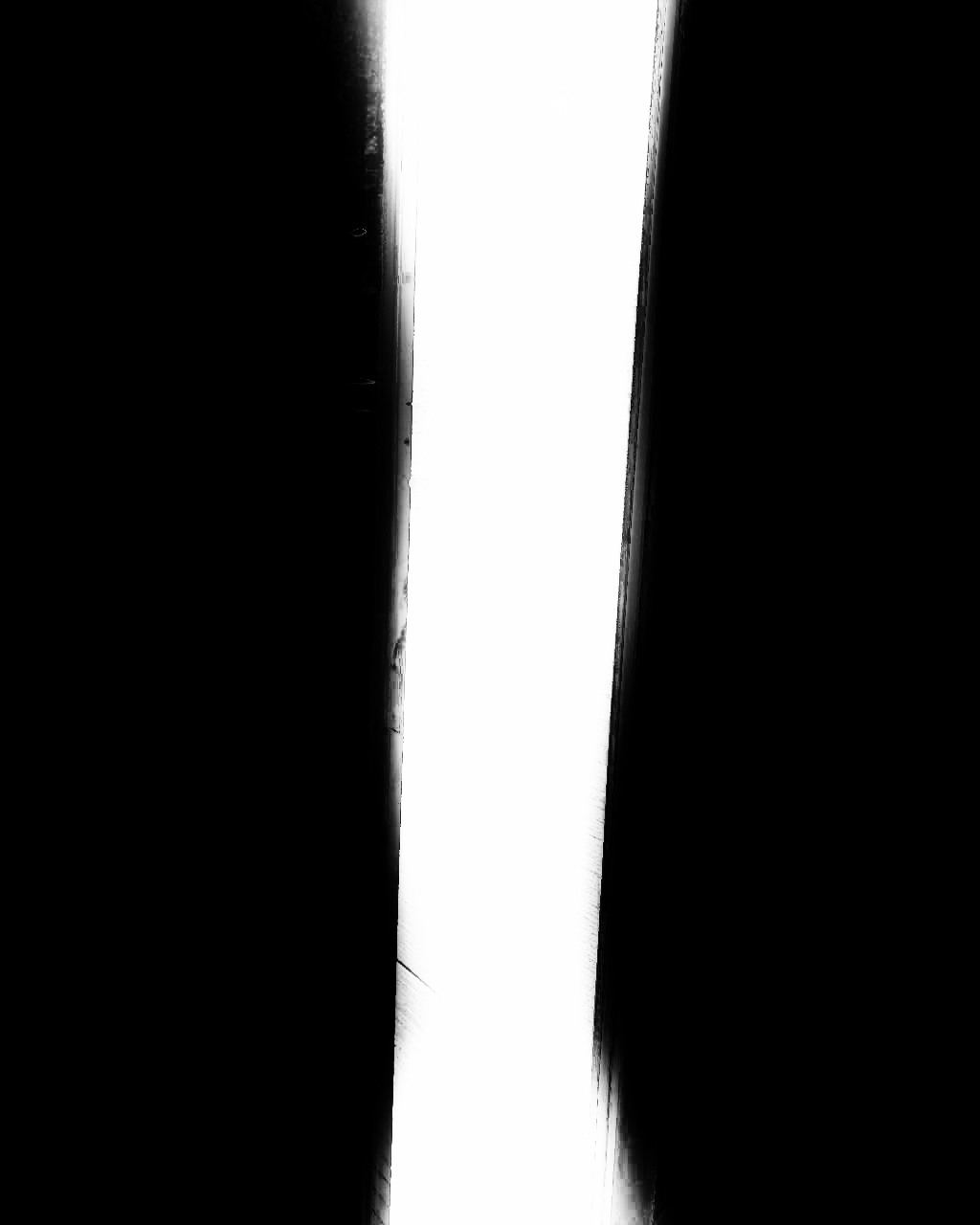}
    
                \includegraphics[width=1\linewidth]{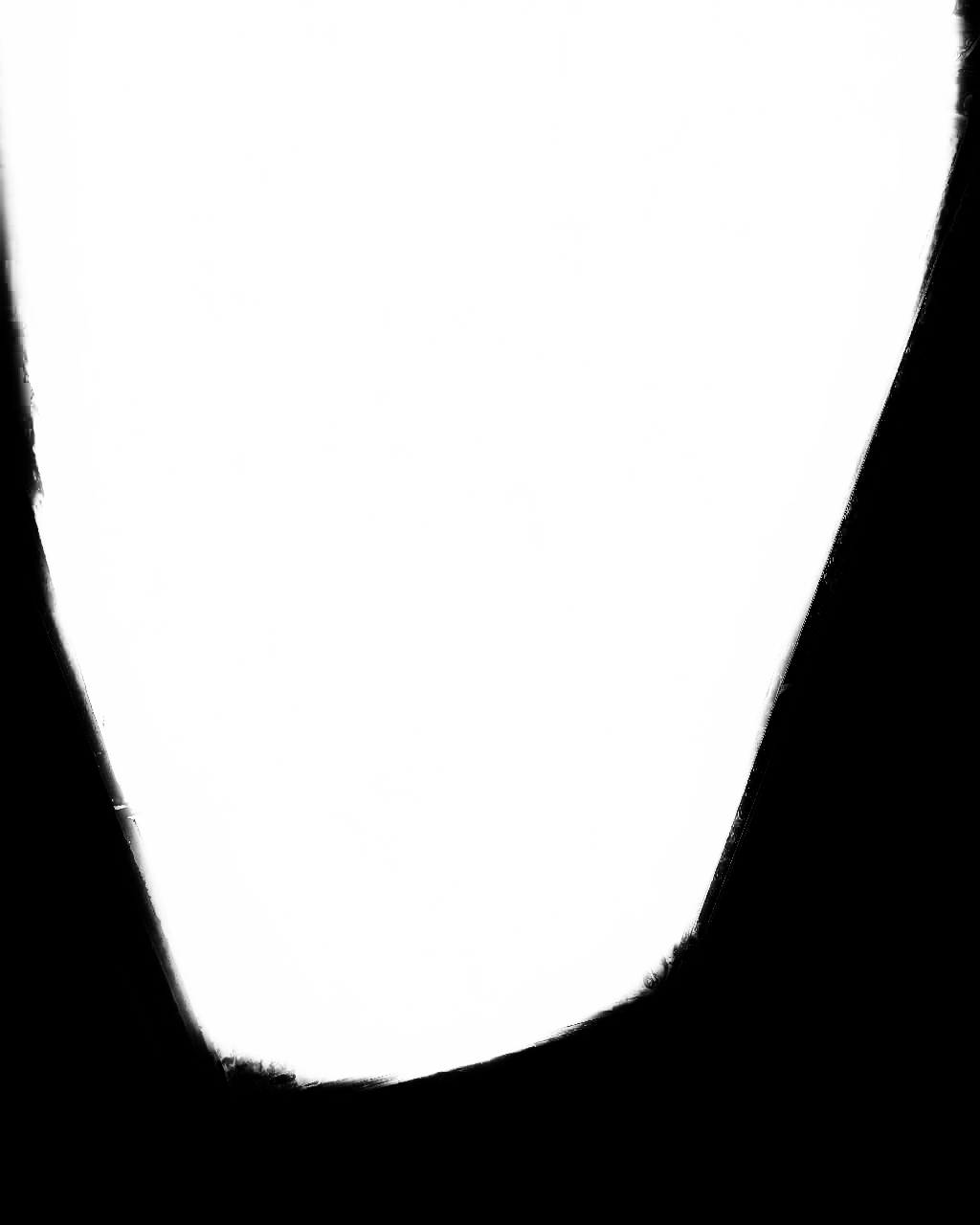}
    
                \includegraphics[width=1\linewidth]{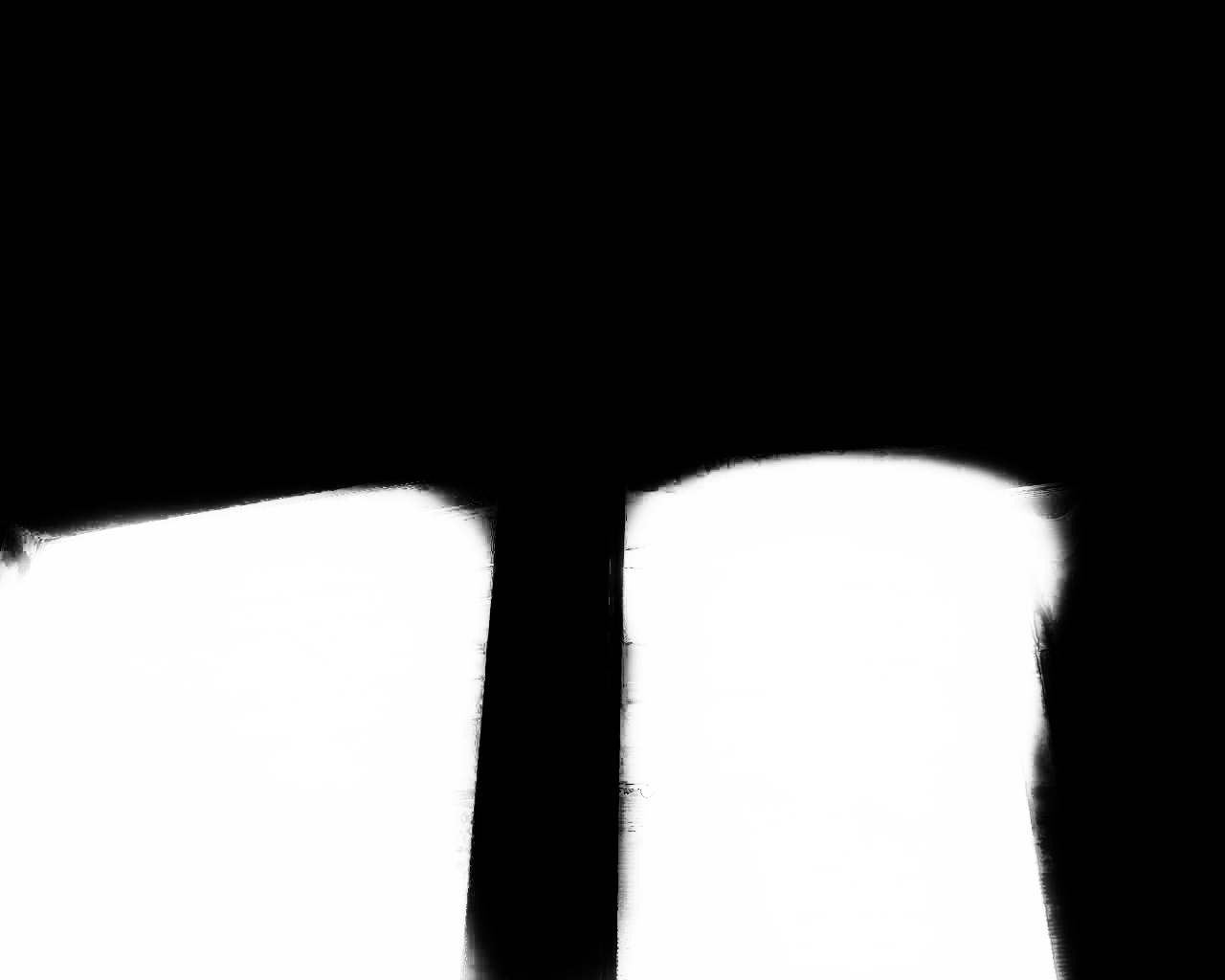}
                
                \includegraphics[width=1\linewidth]{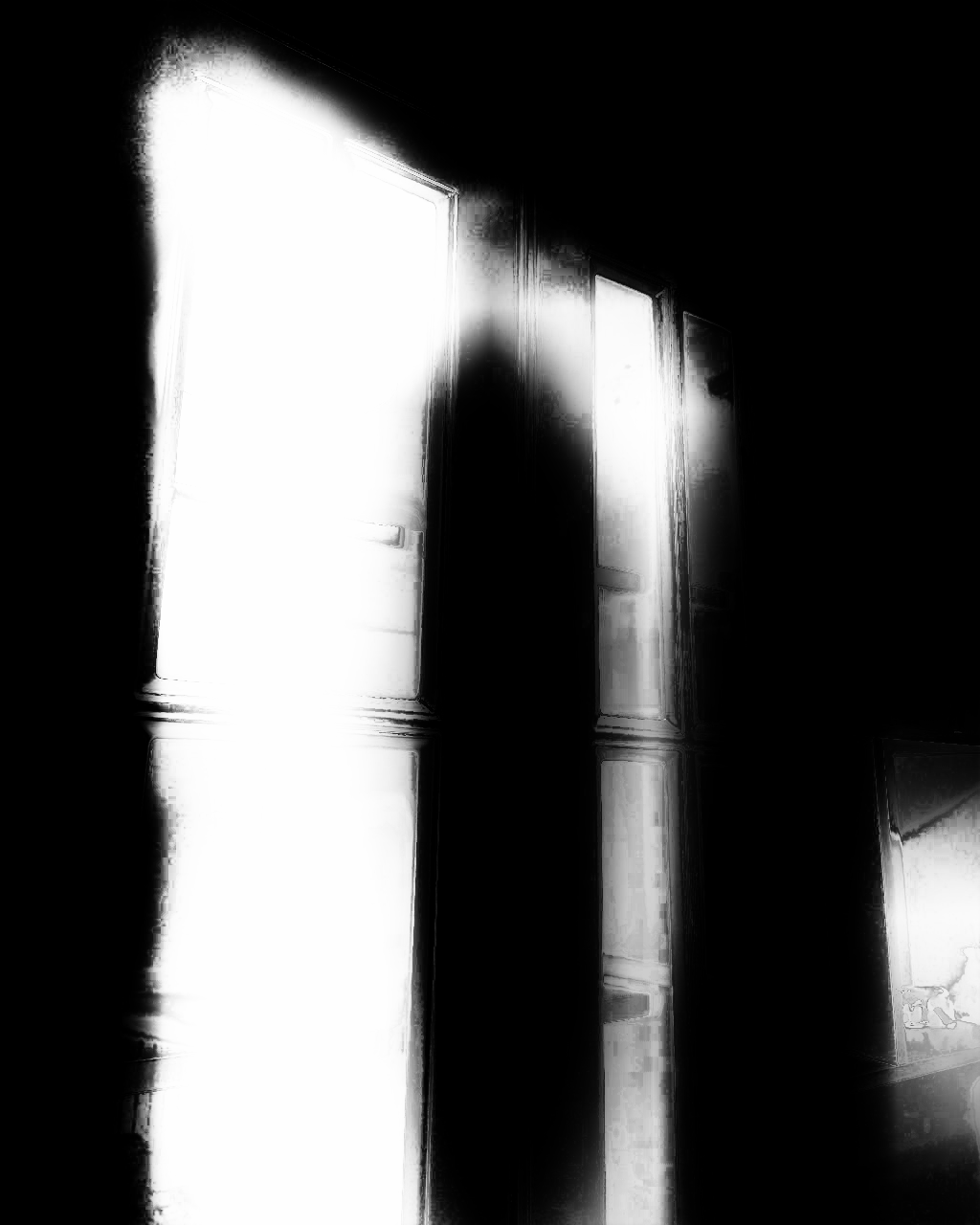}

                \includegraphics[width=1\linewidth]{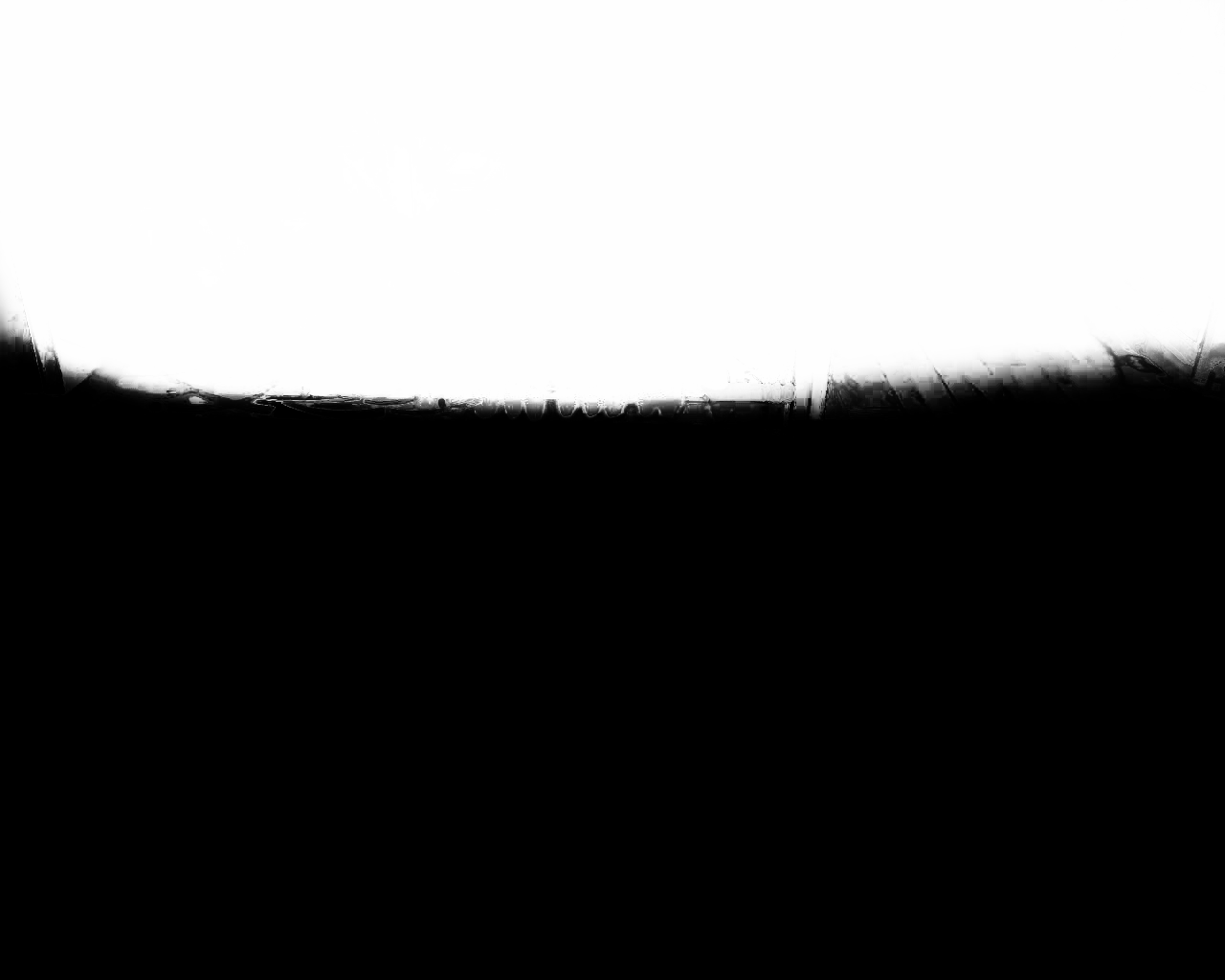}
    
                \includegraphics[width=1\linewidth]{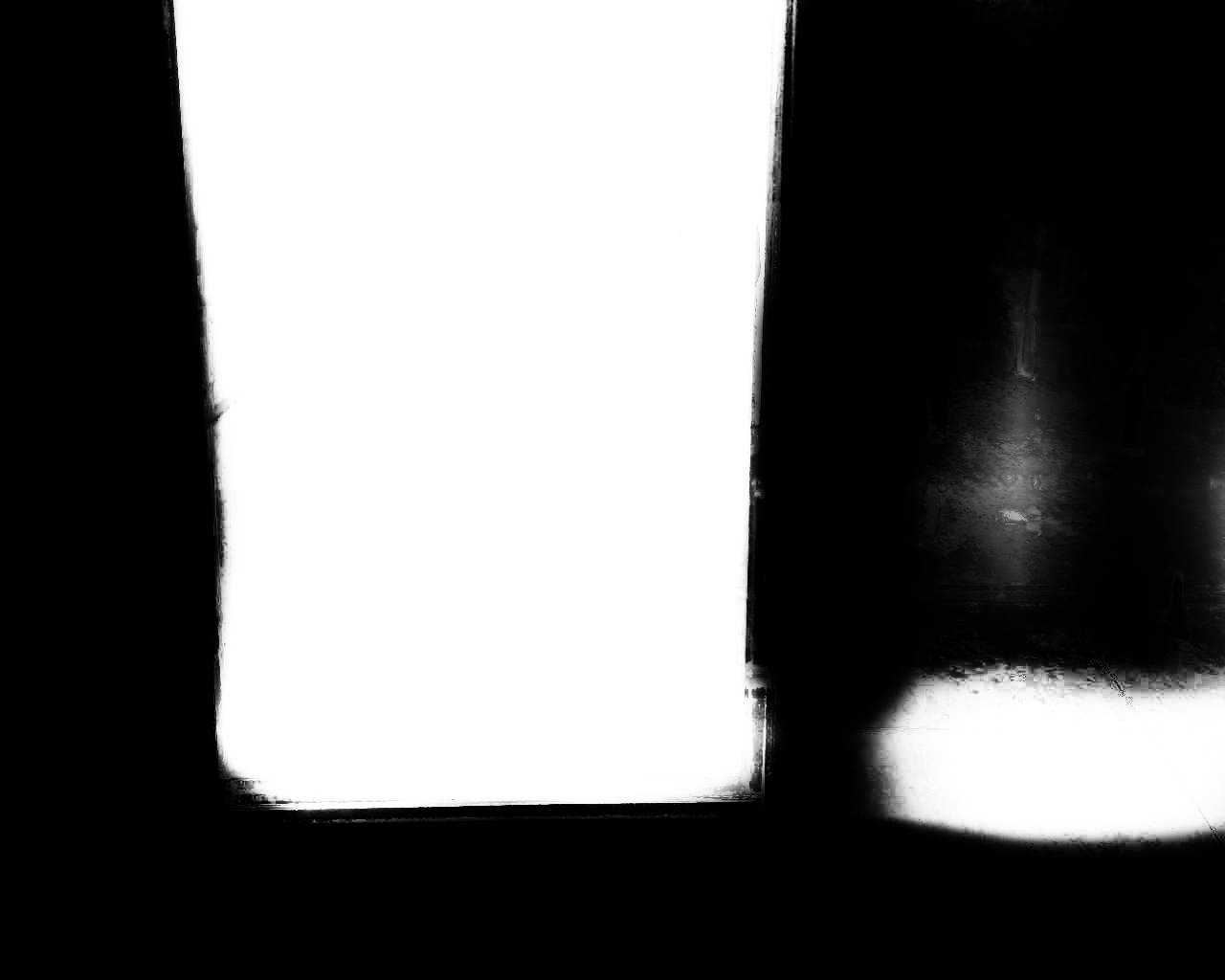}
                
                \includegraphics[width=1\linewidth]{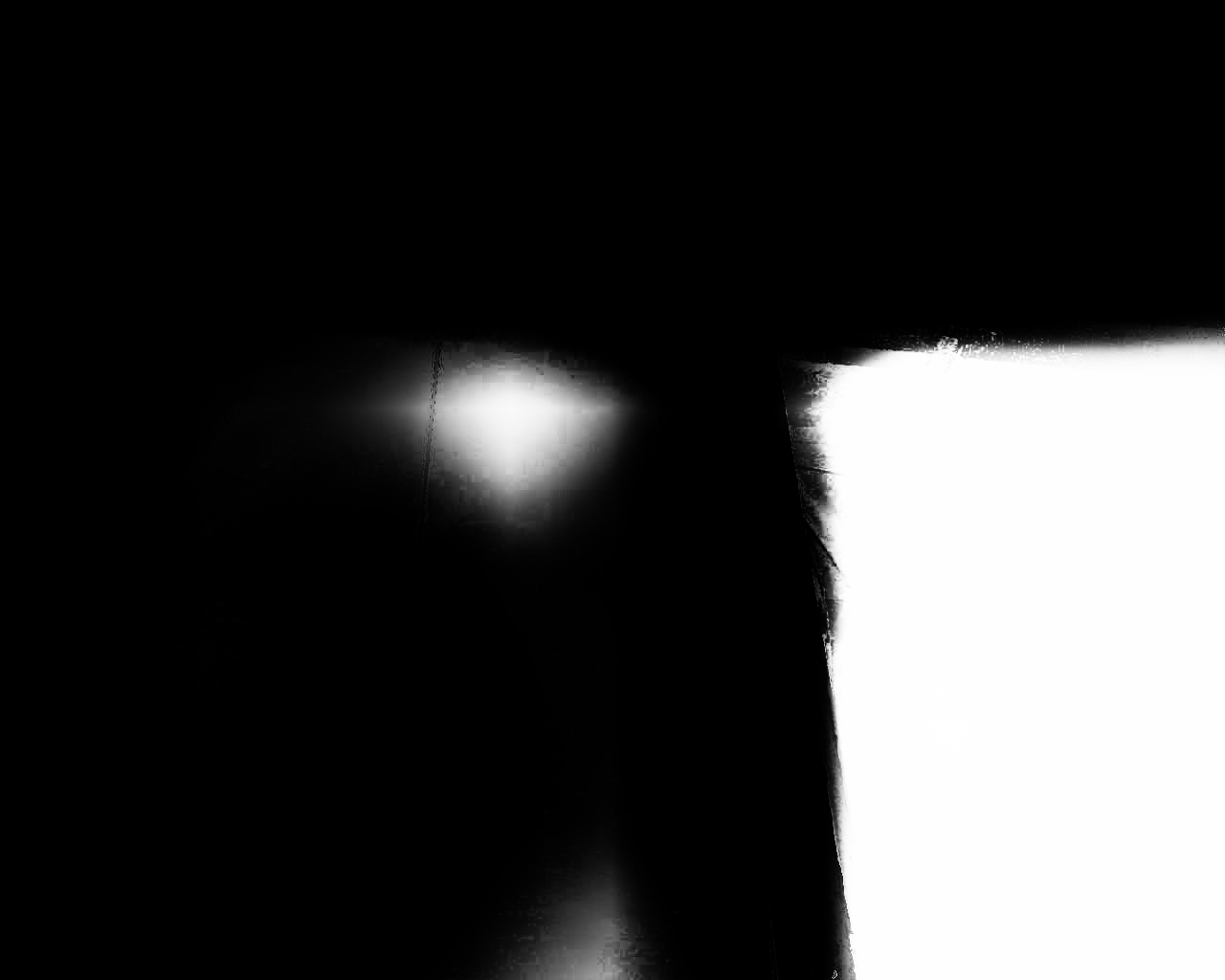}
    
                \includegraphics[width=1\linewidth]{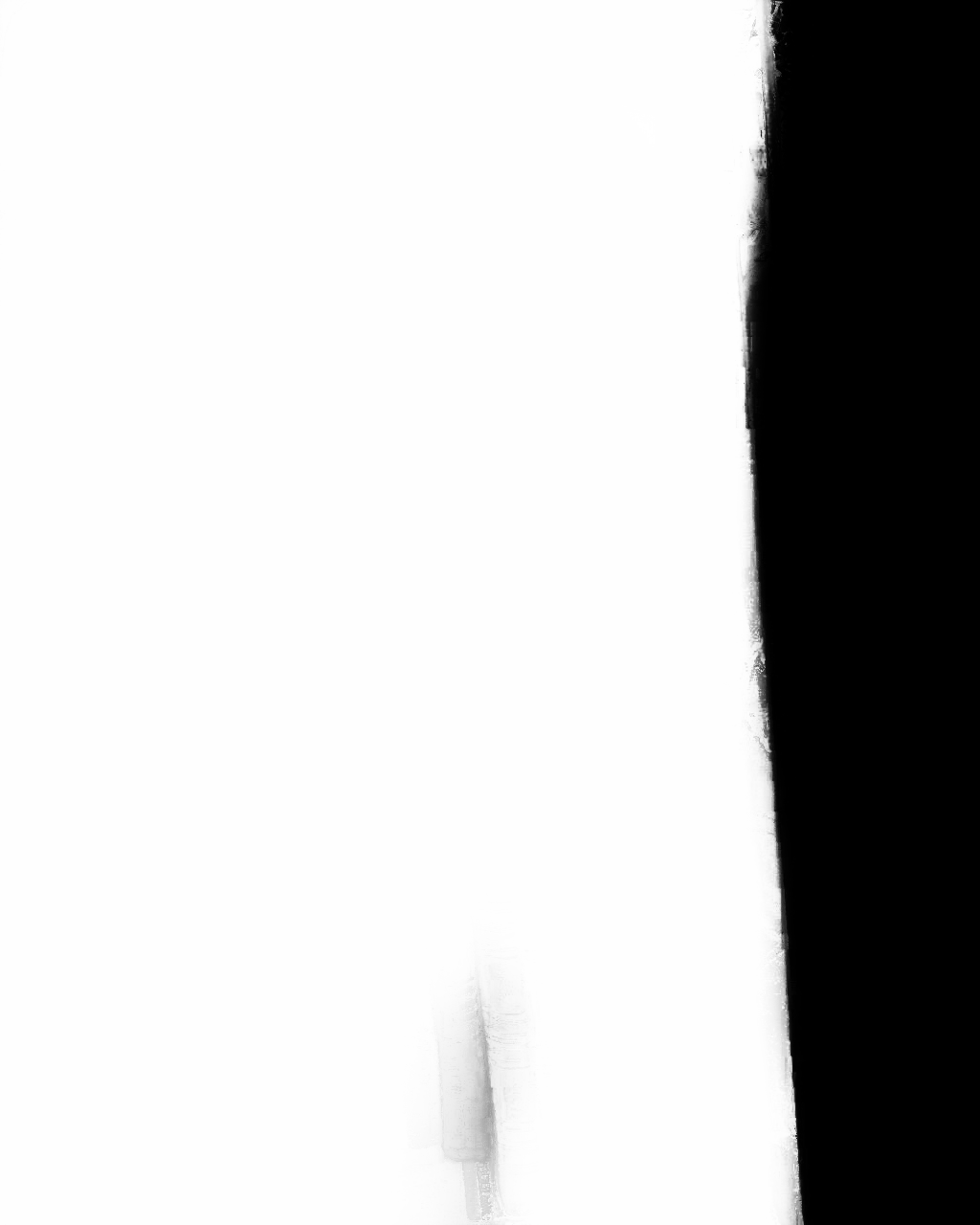}
    
                \includegraphics[width=1\linewidth]{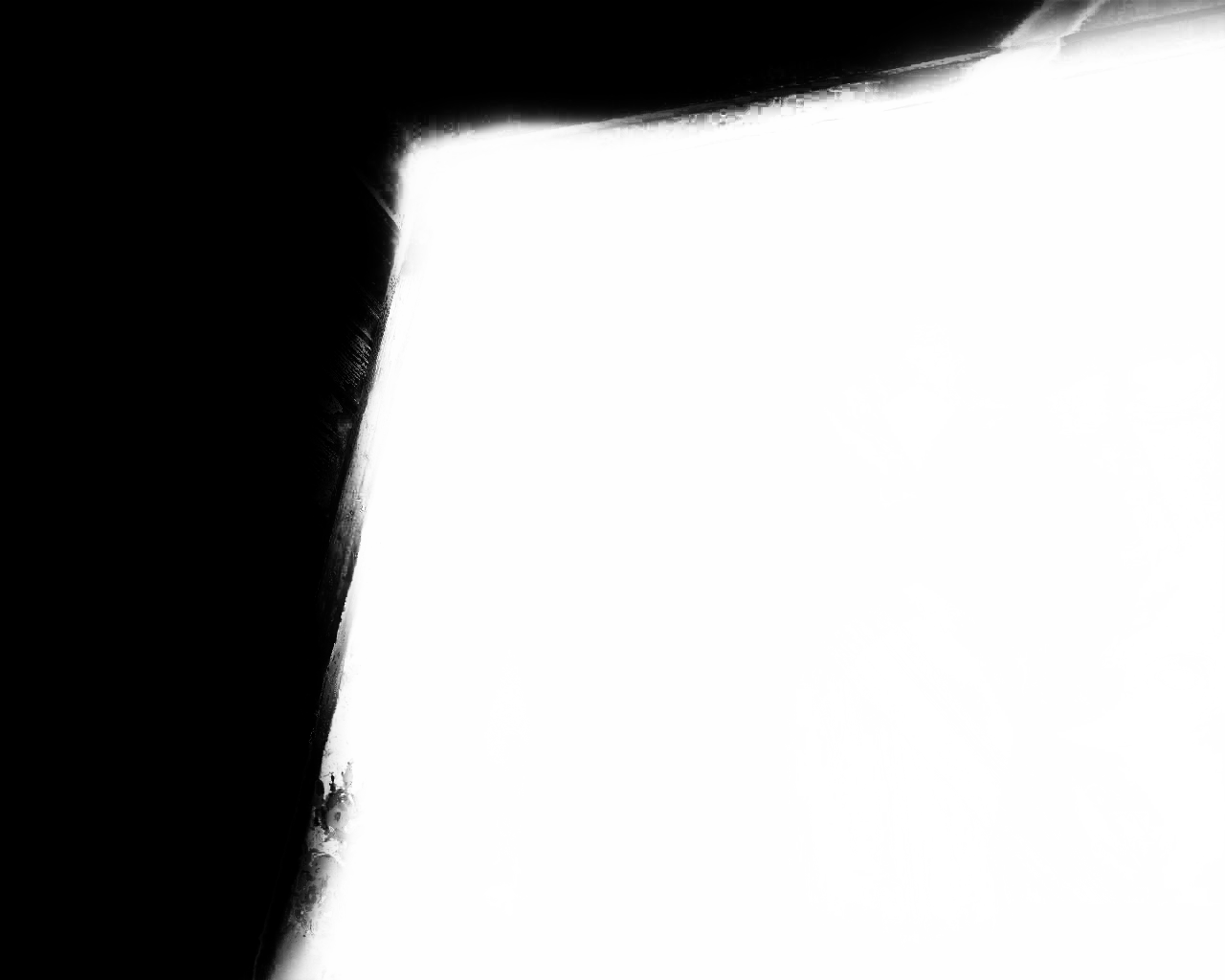}

          \end{minipage}
          }  
          \subfloat[DSD]{
          \begin{minipage}[t]{0.08\textwidth}
                \centering

                \includegraphics[width=1\linewidth]{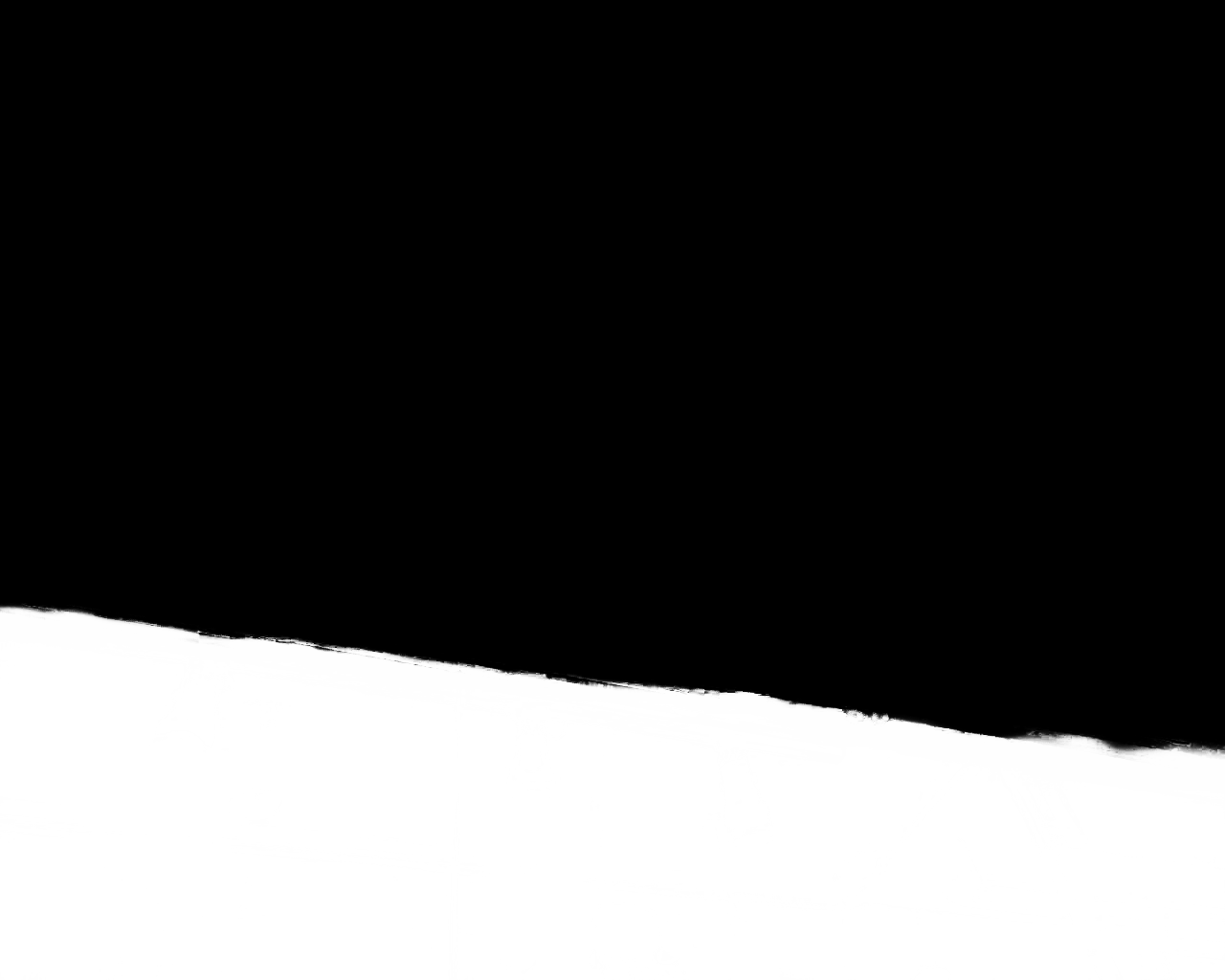}
                
                \includegraphics[width=1\linewidth]{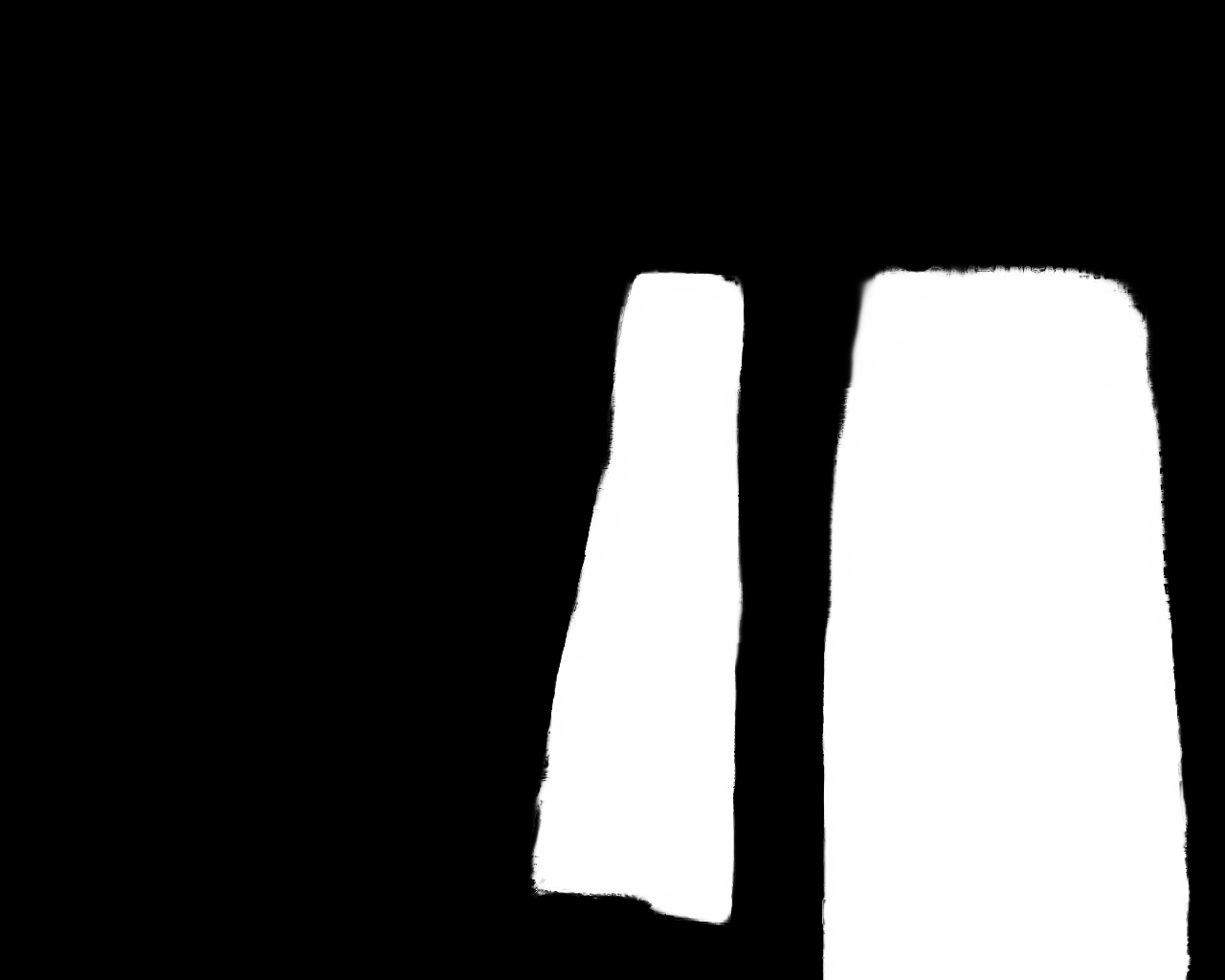}
    
                \includegraphics[width=1\linewidth]{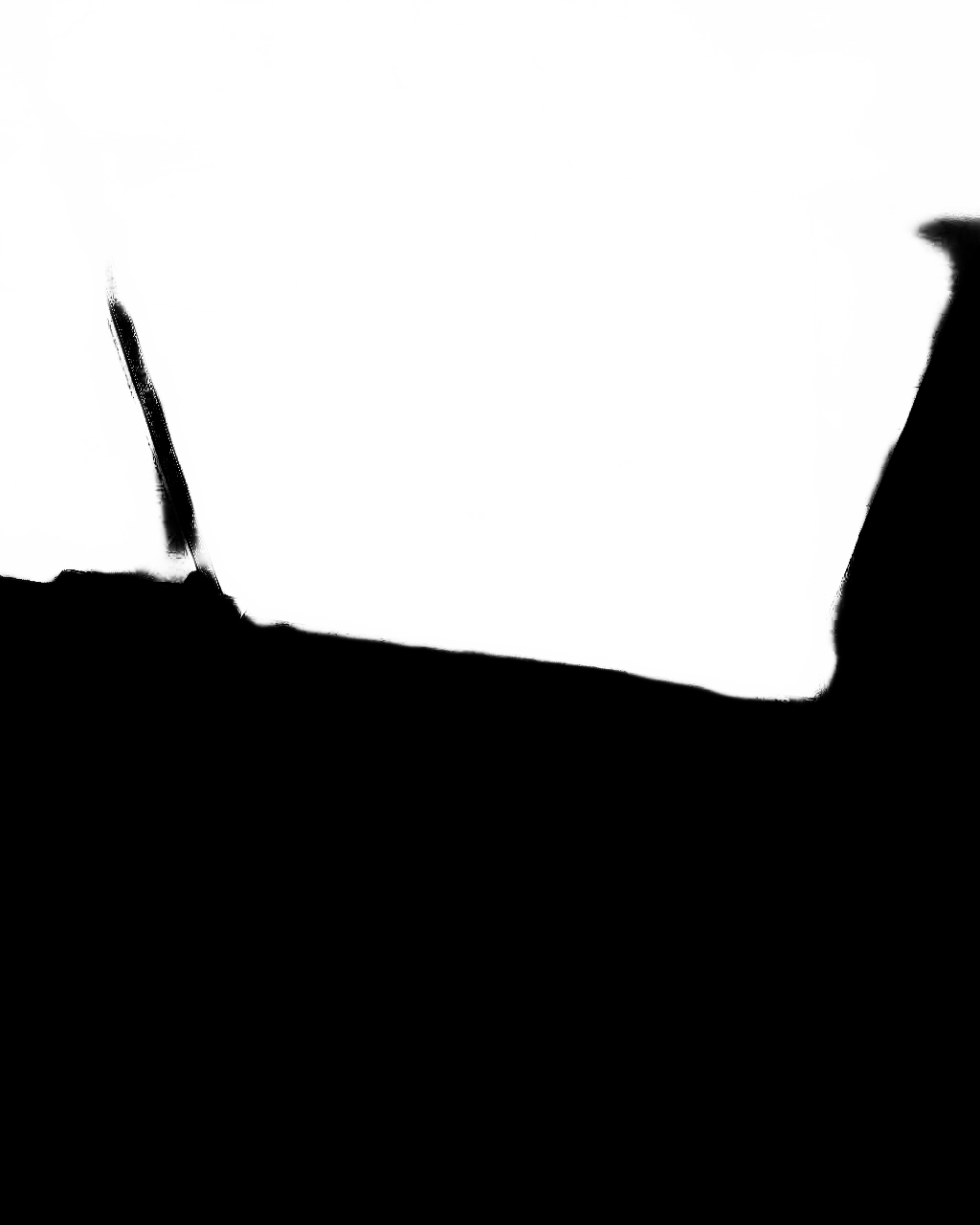}
                
                \includegraphics[width=1\linewidth]{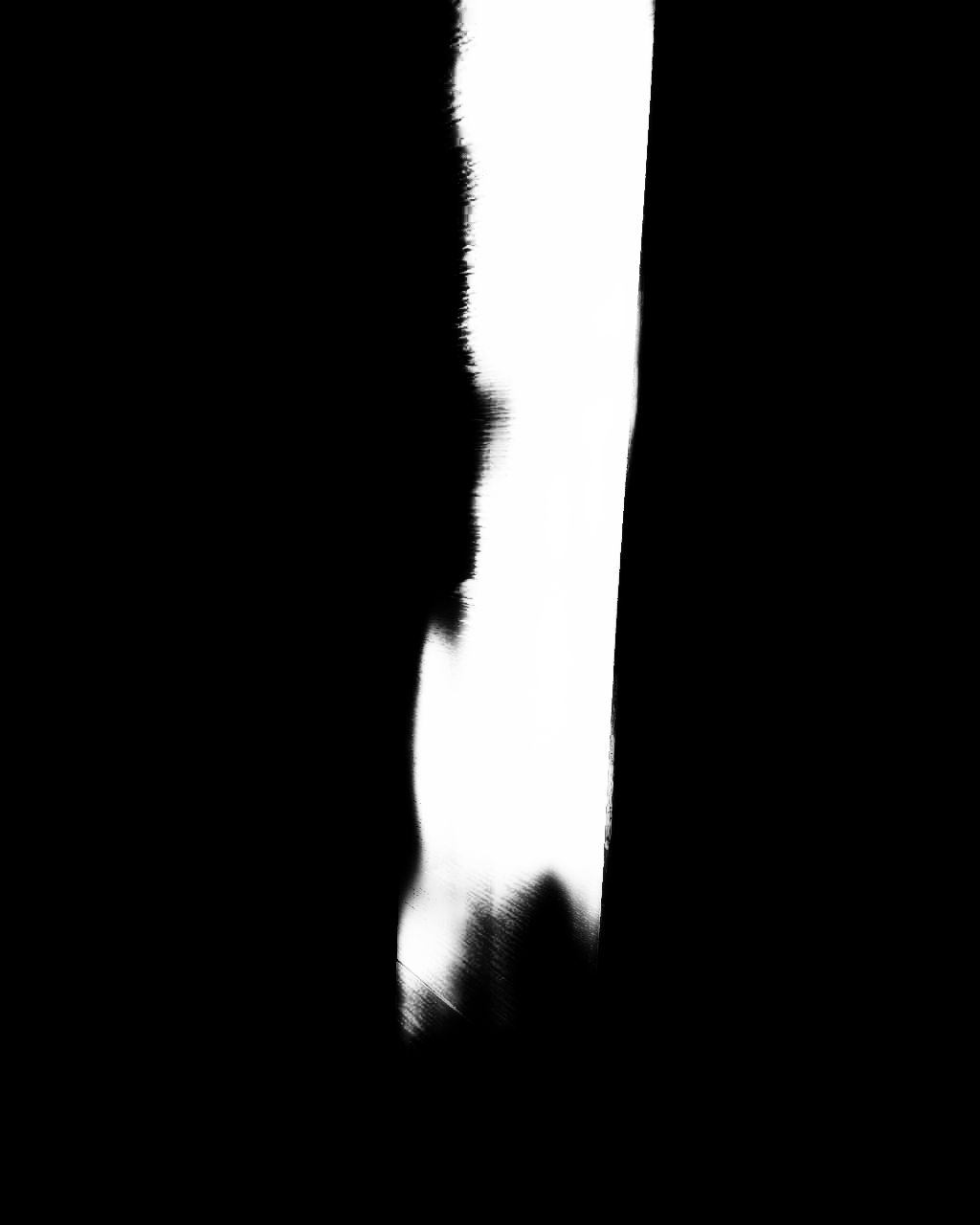}
    
                \includegraphics[width=1\linewidth]{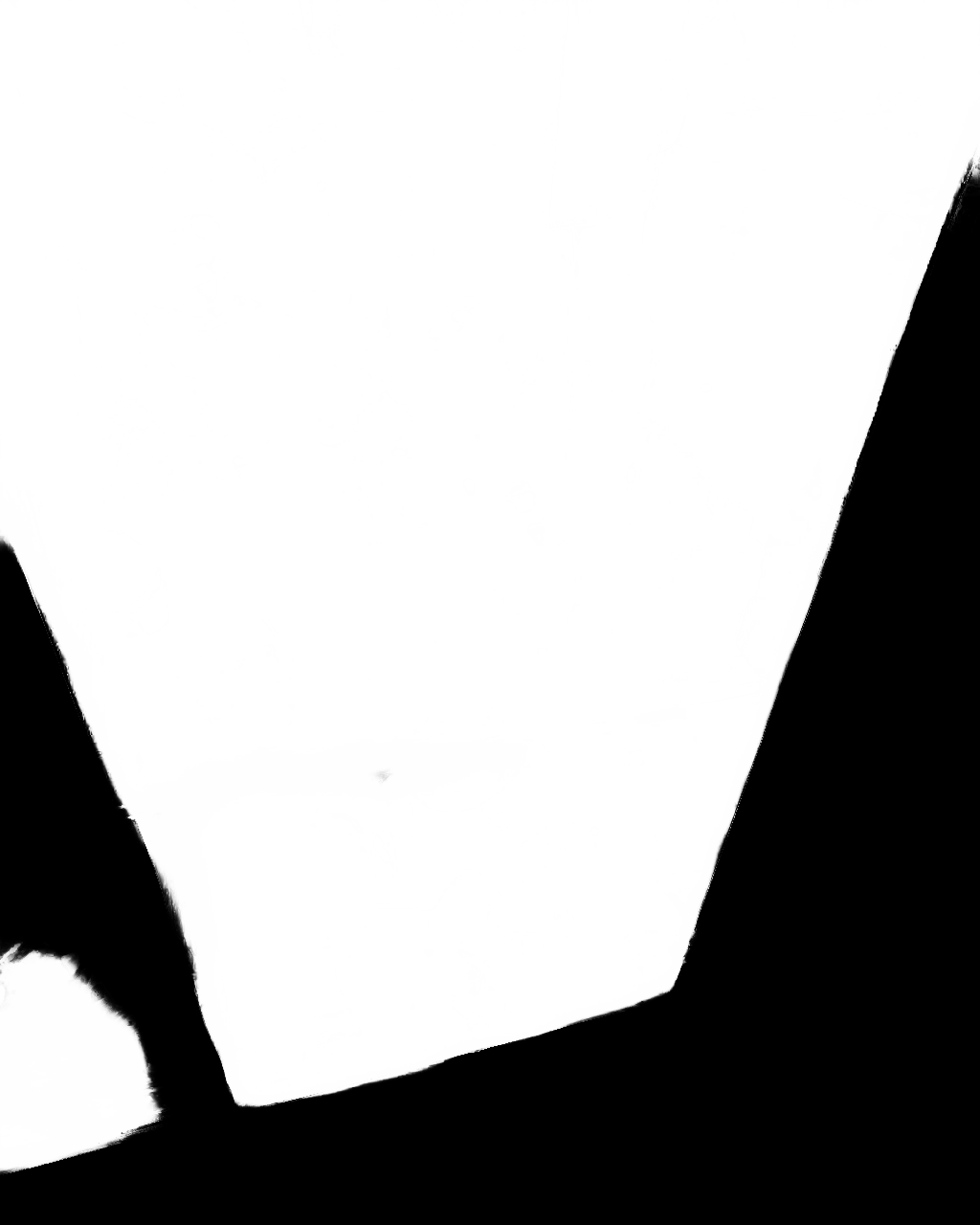}
    
                \includegraphics[width=1\linewidth]{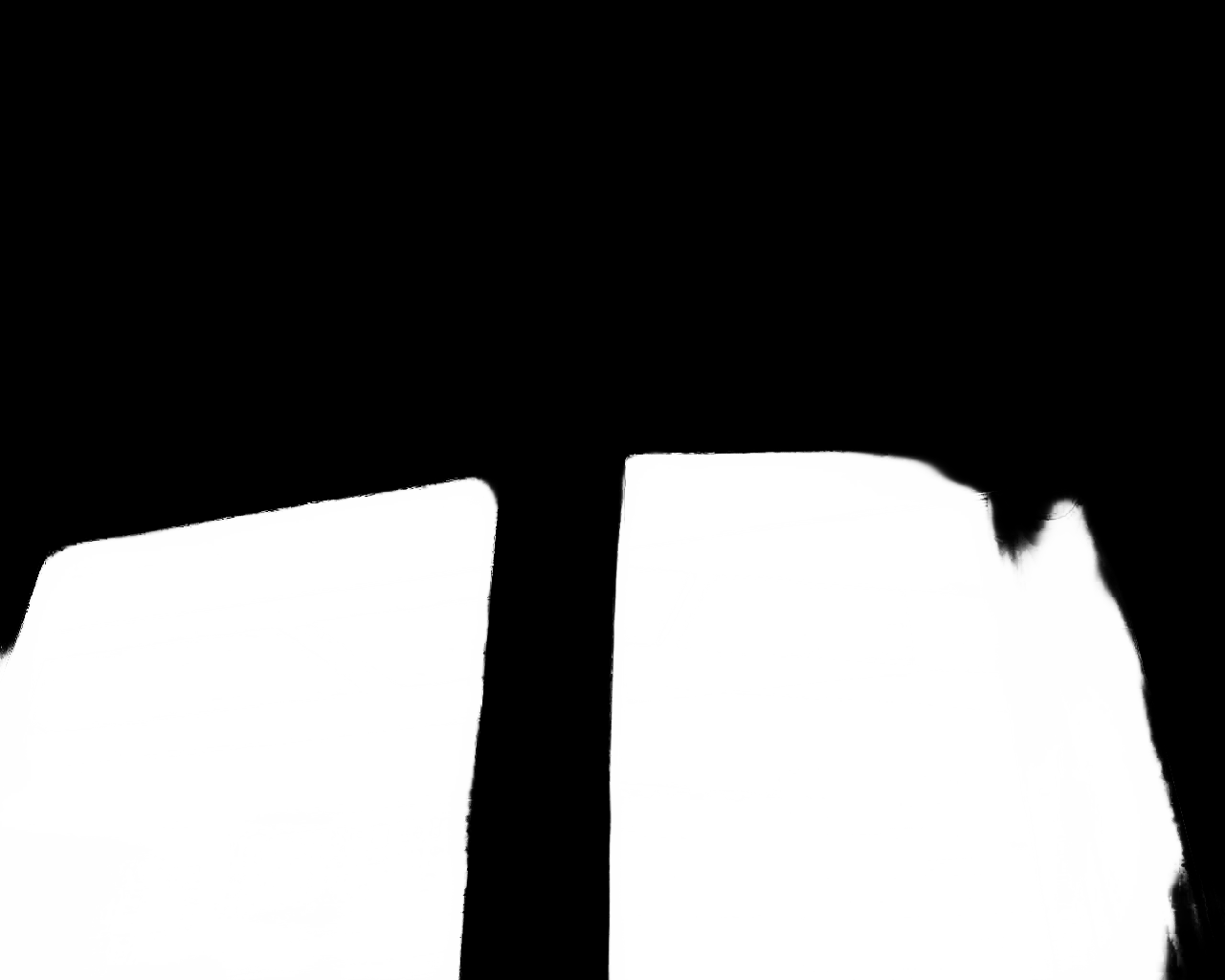}
                
                \includegraphics[width=1\linewidth]{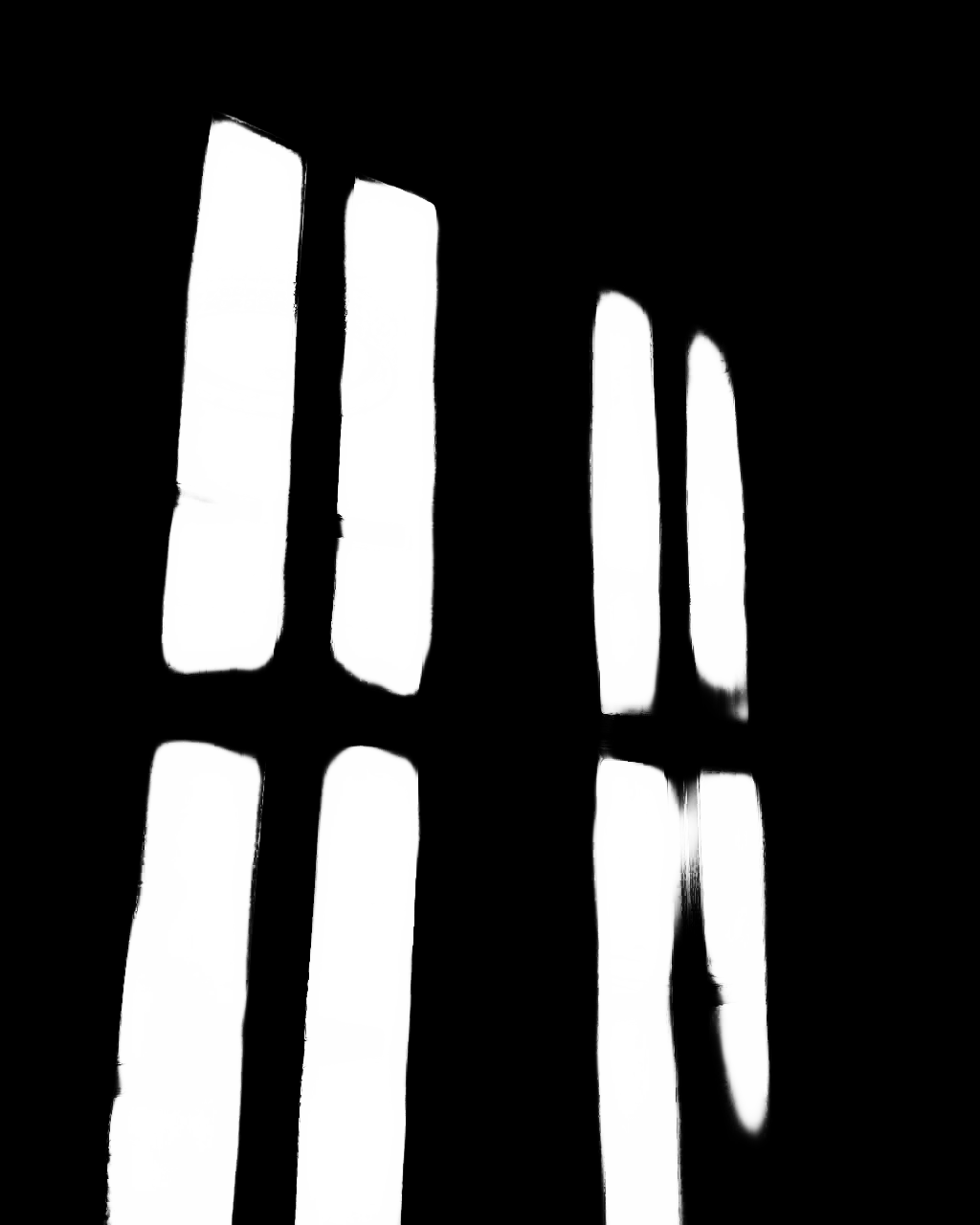}

                \includegraphics[width=1\linewidth]{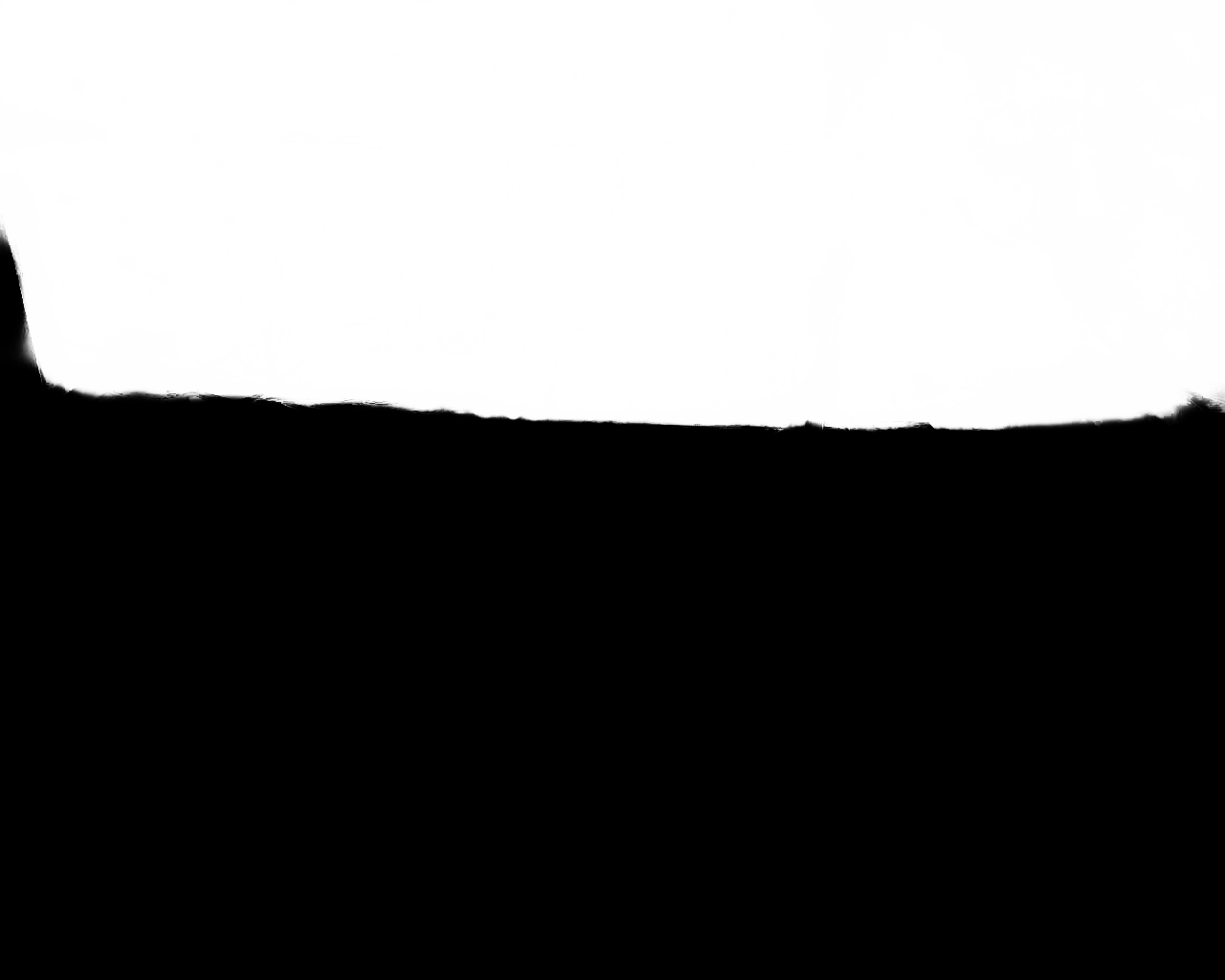}
    
                \includegraphics[width=1\linewidth]{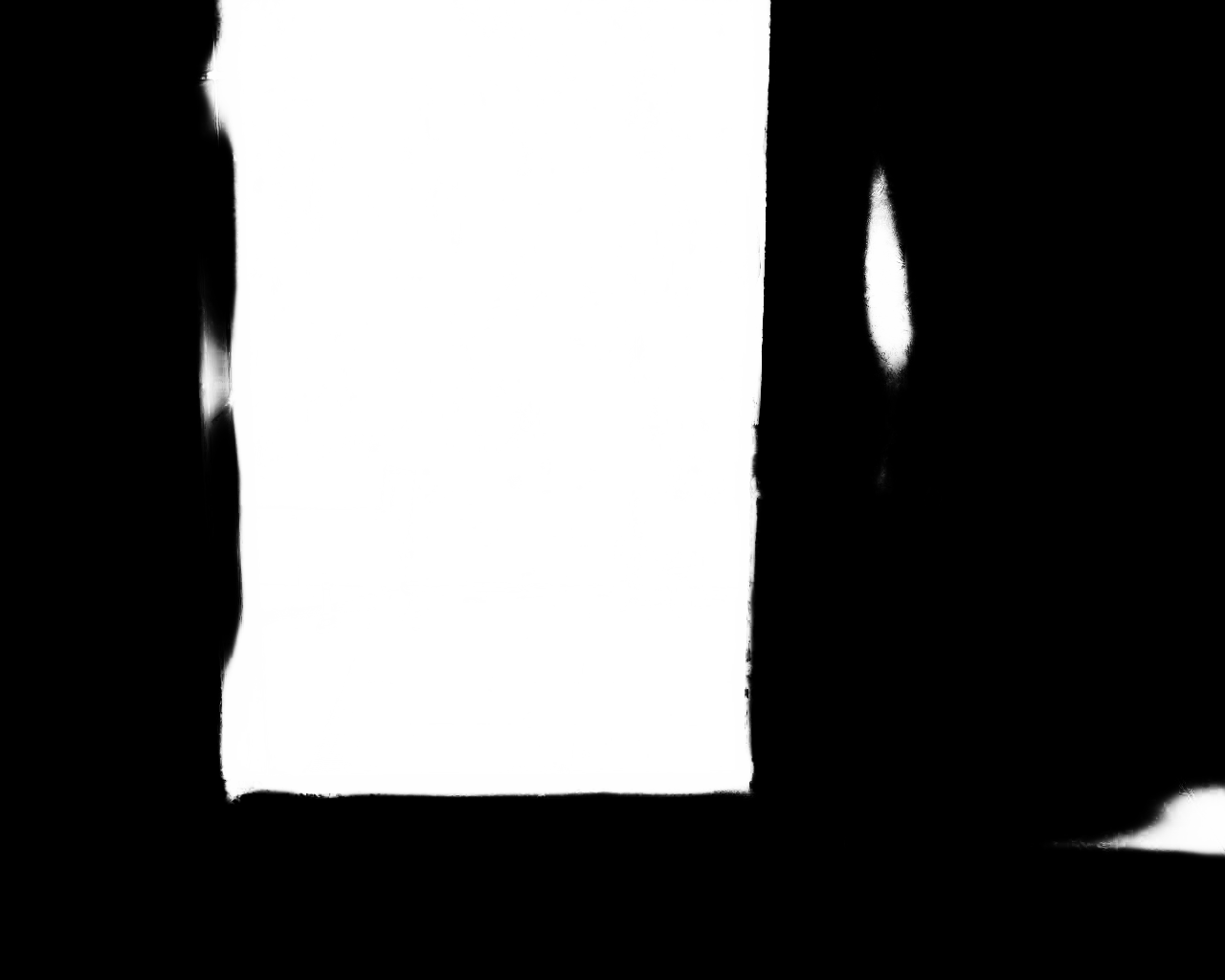}
                
                \includegraphics[width=1\linewidth]{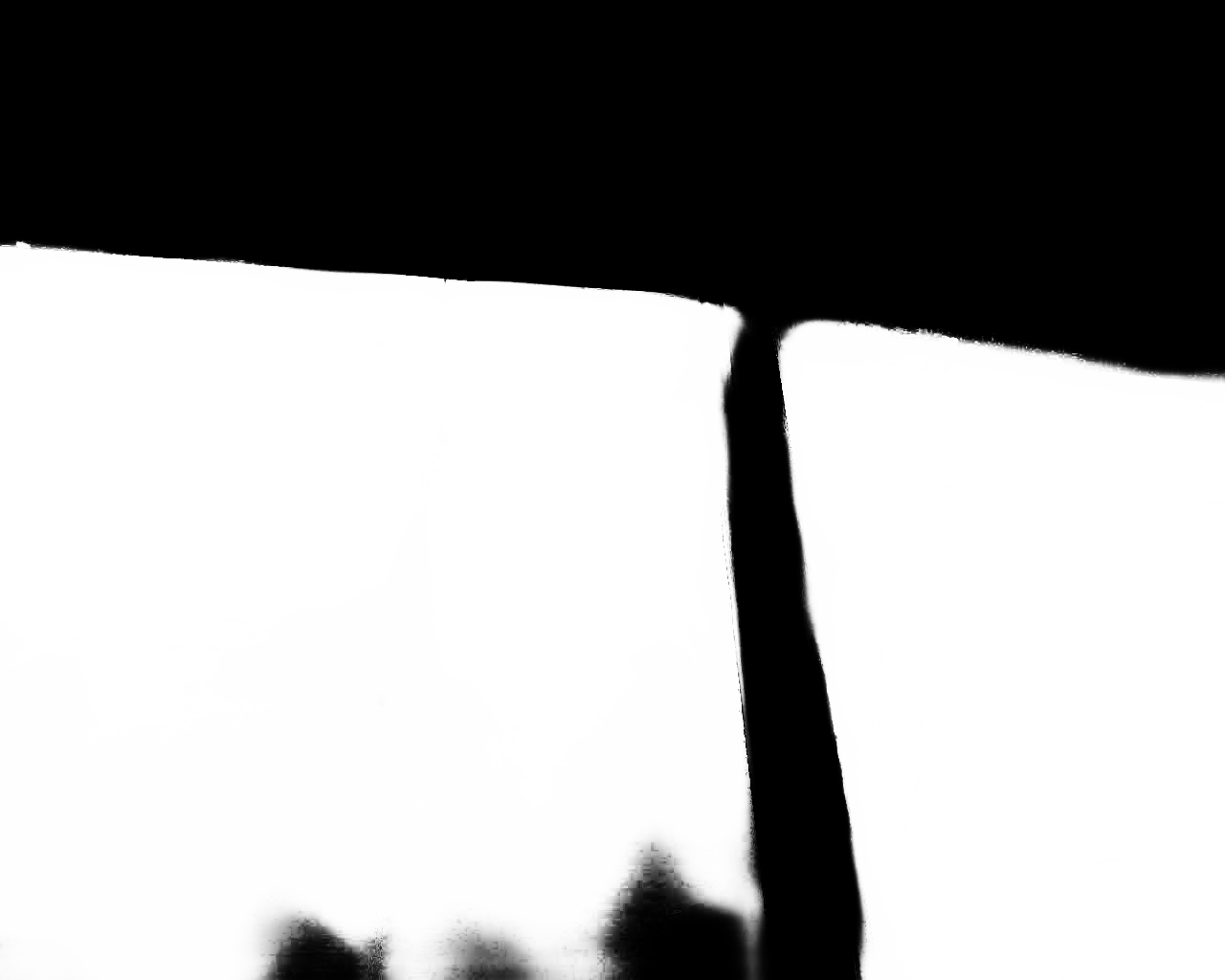}
    
                \includegraphics[width=1\linewidth]{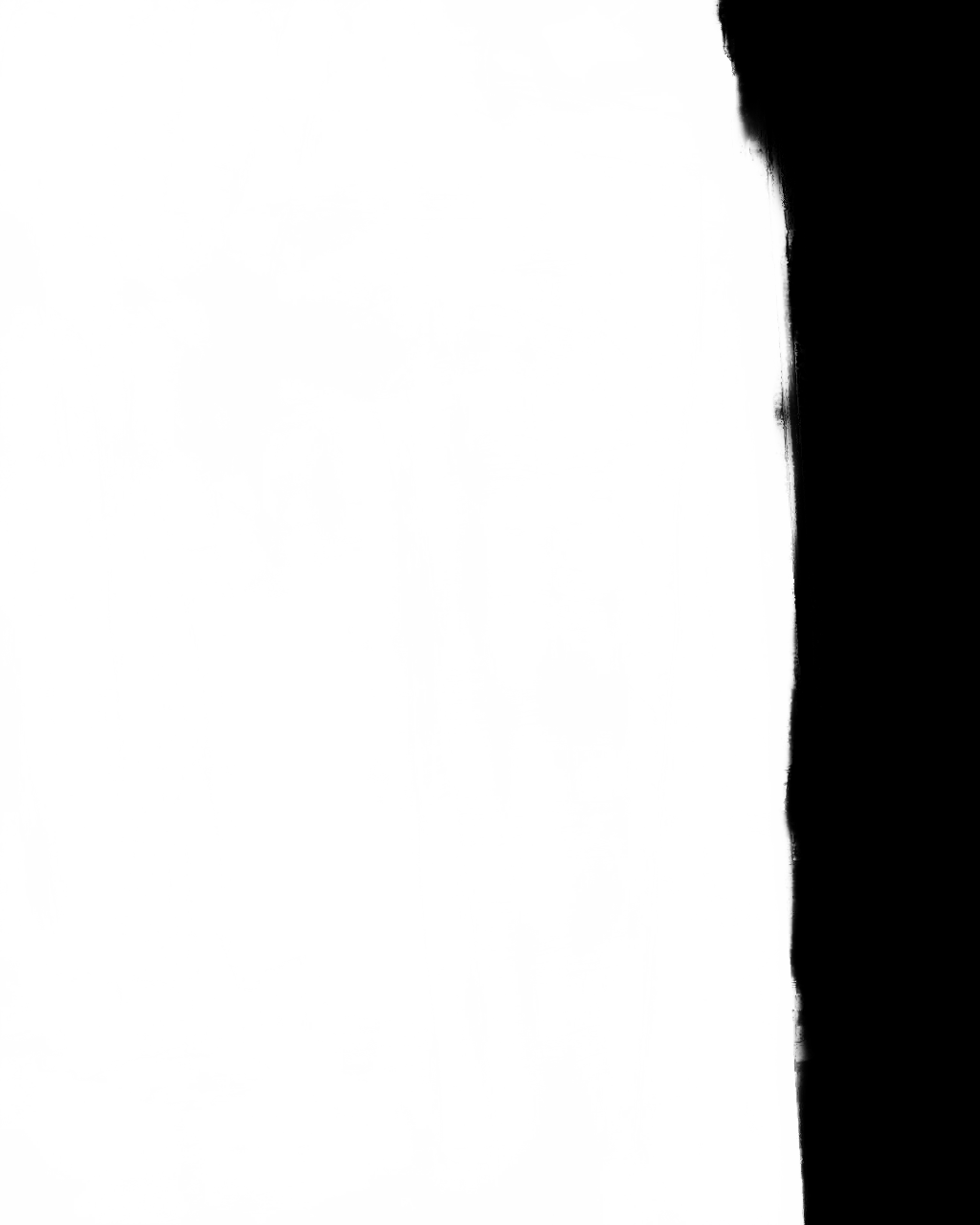}
    
                \includegraphics[width=1\linewidth]{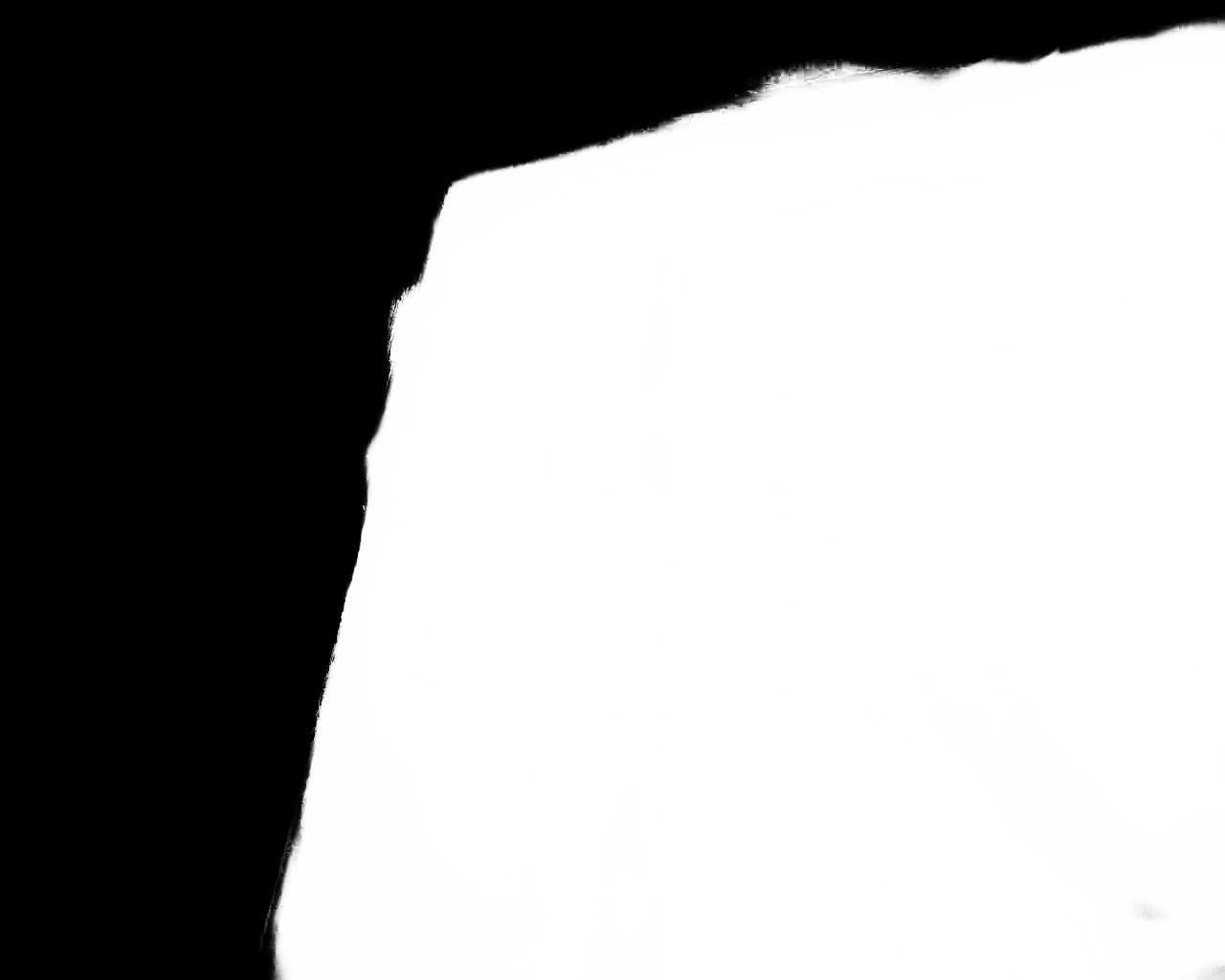}

          \end{minipage}
          }  
          \subfloat[SINet]{
          \begin{minipage}[t]{0.08\textwidth}
                \centering

                \includegraphics[width=1\linewidth]{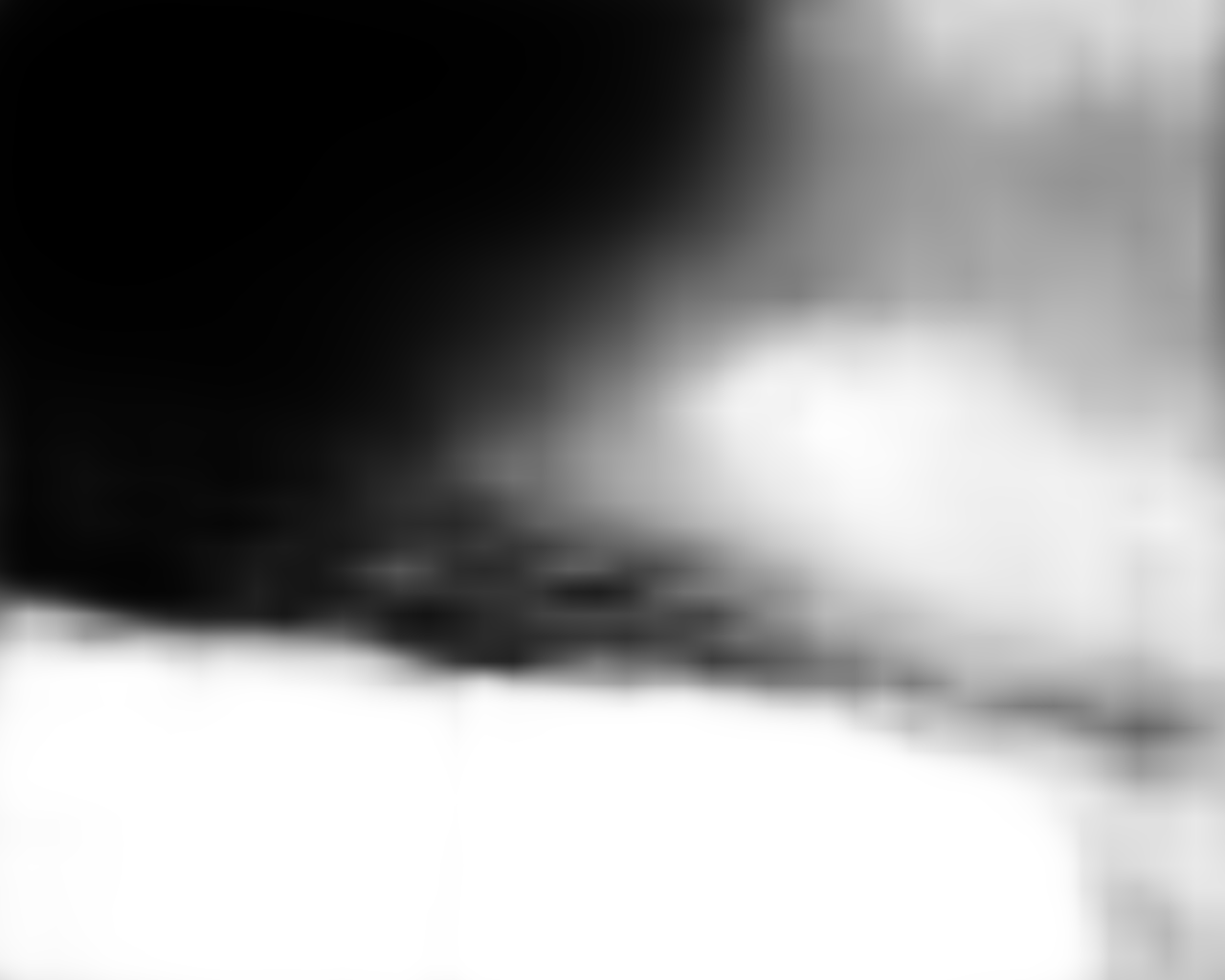}
                
                \includegraphics[width=1\linewidth]{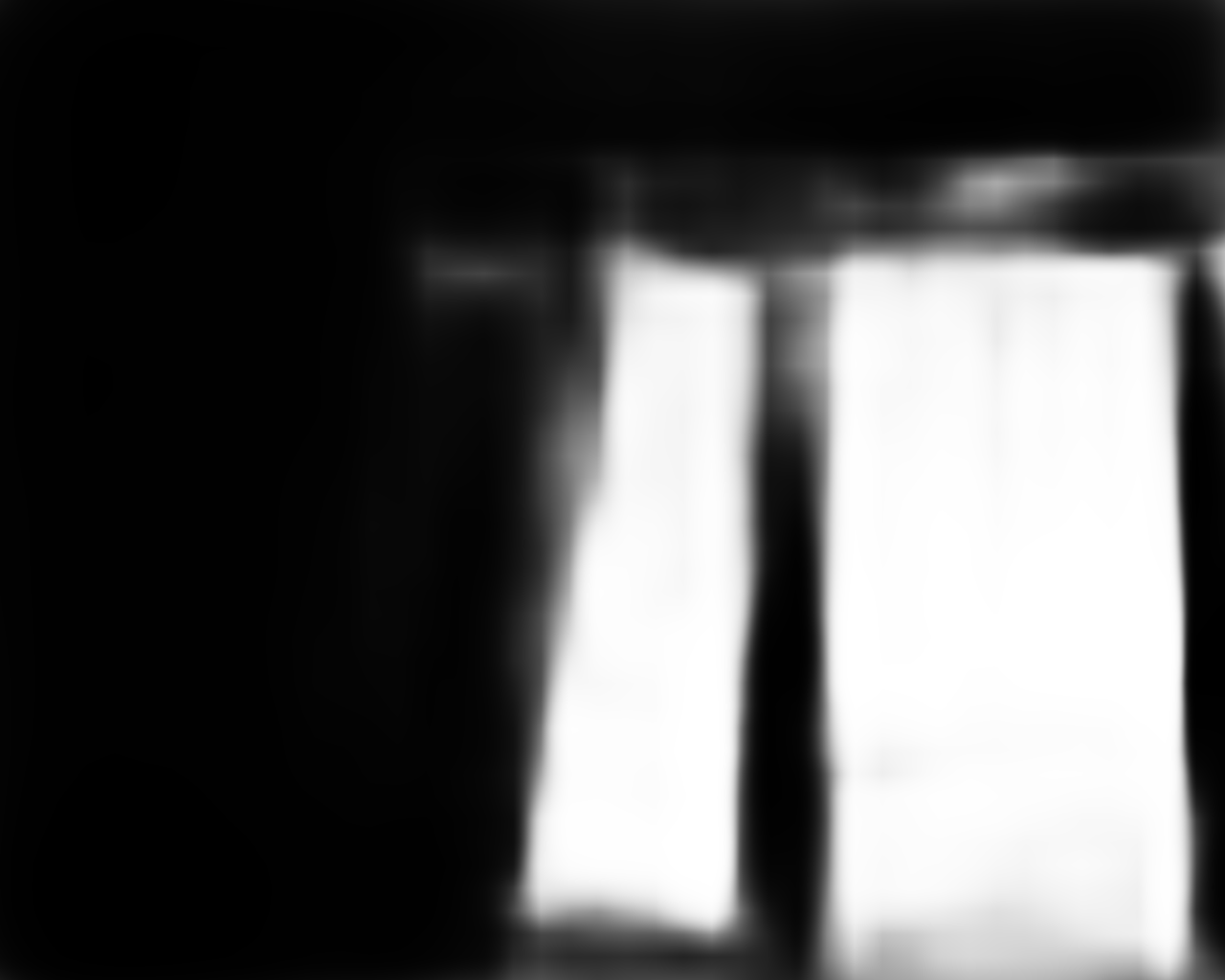}
    
                \includegraphics[width=1\linewidth]{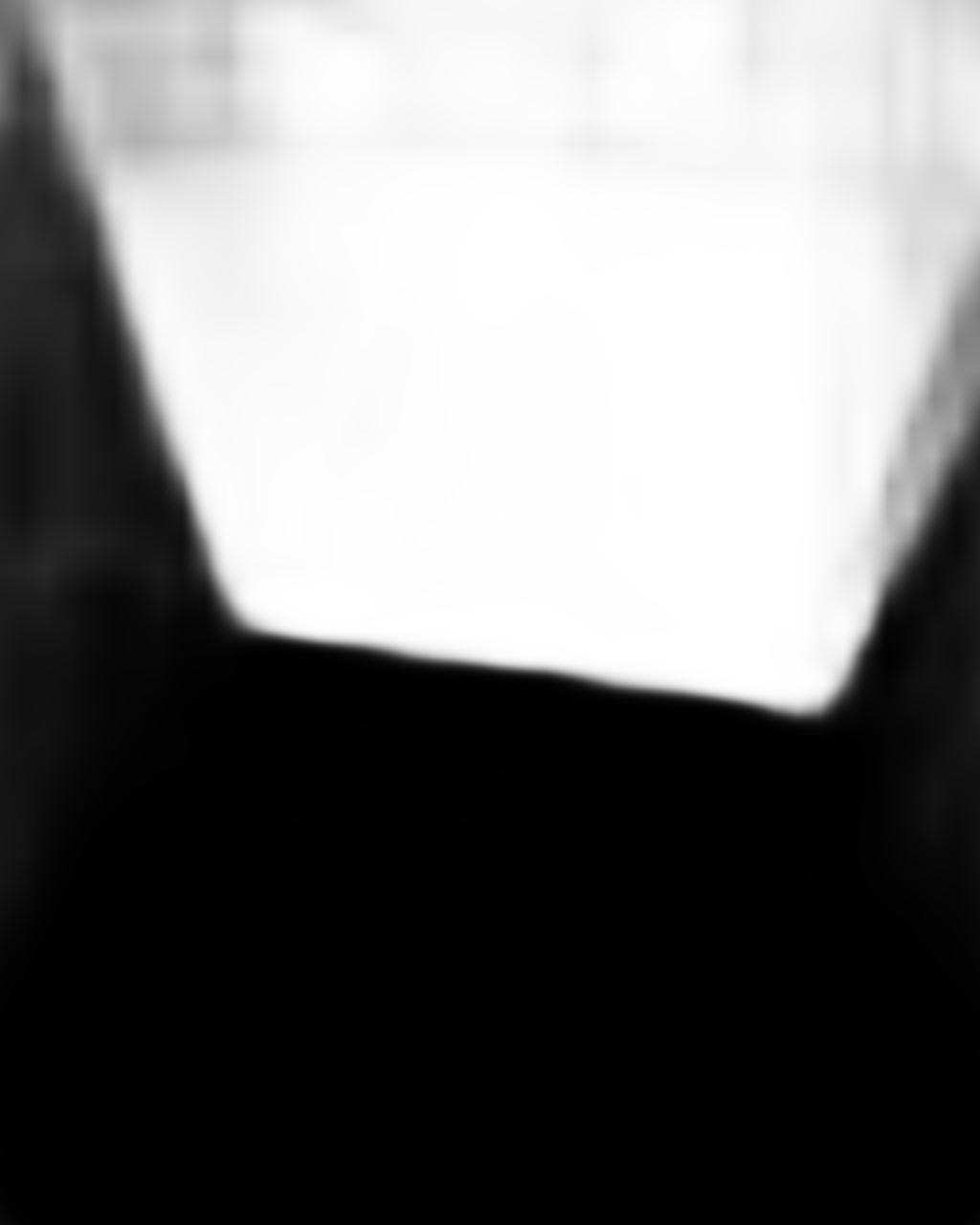}
                
                \includegraphics[width=1\linewidth]{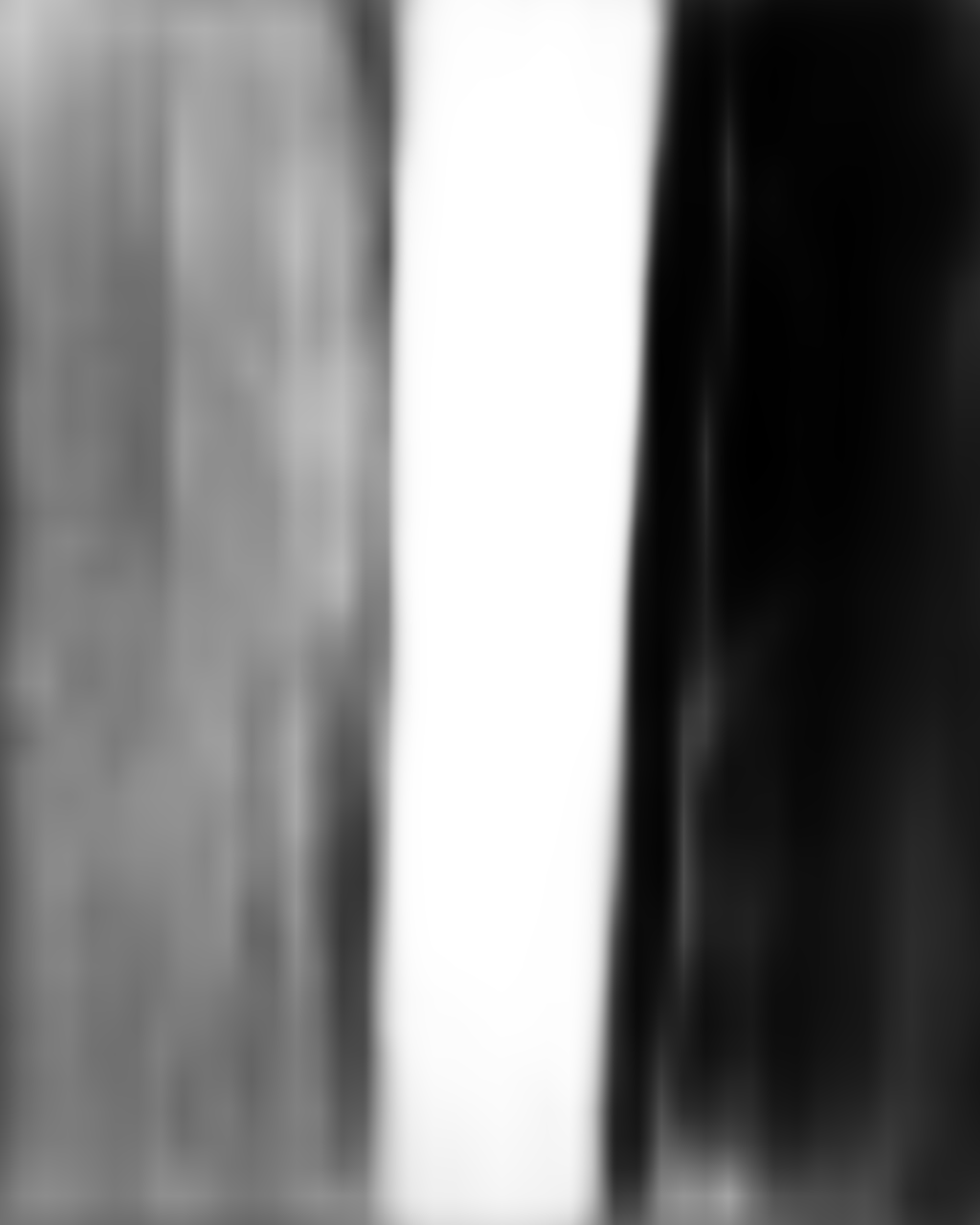}
    
                \includegraphics[width=1\linewidth]{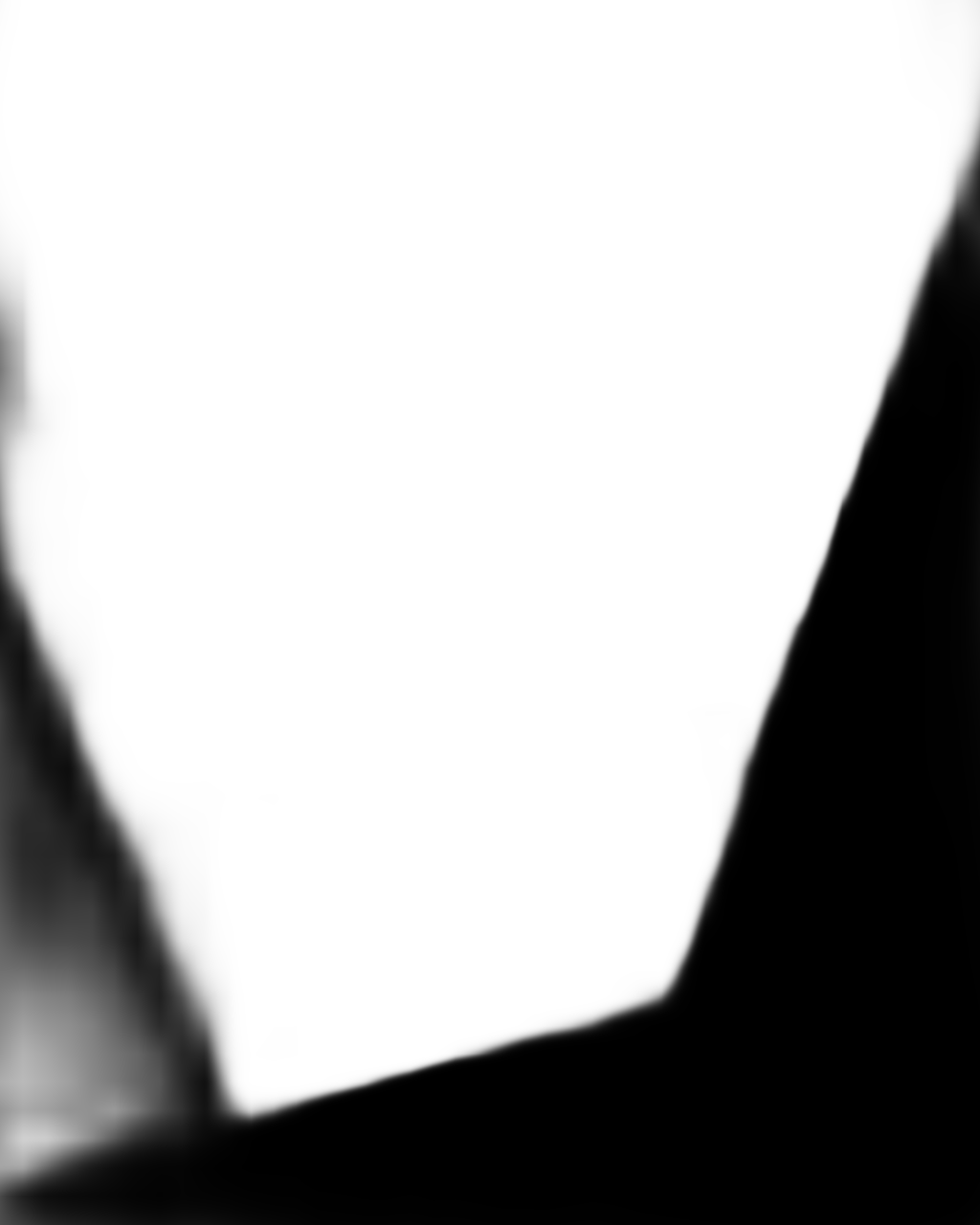}
    
                \includegraphics[width=1\linewidth]{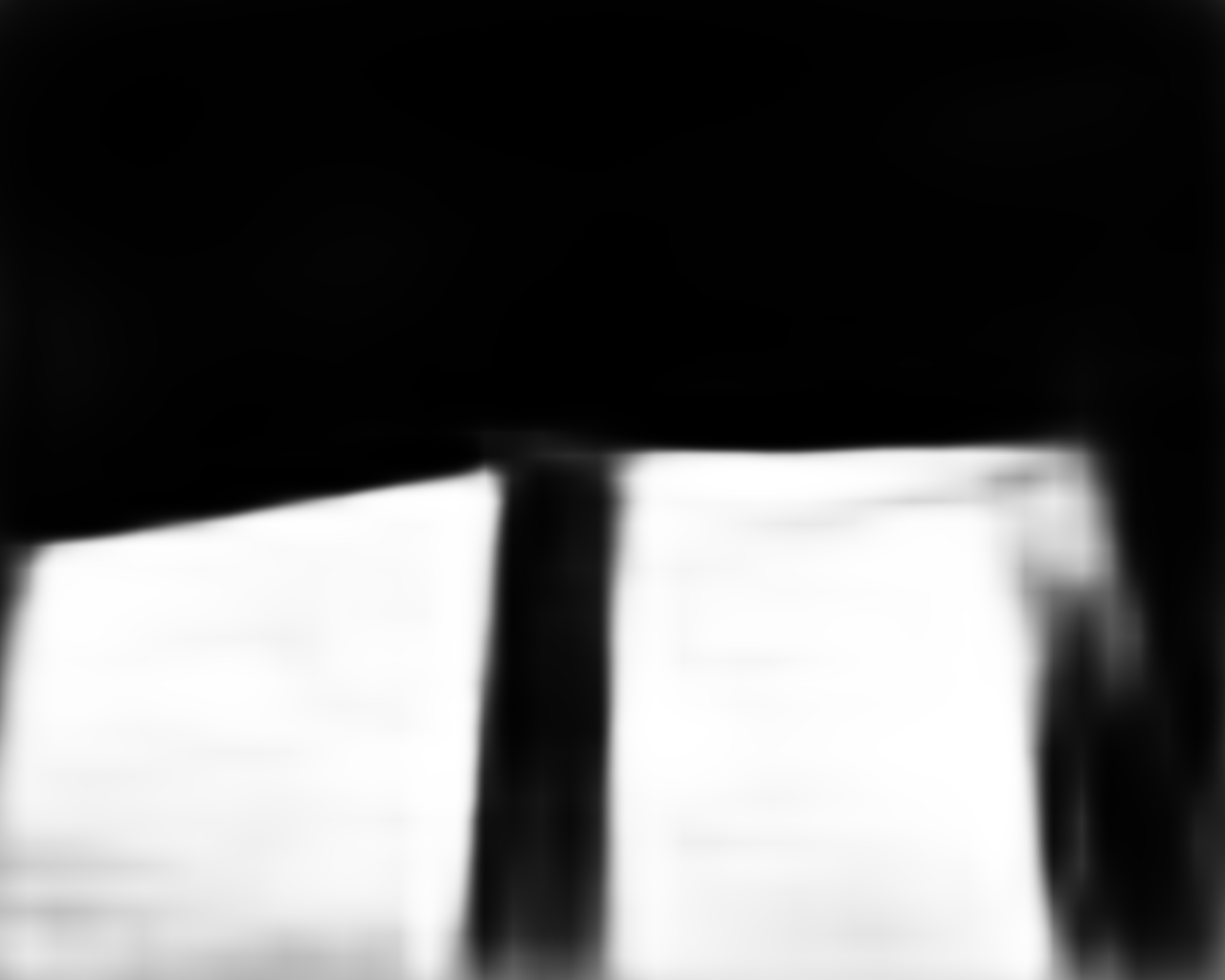}
                
                \includegraphics[width=1\linewidth]{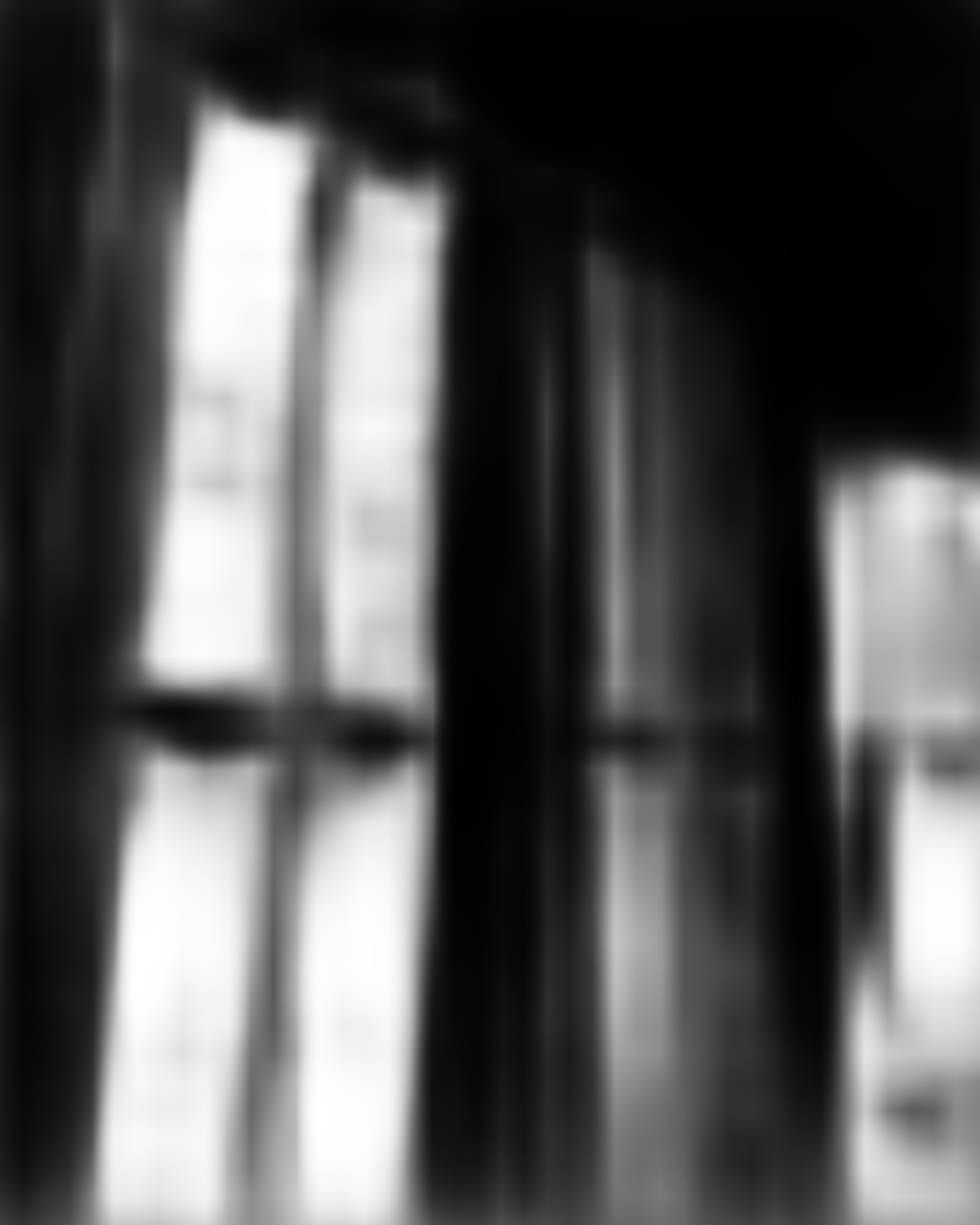}

                \includegraphics[width=1\linewidth]{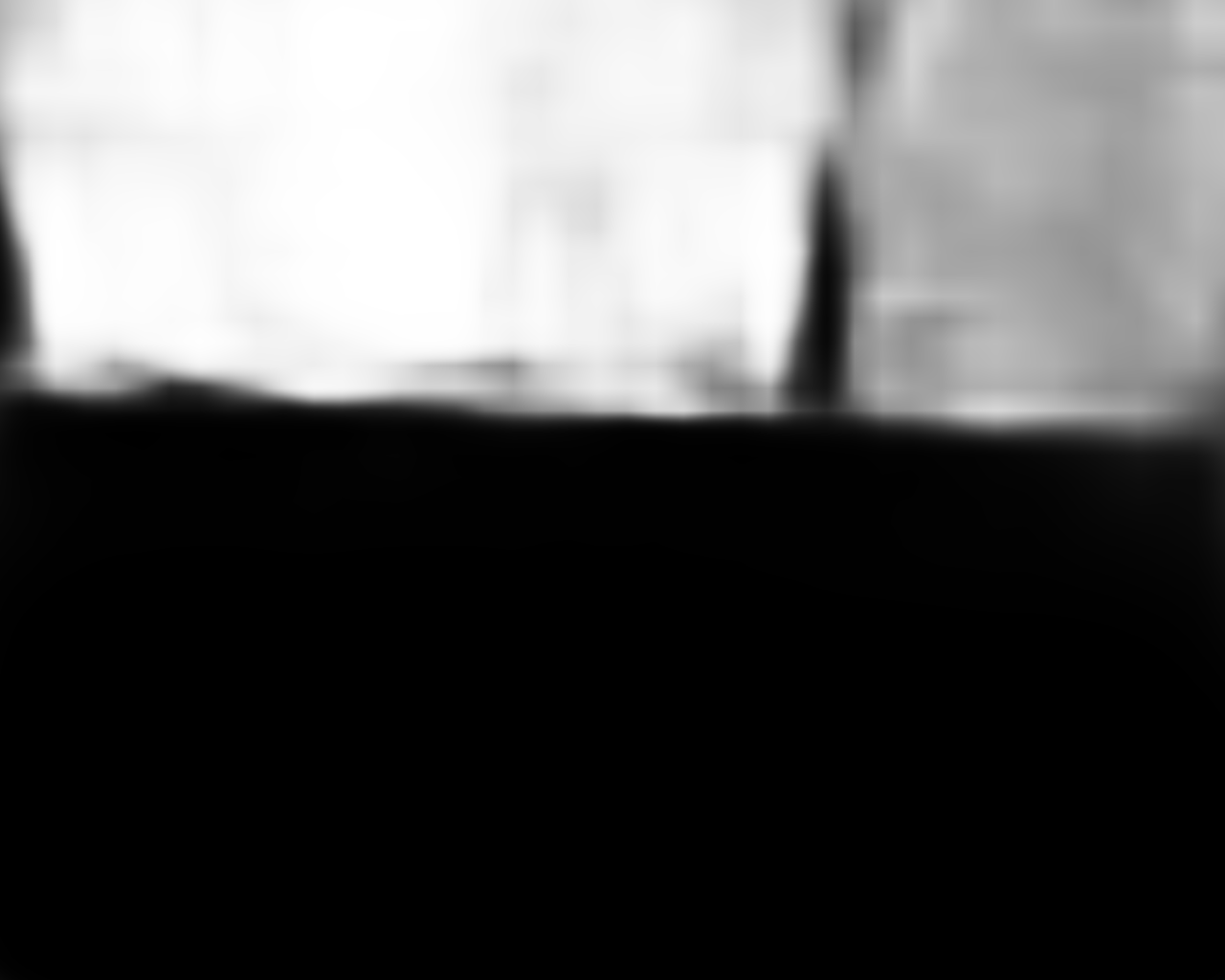}
    
                \includegraphics[width=1\linewidth]{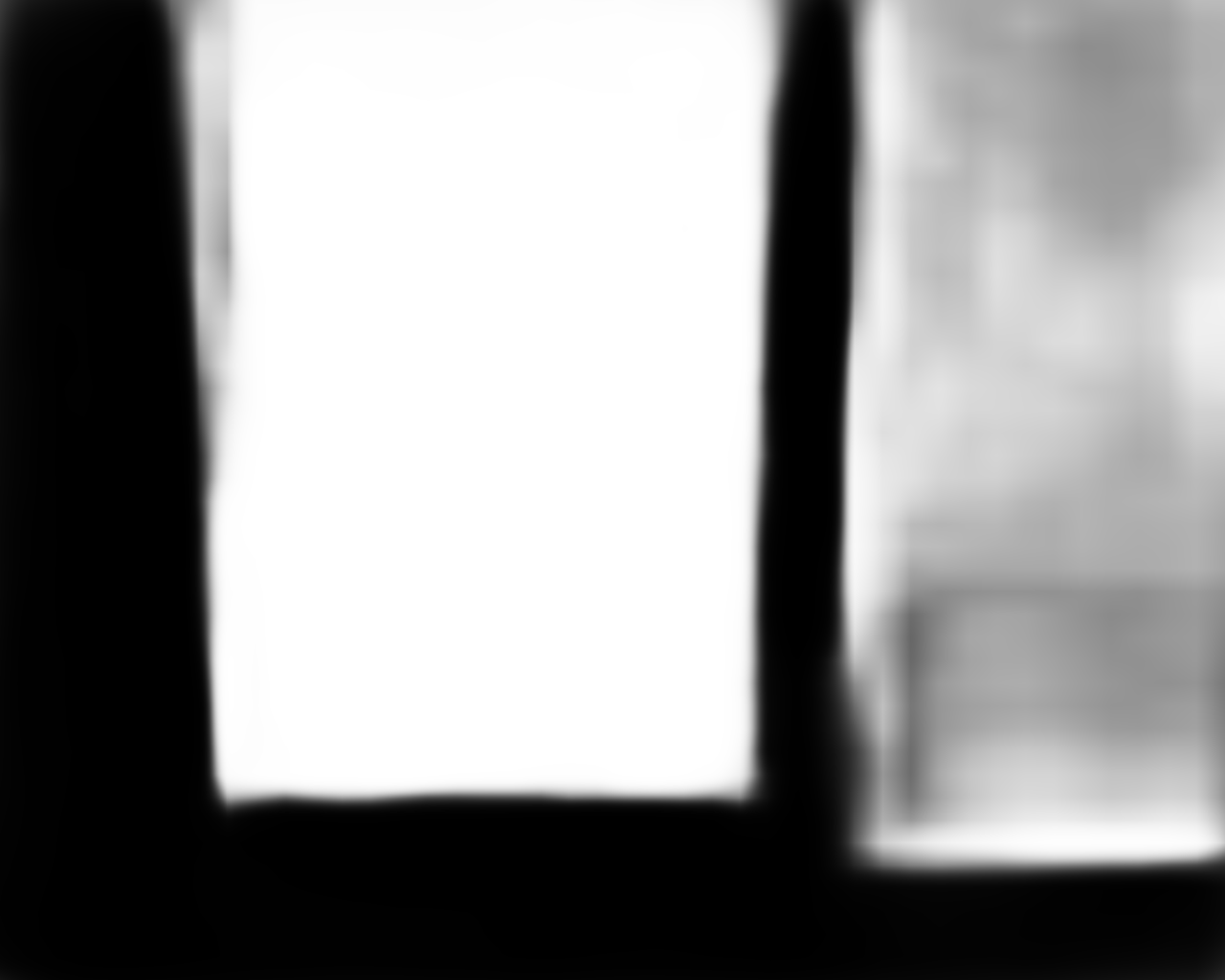}
                
                \includegraphics[width=1\linewidth]{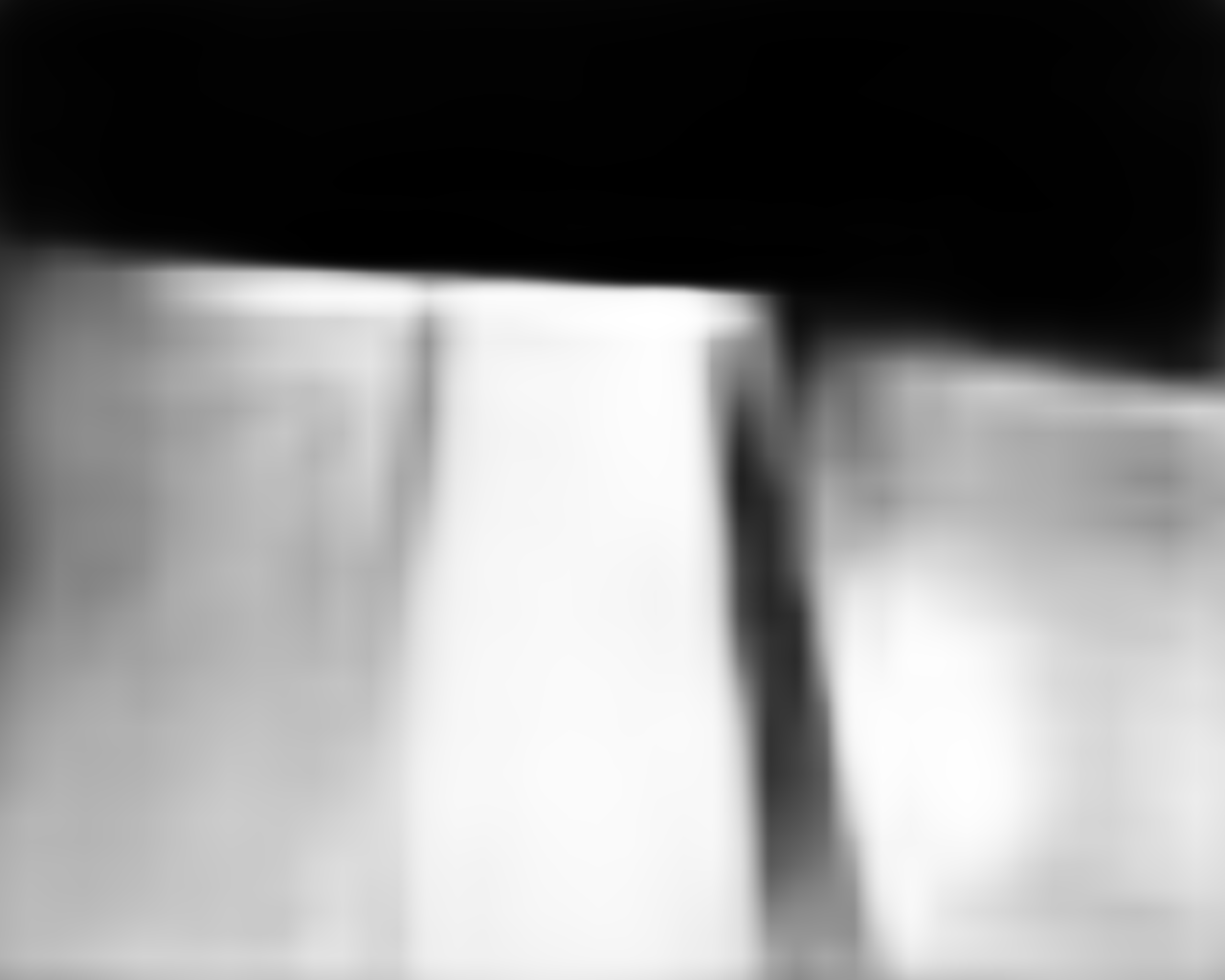}
    
                \includegraphics[width=1\linewidth]{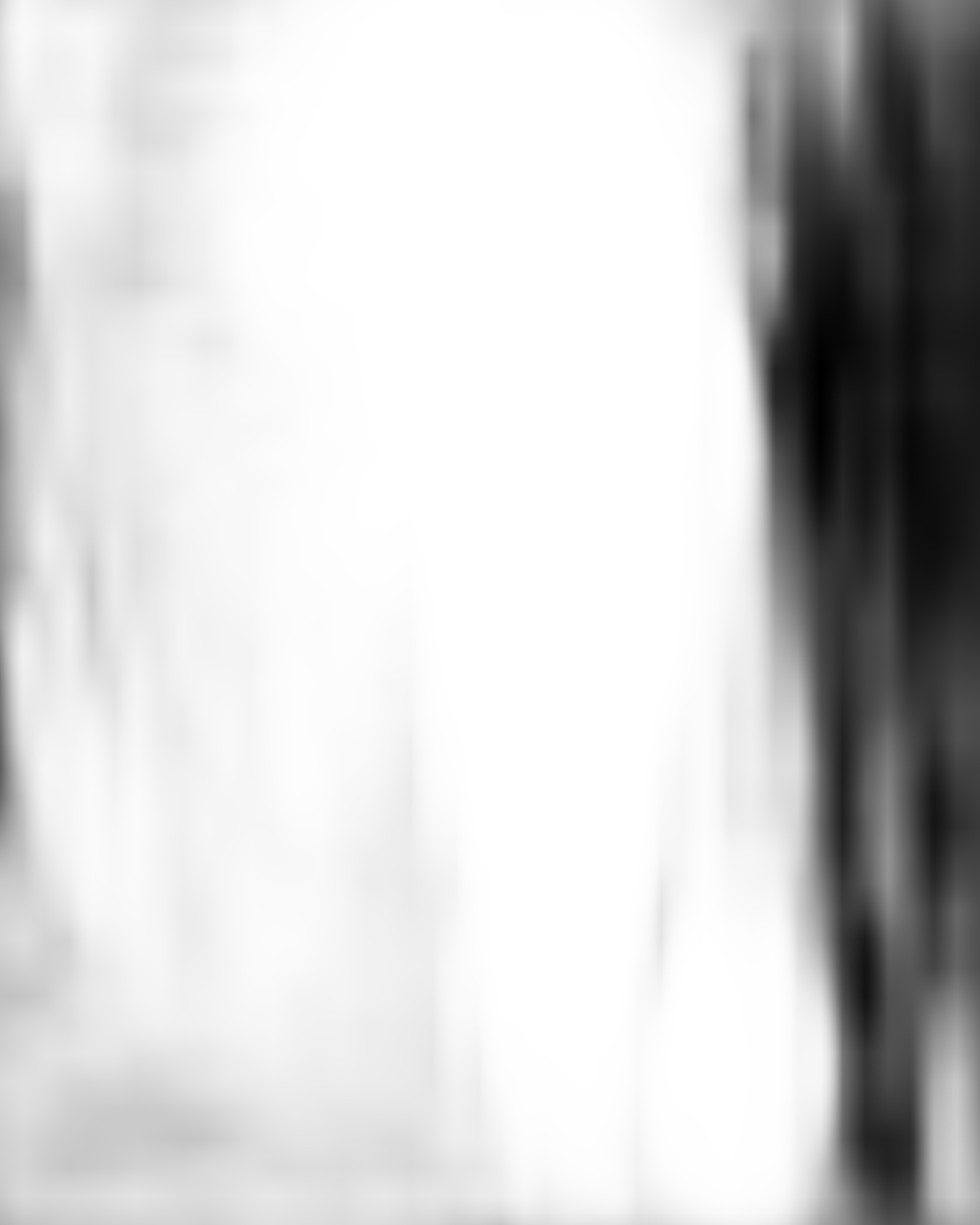}
    
                \includegraphics[width=1\linewidth]{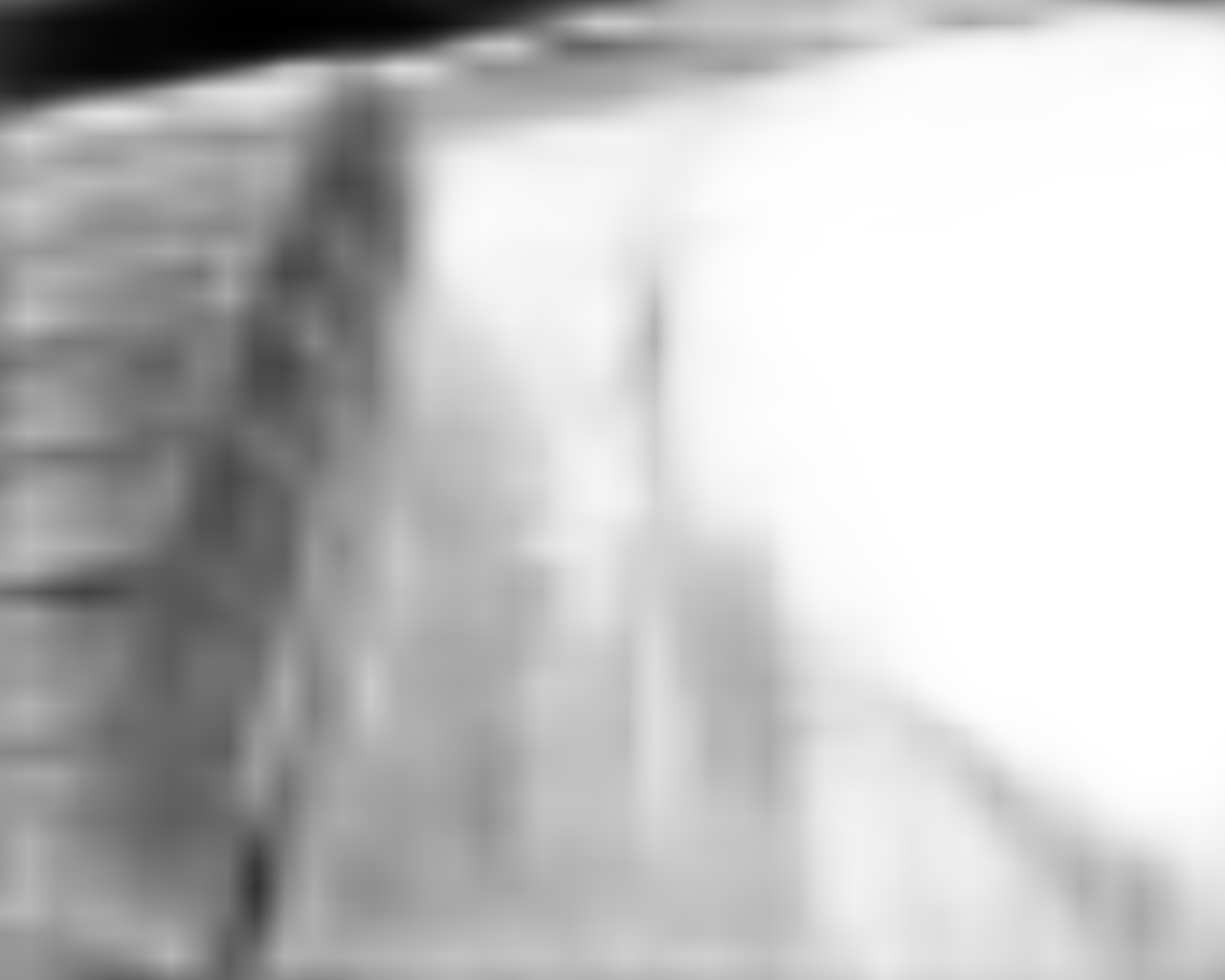}

          \end{minipage}
          }  
          \subfloat[PFNet]{
          \begin{minipage}[t]{0.08\textwidth}
                \centering

                \includegraphics[width=1\linewidth]{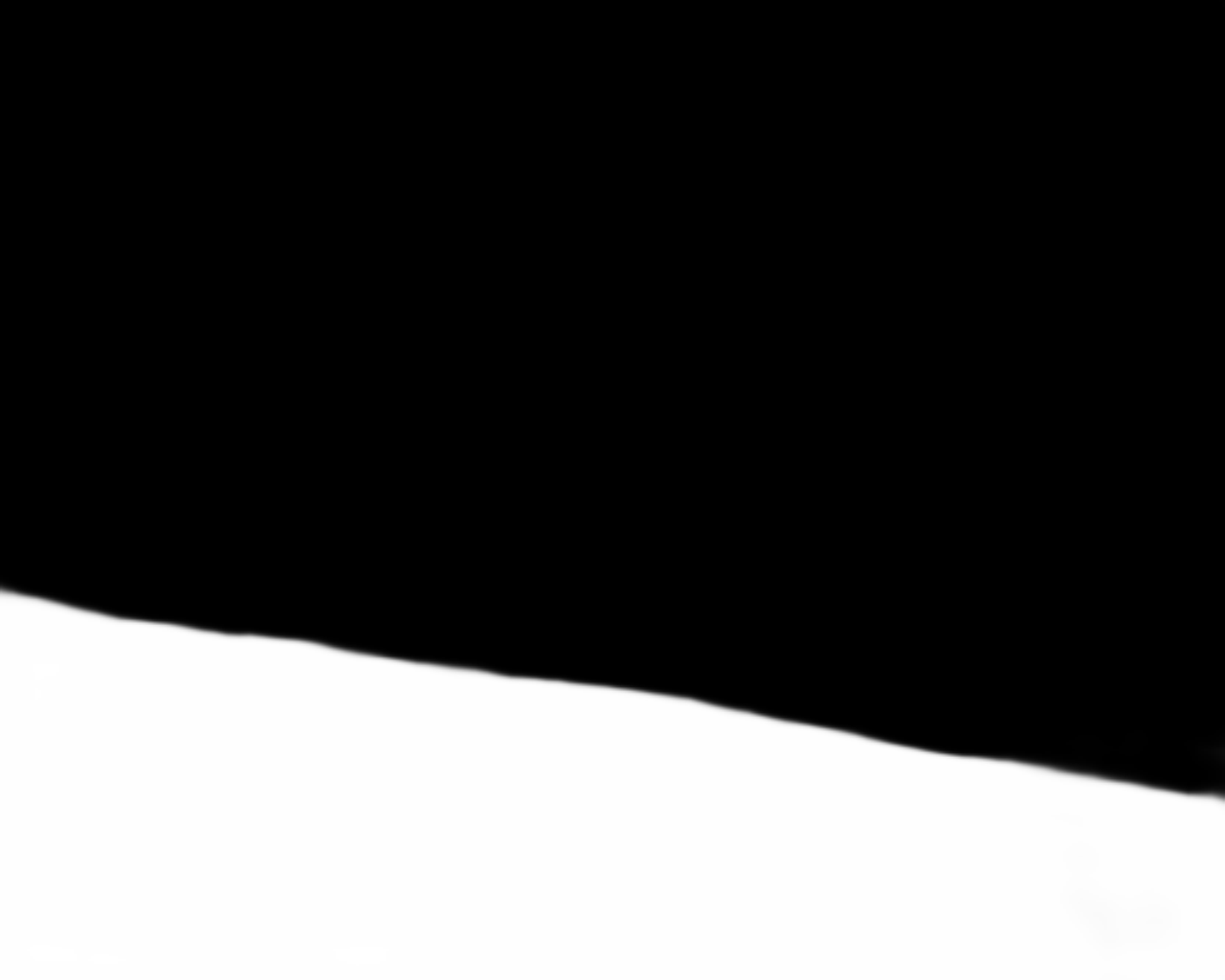}
                
                \includegraphics[width=1\linewidth]{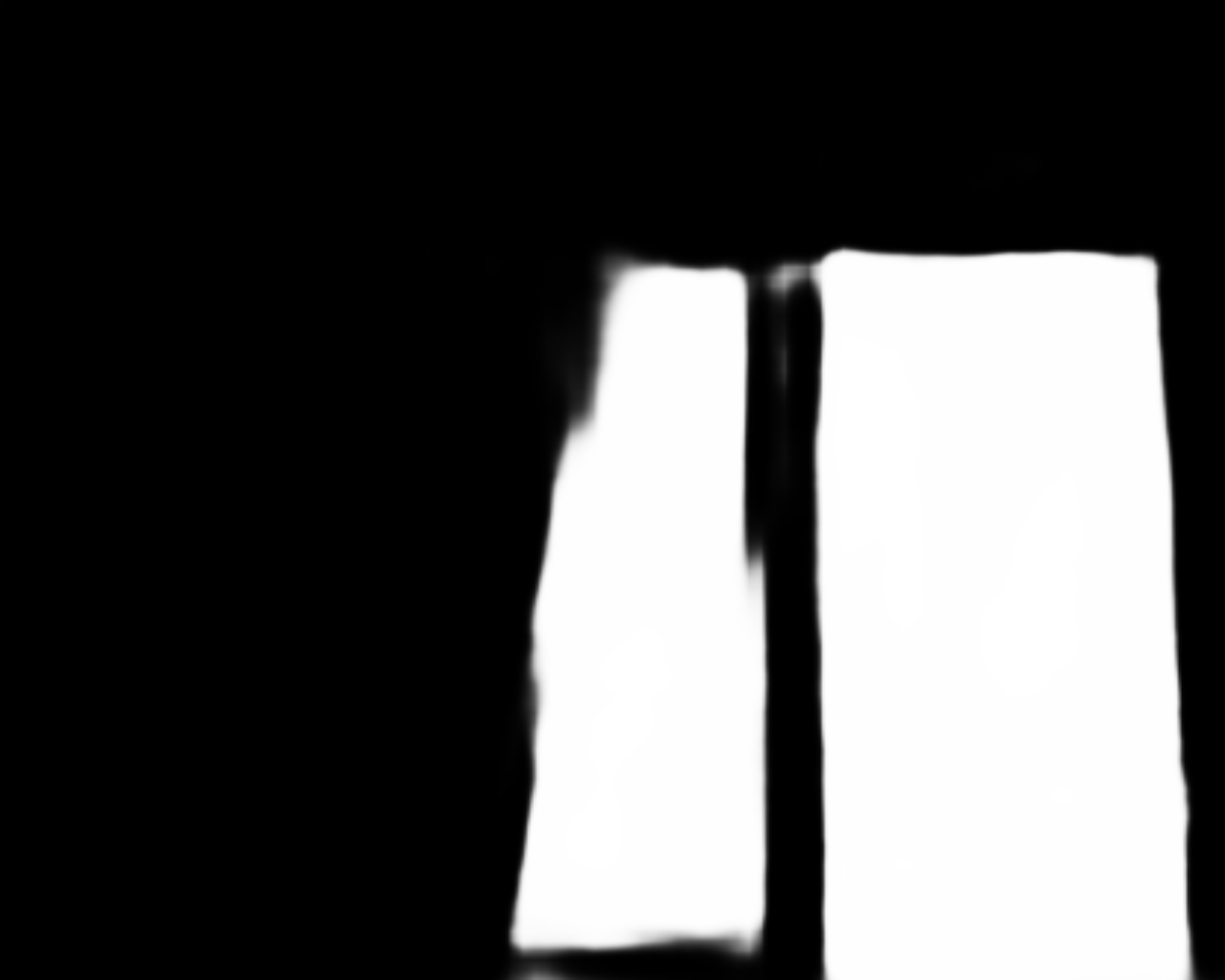}
    
                \includegraphics[width=1\linewidth]{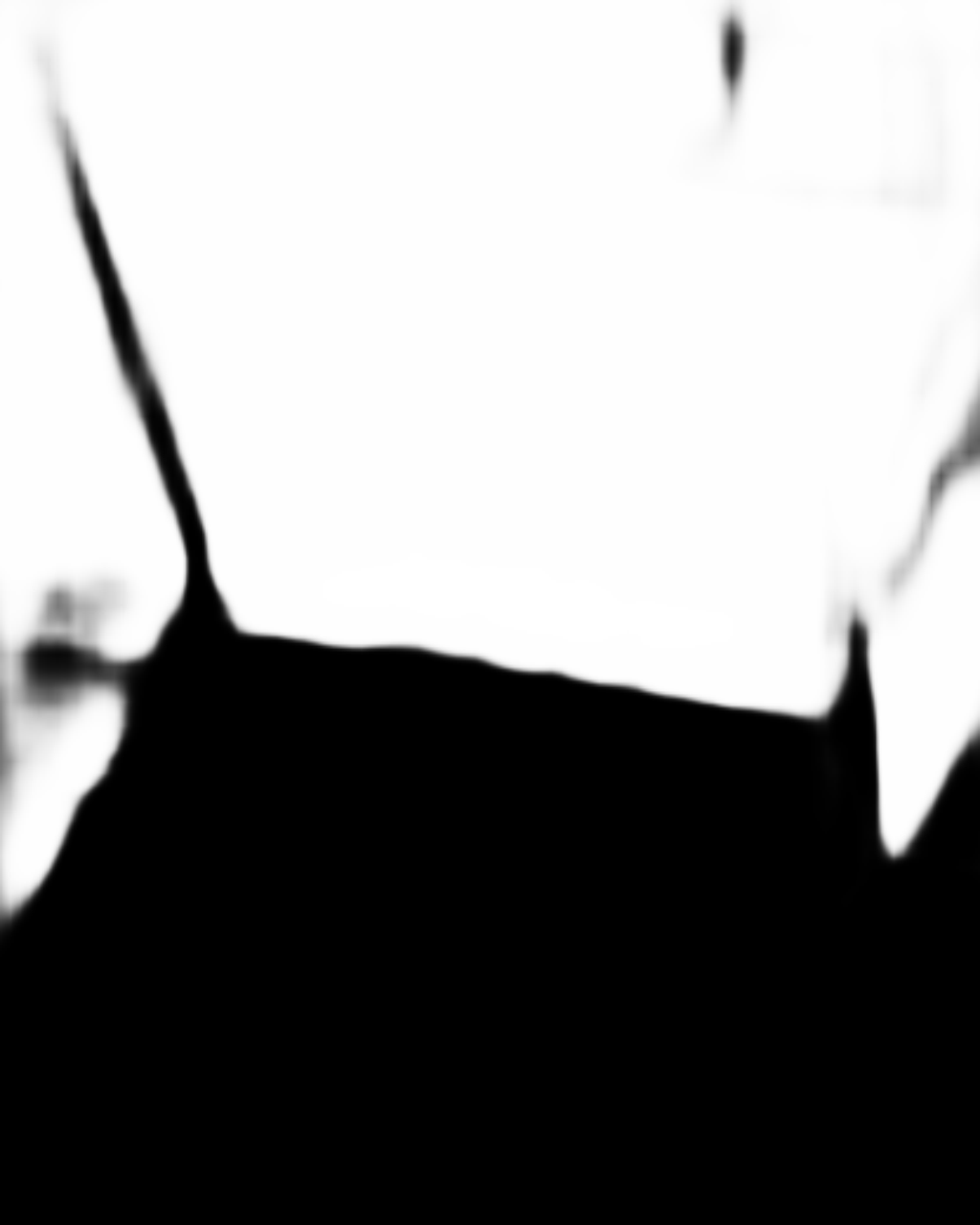}
                
                \includegraphics[width=1\linewidth]{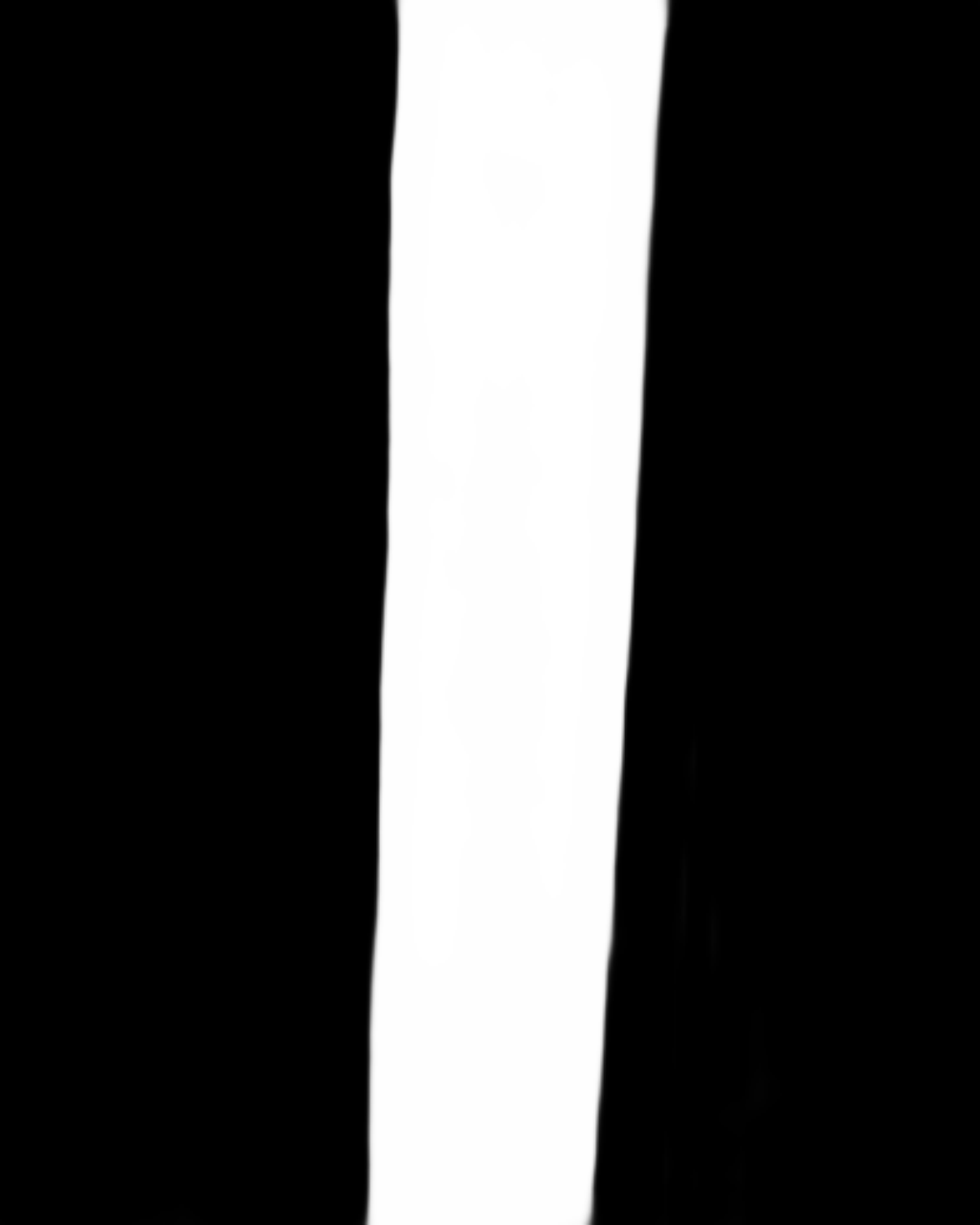}
    
                \includegraphics[width=1\linewidth]{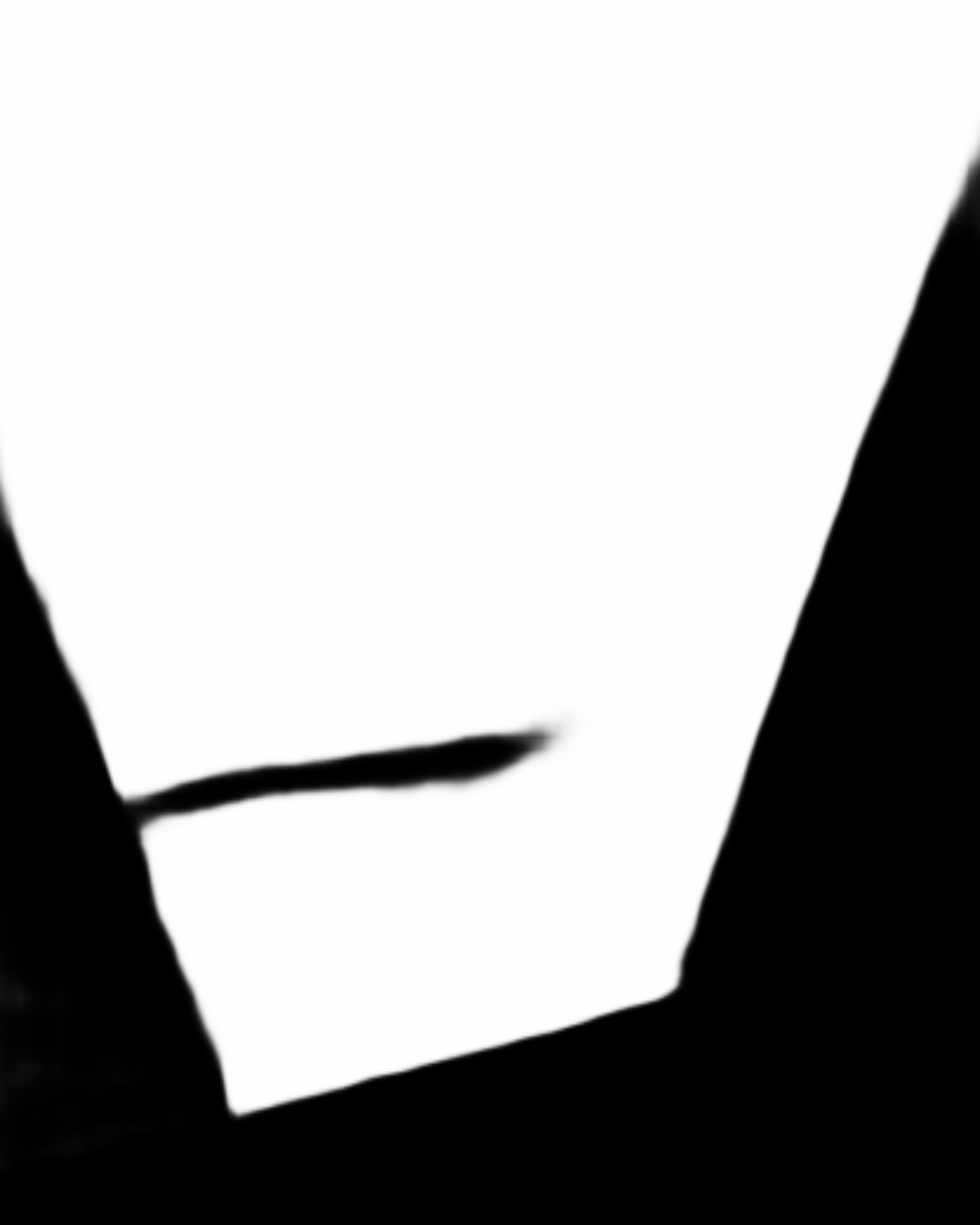}
    
                \includegraphics[width=1\linewidth]{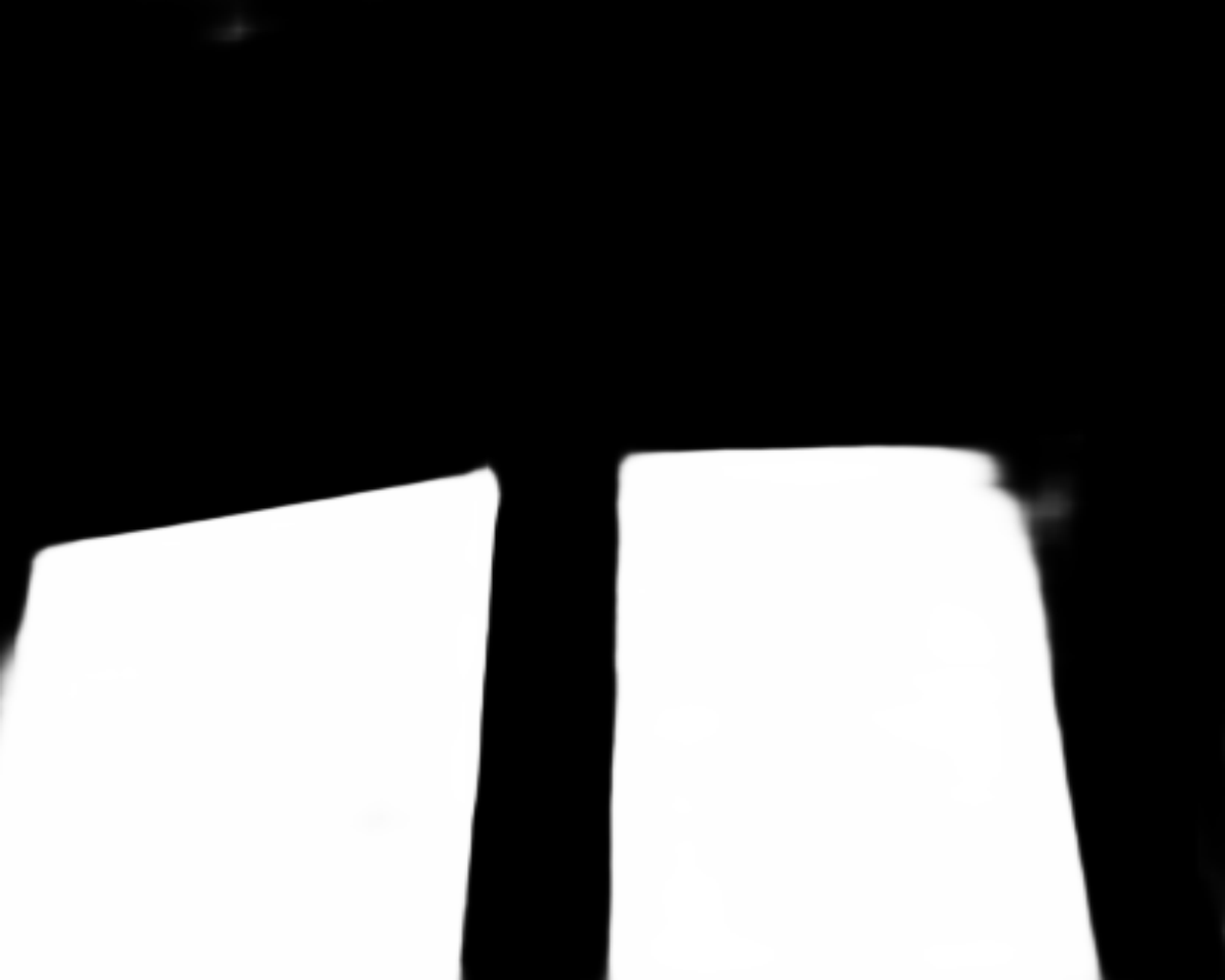}
                
                \includegraphics[width=1\linewidth]{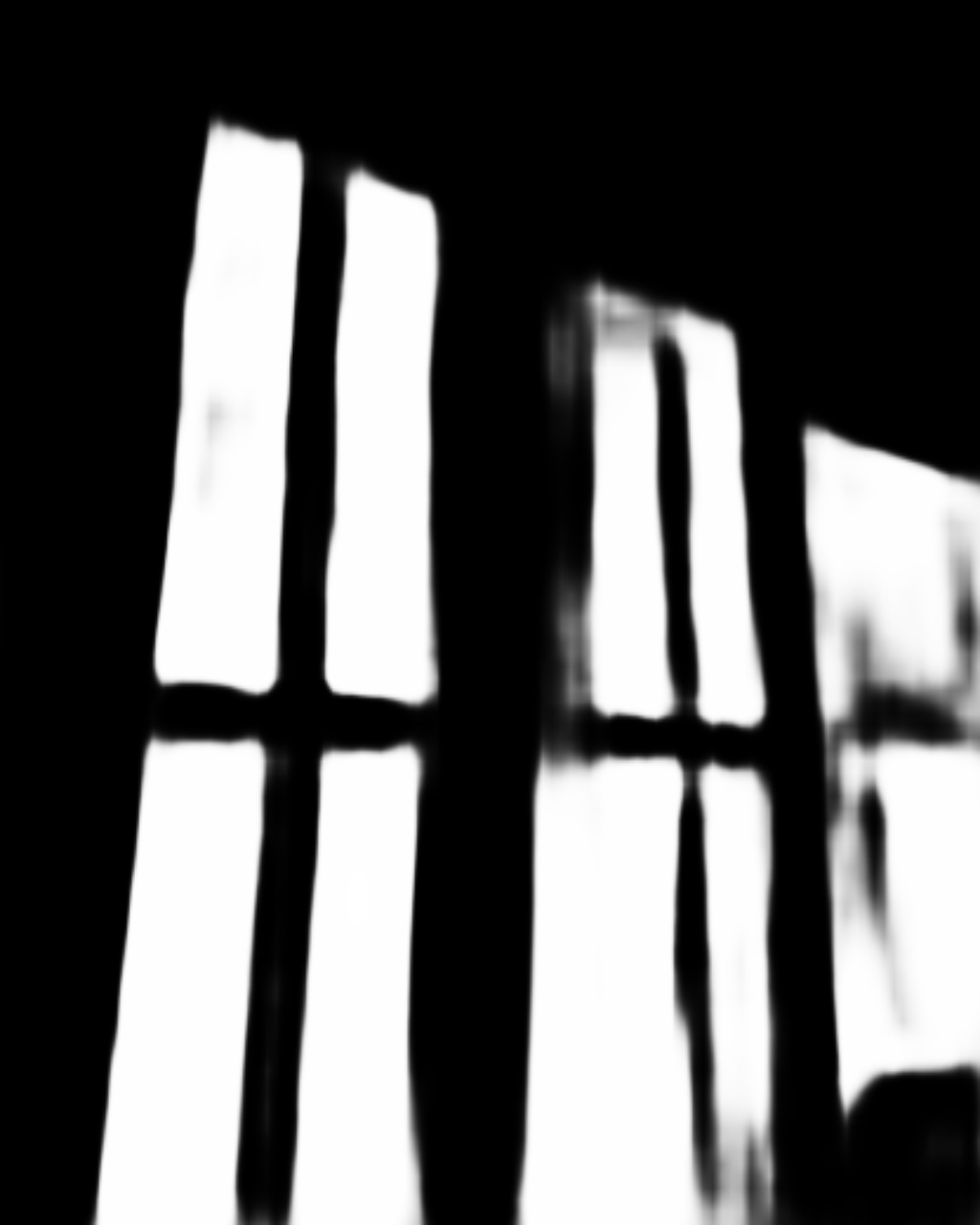}

                \includegraphics[width=1\linewidth]{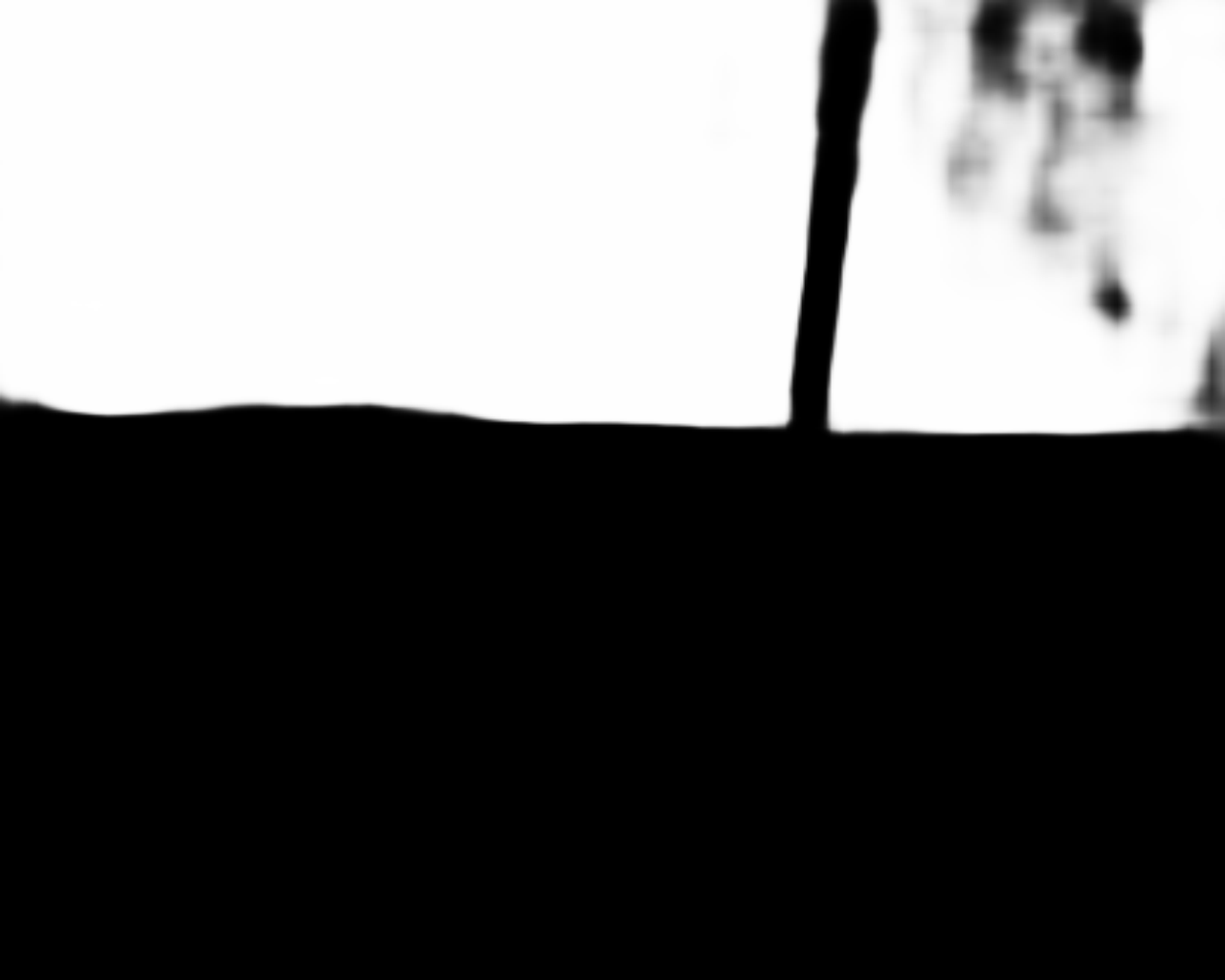}
    
                \includegraphics[width=1\linewidth]{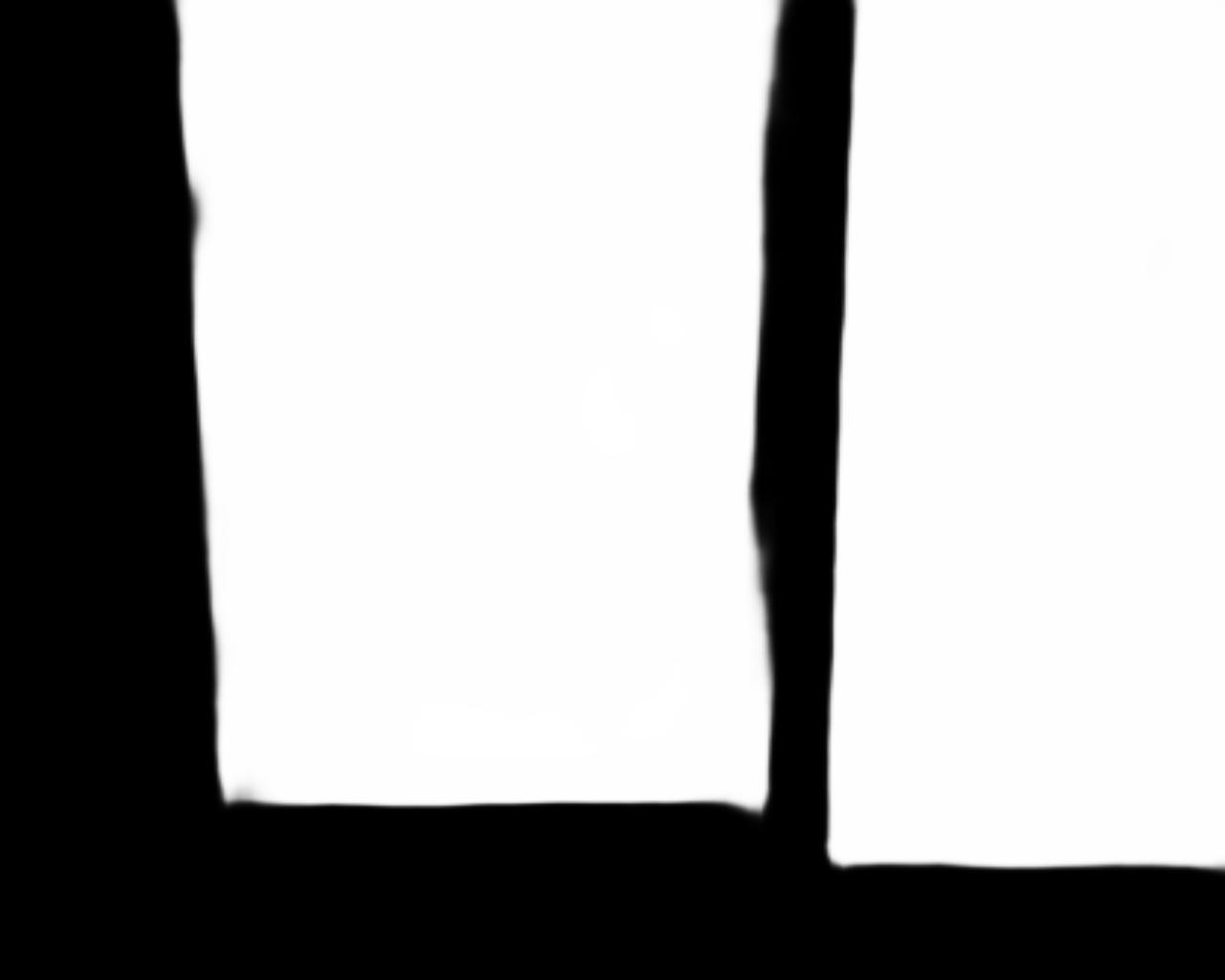}
                
                \includegraphics[width=1\linewidth]{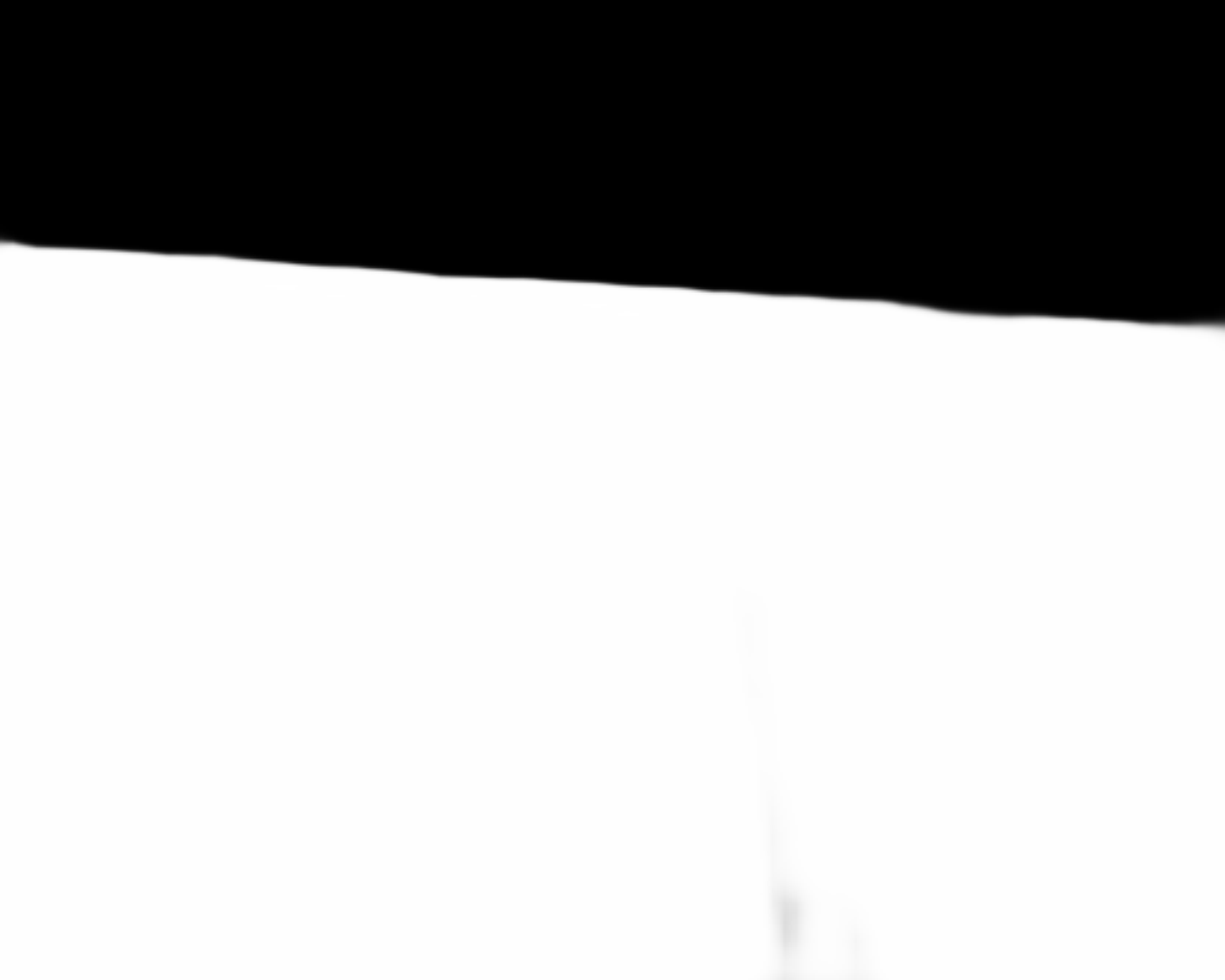}
    
                \includegraphics[width=1\linewidth]{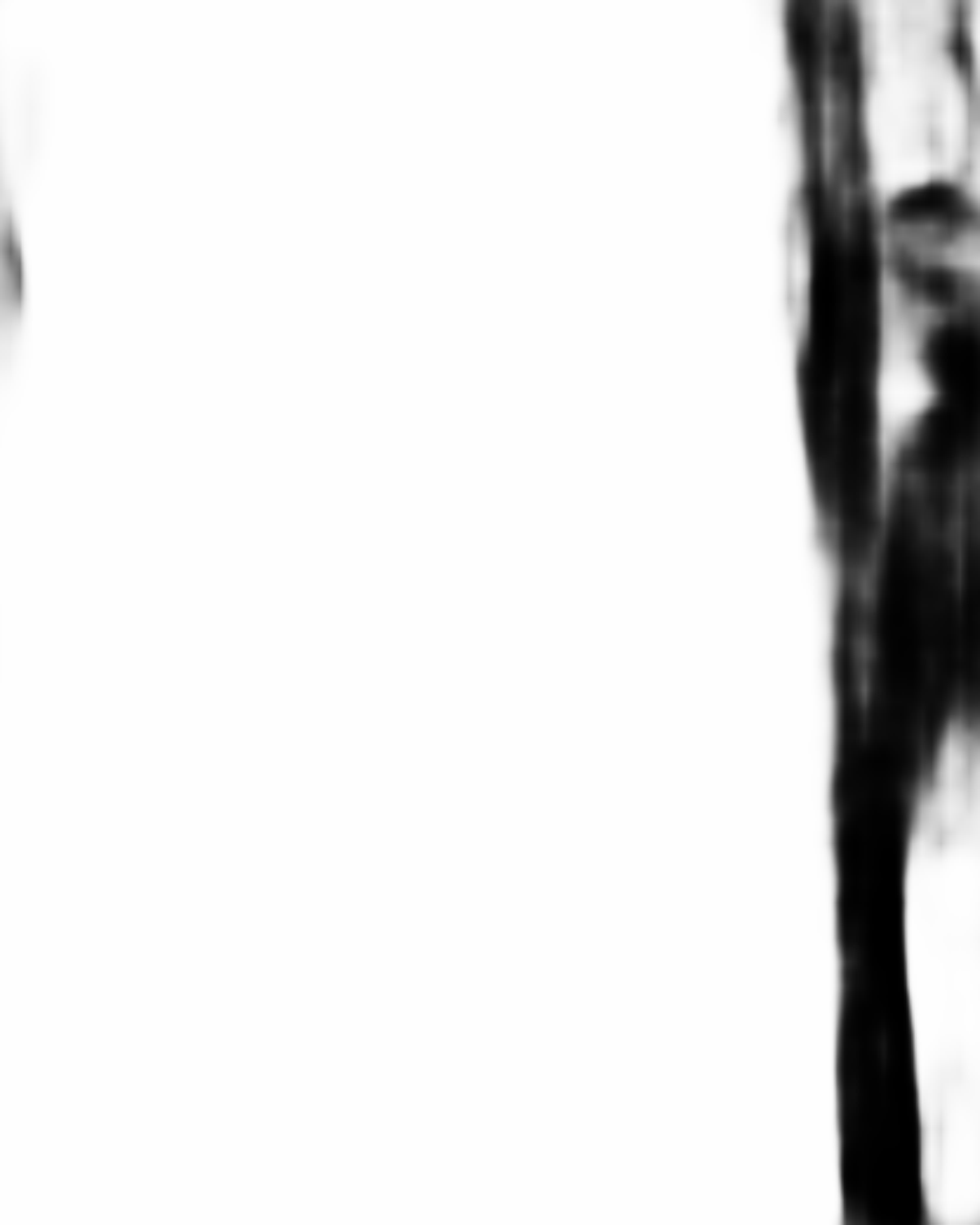}
    
                \includegraphics[width=1\linewidth]{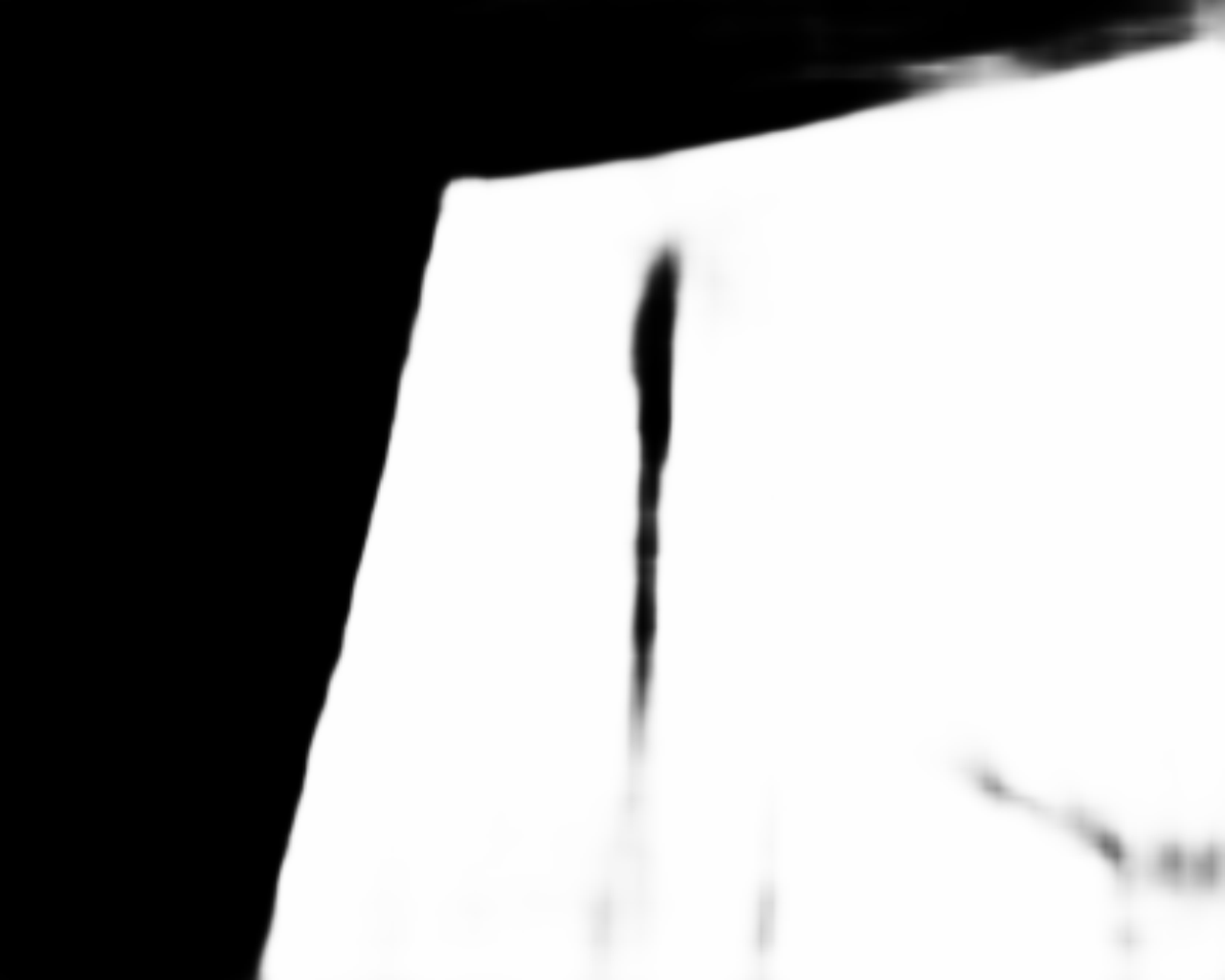}

          \end{minipage}
          }  
          \subfloat[GDNet]{
          \begin{minipage}[t]{0.08\textwidth}
                \centering

                \includegraphics[width=1\linewidth]{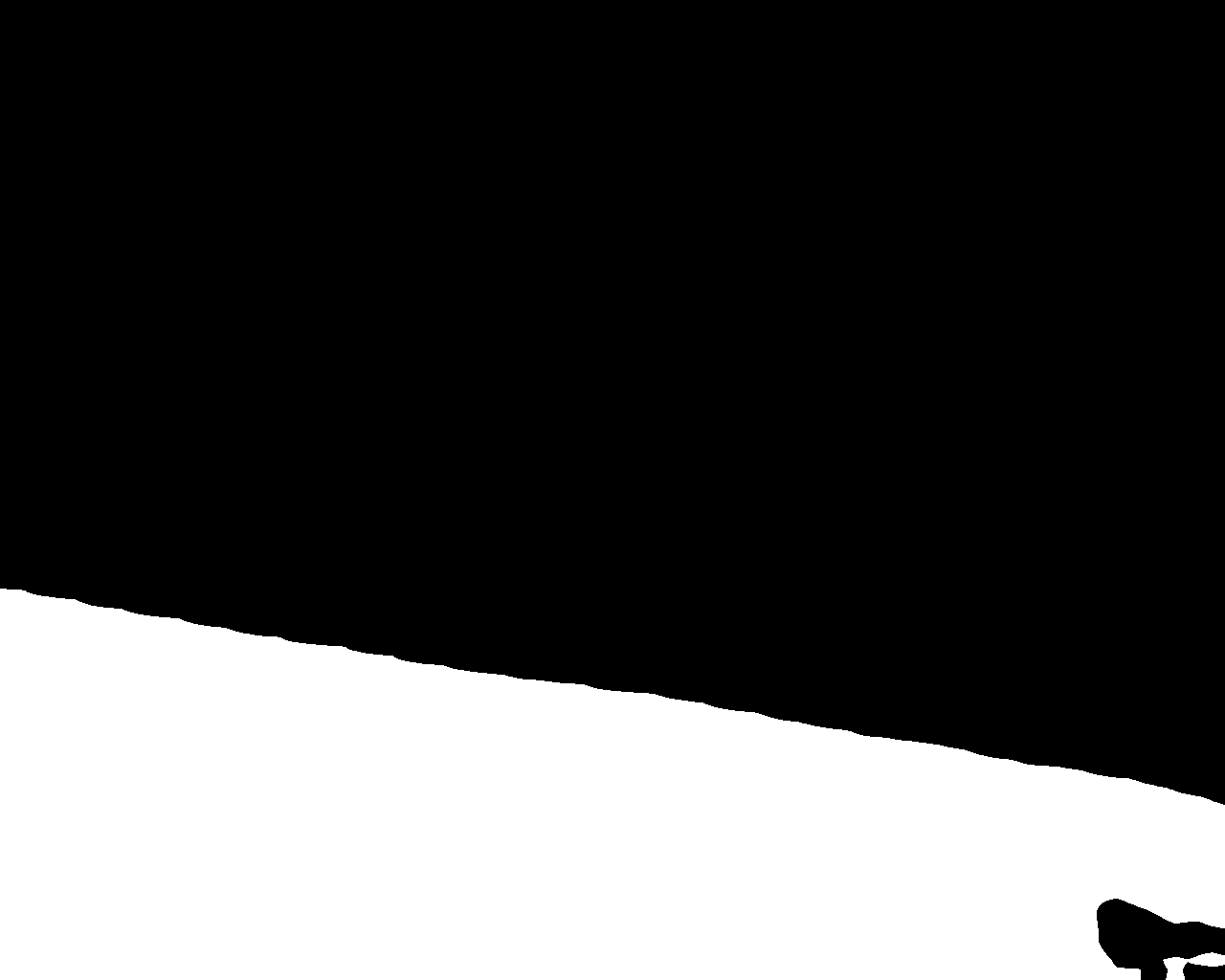}
                
                \includegraphics[width=1\linewidth]{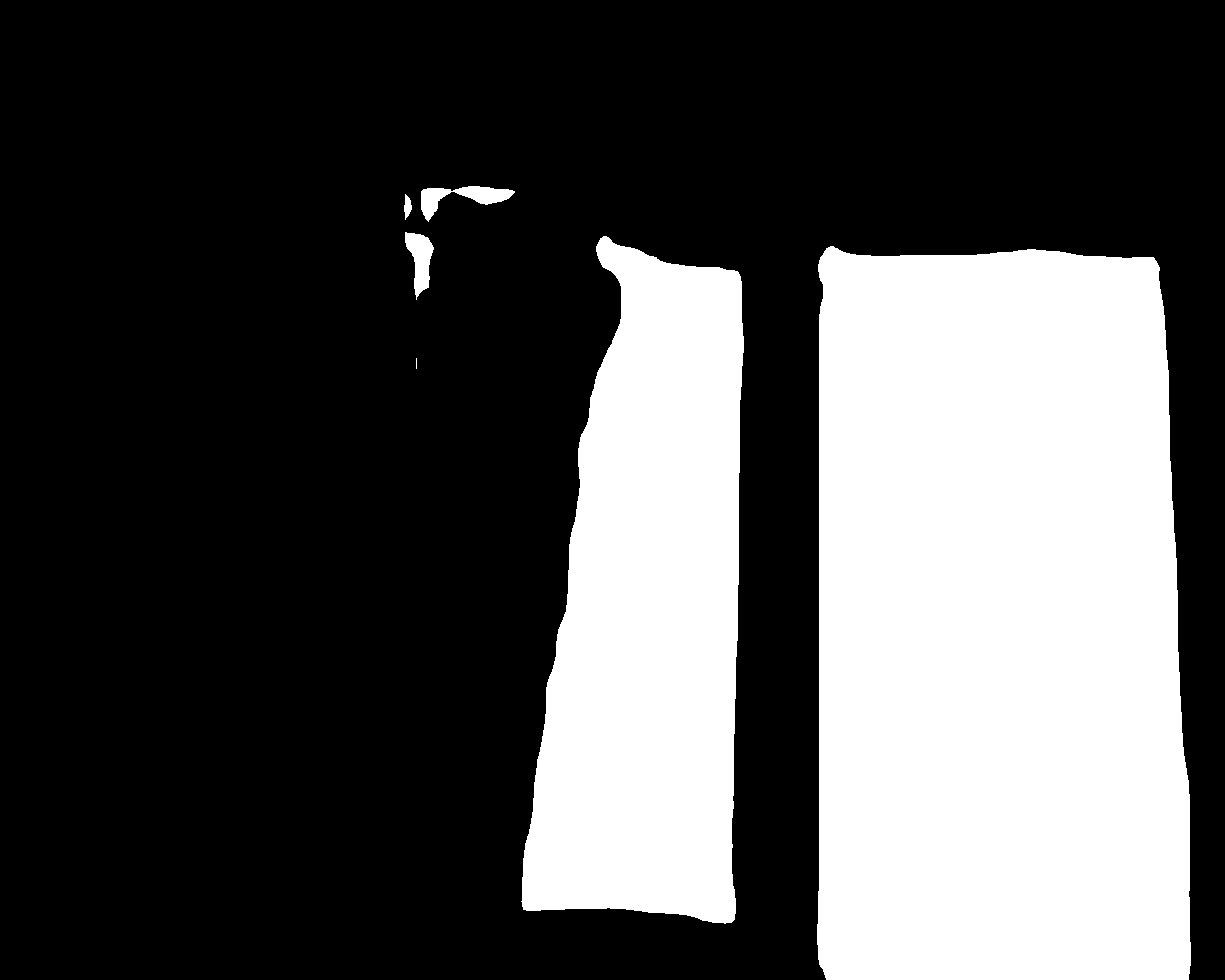}
    
                \includegraphics[width=1\linewidth]{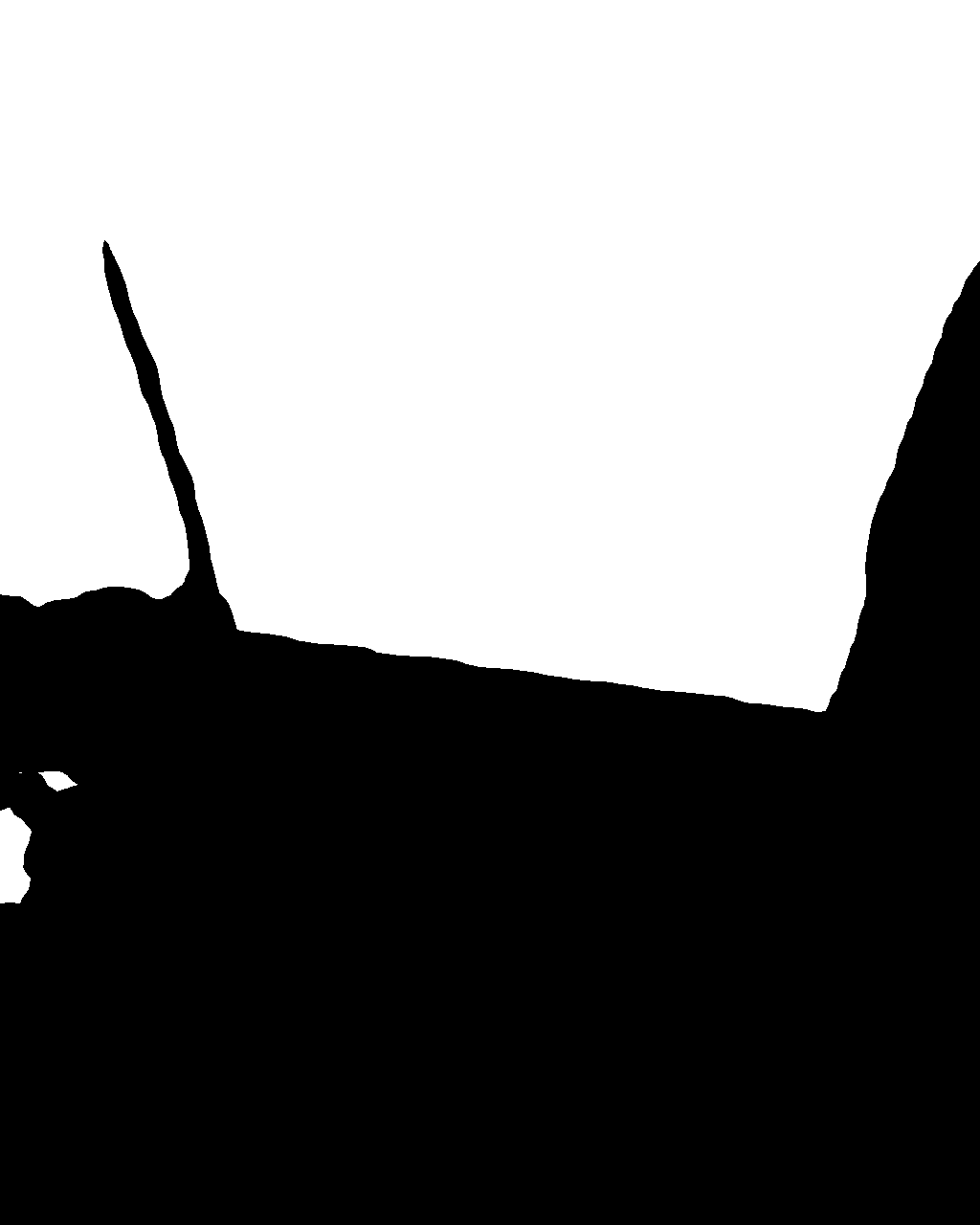}
                
                \includegraphics[width=1\linewidth]{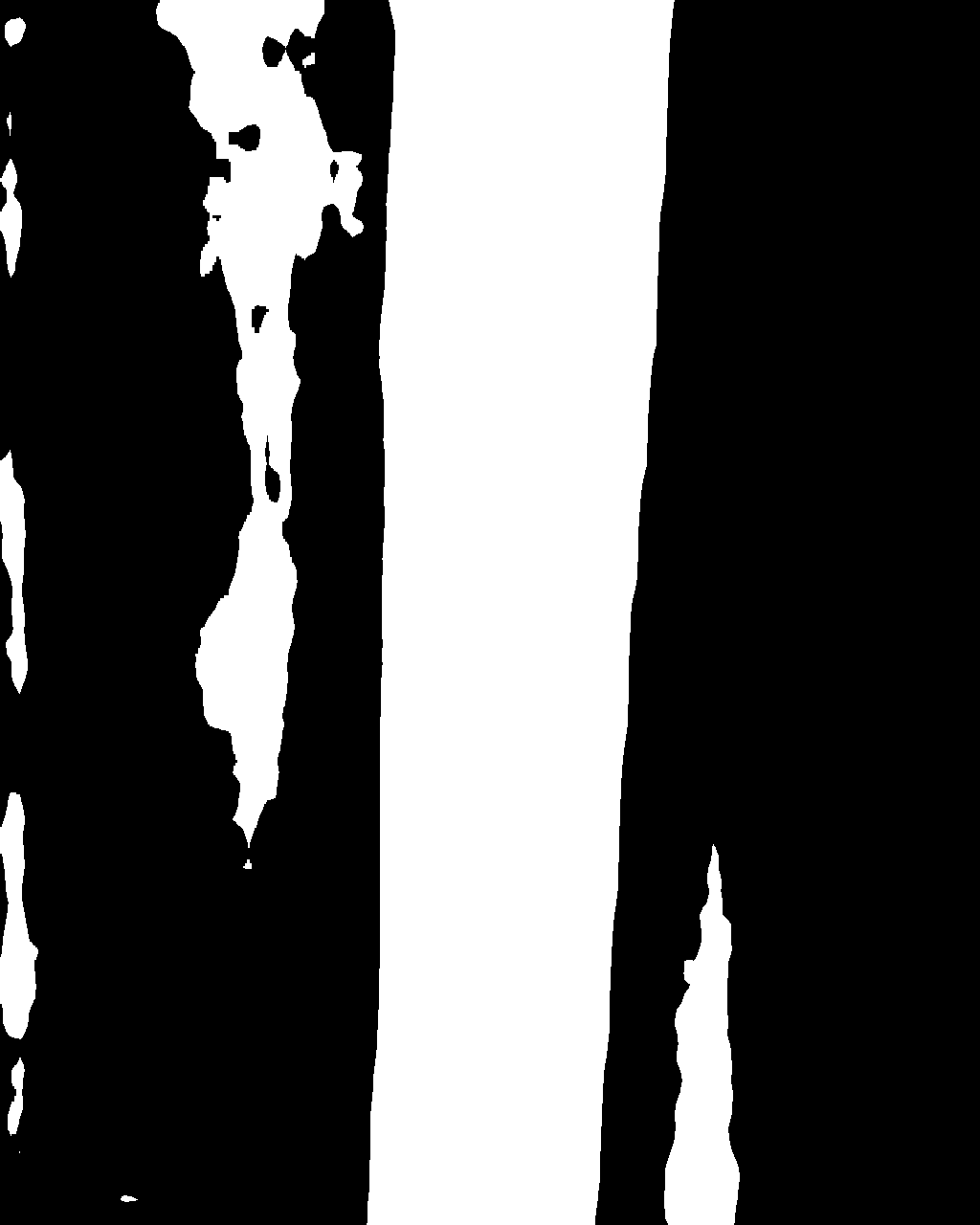}
    
                \includegraphics[width=1\linewidth]{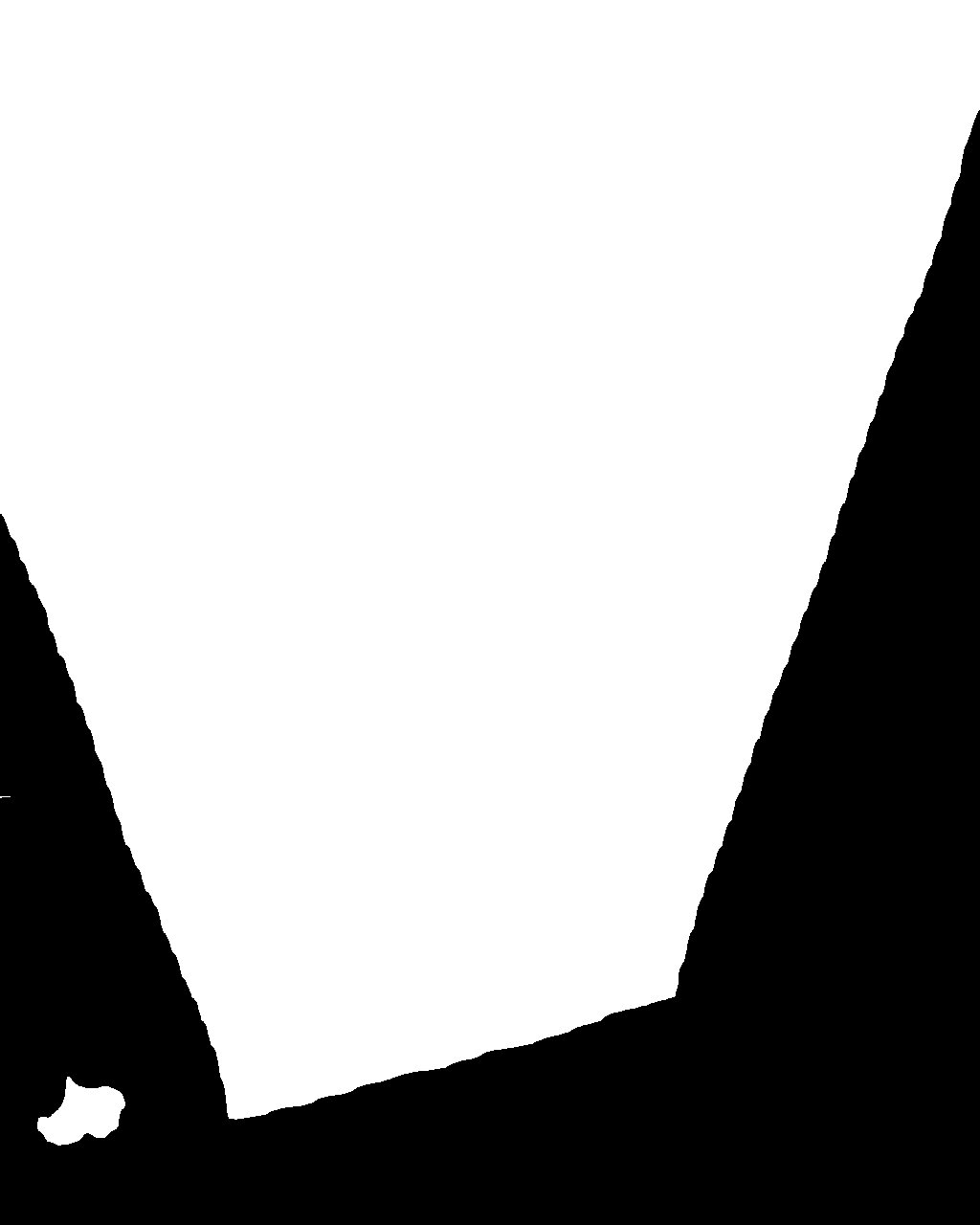}
    
                \includegraphics[width=1\linewidth]{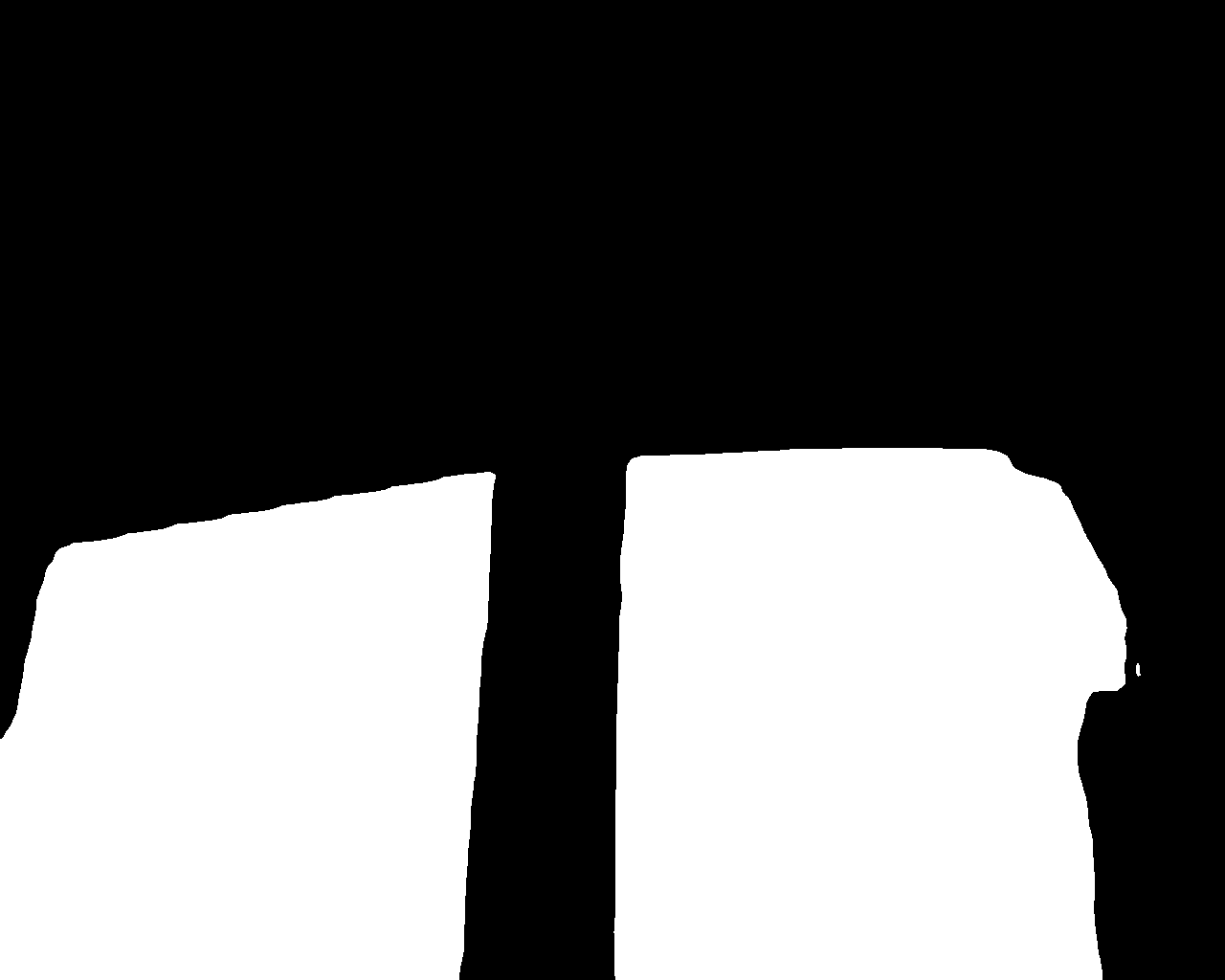}
                
                \includegraphics[width=1\linewidth]{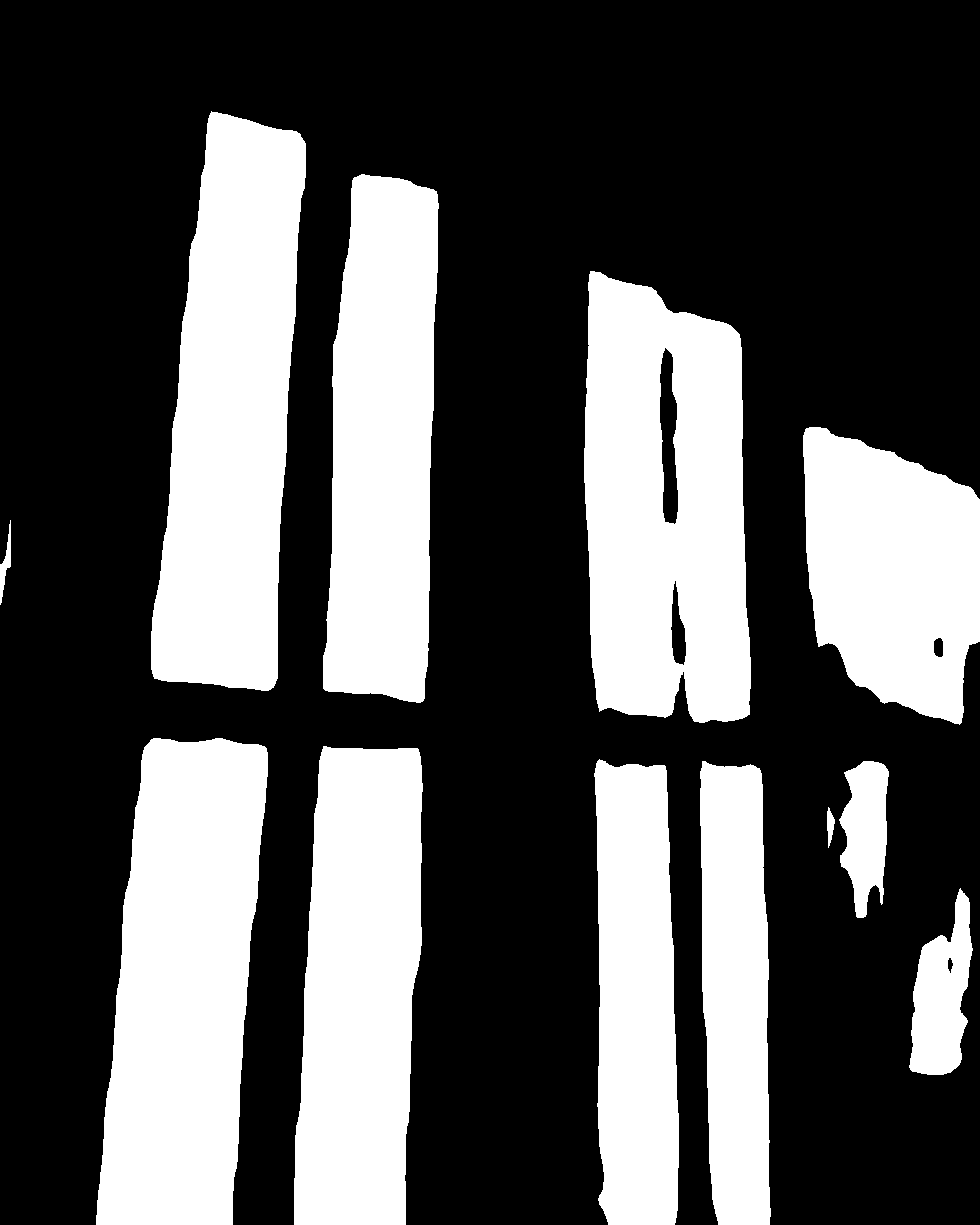}

                \includegraphics[width=1\linewidth]{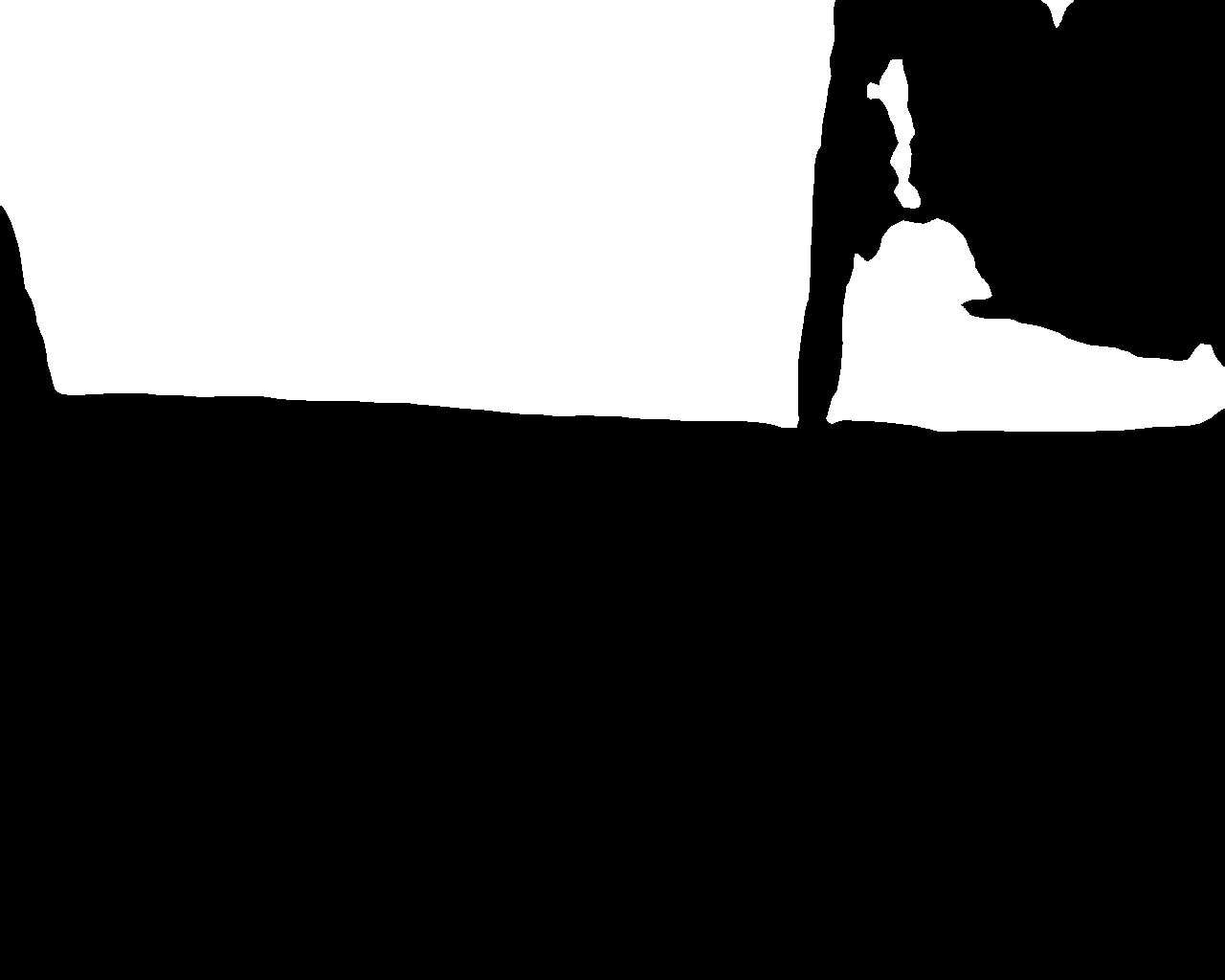}
    
                \includegraphics[width=1\linewidth]{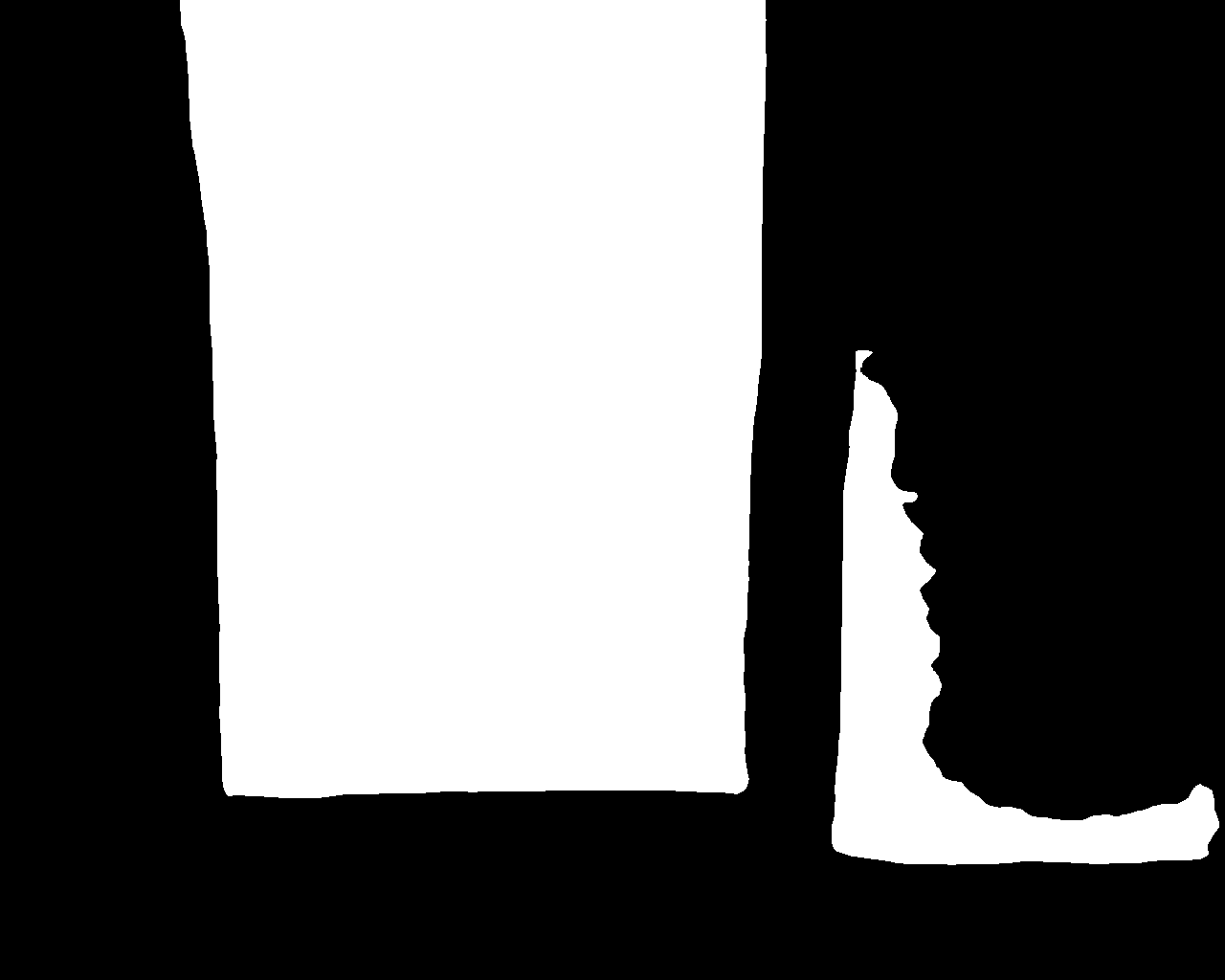}
                
                \includegraphics[width=1\linewidth]{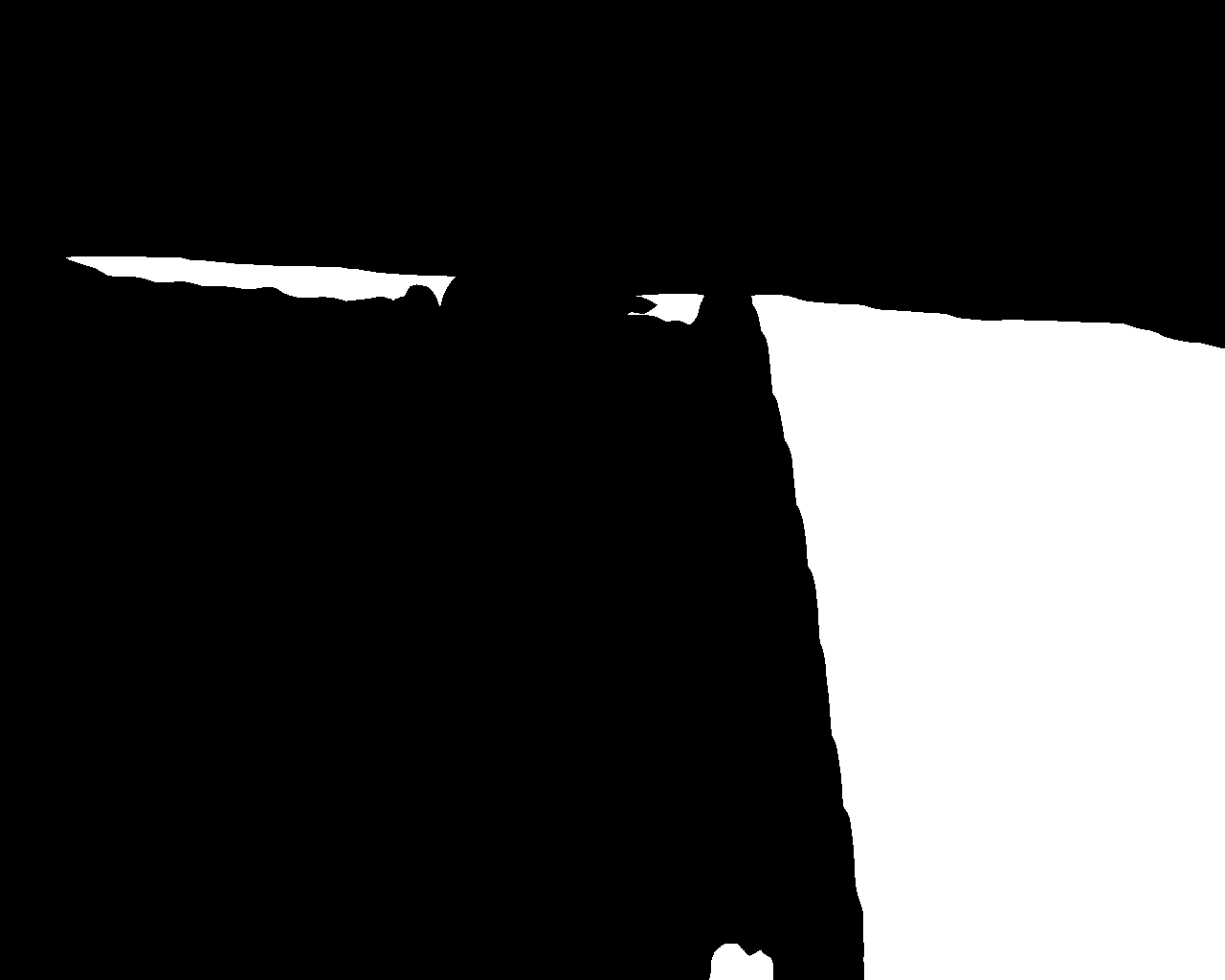}
    
                \includegraphics[width=1\linewidth]{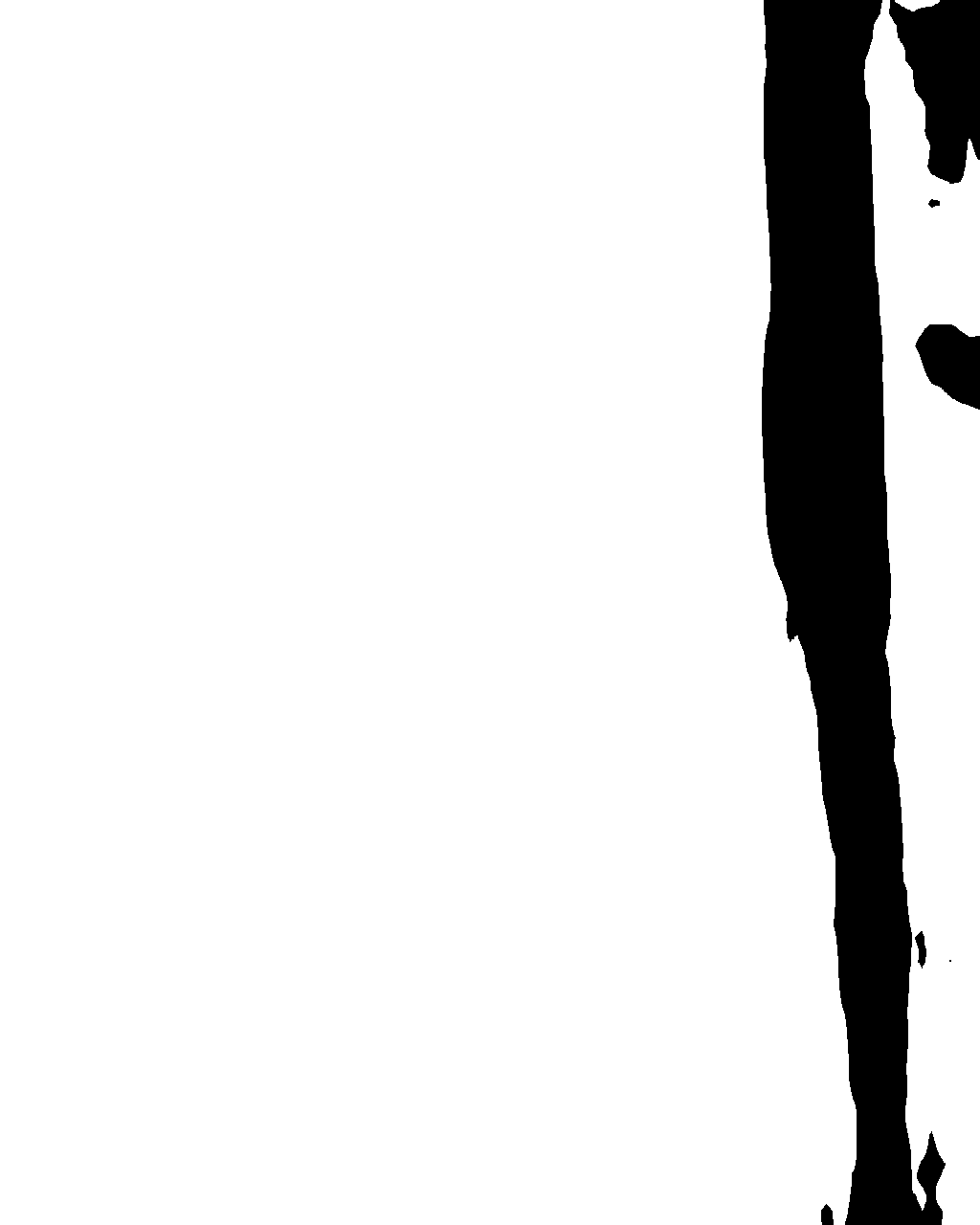}
    
                \includegraphics[width=1\linewidth]{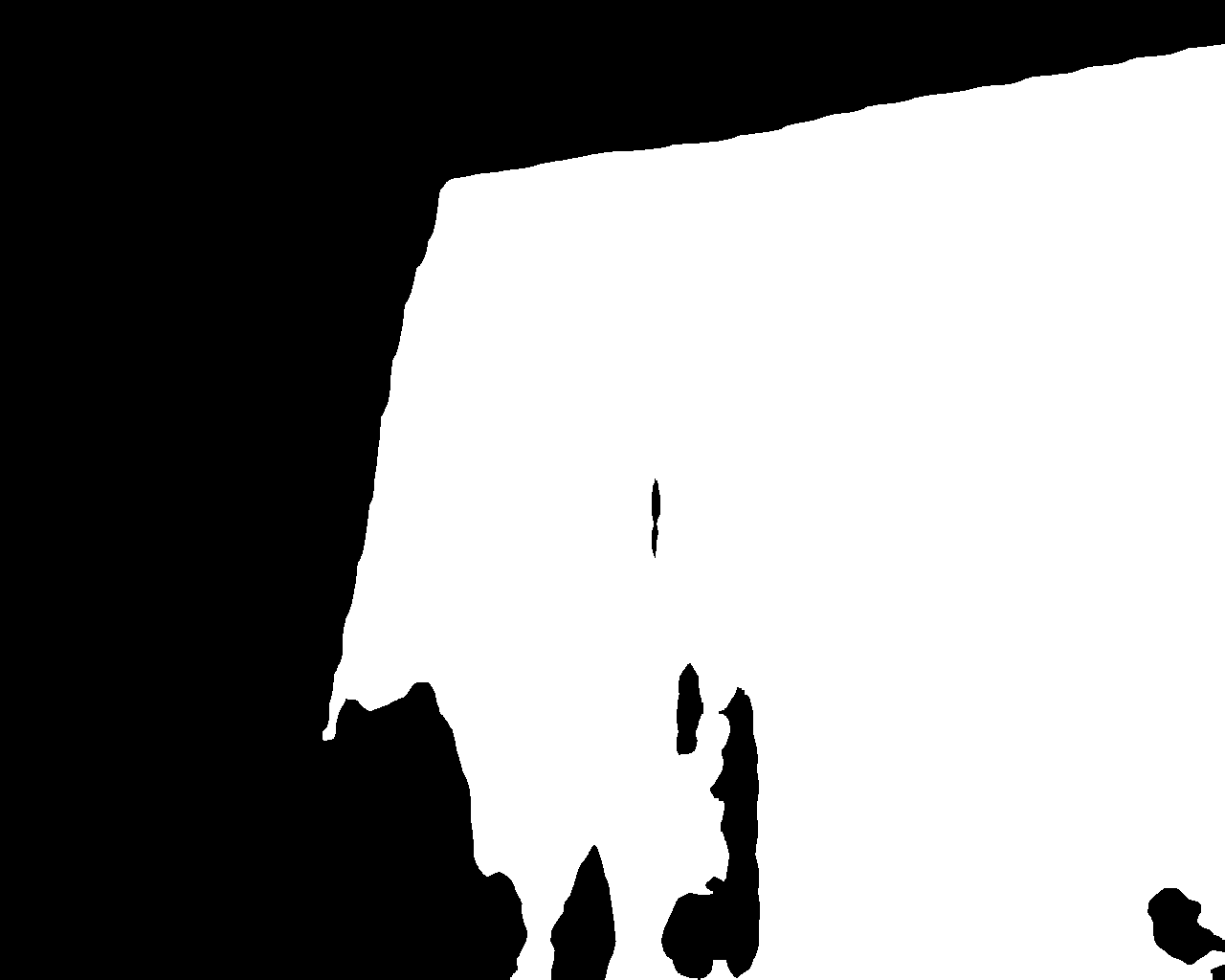}

          \end{minipage}
          }  
          \subfloat[GSD]{
          \begin{minipage}[t]{0.08\textwidth}
                \centering

                \includegraphics[width=1\linewidth]{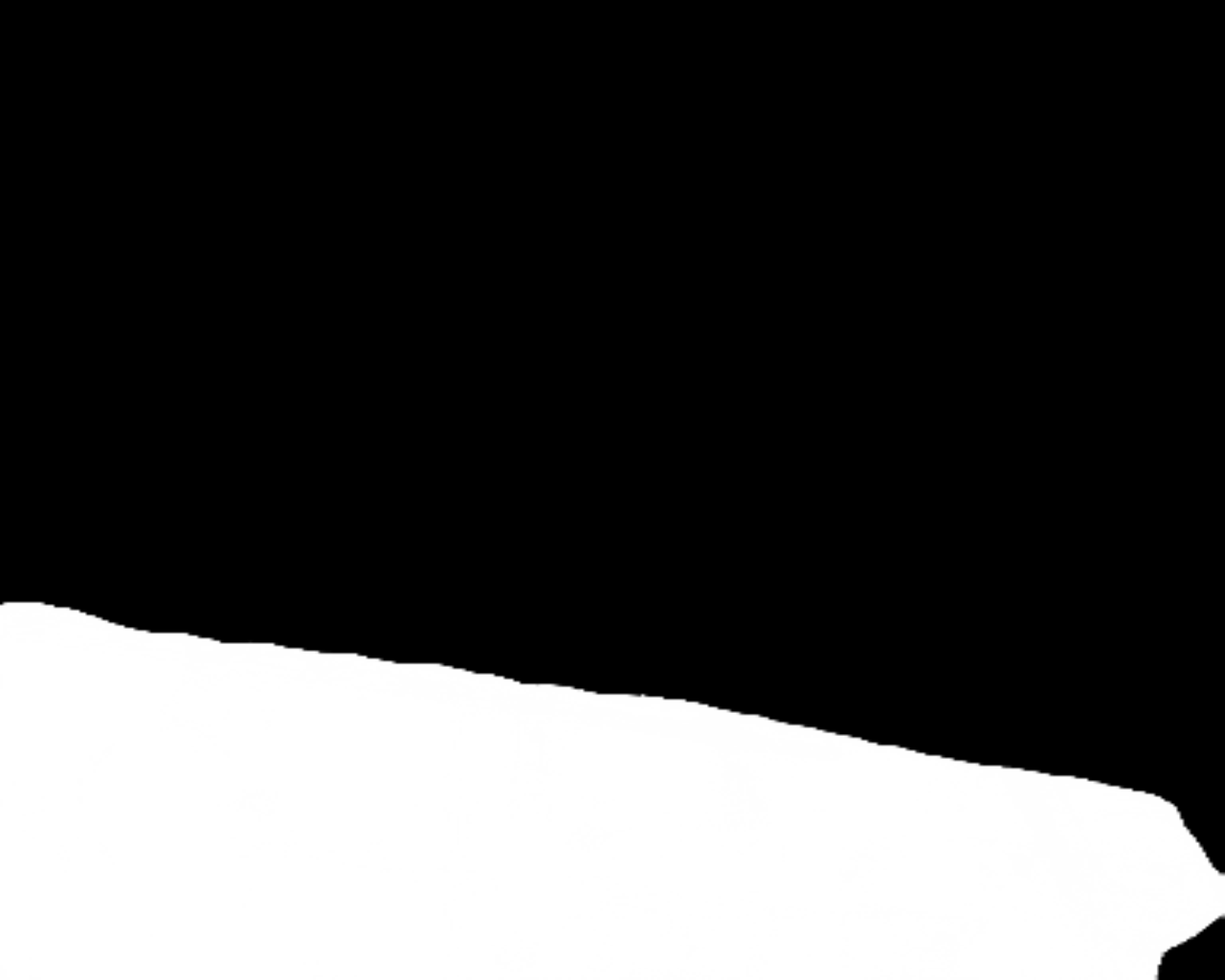}
                
                \includegraphics[width=1\linewidth]{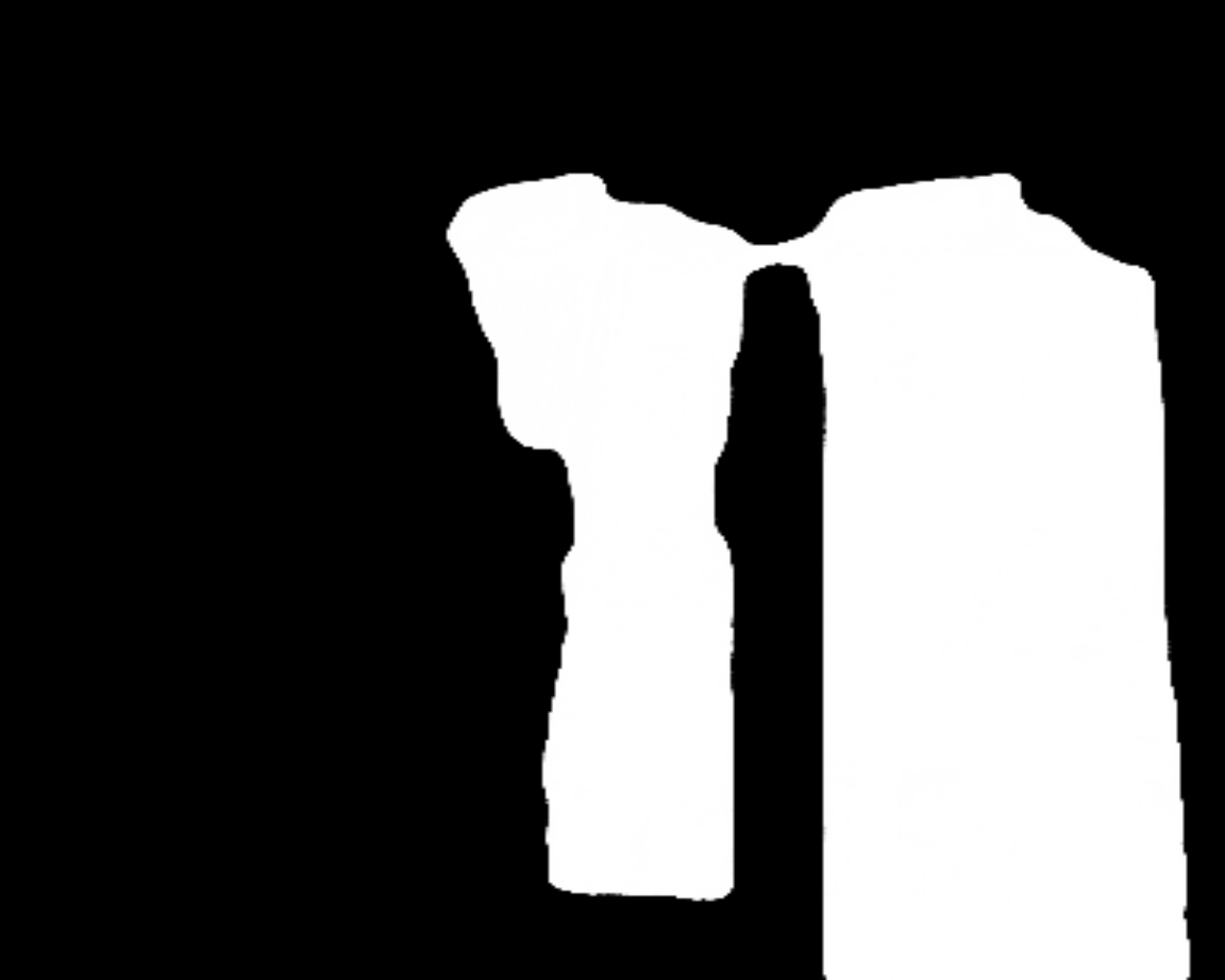}
    
                \includegraphics[width=1\linewidth]{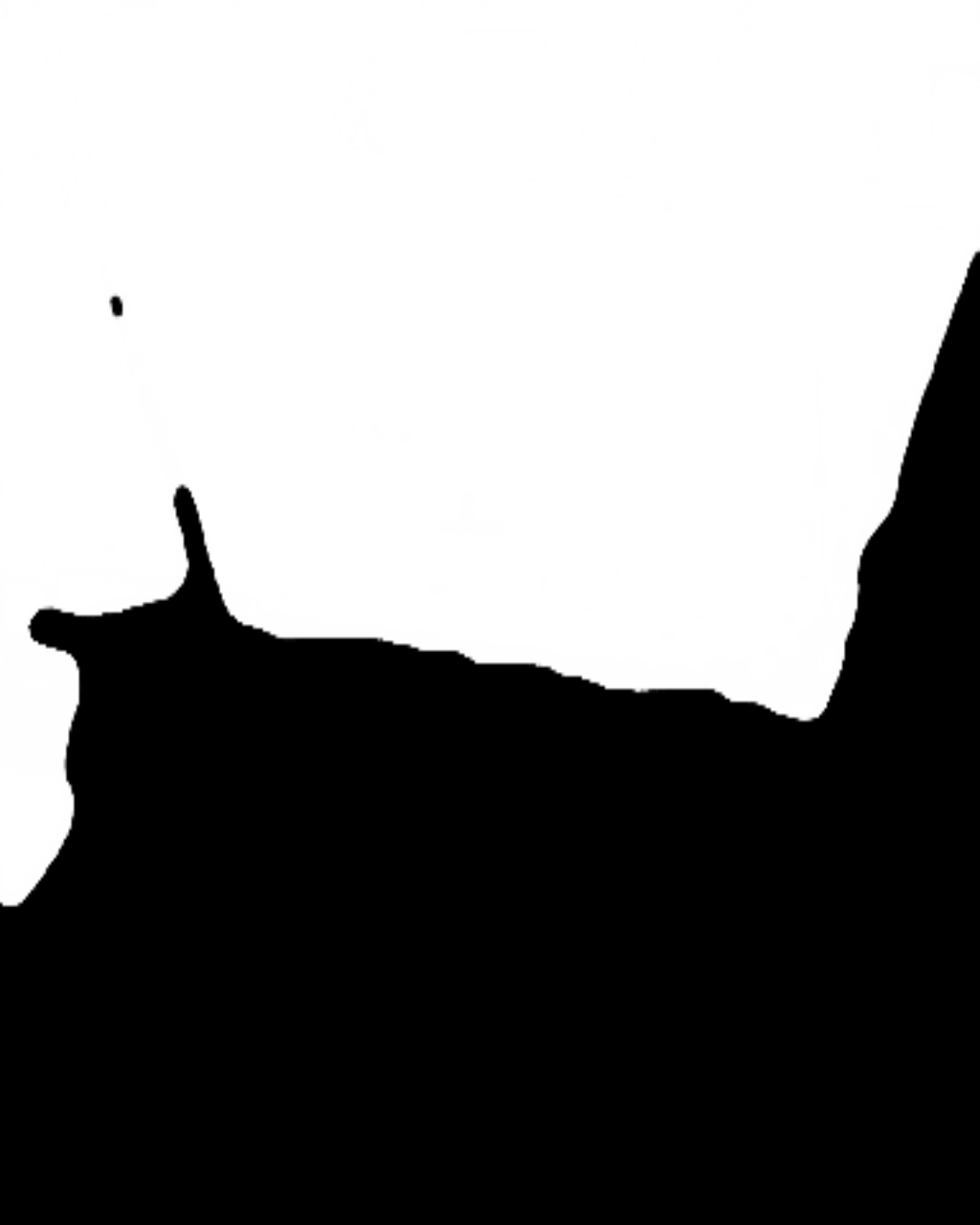}
                
                \includegraphics[width=1\linewidth]{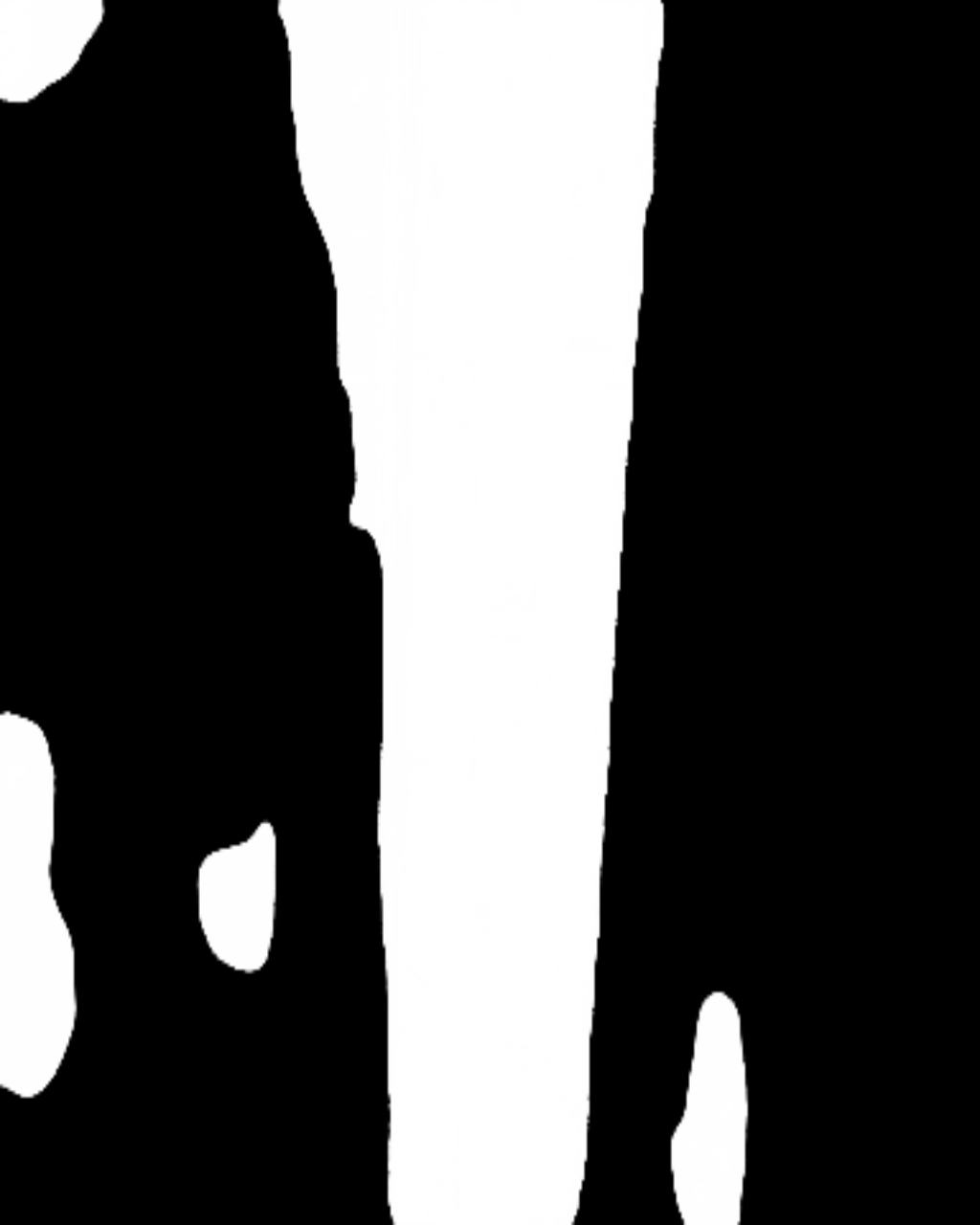}
    
                \includegraphics[width=1\linewidth]{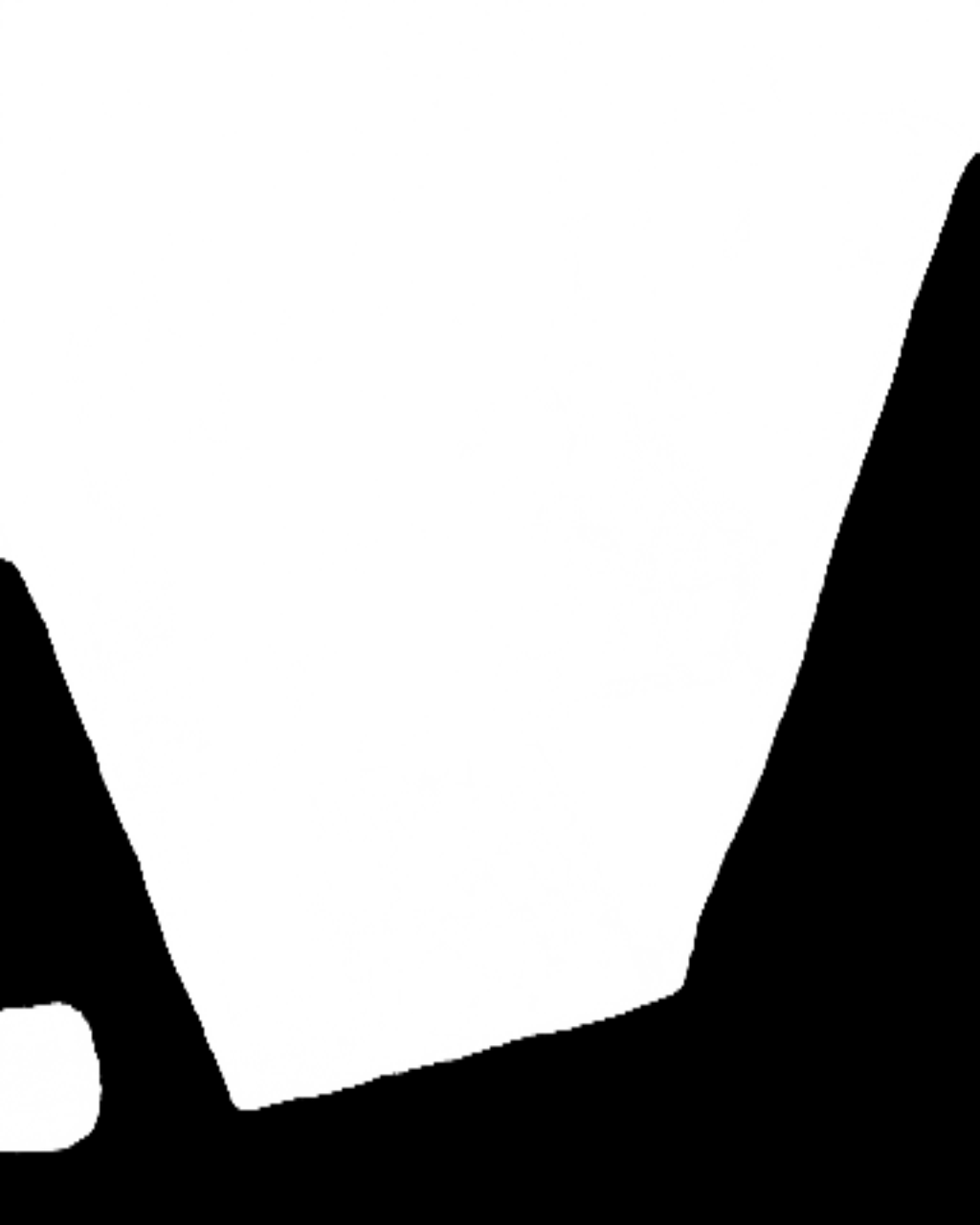}
    
                \includegraphics[width=1\linewidth]{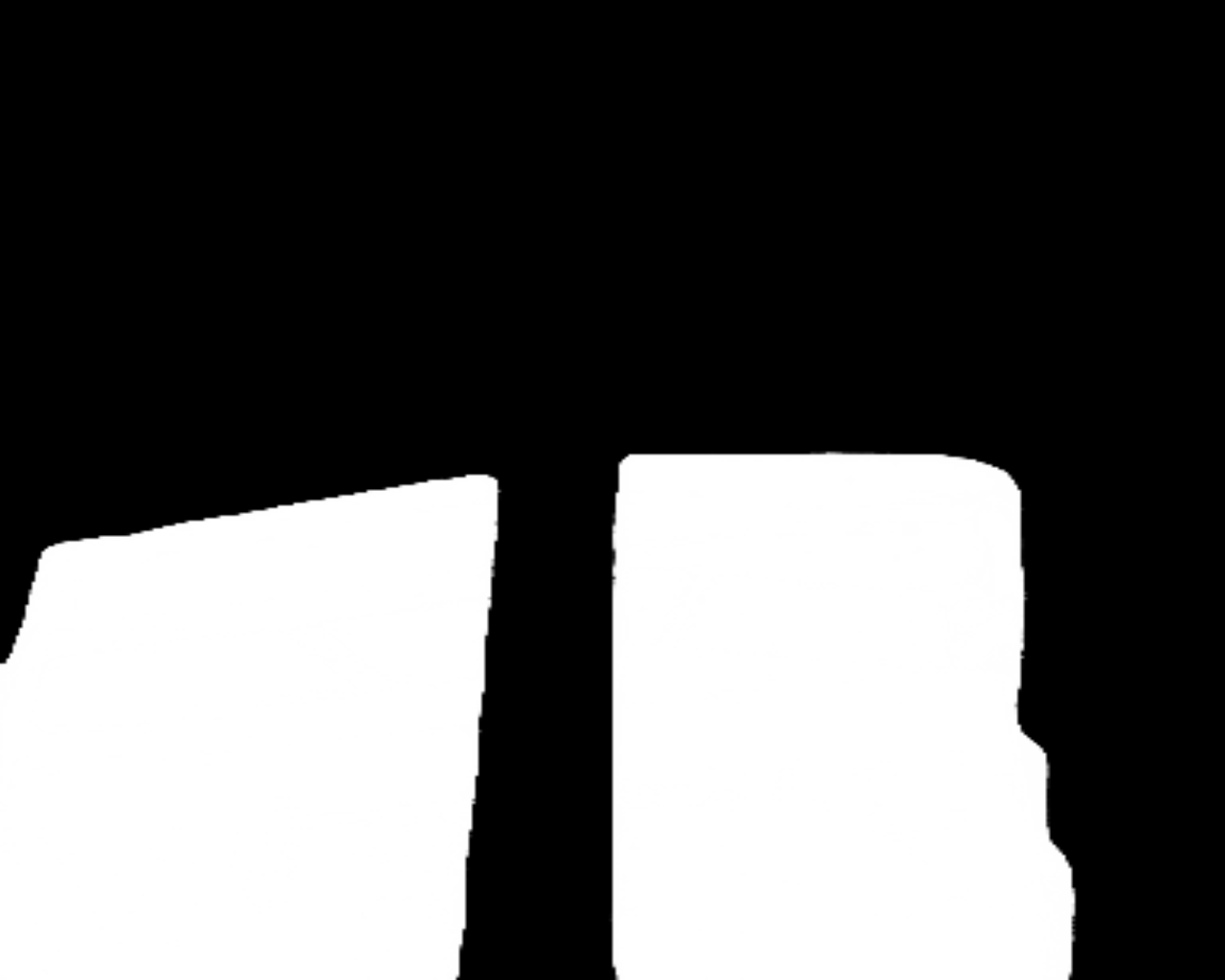}
                
                \includegraphics[width=1\linewidth]{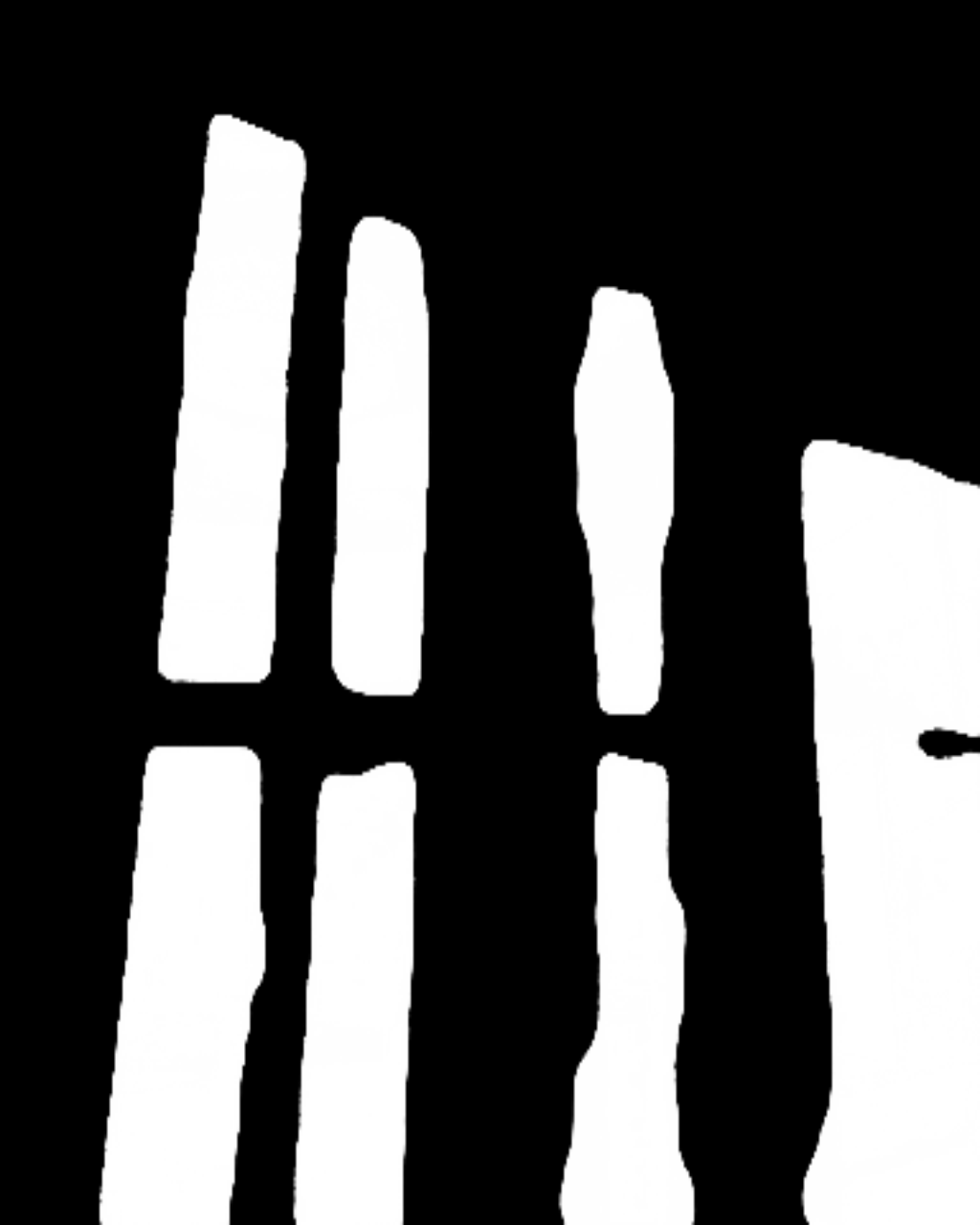}

                \includegraphics[width=1\linewidth]{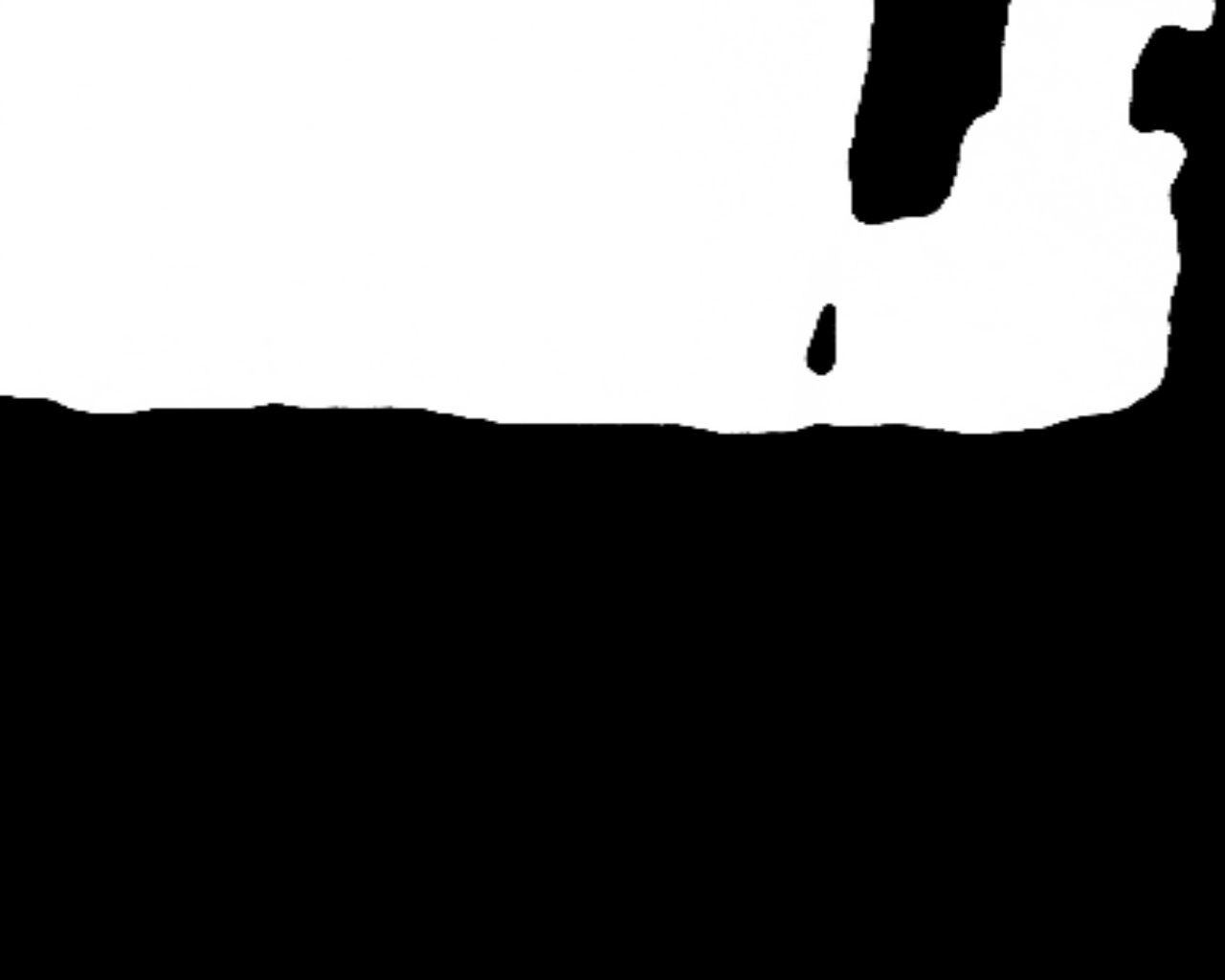}
    
                \includegraphics[width=1\linewidth]{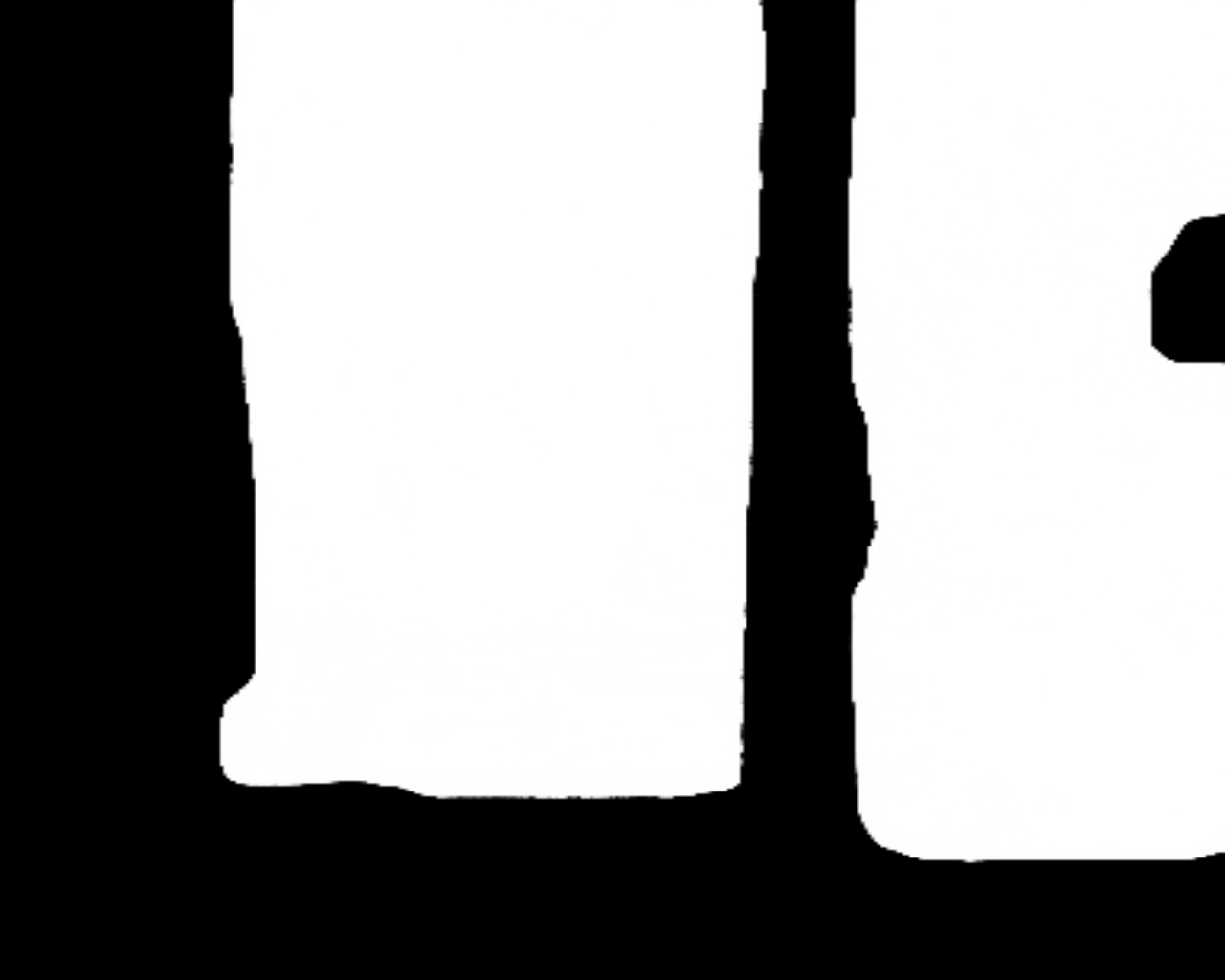}
                
                \includegraphics[width=1\linewidth]{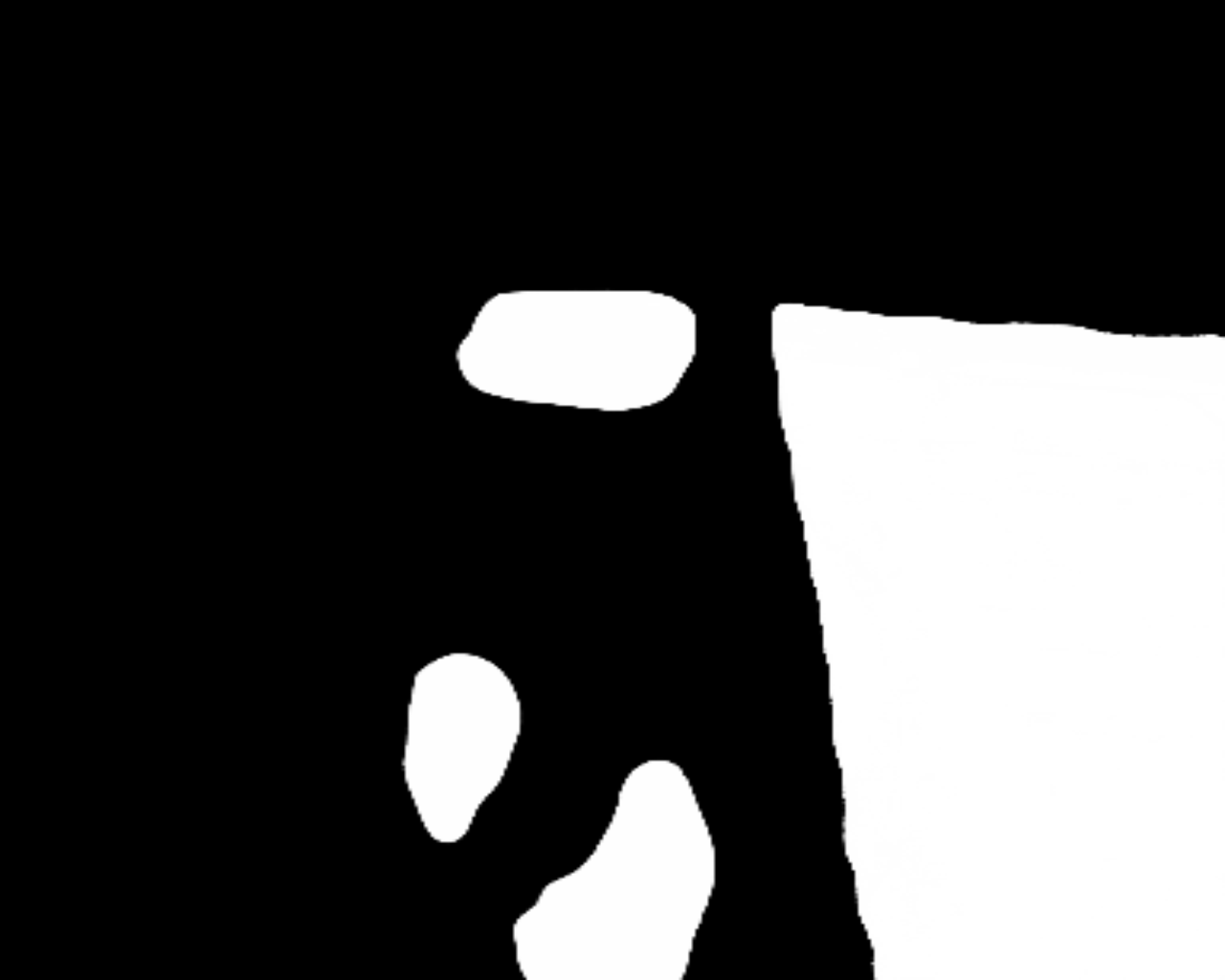}
    
                \includegraphics[width=1\linewidth]{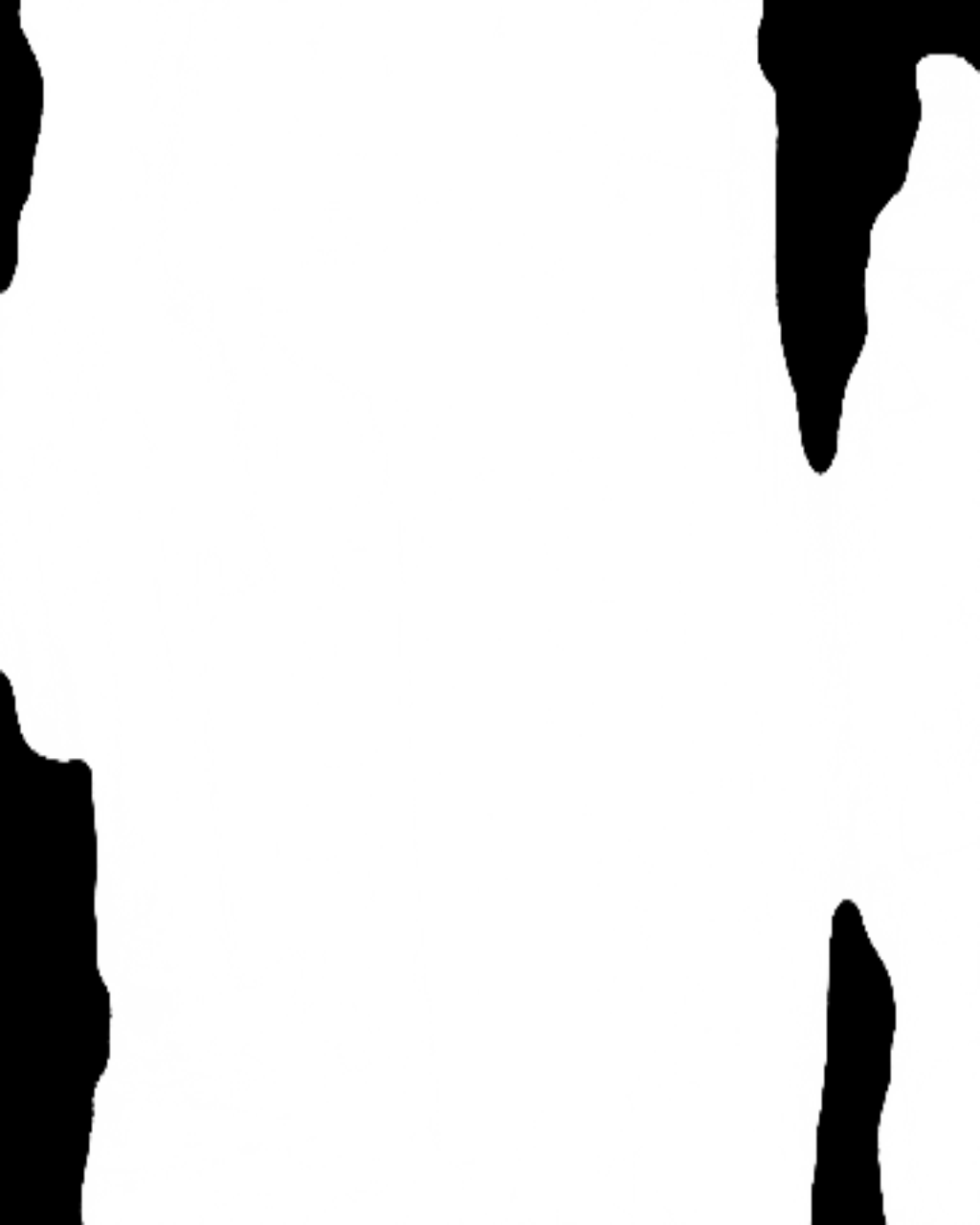}
    
                \includegraphics[width=1\linewidth]{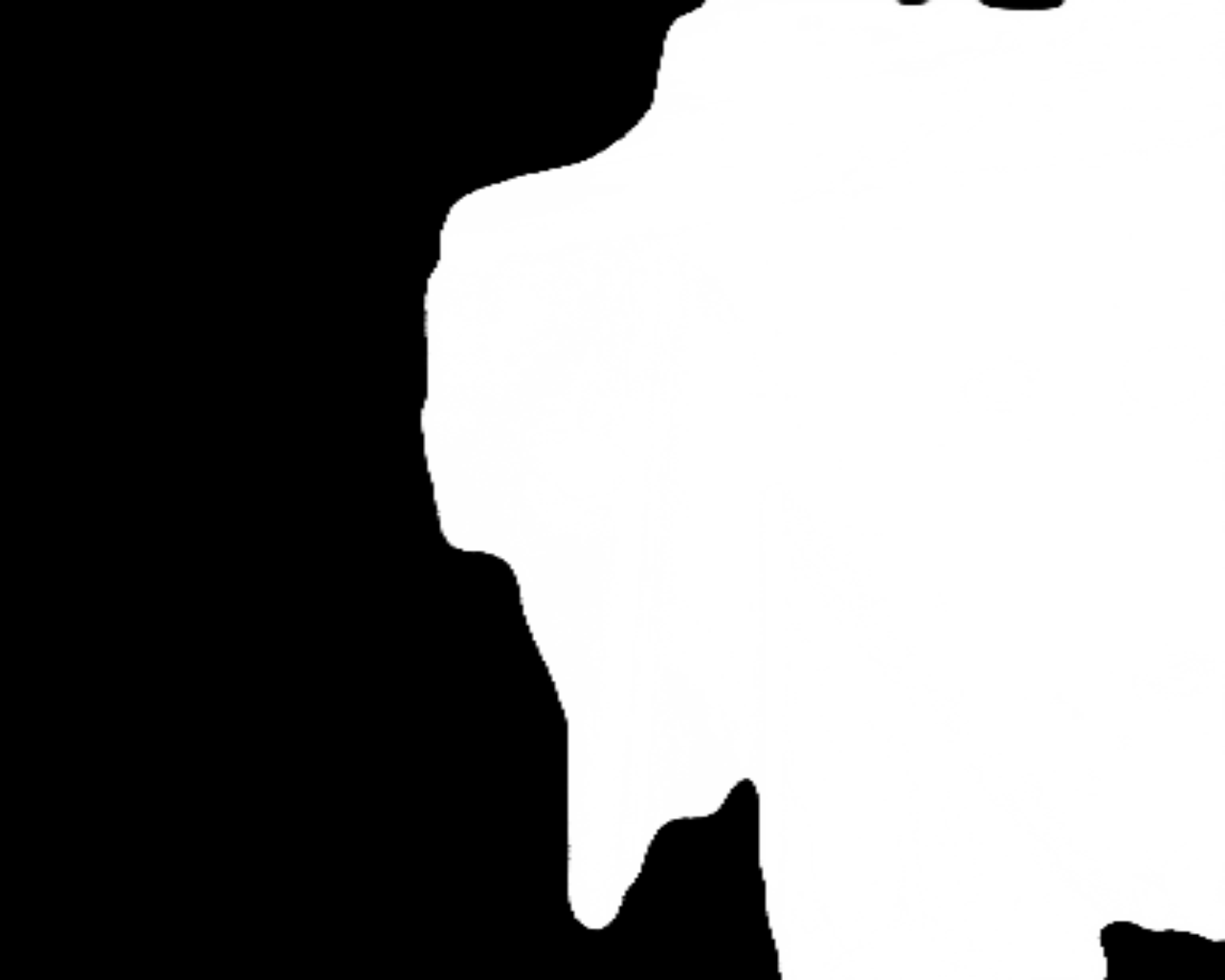}

          \end{minipage}
          }  
          \subfloat[Ours]{
          \begin{minipage}[t]{0.08\textwidth}
                \centering

                \includegraphics[width=1\linewidth]{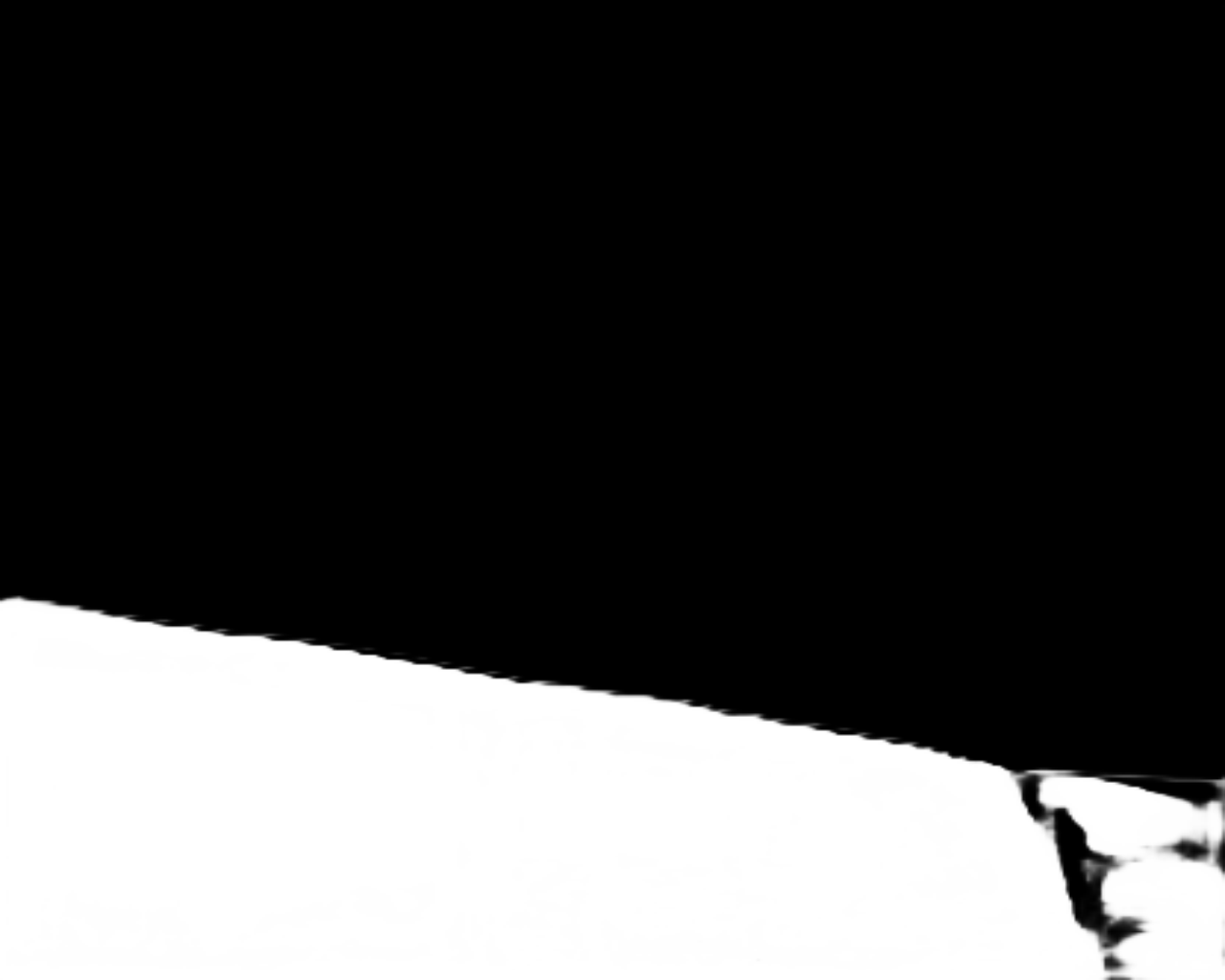}
                
                \includegraphics[width=1\linewidth]{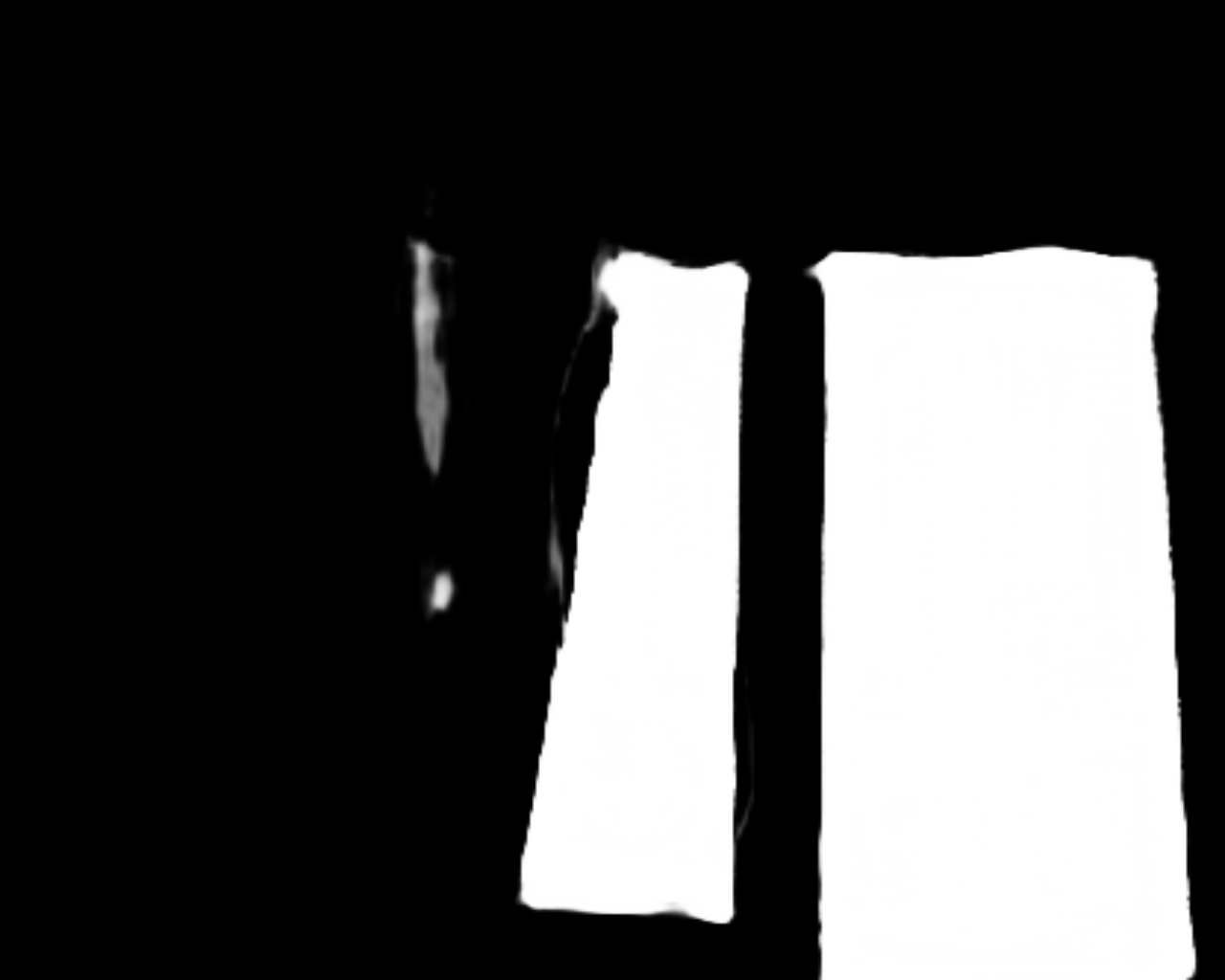}
    
                \includegraphics[width=1\linewidth]{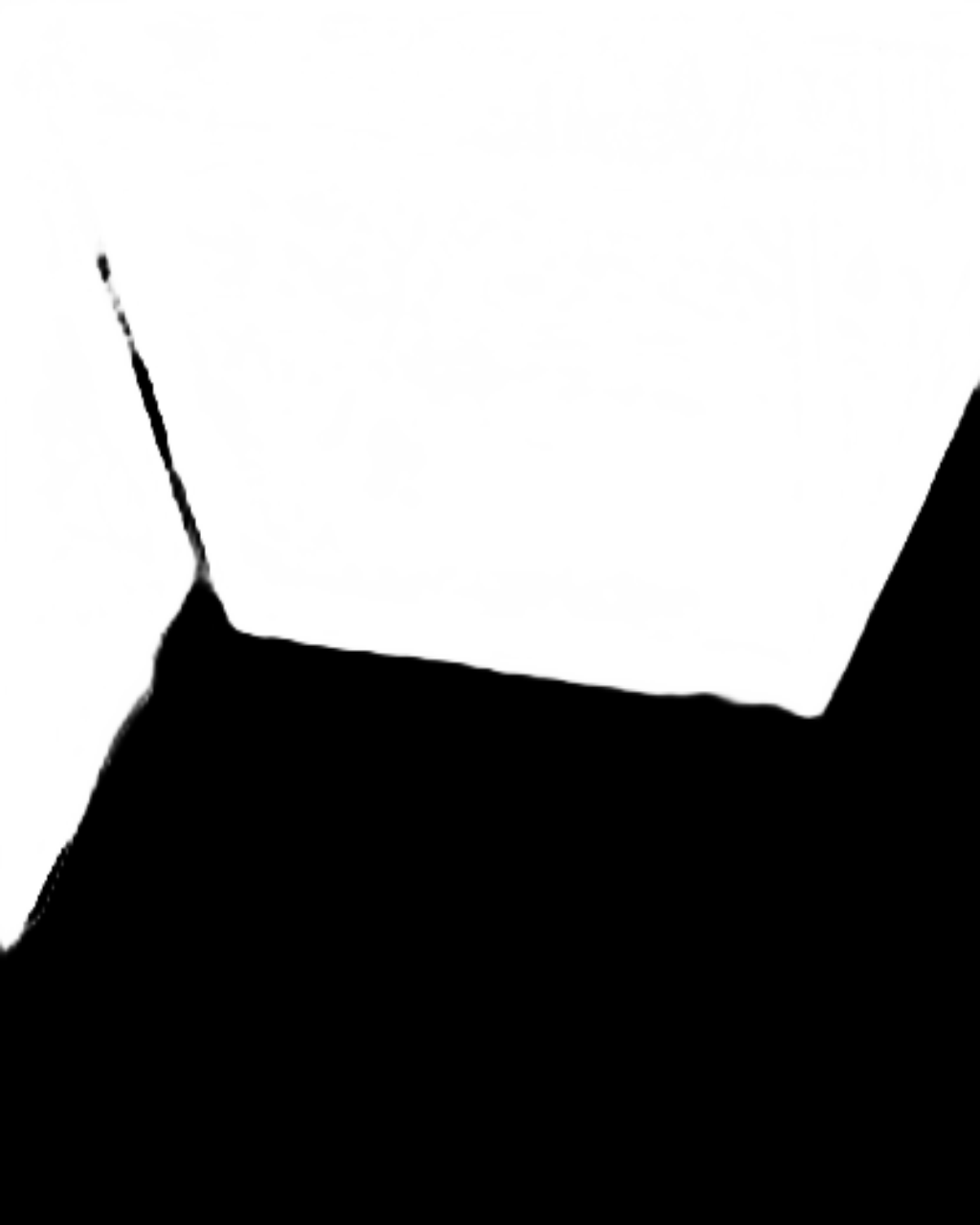}
                
                \includegraphics[width=1\linewidth]{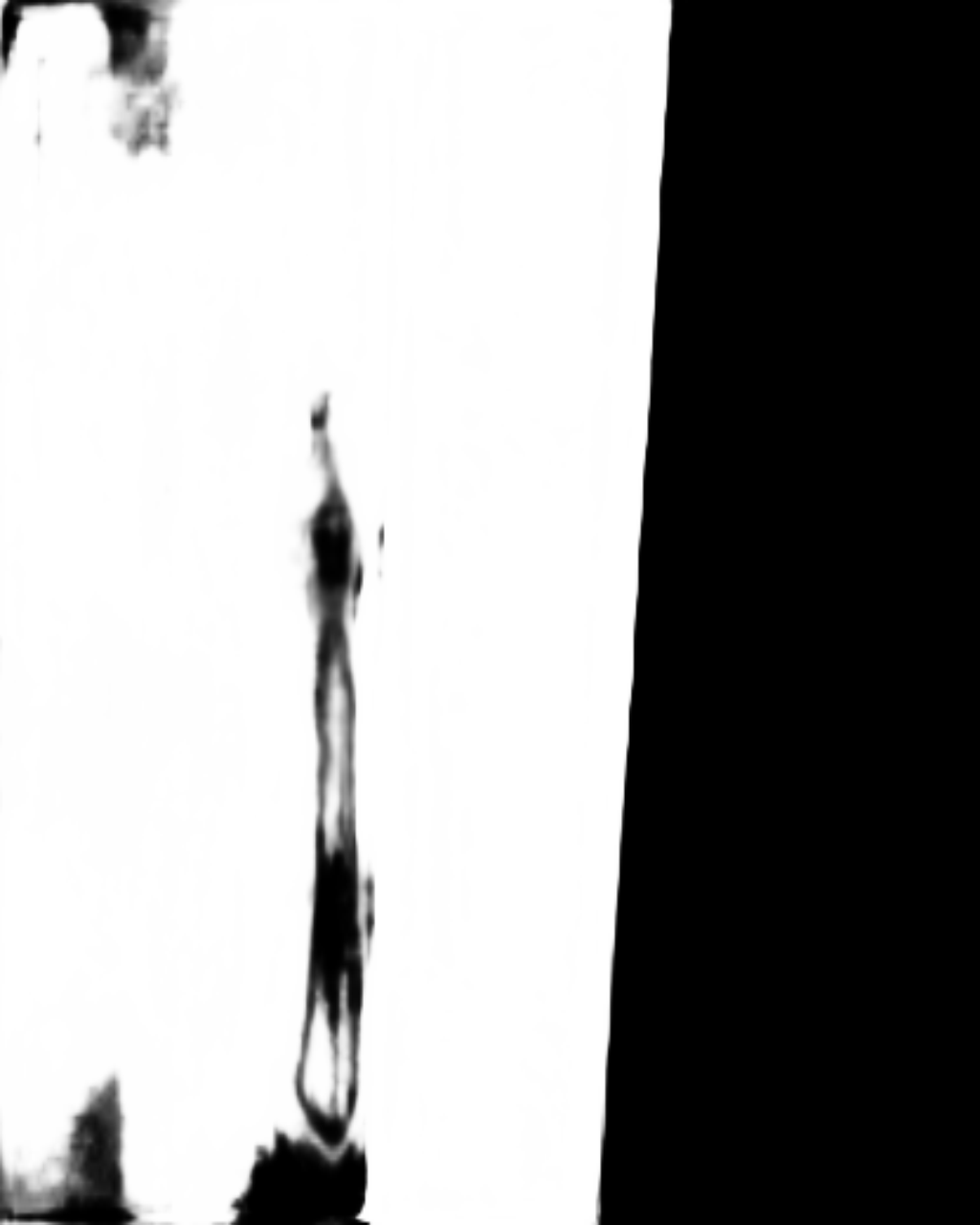}
    
                \includegraphics[width=1\linewidth]{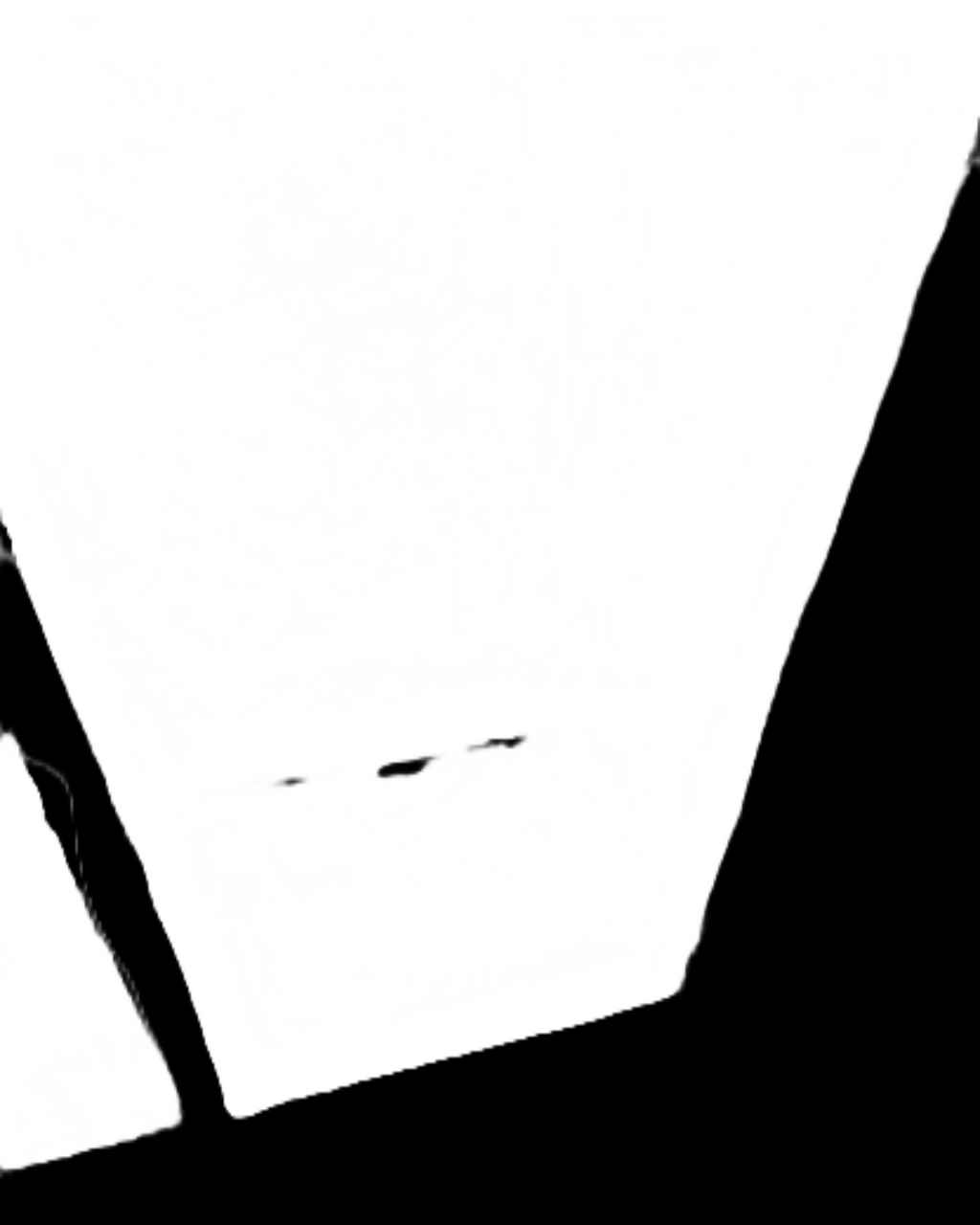}
    
                \includegraphics[width=1\linewidth]{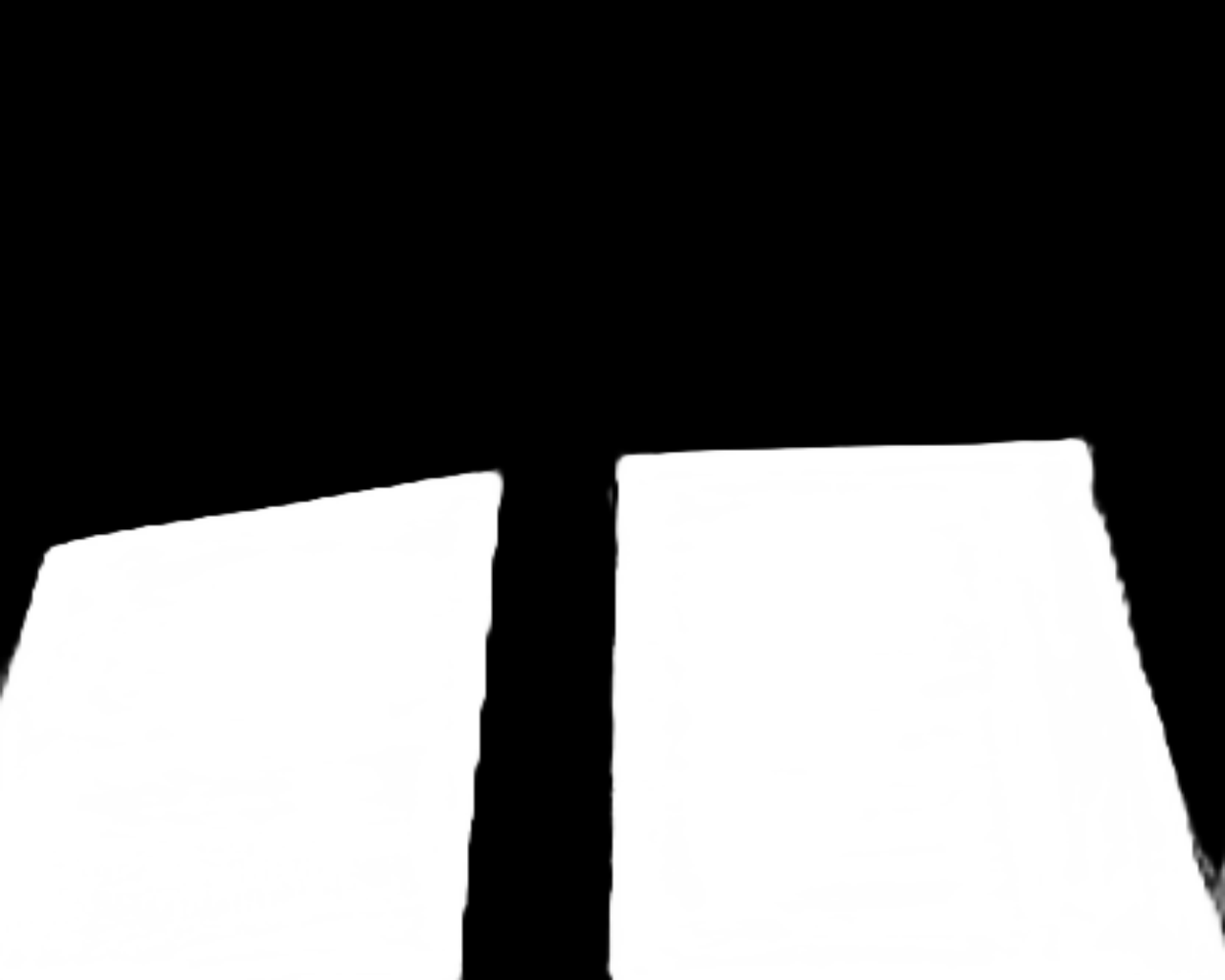}
                
                \includegraphics[width=1\linewidth]{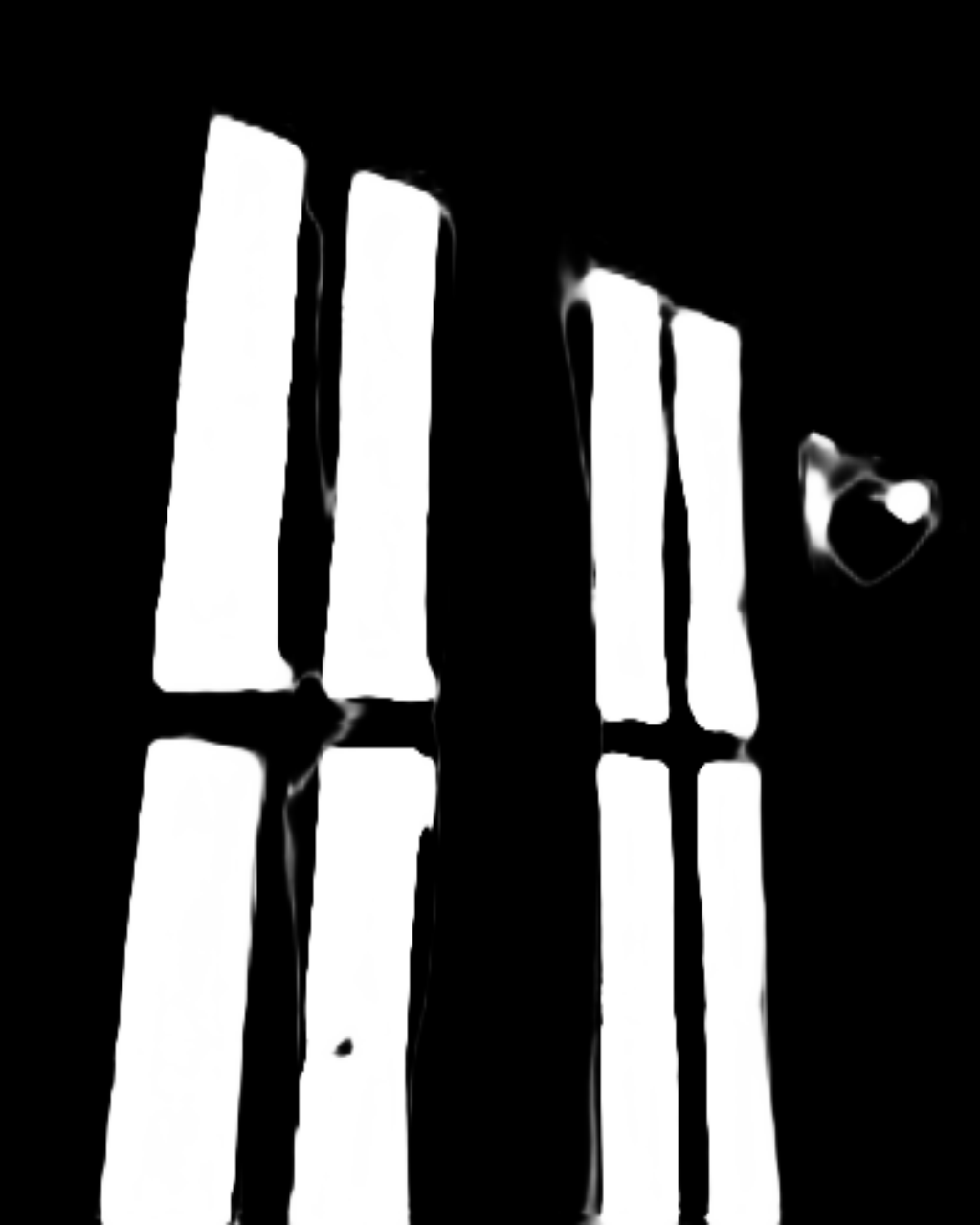}

                \includegraphics[width=1\linewidth]{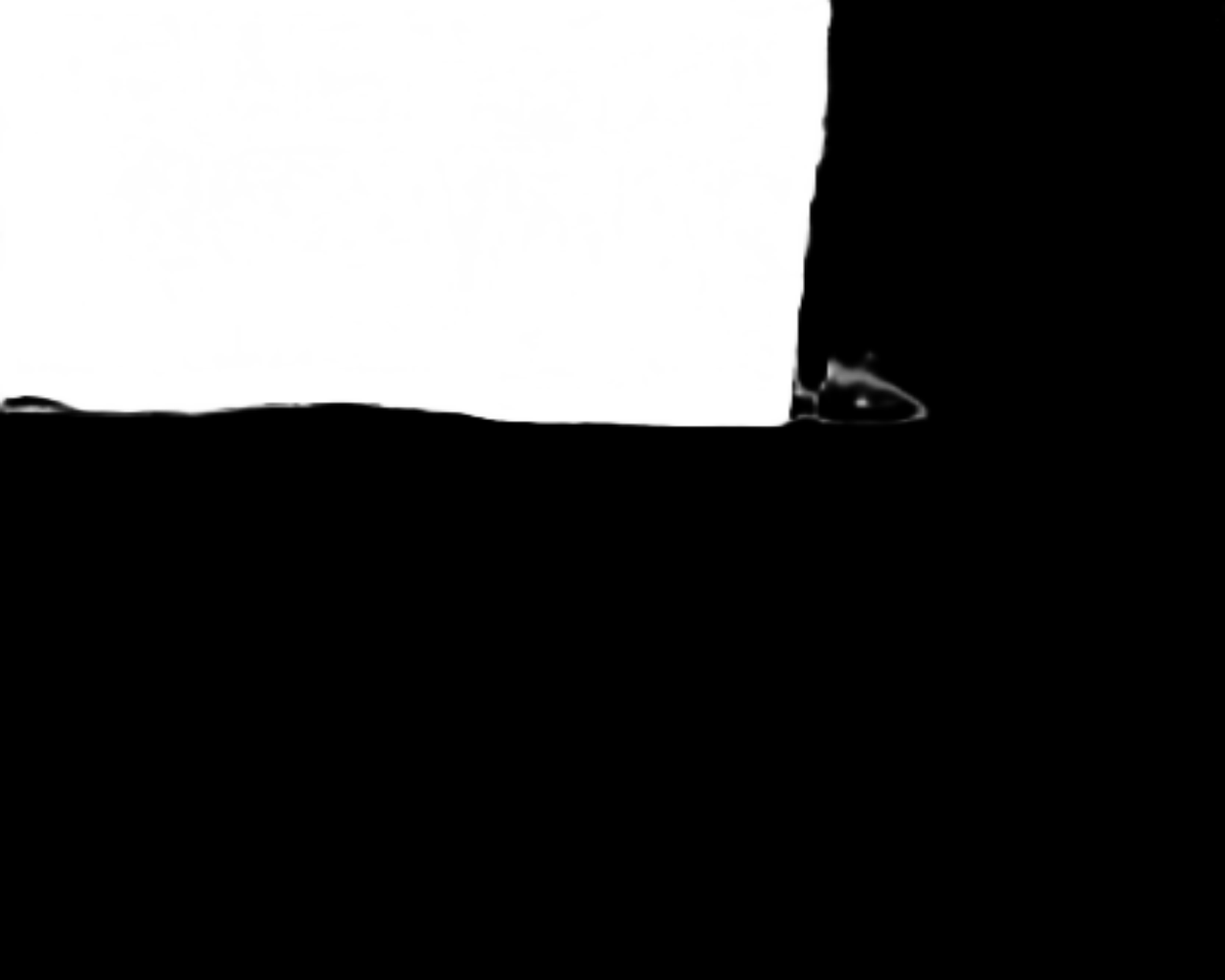}
    
                \includegraphics[width=1\linewidth]{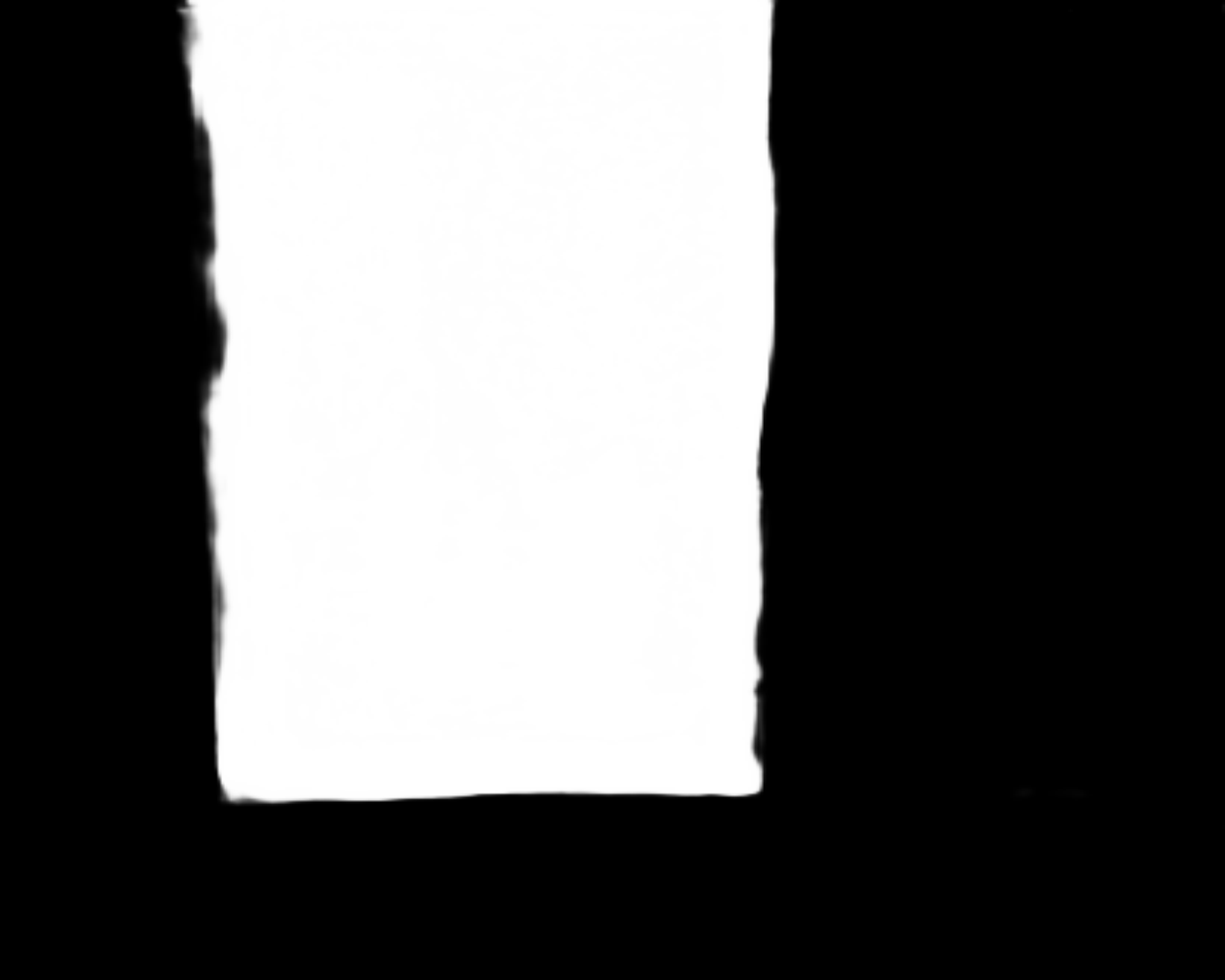}
                
                \includegraphics[width=1\linewidth]{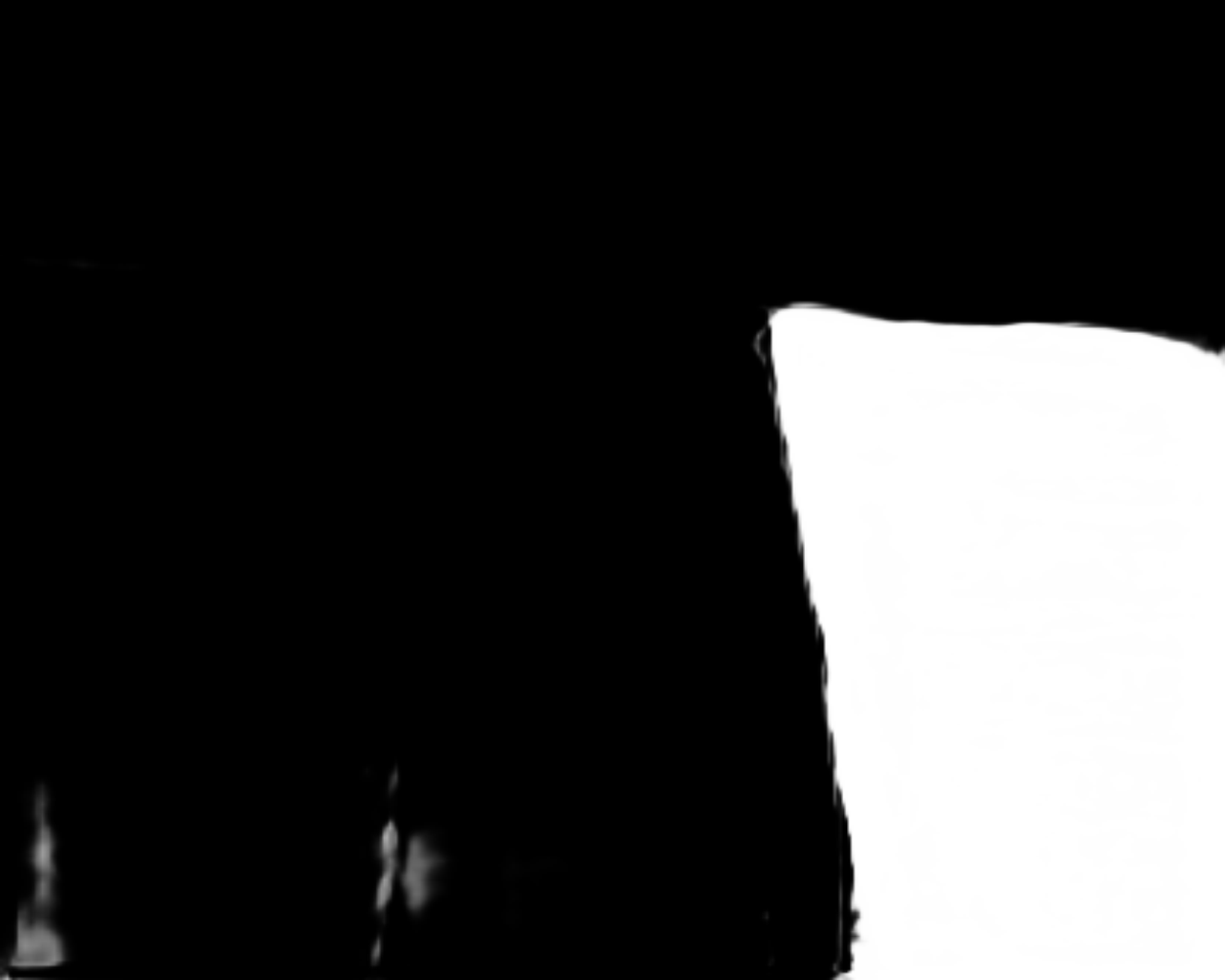}
    
                \includegraphics[width=1\linewidth]{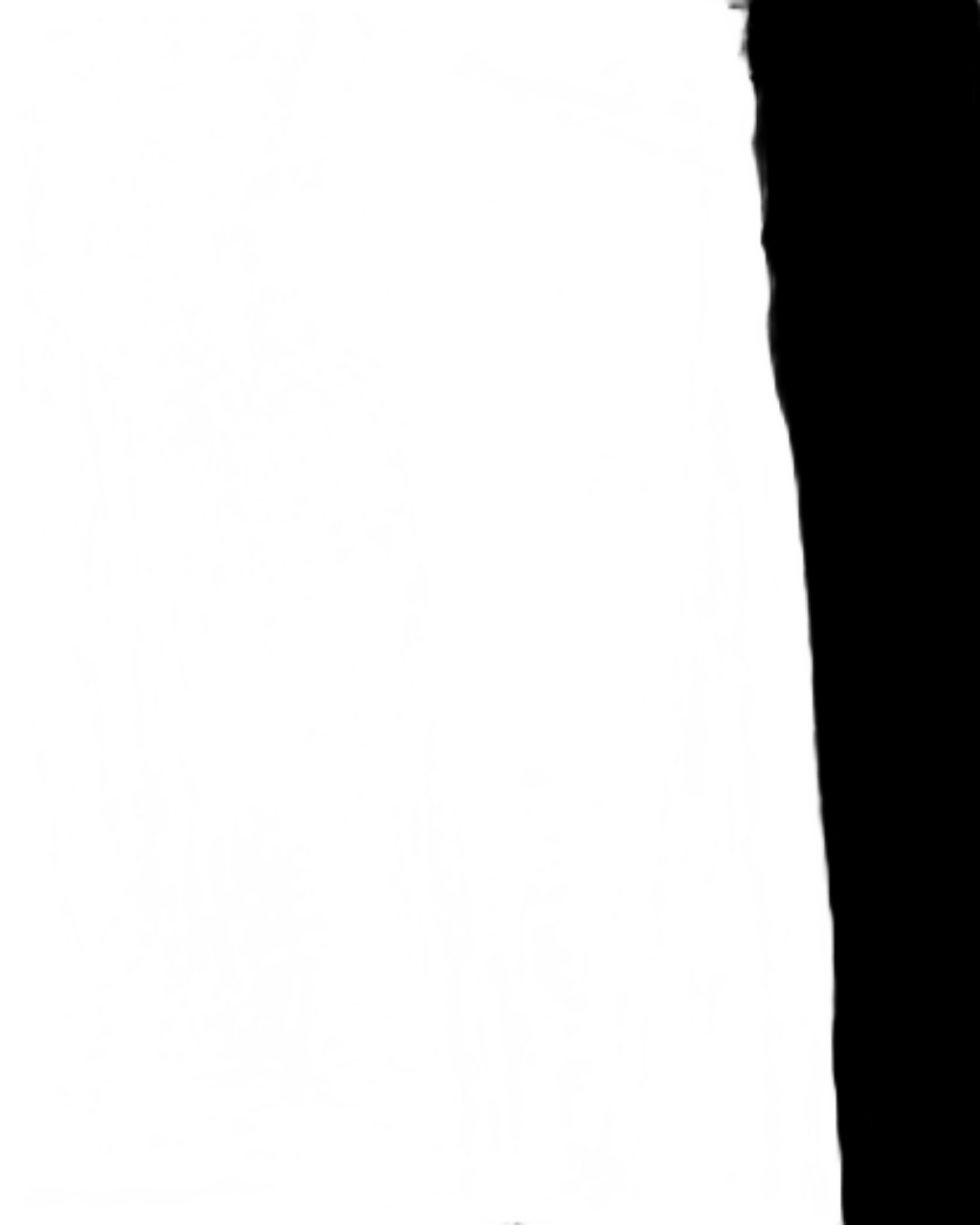}
    
                \includegraphics[width=1\linewidth]{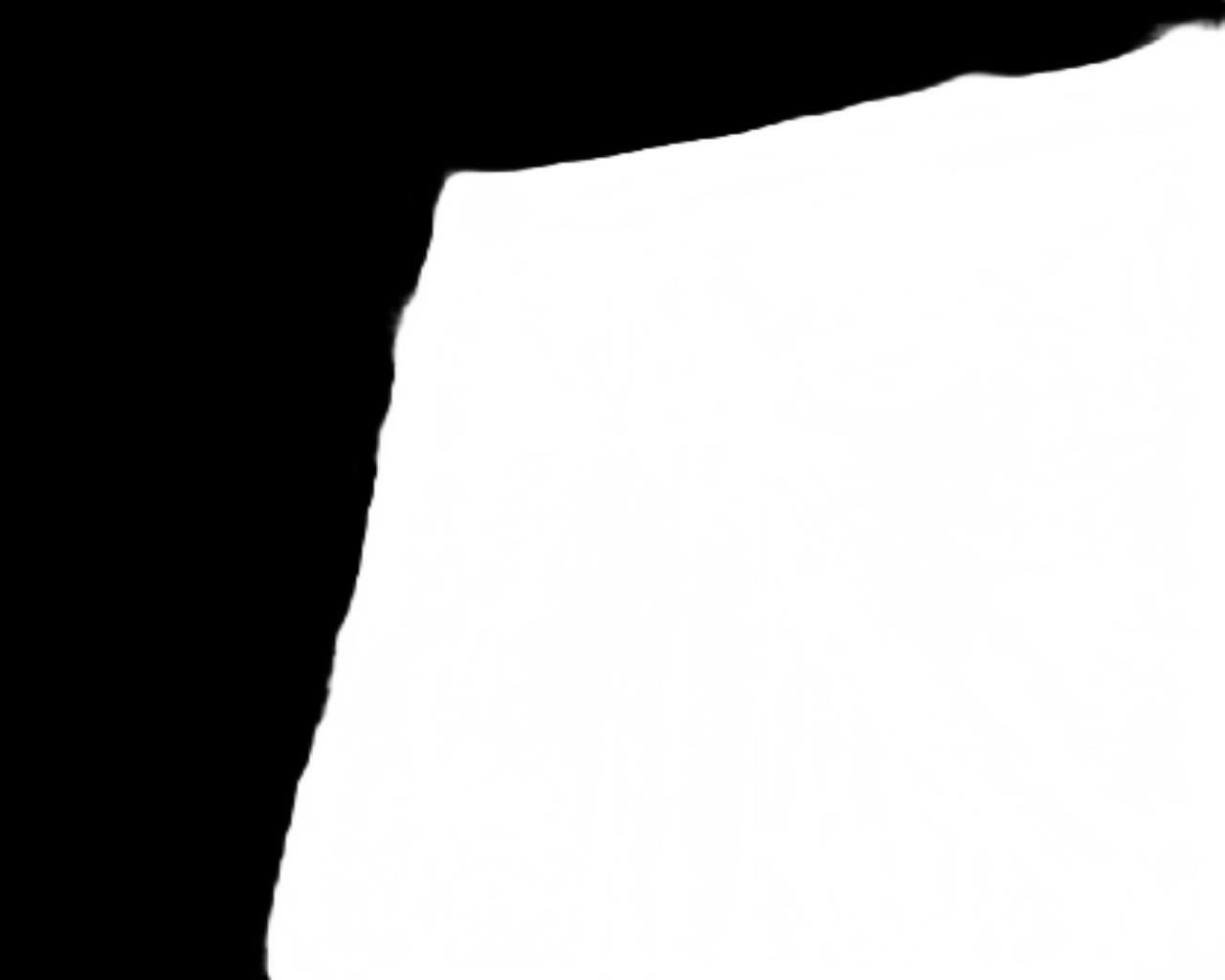}

          \end{minipage}
          }  
        \subfloat[GT]{
          \begin{minipage}[t]{0.08\textwidth}
                \centering

                \includegraphics[width=1\linewidth]{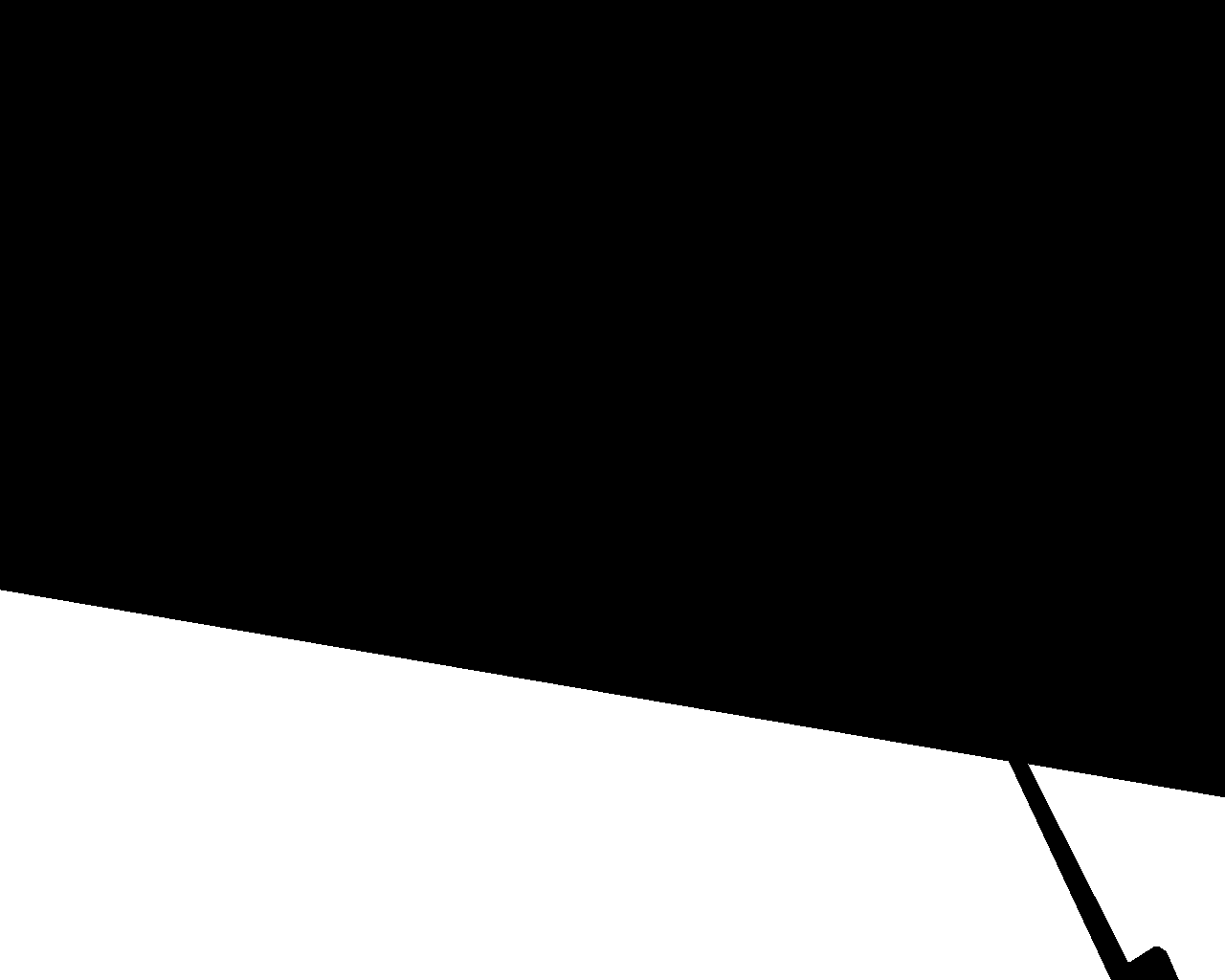}
                
                \includegraphics[width=1\linewidth]{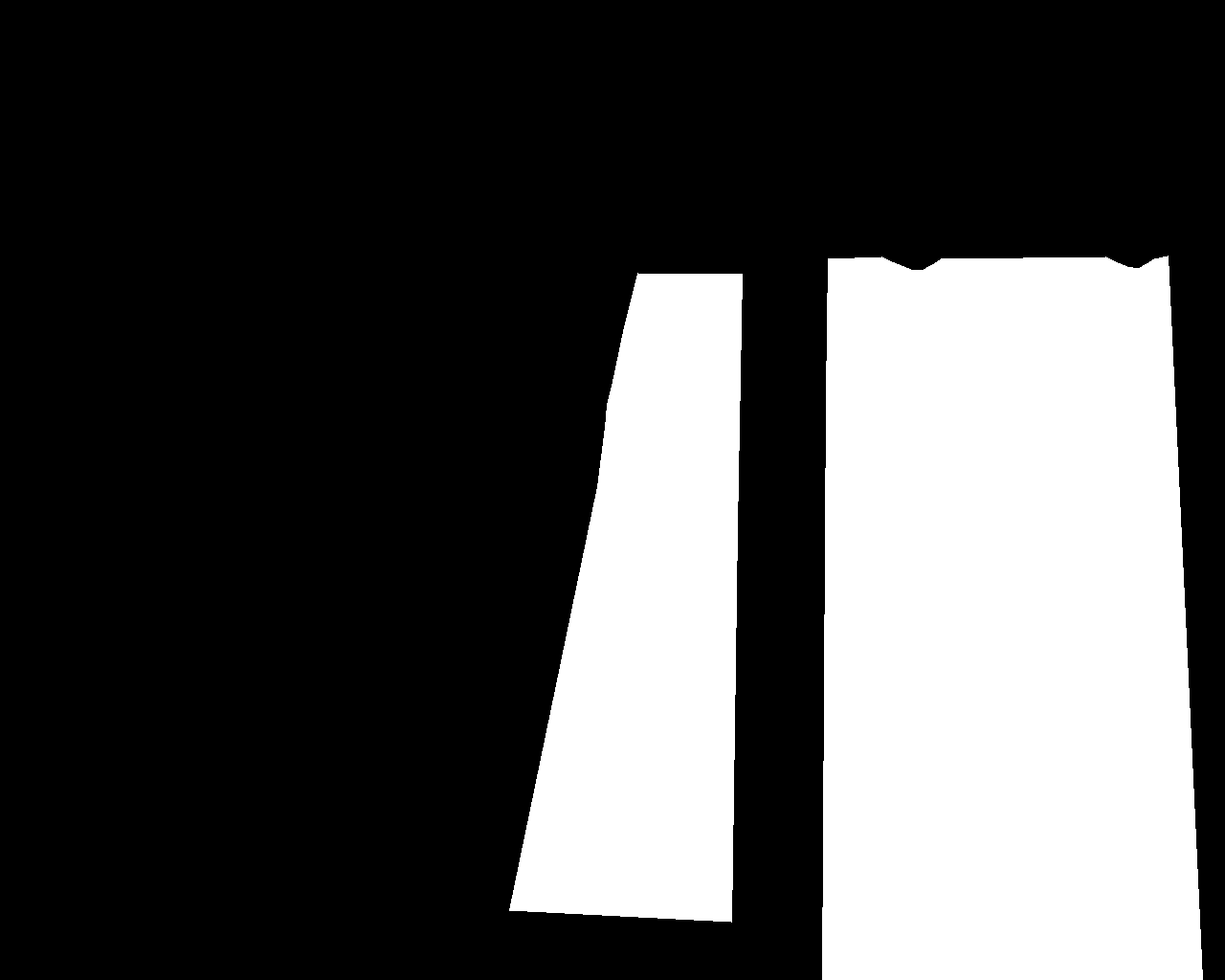}
    
                \includegraphics[width=1\linewidth]{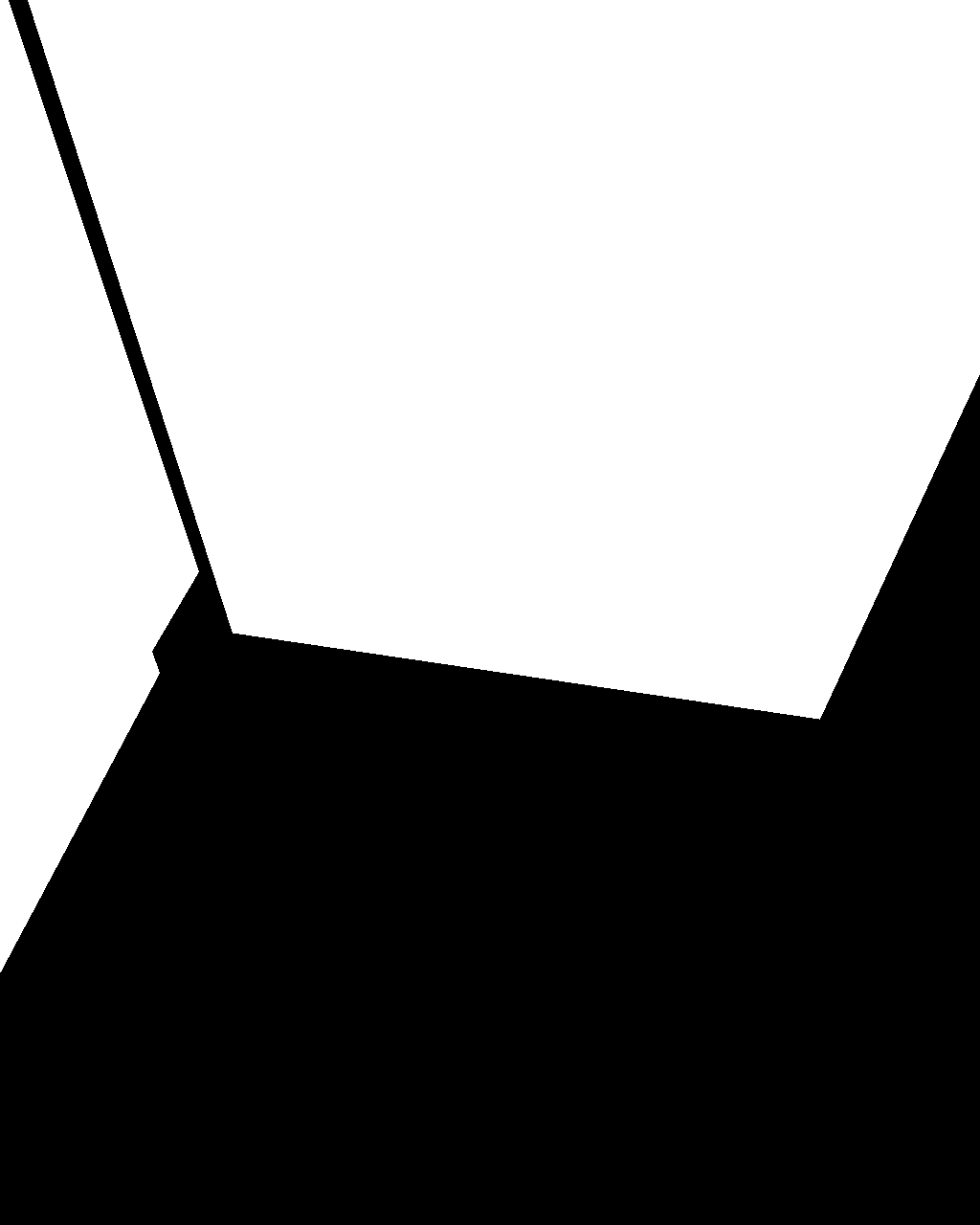}
                
                \includegraphics[width=1\linewidth]{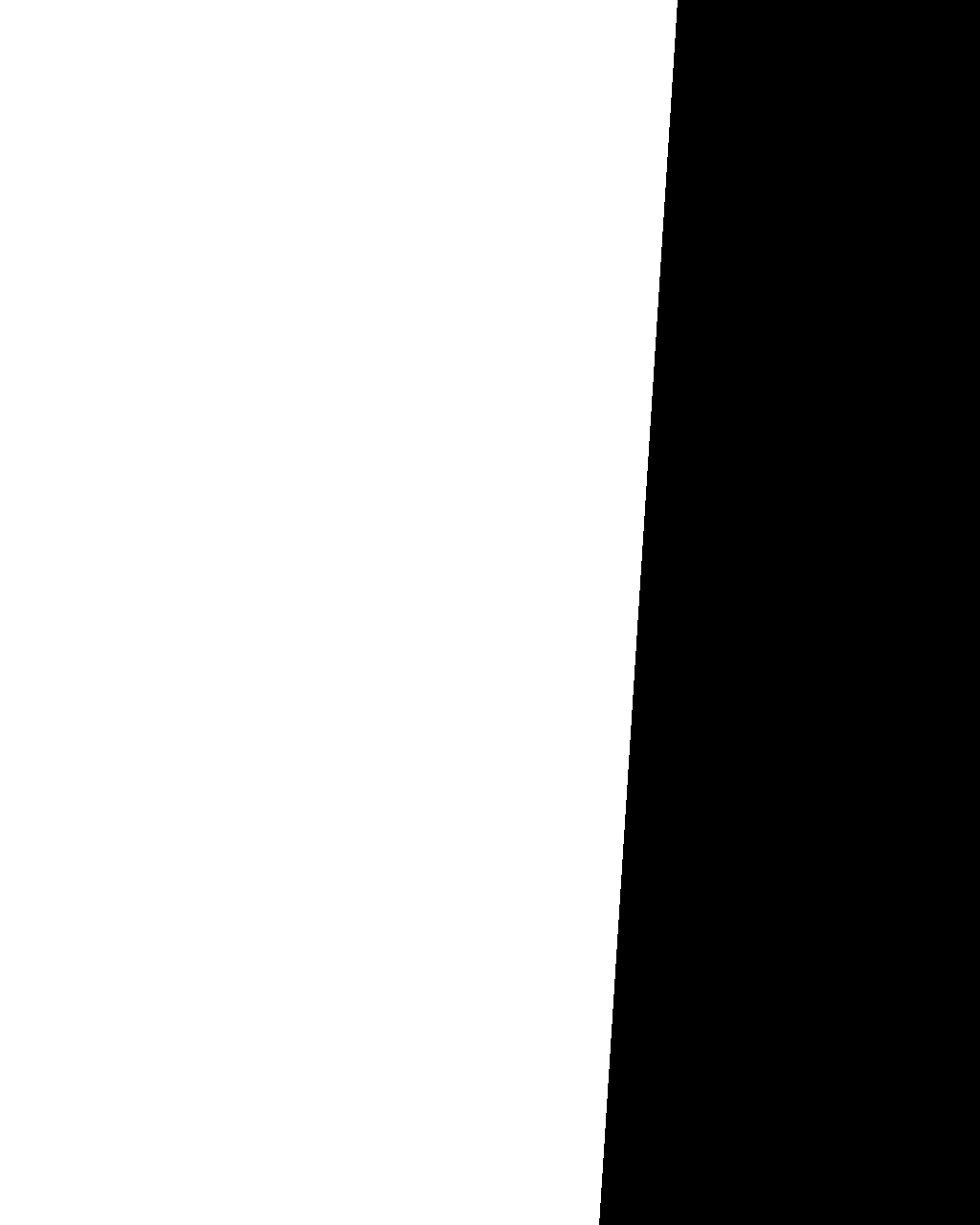}
    
                \includegraphics[width=1\linewidth]{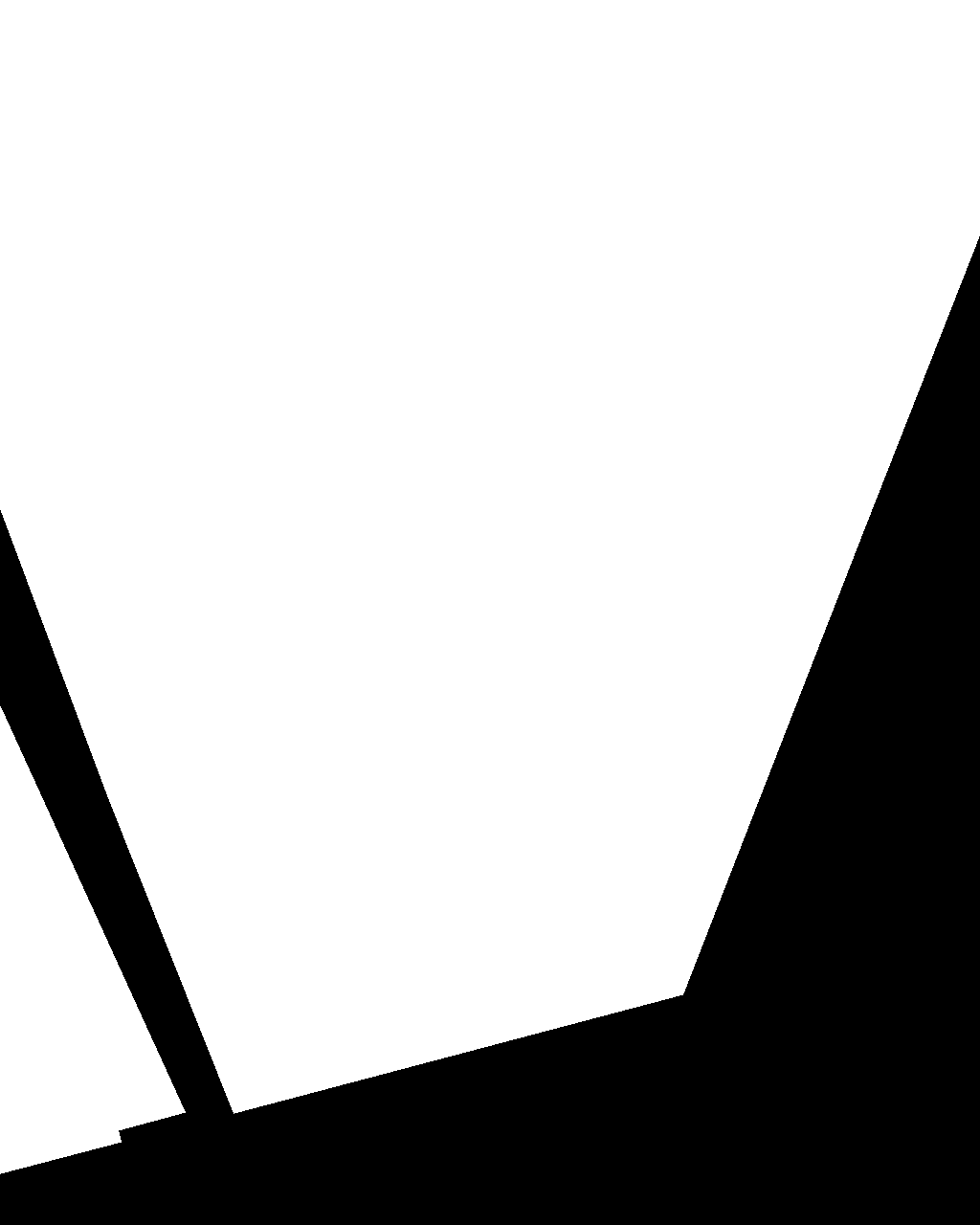}
    
                \includegraphics[width=1\linewidth]{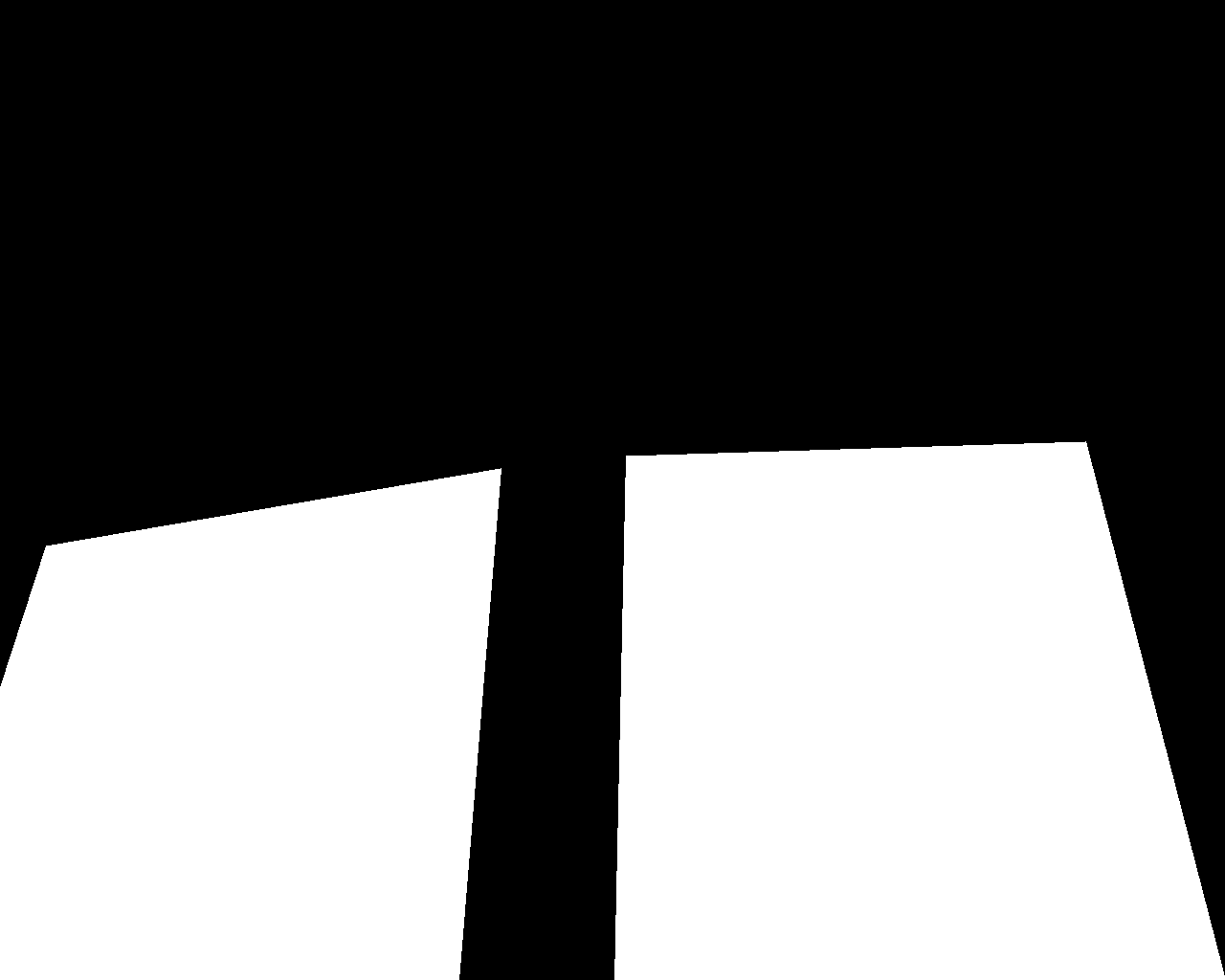}
                
                \includegraphics[width=1\linewidth]{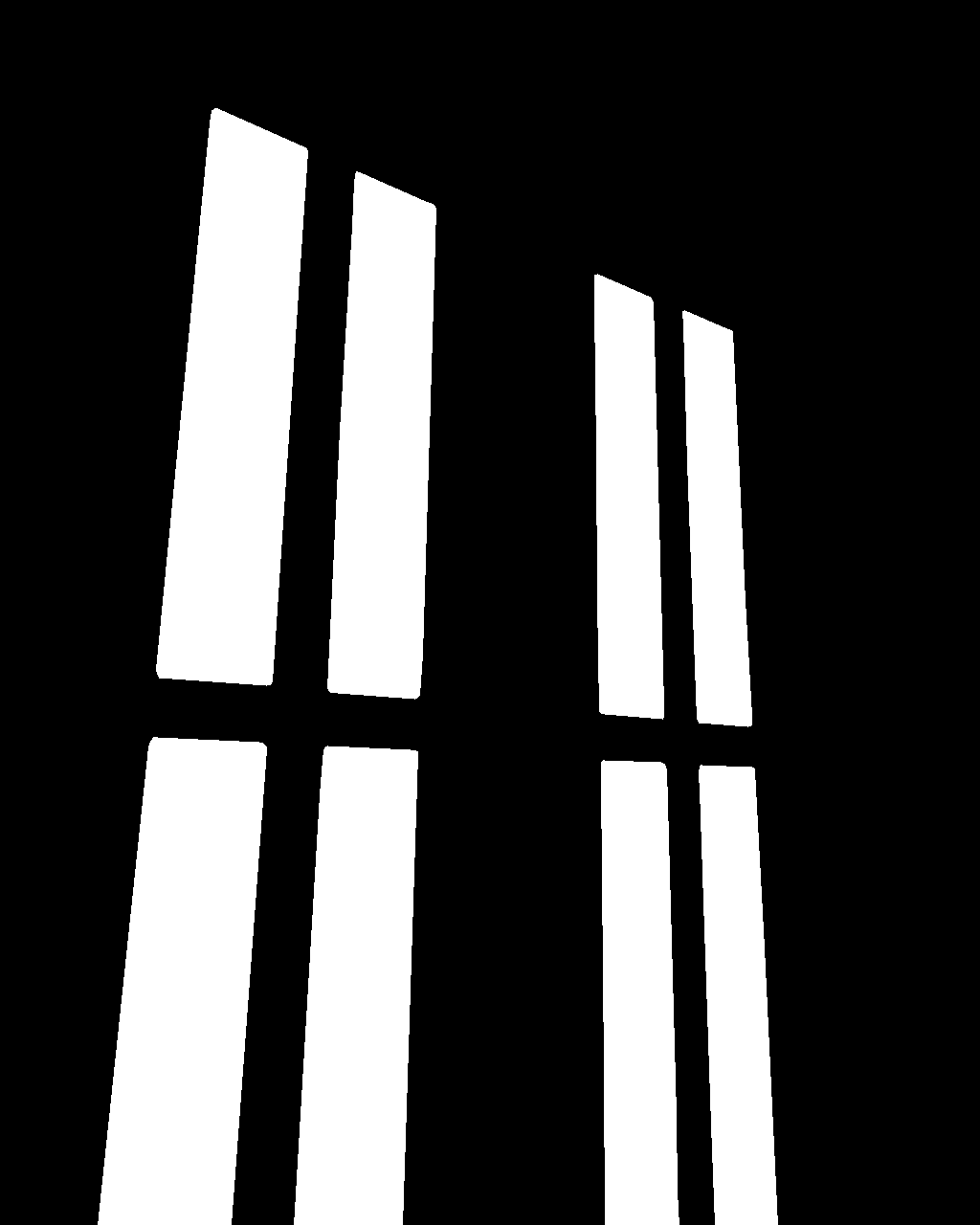}

                \includegraphics[width=1\linewidth]{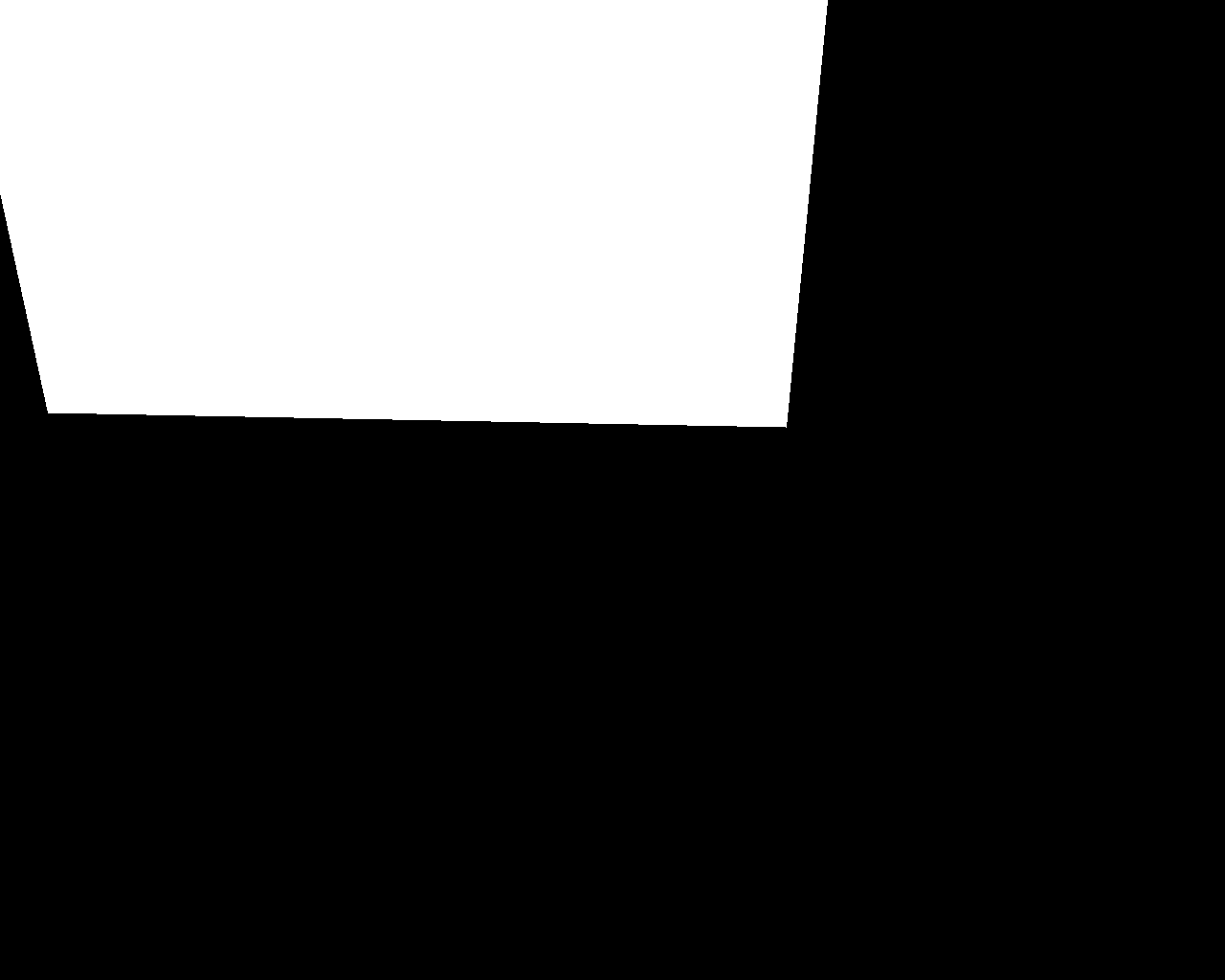}
    
                \includegraphics[width=1\linewidth]{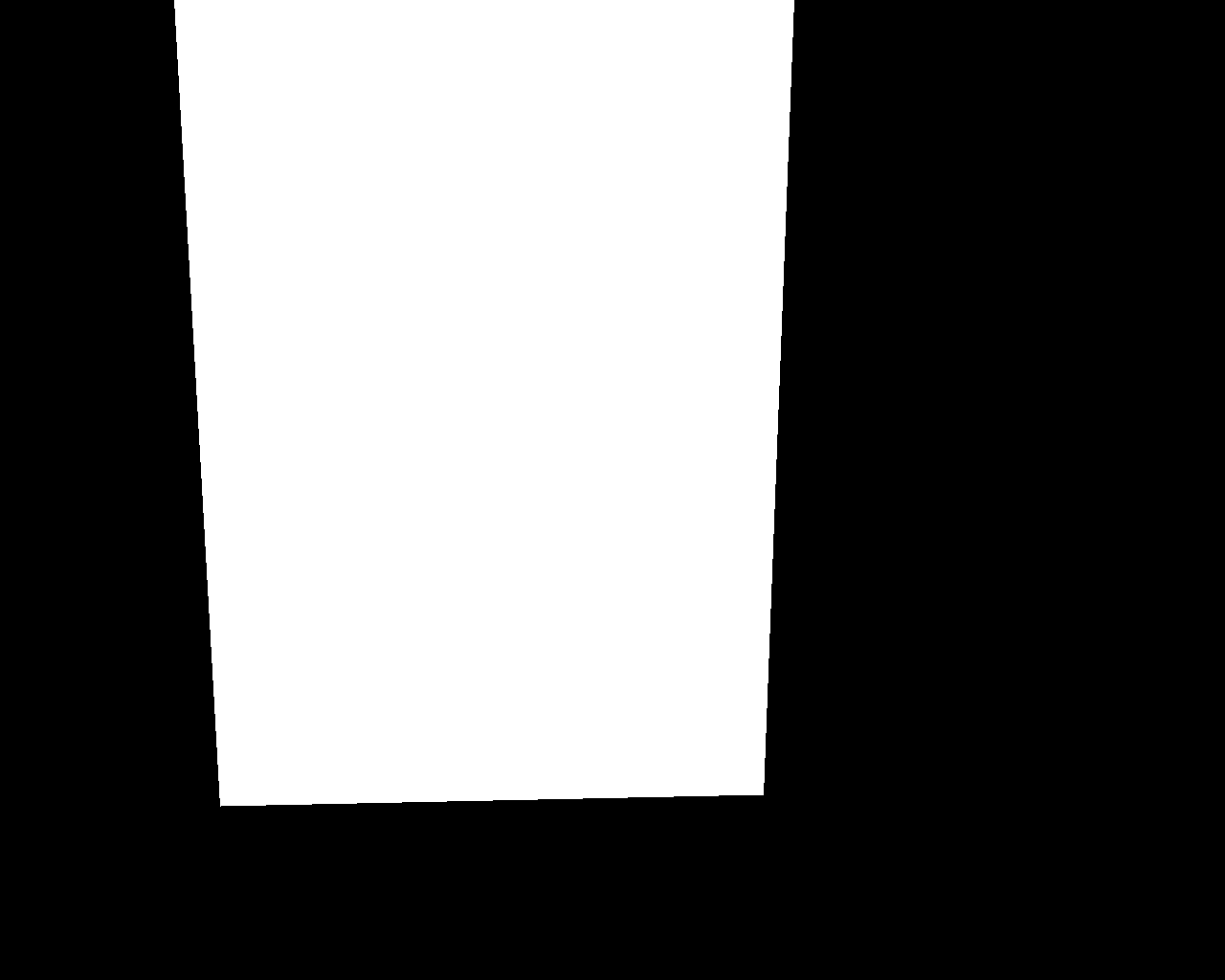}
                
                \includegraphics[width=1\linewidth]{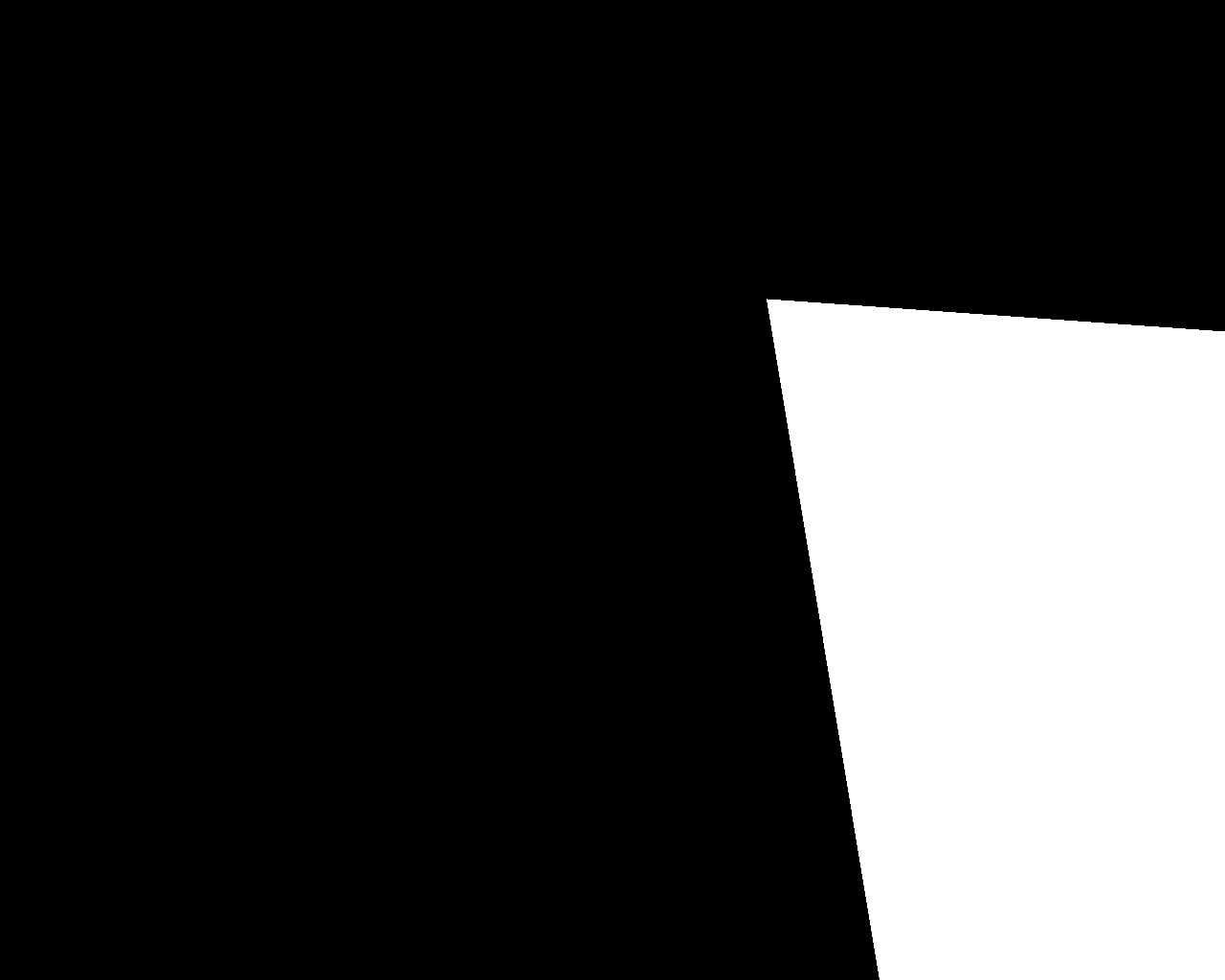}
    
                \includegraphics[width=1\linewidth]{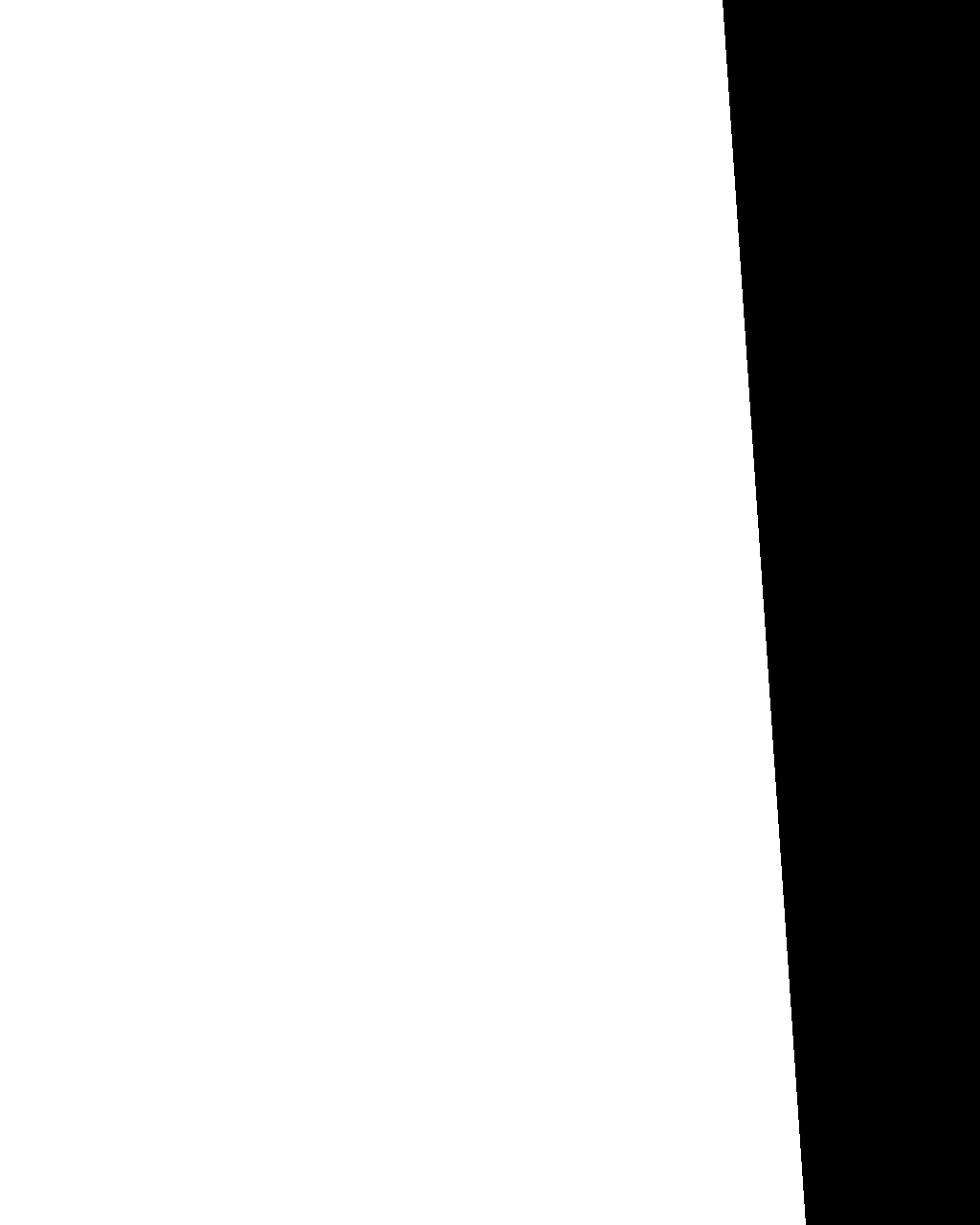}
    
                \includegraphics[width=1\linewidth]{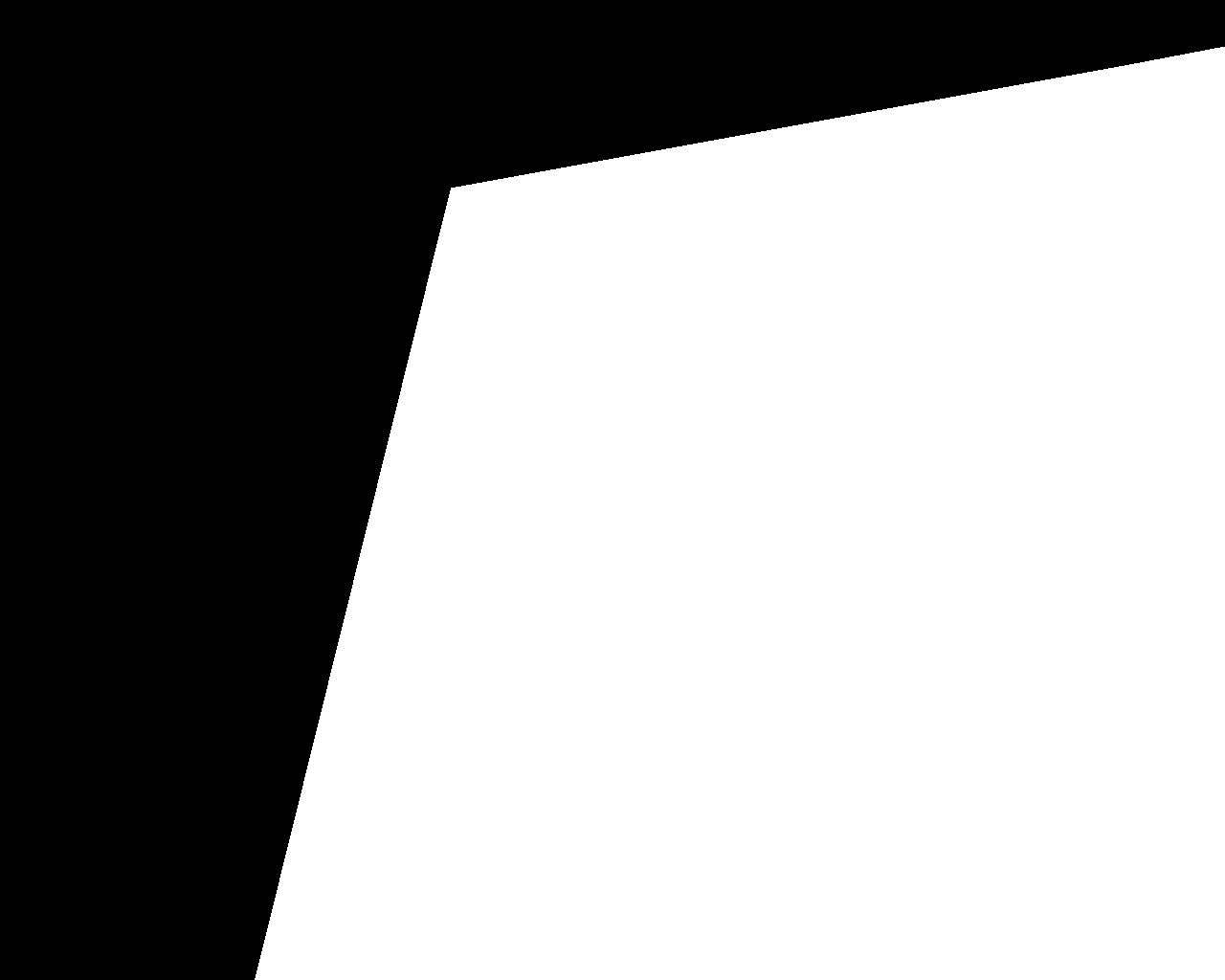}

          \end{minipage}
          } 
    
      \caption{
      Visual comparison of our GlassSegNet with state-of-the-art methods. }
      \label{fig:QE}
\end{figure*}

Besides the quantitative comparison, we illustrate the visualization of prediction results generated by some typical methods and our GlassSegNet in \Cref{fig:QE}. 
First, our GlassSegNet uses the Edge Block to extract edge information. It can preserve the transparent glass boundaries (see the 1st to 3rd rows). 
Second, the process of false-negative mistake correction can enhance the false-negative mistake region to improve the accuracy of glass segmentation (see the 4th to 6th rows). 
Third, we also reduce the impact of false-positive mistake regions by identifying them through our false-positive mistake correction (see the 7th to 10th rows). 
Fourth, the CCSA can effectively identify  large glass regions by capturing  long-range semantic dependencies in terms of spatial position (see the 11th and 12th rows). 
In brief,  our GlassSegNet is designed to imitate the mistake correction behavior of humans, leading to quality glass segmentation results.

We also retrain our GlassSegNet on the GSD dataset \cite{lin2021rich} and present the results in \Cref{tab:GSD}. It can be seen that our GlassSegNet achieves comparable performance against state-of-the-art techniques.  
Compared with the state-of-the-art glass segmentation method PGSNet \cite{yu2022progressive}, our method improves IoU by 2.5\% and reduces BER by 0.40\%. 

\begin{table}[]
 \centering
\begin{tabular}{c|c|ccc}
\hline
\multirow{3}{*}{Methods} & \multirow{3}{*}{Pub.'Year} & \multicolumn{3}{c}{GSD}                                                        \\ \cline{3-5} 
                         &                            & \multicolumn{3}{l}{Trainset:4427 Testset:1044}                                 \\ \cline{3-5} 
                         &                            & \multicolumn{1}{c|}{IoU$\uparrow$}       & \multicolumn{1}{c|}{MAE$\downarrow$}   & BER$\downarrow$  \\ \hline
      TransLab                 & ECCV'20                    & \multicolumn{1}{c|}{78.05}  & \multicolumn{1}{c|}{0.069} & 9.19 \\
      Trans2Seg                & IJCAI'21                   & \multicolumn{1}{c|}{79.65}  & \multicolumn{1}{c|}{0.069} & 8.21 \\ \hline
      GDNet                    & CVPR'20                    & \multicolumn{1}{c|}{82.51}  & \multicolumn{1}{c|}{0.058} & 6.41 \\
      GSD                      & CVPR'21                    & \multicolumn{1}{c|}{83.64}  & \multicolumn{1}{c|}{0.055} & 6.12 \\
      EBLNet                   & ICCV'21                    & \multicolumn{1}{c|}{79.66}       & \multicolumn{1}{c|}{0.068}      & 9.03     \\
      PGSNet                   & IEEE TIP'22                & \multicolumn{1}{c|}{83.65}  & \multicolumn{1}{c|}{0.054} & 6.25 \\ \hline
      GlassSegNet              & Ours                       & \multicolumn{1}{c|}{\textbf{86.15}}  & \multicolumn{1}{c|}{\textbf{0.054}} & \textbf{5.85} \\ \hline
      \end{tabular}
      \caption{Comparison between our GlassSegNet and state-of-the-art glass segmentation methods on GSD \cite{lin2021rich}.}
      \label{tab:GSD}
\end{table}


In addition, we retrain GlassSegNet on the multi-modal glass segmentation dataset RGB-T and RGB-P and present the results in \Cref{tab:MGSD} and \Cref{tab:MGSD1}. It can be seen that our GlassSegNet achieves comparable performance against the single-modal based methods, but it may be inferior to the network using multi-modal information. 

\begin{table}[]
 \centering
      \begin{tabular}{c|c|ccc}
      \hline
      \multirow{3}{*}{Methods} & \multirow{3}{*}{Pub.'Year} & \multicolumn{3}{c}{RGB-T}                                                                     \\ \cline{3-5} 
                               &                            & \multicolumn{3}{c}{Trainset:3070 Testset:1782}                                                \\ \cline{3-5} 
                               &                            & \multicolumn{1}{c|}{IoU$\uparrow$}        & \multicolumn{1}{c|}{MAE$\downarrow$}   & BER$\downarrow$    \\ \hline
      TransLab                 & ECCV'20                    & \multicolumn{1}{c|}{71.28}    & \multicolumn{1}{c|}{0.165}      &   18.17    \\
      Trans2Seg                & IJCAI'21                   & \multicolumn{1}{c|}{67.68}    & \multicolumn{1}{c|}{0.210}      &   22.99     \\ \hline
      GDNet                    & CVPR'20                    & \multicolumn{1}{c|}{89.65}    & \multicolumn{1}{c|}{0.143}      &   6.53     \\
      GSD                      & CVPR'21                    & \multicolumn{1}{c|}{90.47}    & \multicolumn{1}{c|}{0.151}      &   6.94    \\
      EBLNet                   & ICCV'21                    & \multicolumn{1}{c|}{80.54}    & \multicolumn{1}{c|}{0.113}      &   10.30 \\
      PGSNet                   & IEEE TIP'22                & \multicolumn{1}{c|}{91.07}    & \multicolumn{1}{c|}{0.132}      &   6.26   \\ \hline
      RGB-only                 & Arxiv                      & \multicolumn{1}{c|}{88.94}    & \multicolumn{1}{c|}{0.056}      &   6.61  \\
      RGB-T                    & Arxiv                      & \multicolumn{1}{c|}{\textbf{93.80}}    & \multicolumn{1}{c|}{\textbf{0.027}}      &   \textbf{4.07}  \\ \hline
      GlassSegNet              & Ours                       & \multicolumn{1}{c|}{\textcolor{red}{94.19}}   & \multicolumn{1}{c|}{\textcolor{red}{0.051}}       &   \textcolor{red}{6.18}       \\ \hline
      \end{tabular}
      \caption{Comparison between our GlassSegNet and state-of-the-art glass segmentation methods on RGB-T \cite{huo2022glass}.}
      \label{tab:MGSD}
\end{table}

\begin{table}[]
 \centering
      \begin{tabular}{c|c|ccc}
      \hline
      \multirow{3}{*}{Methods} & \multirow{3}{*}{Pub.'Year} & \multicolumn{3}{c}{RGB-P}                                                                    \\ \cline{3-5} 
                               &                            & \multicolumn{3}{c}{Trainset:3207 Testset:1304}                                               \\ \cline{3-5} 
                               &                            & \multicolumn{1}{c|}{IoU$\uparrow$}      & \multicolumn{1}{c|}{MAE$\downarrow$}   & BER $\downarrow$  \\ \hline
      TransLab                 & ECCV'20                    & \multicolumn{1}{c|}{73.59}  & \multicolumn{1}{c|}{0.148} & 15.73 \\
      Trans2Seg                & IJCAI'21                   & \multicolumn{1}{c|}{75.21}  & \multicolumn{1}{c|}{0.122} & 13.23 \\ \hline
      GDNet                    & CVPR'20                    & \multicolumn{1}{c|}{77.64}  & \multicolumn{1}{c|}{0.119} & 11.79 \\
      GSD                      & CVPR'21                    & \multicolumn{1}{c|}{78.11}  & \multicolumn{1}{c|}{0.122} & 12.61 \\
      EBLNet                   & ICCV'21                    & \multicolumn{1}{c|}{68.70}  & \multicolumn{1}{c|}{0.156} & 17.52 \\
      PGSNet                   & IEEE TIP'22                & \multicolumn{1}{c|}{79.18}  & \multicolumn{1}{c|}{0.108} & 11.54 \\ \hline
      RGB-only                 & CVPR'22                    & \multicolumn{1}{c|}{77.22}  & \multicolumn{1}{c|}{0.121} & 12.07  \\
      RGB-P                    & CVPR'22                    & \multicolumn{1}{c|}{\textbf{81.08}}  & \multicolumn{1}{c|}{\textbf{0.091}} & \textcolor{red}{9.63}  \\ \hline
      GlassSegNet              & Ours                       & \multicolumn{1}{c|}{\textcolor{red}{80.92}}  & \multicolumn{1}{c|}{\textcolor{red}{0.101}} & \textbf{6.20}  \\ \hline
      \end{tabular}
      \caption{Comparison between our GlassSegNet and state-of-the-art glass segmentation methods on RGB-P \cite{huo2022glass}.}
      \label{tab:MGSD1}
\end{table}

\subsection{Ablation Study}
\label{sec:Ablat}
GlassSegNet is to segment transparent glass by mimicking the process of human recognition of transparent glass. 
We design the identification stage as a large-field processing and edge processing paradigm based on the observation that humans often recognize transparent glass by context and edge information.
We design the CCSA module as a large-field processing paradigm based on the observation that the contexts in a large receptive field helps to enhance the feature semantics. 
The correction stage is designed to perform the mistake correction for progressively refining the coarse prediction via correcting mistake regions using gained experience. 

We conduct ablation studies to validate the two key components tailored for effective features fusion towards accurate glass segmentation, i.e., the identification stage (IS) and the correction stage (CS), as well as explore the impact of different long-range contextual feature extraction modules, and report the experimental results in \Cref{tab:AS1} and \Cref{tab:AS2}.

\begin{table}[]
 \centering
      \begin{tabular}{cc|ccc}
      \hline
      \multicolumn{2}{c|}{\multirow{2}{*}{Methods}} & \multicolumn{3}{c}{HSO} \\ \cline{3-5} 
      \multicolumn{2}{c|}{}                         & IoU$\uparrow$   & MAE$\downarrow$  & BER$\downarrow$  \\ \hline
      A                     & B                     & 81.34 & 0.105 & 11.57 \\
      B                     & B + IS w/o CCSA       & 83.52 & 0.095 & 11.03 \\
      C                     & B + IS w/o edge       & 83.34 & 0.091 & 10.59 \\
      D                     & B + IS                & 84.42 & 0.092 & 10.50 \\ \hline
      E                     & B + MC w/o FN        & 82.50 & 0.093 & 9.96  \\
      F                     & B + MC w/o FP        & 82.98 & 0.094 & 9.16 \\
      G                     & B + MC               & 84.90 & 0.087 & 8.85  \\ \hline 
      H                     & B + IS + CS (w/o mis-sup)      & 81.59 & 0.100 & 9.42 \\
      I                     & B + IS + CS (w/o edge-sup)     & 83.45 & 0.089 & 9.62 \\
      J                     & B + IS + CS (w/o iou-loss)     & 82.98 & 0.099 & 10.38 \\ \hline
      K                     & B + IS + CS (our GlassSegNet)  & \textbf{85.33}   & \textbf{0.088} & \textbf{8.40} \\ \hline
      \end{tabular}
      \caption{Quantitative ablation results.  Each component in our GlassSegNet contributes to the overall performance.}
      \label{tab:AS1}
\end{table}

\begin{table}[]
 \centering
        \begin{tabular}{c|c|c|ccc}
        \hline
        \multirow{2}{*}{Methods} & \multirow{2}{*}{Pub.'Year} & \multirow{2}{*}{Module} & \multicolumn{3}{c}{HSO} \\ \cline{4-6} 
                                 &                            &                         & IoU$\uparrow$    & MAE$\downarrow$    & BER$\downarrow$   \\ \hline
        DenseASPP \cite{yang2018denseaspp}                 & CVPR'18                    & DenseASPP                & 82.13  & 0.090  & 8.64  \\
        SoINet  \cite{su2019selectivity}                 & ICCV'19                    & TDS                     & 81.97  & 0.097  & 9.55  \\
        TransLab   \cite{xie2020segmenting}              & ECCV'20                    & ASPP                    & 81.34  & 0.093  & 9.24  \\
        GDNet \cite{mei2020don}                   & CVPR'20                    & LCFI                    & 84.80  & 0.097  & 9.27  \\
        GSD     \cite{lin2021rich}                 & CVPR'21                    & RCAM                    & 85.20  & 0.091  & 8.86  \\
        MirrorNet \cite{yang2019my}               & ICCV'19                    & CCFE                    & 82.30  & 0.090  & 8.86  \\
        DCENet  \cite{mei2021exploring}                 & TCSVT'21                   & DCE                     & 82.80  & 0.089  & 8.51  \\
        PIDNet   \cite{su2021pixel}                & ICCV'21                    & CDCM                    & 85.20  & 0.092  & 8.92  \\
        CCNet   \cite{huang2019ccnet}                &  ICCV'19         &  RCCA      & 84.57  & 0.095  & 9.51  \\
        GlassSegNet              & Ours                        & CCSA                    & \textbf{85.33}  & \textbf{0.088}  & \textbf{8.40}  \\ \hline
        \end{tabular}
      \caption{Comparison under different long-range contextual feature extractors.}
      \label{tab:AS2}
\end{table}

1) The effectiveness of the identification stage: We first define and train a base model ``B'' (i.e., A in \Cref{tab:AS1}) which is based on GlassSegNet. 
Starting from the base model, we reintroduce the CCSA module and the edge block in the identification stage (B, C), respectively. Then both modules are used in the identification stage (D). 
From the results, we observe that: 
(i) the identification stage can help boost the segmentation performance largely (i.e., D is better than A)
 as IS uses long-range contextual information and glass boundary information to locate the region of glass;
(ii) applying long-range contextual feature extractors on the different level features before fusing them together is helpful (i.e., B is better than A), indicating that long-range contextual feature extractors are essential for localizing the glass region; 
(iii) edge information can help preserve the glass boundary (i.e., C is better than A). 

2) The effectiveness of the correction stage: Compared to the base model (i.e., A in \Cref{tab:AS1}), FP branch (E) and FN branch (F) in the mistake correction module perform similarly. The mistake correct module, which contains FP branch and FN branch, can greatly improve the segmentation performance (i.e., G is better than A).

3) The effectiveness of different loss functions: From  \Cref{tab:AS1}, we know that our loss function performs well for network training.

\subsection{Different Long-range Contextual Feature Extractors}
 Due to the diverse appearance of glass, the glass segmentation task relies heavily on contextual information. Existing glass segmentation methods capture contextual information through well-designed large-scale contextual feature extractors. These feature extractors are often based on improvements of the ASPP module. For example, GDNet \cite{mei2020don} design the LCFI module by introducing spatially separated convolution into the ASPP, to reduce the number of parameters. GSD \cite{lin2021rich} further expands the receptive field with the RCAM module which introduces an additional pooling branch. Our CCSA module aggregates global contexts based on the non-local module \cite{wang2018non} and reduces the number of parameters by criss-cross and strip pooling operations. We compared with some long-range contextual feature extractors from prior works \cite{yang2018denseaspp, su2019selectivity, xie2020segmenting, mei2020don, lin2021rich, yang2019my, mei2021exploring, su2021pixel, wang2018non, huang2019ccnet, fu2019dual} in \Cref{tab:AS2}. We can observe that our GlassSegNet achieve better results than other methods, showing the superiority of our method over them. 

\subsection{Failure Cases}
\begin{figure}[htb]
      \centering
      \subfloat[Input]{\label{image}
      \begin{minipage}[t]{0.15\textwidth}
      \centering
      \includegraphics[width=1\linewidth]{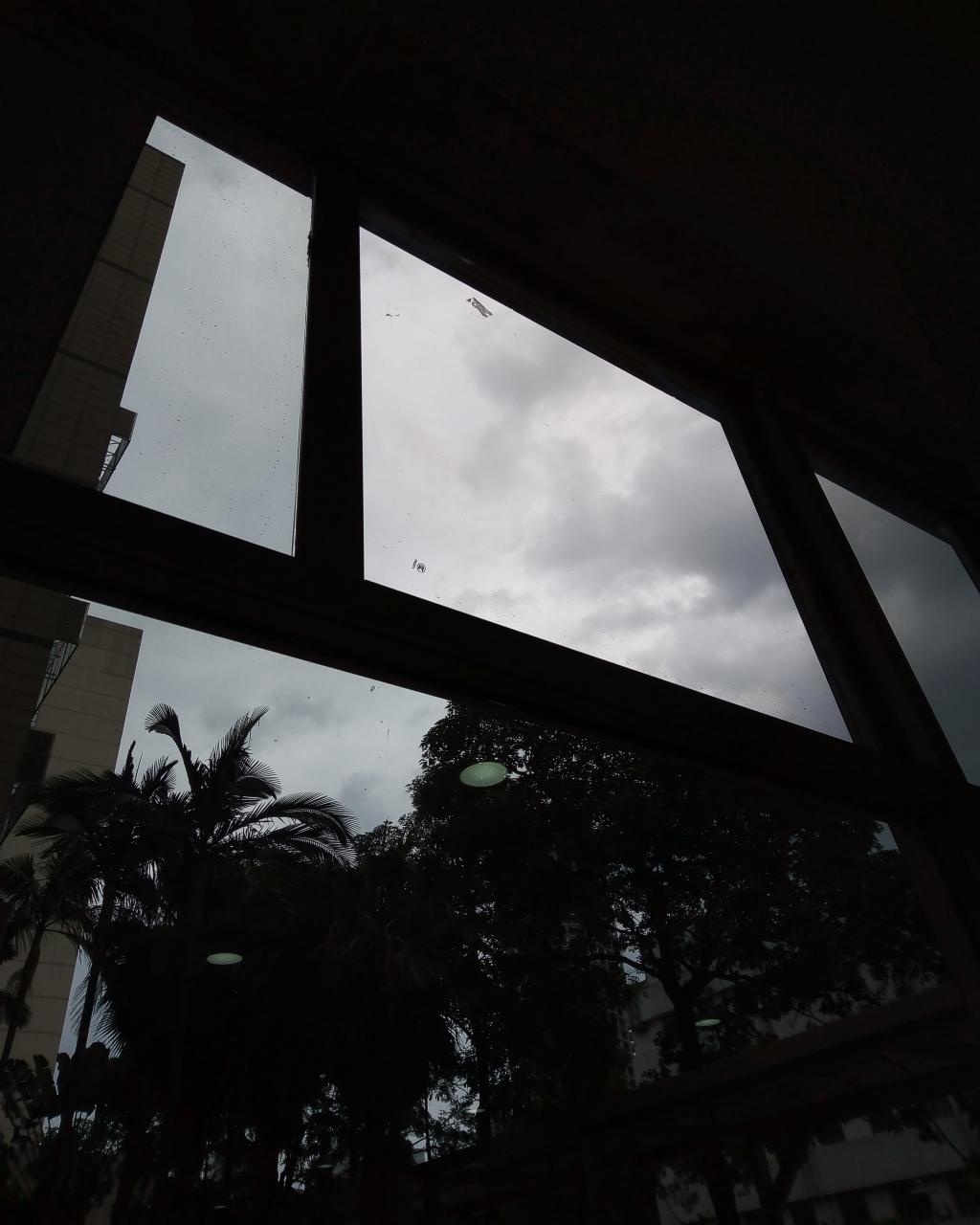}
    
      \includegraphics[width=1\linewidth]{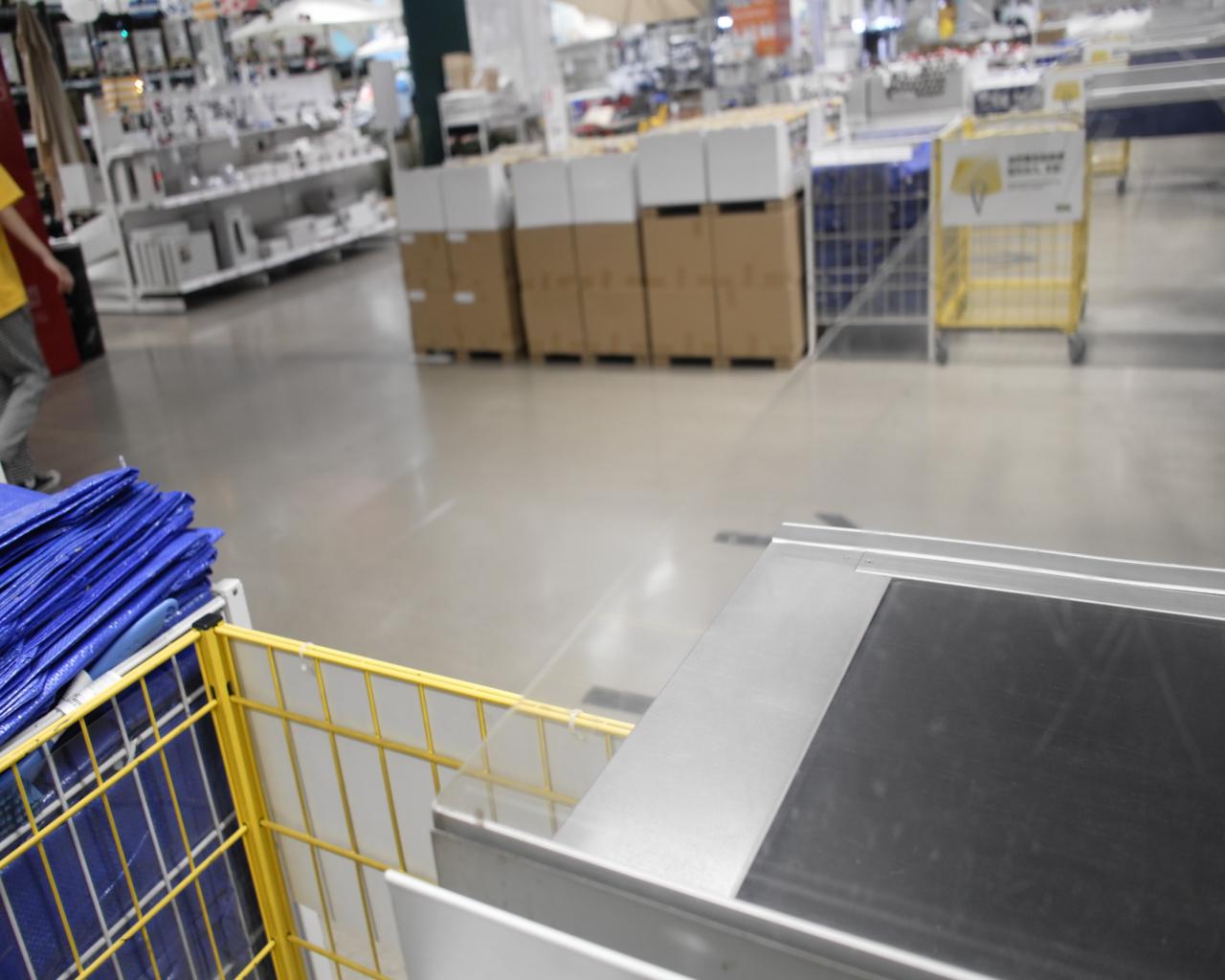}
      \end{minipage}
      }
      \subfloat[GT]{\label{FcGT}
      \begin{minipage}[t]{0.15\textwidth}
      \centering
      \includegraphics[width=1\linewidth]{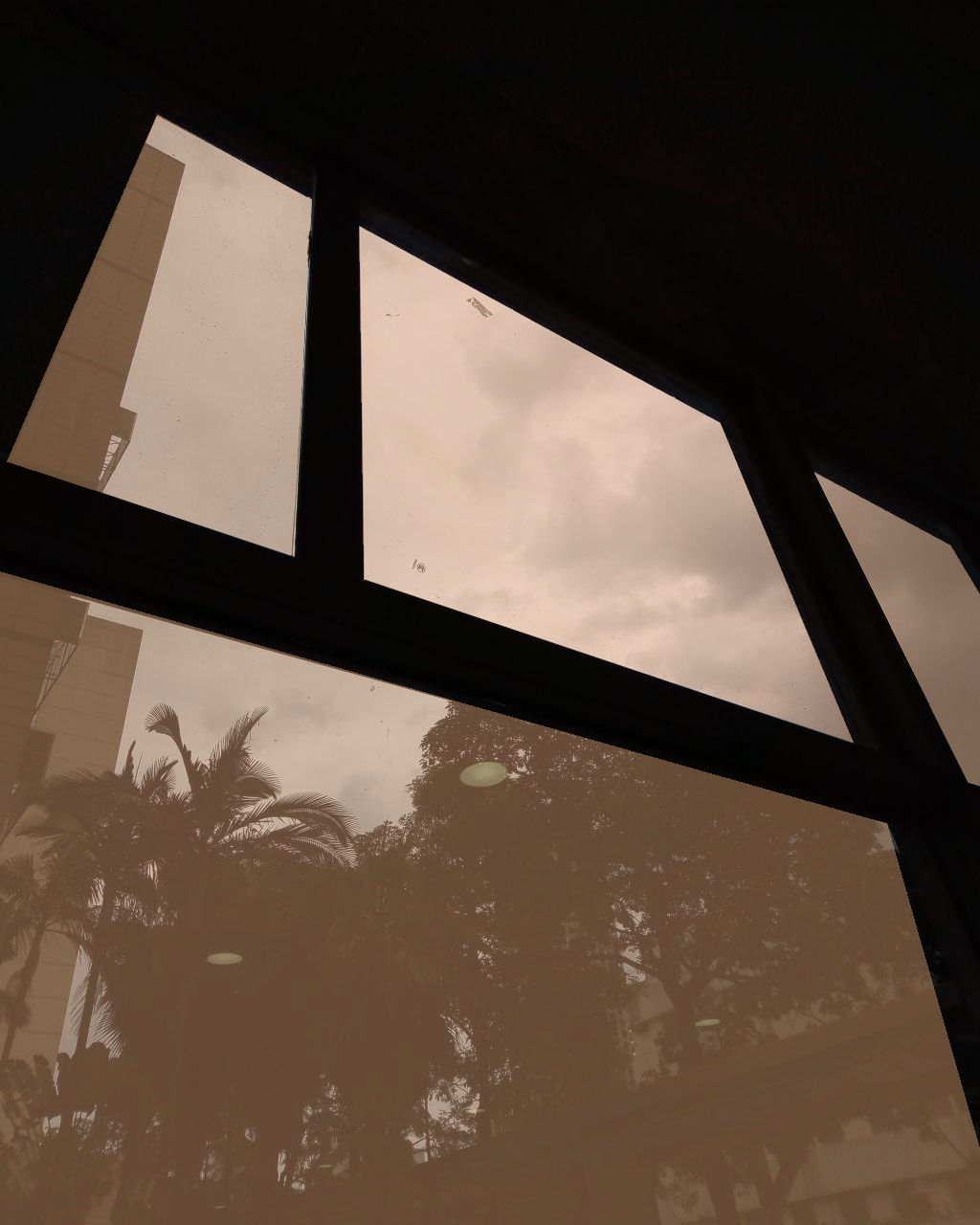}
    
      \includegraphics[width=1\linewidth]{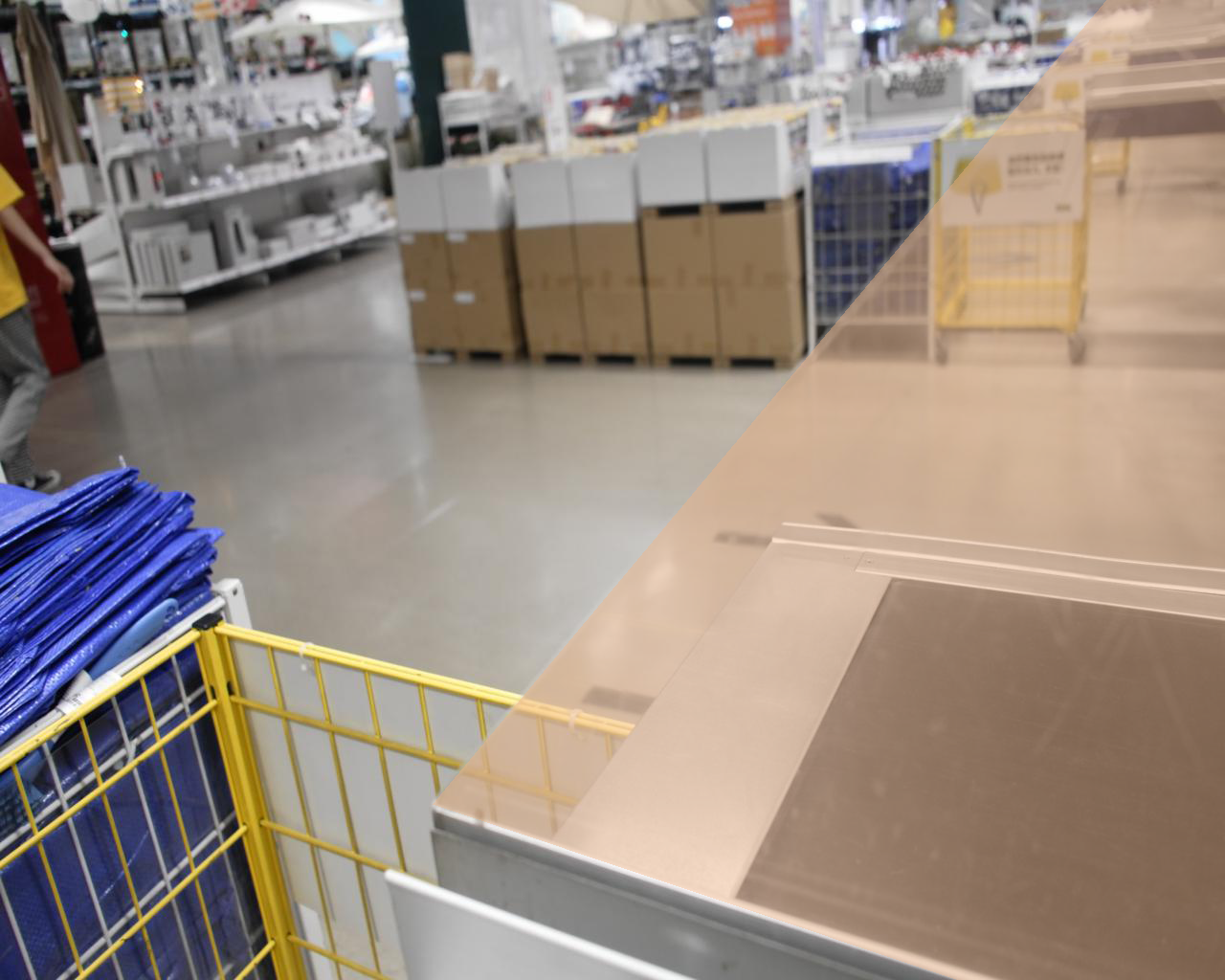}
      \end{minipage}
      }     
      \subfloat[Ours]{\label{result}
      \begin{minipage}[t]{0.15\textwidth}
      \centering
      \includegraphics[width=1\linewidth]{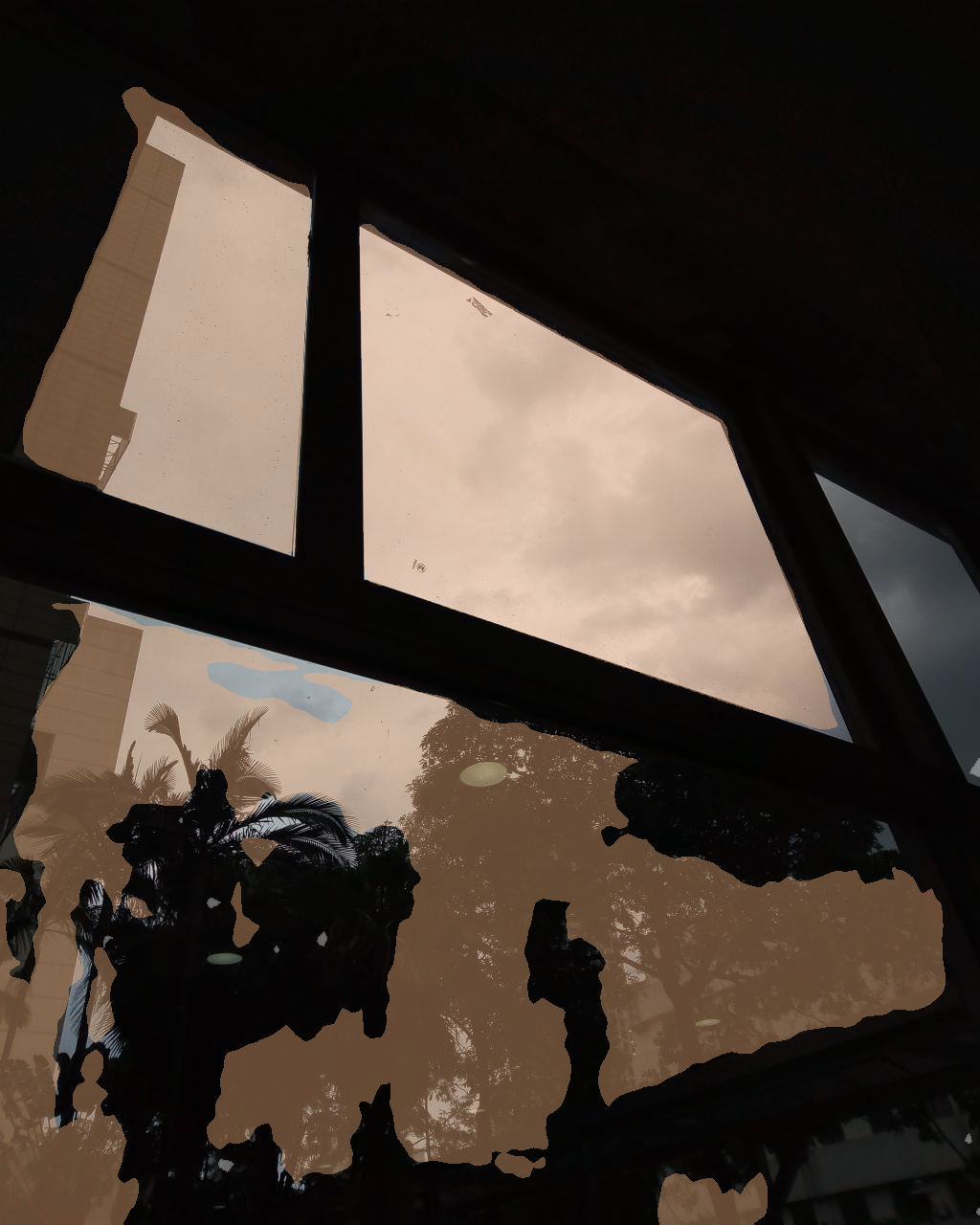}
    
      \includegraphics[width=1\linewidth]{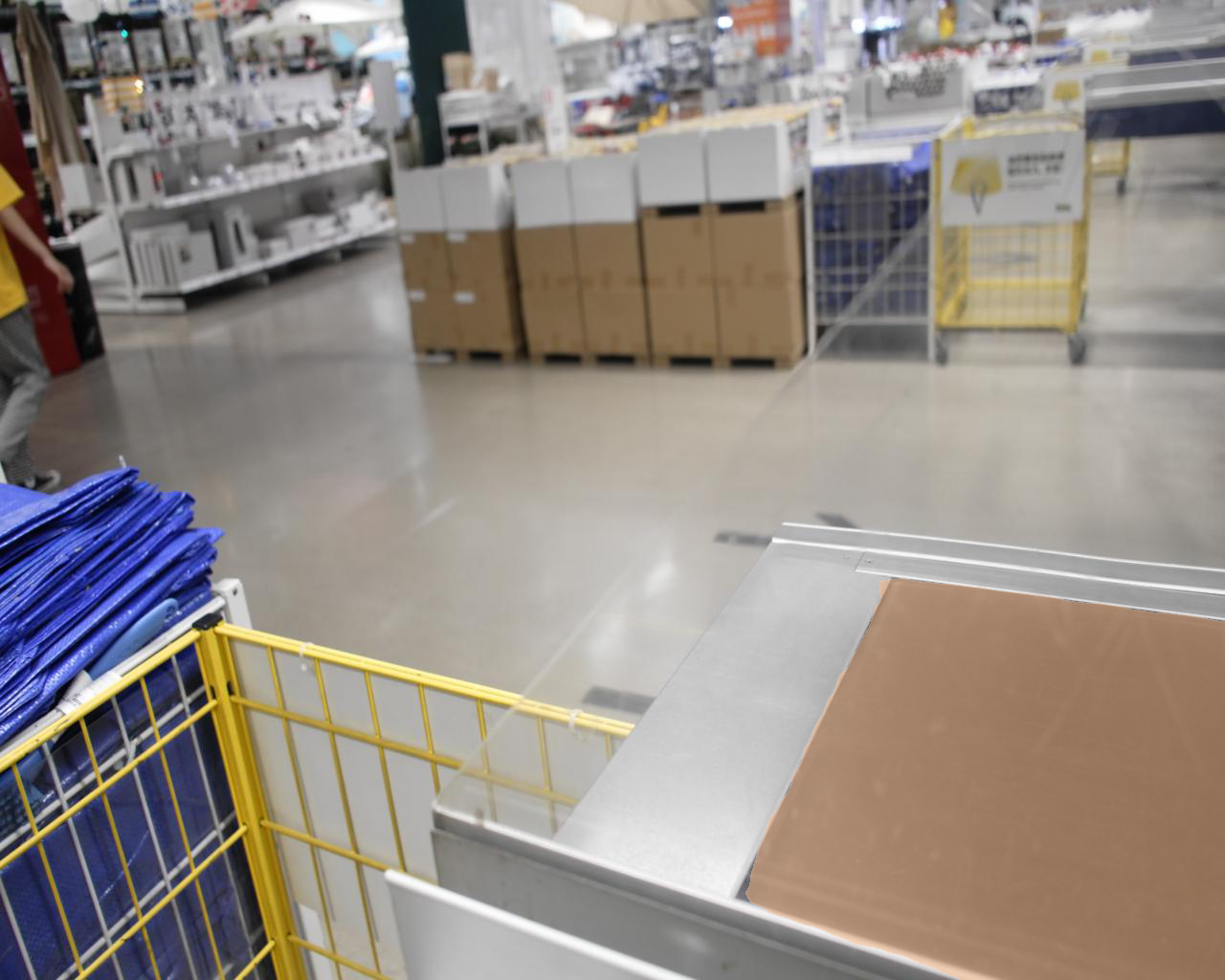}
      \end{minipage}
      } 
      \caption{Failure cases. First row: the glass is almost indistinguishable at low ambient brightness. Second row: the network focuses on the regions with excessive contrasts and ignores the whole glass.}
      \label{fig:FC}
\end{figure}

Although producing satisfactory results on glass segmentation, our GlassSegNet may fail to detect the glass with very similar brightness to the background. 
Poor detection performance occurs for images where parts of the glass region have excessive contrast with other glass regions.
As shown in \Cref{fig:FC}, this is likely due to the lack of related training samples. 


\section{Conclusion}
\label{sec:concl}
In this paper, we propose a glass segmentation network (GlassSegNet) for glass segmentation from  single RGB images. 
Motivated by the process of human recognition of transparent glass, we design a two-stage recognition process, i.e., the identification stage (IS) and the correction stage (CS), to accurately segment the transparent glass. 
The IS is designed to simulate the detection process of human recognition for identifying transparent glass by global context and edge information, and the CS is then used to perform the mistake correction for progressively refining the coarse prediction via correcting mistake regions using gained experience.   
Experimental results show that our model yields new state-of-the-art performance on the benchmark datasets. This in turn answers the raised question in Section \ref{sec:intro}. That is, the mistake correction behavior can successfully benefit the performance boost in glass segmentation.

\bibliographystyle{IEEEtran}
\bibliography{egbib}

\vspace{-10mm}
\begin{IEEEbiography}[{\includegraphics[width=1in,height=1.1in,clip,keepaspectratio]{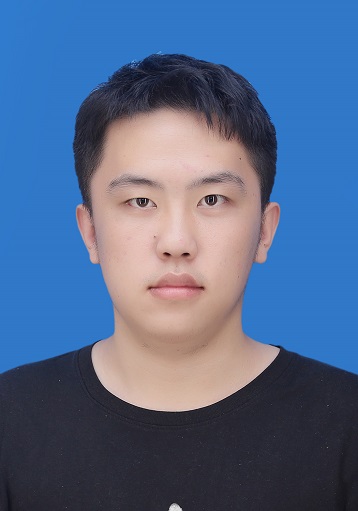}}]{Chengyu Zheng}
is a Ph.D candidate in the School of Computer Science and Technology, Nanjing University of Aeronautics and Astronautics, China, from 2020. His research interests include deep learning, image processing and computer vision.
\end{IEEEbiography}
\vspace{-10mm}

\begin{IEEEbiography}[{\includegraphics[width=1in,height=1.1in,clip,keepaspectratio]{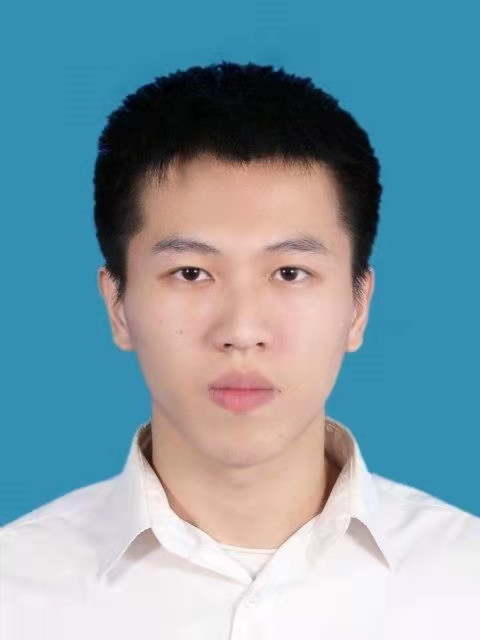}}]{Peng Li}
is a Ph.D candidate in the School of Computer Science and Technology, Nanjing University of Aeronautics and Astronautics, China, from 2020. His research interests include deep learning, image processing and computer vision.
\end{IEEEbiography}

\vspace{-10mm}
\begin{IEEEbiography}[{\includegraphics[width=1in,height=1.1in,clip,keepaspectratio]{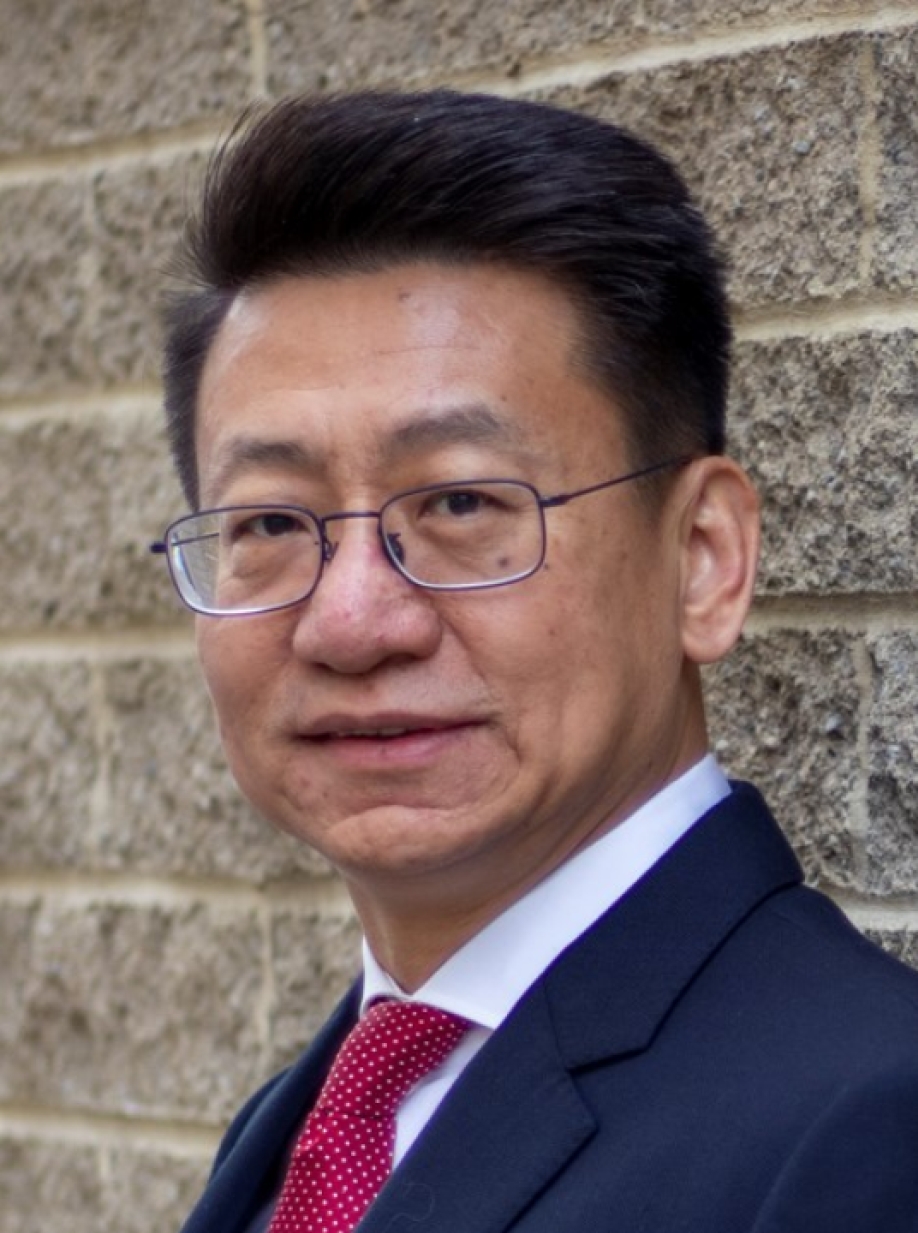}}]{Xiao-Ping Zhang}
received B.S. and Ph.D. degrees from Tsinghua University, in 1992 and 1996, respectively, both in Electronic Engineering. He holds an
MBA in Finance, Economics and Entrepreneurship
with Honors from the University of Chicago Booth
School of Business, Chicago, IL. Since Fall 2000, he has been with the Department of Electrical, Computer and Biomedical Engineering,
Ryerson University, Toronto, ON, Canada, where he is currently a Professor and the Director of the Communication and Signal Processing Applications Laboratory. He has served as the Program Director of Graduate Studies. He is cross-appointed to the Finance Department at the Ted Rogers School of Management, Ryerson University. He was a Visiting Scientist with the Research Laboratory of Electronics, Massachusetts Institute of Technology, Cambridge, MA, USA, in 2015 and 2017. He is a frequent consultant for biotech companies and investment firms. His research interests include sensor networks and IoT, machine learning, statistical signal processing, image and multimedia content analysis, and applications in big data, finance, and marketing.

Dr. Zhang is Fellow of the Canadian Academy of Engineering, Fellow
of the Engineering Institute of Canada, Fellow of the IEEE, a registered
Professional Engineer in Ontario, Canada, and a member of Beta Gamma
Sigma Honor Society. He is the general Co-Chair for the IEEE International
Conference on Acoustics, Speech, and Signal Processing, 2021. He is the
general co-chair for 2017 GlobalSIP Symposium on Signal and Information
Processing for Finance and Business, and the general co-chair for 2019
GlobalSIP Symposium on Signal, Information Processing and AI for Finance
and Business. He is an elected Member of the ICME steering committee. He is the General Chair for the IEEE International Workshop on Multimedia Signal Processing, 2015. He is the Publicity Chair for the International Conference on Multimedia and Expo 2006, and the Program Chair for International
Conference on Intelligent Computing in 2005 and 2010. He served as a
Guest Editor for Multimedia Tools and Applications and the International
Journal of Semantic Computing. He was a tutorial speaker at the 2011
ACM International Conference on Multimedia, the 2013 IEEE International
Symposium on Circuits and Systems, the 2013 IEEE International Conference on Image Processing, the 2014 IEEE International Conference on Acoustics, Speech, and Signal Processing, the 2017 International Joint Conference on Neural Networks and the 2019 IEEE International Symposium on Circuits and Systems. He serves IEEE JOURNAL OF SELECTED TOPICS IN SIGNAL PROCESSING as the Editor-in-Chief from January 2022. He is the Senior Area Editor for the IEEE TRANSACTIONS ON SIGNAL PROCESSING and the IEEE TRANSACTIONS ON IMAGE
PROCESSING. He was Associate Editor for the IEEE TRANSACTIONS ON
IMAGE PROCESSING, the IEEE TRANSACTIONS ON MULTIMEDIA,
the IEEE TRANSACTIONS ON CIRCUITS AND SYSTEMS FOR VIDEO
TECHNOLOGY, the IEEE TRANSACTIONS ON SIGNAL PROCESSING,
and the IEEE SIGNAL PROCESSING LETTERS. He received 2020 Sarwan
Sahota Ryerson Distinguished Scholar Award - the Ryerson University highest honor for scholarly, research and creative achievements. He is selected as IEEE Distinguished Lecturer by the IEEE Signal Processing Society for the term 2020 to 2021, and by the IEEE Circuits and Systems Society for the term 2021 to 2022.
\end{IEEEbiography}
\vspace{-10mm}
\begin{IEEEbiography}[{\includegraphics[width=1in,height=1.1in,clip,keepaspectratio]{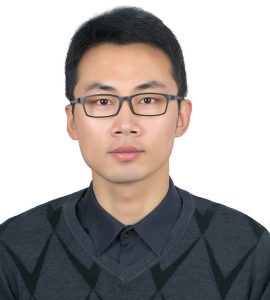}}]{Xuequan Lu} received the Ph.D. degree from Zhejiang University, China, in June 2016. He spent more than two years as a Research Fellow in Singapore. He is currently an Assistant Professor with the School of Information Technology, Deakin University, Australia. His research interests mainly fall into the category of visual computing, for example, geometry modeling, processing, and analysis, animation/simulation, 2D data processing, and analysis. More information can be found at http://www.xuequanlu.com.
\end{IEEEbiography}
\vspace{-10mm}

\begin{IEEEbiography}[{\includegraphics[width=1in,height=1.1in,clip,keepaspectratio]{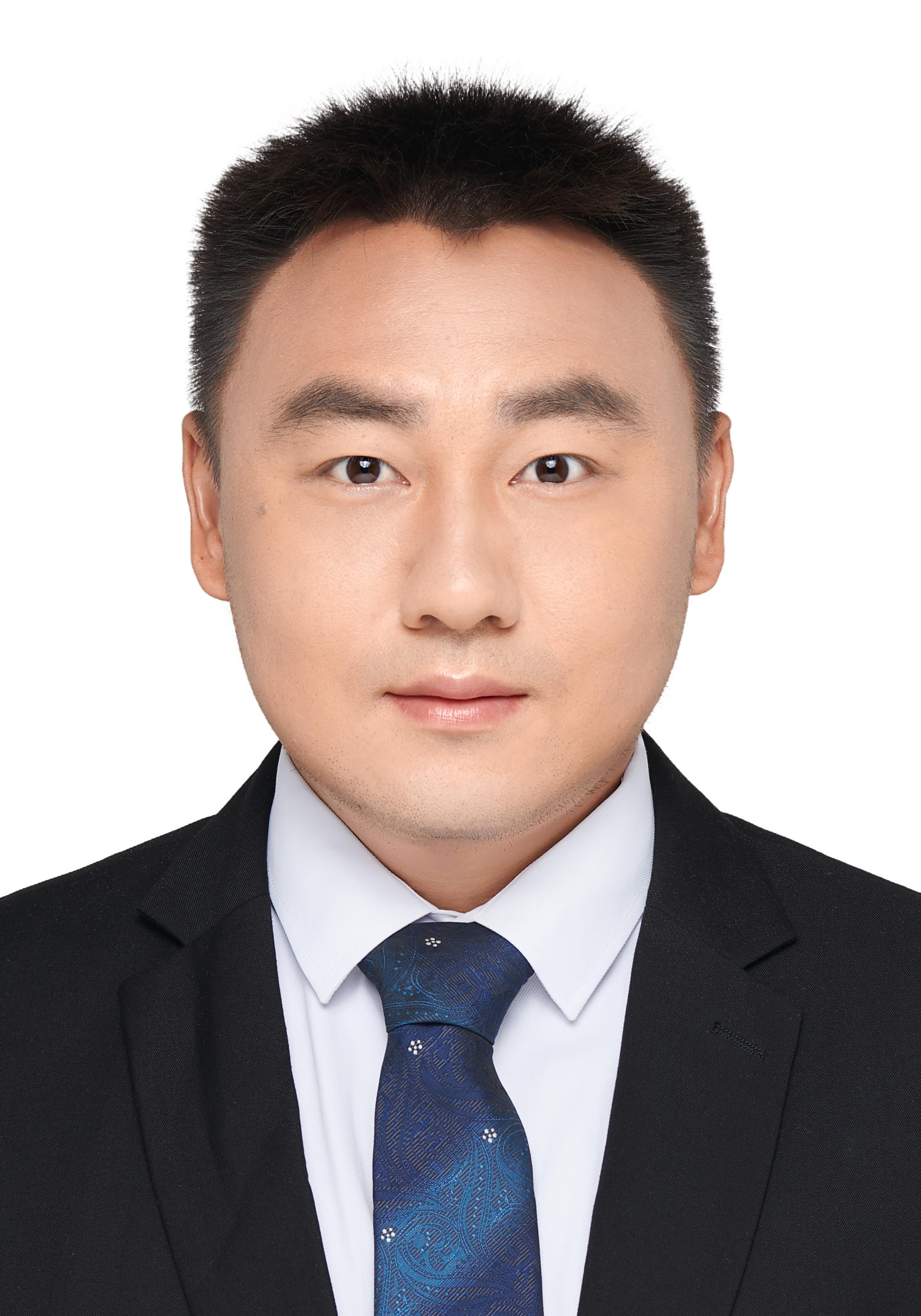}}]{Mingqiang Wei}
received his Ph.D degree (2014) in Computer Science and Engineering from the Chinese University of Hong Kong (CUHK). He is Professor at the School of Computer Science and Technology, Nanjing University of Aeronautics and Astronautics (NUAA). Before joining NUAA, he served as an assistant professor at Hefei University of Technology, and a postdoctoral fellow at CUHK. He was a recipient of the CUHK Young Scholar Thesis Awards in 2014. He is now an Associate Editor for ACM TOMM and The Visual Computer, Journal of Electronic Imaging, and a Guest Editor for IEEE Transactions on Multimedia. His research interests focus on 3D vision, computer graphics, and deep learning.
\end{IEEEbiography}

\end{document}